\documentclass[journal]{IEEEtran}

\usepackage{amssymb}
\usepackage{supertabular}

\usepackage{tikz}

\usepackage{pifont}

\usepackage{bbm}
\usepackage{rotating}
\usepackage{dsfont}

\usepackage{bm}
\usepackage{makecell}
\usepackage{array}
\usepackage{eucal}
\usepackage{graphicx}
\usepackage{epstopdf}
\usepackage{amsfonts,amsthm}
\usepackage{setspace}
\usepackage{cite}
\usepackage[colorlinks,linkcolor=blue,anchorcolor=blue,citecolor=blue]{hyperref}
\usepackage{subfig}
\usepackage{subfloat}
\usepackage{multirow,booktabs}

\usepackage{amsmath,mathrsfs}
\usepackage[misc]{ifsym}

\usepackage{amsmath,graphicx}
\usepackage{epstopdf}
\usepackage{amsfonts,amsthm}
\usepackage{algorithmic}
\usepackage[ruled,linesnumbered,boxed]{algorithm2e}
\usepackage{setspace}
\usepackage{cite}
\usepackage[colorlinks,linkcolor=blue,anchorcolor=blue,citecolor=blue]{hyperref}
\usepackage{subfig}
\usepackage{subfloat}
\usepackage{multirow,booktabs}

\usepackage{amsmath}
\usepackage{amssymb}
\usepackage[misc]{ifsym}
\usepackage{url}

\usepackage{color,soul}
\definecolor{myred}{RGB}{255,0,0}
\definecolor{myblue}{RGB}{0,0,255}
\usepackage{threeparttable}

\usepackage{hyperref}

\hypersetup{hypertex=true,
            colorlinks=true,
            linkcolor=red,
            anchorcolor=green,
            citecolor=blue}

\newtheorem{proposition}{Proposition}[section]

\newtheorem{Definition}{Definition}[section]
\newtheorem{Theorem}{Theorem}[section]
\newtheorem{Remark}{Remark}[section]
\newtheorem{Lemma}{Lemma}[section]
\newtheorem{Assumption}{Assumption}[section]

\newcommand{\tabincell}[2]{\begin{tabular}{@{}#1@{}}#2\end{tabular}}

\ifCLASSINFOpdf
\else
\fi

\begin{document}

\title{Accelerating %Randomized Algorithms for Low-Rank Matrix
%Accelerated
 Large-Scale %Nonconvex
Regularized
 High-Order
 Tensor Recovery}
%%%%%%%%%%%%%%%%%%%%%%%%%%%%%%%%%%%%%%%%%%%%%%%%%%%%%%%%%%%%

\vspace{-1.75cm}
\author{
%Wenjin~Qin,  ~Hailin~Wang,~\IEEEmembership{Student Member,~IEEE,}
%~Jianjun~Wang,~\IEEEmembership{Member,~IEEE,}~Xuequan~Lu,\\  ~Wangmeng~Zuo, ~\IEEEmembership{Senior Member,~IEEE,} %
%~and~Tingwen~Huang,~\IEEEmembership{Fellow,~IEEE}
Wenjin~Qin,  ~Hailin~Wang,~\IEEEmembership{Student Member,~IEEE,}
~Jingyao~Hou,
~Jianjun~Wang,~\IEEEmembership{Member,~IEEE}
%%%%%%%%%%%%%%%%%%%%%%%%%%%%%%%%%%%%%%%%%%%%%%%%%%%%%%%%%%%%%%%%%%%%%%%%%%%%%%%%%%%%%%%%%
\vspace{-0.75cm}
\thanks{
%This work was supported in part by the National Key Research and Development Program of China under Grant 2023YFA1008502; in part by the National Natural Science Foundation of China's Regional Innovation Development Joint Fund under Grant U24A2001; in part by the Natural Science Foundation of Chongqing, China, under Grant
%CSTB2023NSCQ-LZX0044; in part by the National Natural Science Foundation of China under Grant 12301594, Grant 12201505, %12071380,
%Grant 12101512; % and
% in part by the Chongqing Talent Project, China, under Grant cstc2021ycjh-bgzxm0015;
%in part by the Fundamental Research Funds for the Central Universities under Grant SWU-KR25013;
% and in part by  the Initiative Projects for Ph.D. in China West Normal University under Grant 22kE030.
%%
% (Corresponding author: Jianjun Wang.)
%%%%%%%%%%%%%%%%%%%%%%%%%%%%%%%%%%%%%%%%%%%%%%%%%%%%%%%%%%%%%%%%%%%%%%%%%%%%%%%%%%%%%%%%%%%%%%%%%%%%%%%%%%%%%%%%%%%%%%%%%%%%%%%%%%%%%%%%%%%%%
%This work was supported in part by the National Key Research and Development Program of China under Grant 2023YFA1008502; in part by the National Natural Science Foundation of China's Regional Innovation Development Joint Fund under Grant U24A2001; in part by the Natural Science Foundation of Chongqing, China, under Grant
%CSTB2023NSCQ-LZX0044; in part by the National Natural Science Foundation of China under %Grant 12071380, Grant 12101512;
%Grant 12301594, Grant 12201505, %12071380,
%Grant 12101512;
%and in part by the Chongqing Talent Project, China, under Grant cstc2021ycjh-bgzxm0015. (Corresponding author: Jianjun Wang.)
This work was supported in part by the National Key Research and Development Program of China under Grant 2023YFA1008502; in part by the National Natural Science Foundation of China's Regional Innovation Development Joint Fund under Grant U24A2001; in part by the Natural Science Foundation of Chongqing, China, under Grant
CSTB2023NSCQ-LZX0044; in part by the National Natural Science Foundation of China under Grant 12301594, Grant 12201505, %12071380,
Grant 12101512; % and
 in part by the Chongqing Talent Project, China, under Grant cstc2021ycjh-bgzxm0015;
in part by the Fundamental Research Funds for the Central Universities under Grant SWU-KR25013;
 and in part by  the Initiative Projects for Ph.D. in China West Normal University under Grant 22kE030.
(Corresponding author: Jianjun Wang.)
%%%%%%%%%%%%%%%%%%%%%%%%%%%%%%%%%%%%%%%%%%%%%%%%%%%%%%%%%%%%%%%%%%%%%%%%%%%%%%%%%%%%%%%%%%%%%%%%%%%%%%%%%%%%%%%%%%%%%%%%%%%%%%%%%%%%%%%%%%%%%
%
}
\thanks{Wenjin Qin and Jianjun Wang are with the School of Mathematics and Statistics, Southwest University, Chongqing 400715, China (e-mail:
qinwenjin2021@163.com, wjj@swu.edu.cn).
Hailin Wang is with the School of Mathematics and Statistics, Xi'an
Jiaotong University, Xi'an 710049, China (e-mail: wanghailin97@163.com).
Jingyao  Hou  is with the School of Mathematics and Information,
China West Normal University, Nanchong 637009, China
(e-mail: hjy17623226280@163.com).
}
\vspace{-0.5cm}
}
\markboth{Journal of \LaTeX\ Class Files,~Vol.~, No.~, XX~XXXX}%
{Shell \MakeLowercase{\textit{et al.}}: Bare Demo of IEEEtran.cls for IEEE Journals}
%  \vspace{-0.5cm}
%%%%%%%%%%%%%%%%%
\maketitle
\begin{abstract}

Currently, existing tensor recovery methods  fail to recognize
the impact of %the ramifications of
tensor scale variations % fluctuations
on their structural characteristics. %structural properties. % structural characteristics.
Furthermore,
existing studies %tensor recovery methods
face prohibitive computational costs
when dealing with large-scale high-order tensor data.
%When dealing with large-scale high-order tensor data,
%existing tensor recovery methods face prohibitive computational costs. Furthermore,
%    %numerous studies have overlooked the impact of tensor dimension variations on their structural properties.
%they %existing studies
%remain in a state of unawareness regarding the impact of %the ramifications of
%tensor scale variations % fluctuations
%on their structural characteristics. %structural properties. % structural characteristics.
%%%%%%%%%%%%%%%%%%%%%%%%%%%%%%%%%%%%%%%%%%%%%%%%%%%%%%%%%%%%%%%%%%%%%%%%%%%%%%%%%%%%%%%%%%
%In this article,
To alleviate these issue,
assisted by  the Krylov subspace iteration, %method,
 block Lanczos bidiagonalization process, % approach
and random projection strategies,   %techniques,
%we
this article
first devises %develops %devises %design
 two fast and accurate %efficient %an efficient and fast
randomized algorithms for %fixed-precision
 \textit{low-rank  tensor approximation} (LRTA) problem.
%Theoretical bounds on the accuracy of the Frobenius norm error estimate is established.
Theoretical bounds on the accuracy of the approximation error estimate are established.
%On this basis,
Next, we  develop %investigate
 a novel generalized % unified
  nonconvex  modeling %
  framework tailored to %for
  large-scale tensor recovery,
  in which
  % new regularization  %prior representation
%  paradigm is exploited
%  to encode %two
%  insightful tensor priors simultaneously.
a new regularization
paradigm is exploited
 to achieve insightful  prior representation for large-scale %high-order
 tensors.
 %%% i.e., global low-rankness and local smoothness.
%   the RTC  problem within
% the algebraic foundation of  \textit{tensor singular value decomposition} (T-SVD). %framework.
 %%%%%%%%%%%%%%%%%%%%%%%%%%%%%%%%%%%%%%%%%%%%%%%%%%%%%%%%%%%%%%%%%%%%%%%%%%%%%%%%%%%%%%%%%%%%%%%%%%%%%%%%%%%%%%%%
 %Based on the newly proposed %randomized  LRTA strategy and %non-convex
On the basis of the above, % On this basis,
  we further   %have
 investigate  % developed
 new unified nonconvex models
 and %algorithms
  efficient optimization algorithms, respectively,
 for several typical high-order  tensor recovery tasks
 in unquantized and quantized situations.
 %On the basis of the above, we further investigate new unified nonconvex models and efficient optimization algorithms, respectively, for several %typical tasks in both unquantized and quantized high-order tensor recovery.
 %%%%%%%%%%%%%%%%%%%%%%%%%%%%%%%%%%%%%%%%%%%%%%%%%%%%%%%%%%%%%%%%%%%%%%%%%%%%%%%%%%%%%%%%%%%%%%%%%%%%%%%%%%%%%%%%%%%%%%%%%%%%
% (e.g., \textit{unquantized and quantized tensor completion}, \textit{tensor robust principal component analysis}).
 %
% Herein,  in virtue of  \textit{alternating direction method of multipliers} (ADMM)  method, %framework,
%an efficient %effective
%optimization algorithm with convergence guarantees is derived to solve the formulated RTC model
%incorporating novel generalized nonconvex  regularizers.
%%%%%%%%%%%%%%%%%%%%%%%%%%%%%%%%%%%%%%%%%%%%%%%%%%%%%%%%%%%%%%%%%%%%%%%%%%%%%%%%%%%%%%%%%%%%%%%%%%%%%%%%%%%%
To  render  the proposed algorithms  practical and efficient  for large-scale tensor data,
%the designed  randomized approximation scheme
%the designed  fixed-accuracy
the  proposed randomized LRTA schemes are integrated into %%are seamlessly incorporated into
their %its
 central and  time-intensive computations. % %major and time-consuming calculations.
%%%%%%%%%%%%%%%%%%%%%%%%%%%%%%%%%%%%%%%%%%%%%%%%%%%%%%%%%%%%%%%%%%%%%%%%%%%%%%%%%%%%%%%%%%%%%%%%%%%
Finally, we conduct  extensive experiments  on various %real-world tensor data,
large-scale tensors, %multi-dimensional data,
 % and the experimental
whose results  demonstrate  the  practicability, effectiveness and superiority of the proposed method %algorithms
in comparison with some state-of-the-art approaches.
%%
%the proposed method outperforms other state-of-the-art  approaches in terms of both computational efficiency and estimated precision.
\end{abstract}
\vspace{-0.1cm}
\IEEEpeerreviewmaketitle

\vspace{-0.403cm}
\section{\textbf{Introduction}}

\IEEEPARstart{T}{he} era of ``Big Data" has witnessed
massive amounts of  multi-dimensional, multi-modal  datasets in numerous modern  applications, which
%
%The term ``Big Data" refers to multi-dimensional, multi-modal datasets that
are characterized by
(i) large Volume, (ii) high Velocity, and (iii) high Veracity,  and (iv) high Variety \cite{cichocki2016tensor,feldman2020turning55}.
% high velocity, high veracity, high variety, high volume
%Characteristics of these big data can be encapsulated by
% the following   ``V"s: high Volume, high Variety, high Veracity, and  high  Velocity.
%
Each of the ``V" features represents a research challenge in its own right.
%%%%%%%%%%%%%%%%%%%%%%%%%%%%%%%%%%%%%%%%%%%%%%%%%%%%%%%%%%%%%%%%%%%%%%%%%%%%%%%%%%%%%%%%%%%%%%%%%%%%
%Importantly, %For example,
For instance,
large Volume %implies the need for approaches  that are scalable;
%underscores the imperative for
demands scalable strategies that can adapt to increasing data scales;
high Velocity  requires
% the processing of big data  with a  high speed;
%is related to the processing of stream of data in near real-time;
high-efficiency processing mechanisms to ensure timely analysis;
high Veracity calls for robust and reliable %predictive
methods %algorithms
for noisy, incomplete and/or inconsistent data.
%often compromised by noisy, incomplete, or inconsistent data, demands sophisticated, predictive methods to extract reliable insights.
% high Variety %require integration across different types of data.
% demands the fusion of different data types.
%%%%%%%%%%%%%%%%%%%%%%%%%%%%%%%%%%%%%%%%%%%%%%%%%%%%%%%%%%%%%%%%%%%%%%%%%%%%%%%%%%%%%%%%%%%%%%%%%%%
%As such, it has become an increasingly pressing challenge to
%develop new innovative solutions and technologies that can capture, manage,
%and process the data within a tolerable elapsed time.
%develop efficient and effective computational tools that can automatically extract the hidden structures and useful information from such data for % various computer vision tasks.
%Many challenging problems for big data are related to capture, manage, %search, visualize, cluster, classify, assimilate, merge,
% and process the data within a tolerable elapsed time, hence demanding new innovative solutions and technologies.
%
%Consequently,
As such, developing innovative solutions and technologies to %capture, manage,
process and analyze  big data within an acceptable timeframe
%has become  an increasingly urgent challenge.
has become increasingly urgent  and crucial.

%Such massive datasets may have billions of entries and are typically represented in the form of huge block matrices and/or tensors.

Tensors % or multi-dimensional arrays
 possess the capability to represent
% are  the natural representation format of
 %a broad spectrum of
a wide range of %real-world
     %multi-modal and multi-relational
 big data with multi-dimensional/modal/relational/view characteristics.
 %, encompassing   multi-frame/spectral/view data, network flow data, etc.
 Compared with vector/matrix structure-based
representations, tensor can more faithfully and accurately
%deliver inherent %multidimensional  structural characteristics  embedded within the data.
capture  essential structures and correlations  underlying  complex big data.
%%%%%%%%%%%%%%%%%%%%%%%%%%%%%%%%%%%%%%%%%%%%%%%%%%%%%%%%%%%%%%%%%%%%%%%%%%%%%%%%%%%%%%%%%%%%%
%
%As tensors possess the capability to
% As %Since
% tensors  can   capture  essential structures and correlations underlying  complex big data,
%%%%%%%%%%%%%%%%%%%%%%%%%%%%%%%%%%%%%%%%%%%%%%%%%%%%%%%%%%%%%%%%%%%%%%%%%%%%%%%%%%%%%%%%%%%%%%%% %
Thus,
tensor-based approaches and  theories %technologies
 have %recently
been  developed for  data processing and analysis in recent years \cite{liu2021tensors, liu2022tensor}.
% capture the inherent relationships and structures within complex data
%has been recently attracting much attention
%have recently garnered substantial attention from  engineers and researchers.
Among them, tensor recovery is the most fundamental and %critical
 crucial
 task,
%
%%% ############################################################################################################
%Tensor recovery
%(e.g., \cite{ %yang2022robust,lou2019robust,
% goldfarb2014robust, zhao2015bayesian, xie2017kronecker, lu2019tensor,yu2019tensor,
%  zheng2021fullyAAAI, deng2022new, luo2023low  %, hou2025robust,  %,
% %  luo2025lowgl
%%goldfarb2014robust,huang2015provable,zhao2015bayesian,chen2019nonconvex,
%% yang2022robust, liu2024fully, lou2019robust
%%huang2020robust, chen2021auto,liu2021simulated,li2021robust,li2023robust,
%%he2022coarse, liu2024fully,
%}), serving
%%%%%%%%%%%%%%%%%%%%%%%%%%%%%%%%%%%%%%%%%%%%%%%%%%%%%%%%%%%%%%%%%%%%%%%%%%%%%%%%%%%%%%%%%%%%%%%%%%%%%%%%%%%%%%%%%%%%%%%%%
%serves as a cornerstone of tensor processing and analysis,
 %addressing
 %
 %%%%%%%%%%%%%%%%%%%%%%%%%%%%%%%%%%%%%%%%%%%%%%%%%%%%%%%%%%%%%%%%%%%%%%%%%%%%%%%%%%%%%%%%%%%%%%%%%%%%%%%%%%%%%%%%%%%%%%
 which addresses the  challenge of reconstructing   the original tensor from its imperfect
counterpart  degraded by  noise/outliers corruption, information loss, and other factors
\cite{
 han2022optimal, cai2023generalized,  cai2022provable1111,  tong2022scaling,  qin2024guaranteed,
lu2019tensor,  zhang2020low,  hou2021robust, wang2022tensor,
wang2023guaranteed,
 %qin2023nonconvex, zhao2022robust,
 %zhang2023tensor,  gao2023tensor, yu2024generalized,
luo2023low, qin2023nonconvex}.
%noise pollution, abnormal interference, information loss, and other factors.
%plagued by diverse degradation factors.
%%%%%%%%%%%%%%%%%%%%%%%%%%%%%%%%%%%%%%%%%%%%%%%%%%%%%%%%%%%%%%%%%%%%%%%%%%%%%%%%%%%%%%%%%%%%%%%%%%%%%%%%%%%%%%%%%%%%%%%
%In recent years,  this research topic has demonstrated remarkable progress across various domains.
This research topic has increasingly drawn significant  attention %interest
%have recently garnered substantial attention from  engineers and researchers.
across various domains,
%%%%%%%%%%%%%%%%%%%%%%%%%%%%%%%%%%%%%%%%%%%%%%%%%%%%%%%%%%%%%%%%%%%%%%%%%%%%%%%%%%%%%%%%%%%%%%%%%%%%%%%%%%%%%%%%%%%%%%%%%
such as statistics \cite{han2022optimal}, medicine \cite{ShiYongyi2024}, transportation \cite{chen2021bayesian},
 wireless communications \cite{zhang2024integrated}, remote sensing \cite{he2020non},
 machine learning \cite{cai2022provable1111}, signal/image processing \cite{cichocki2015tensor, qin2023nonconvex
 },
and computer vision \cite{panagakis2021tensor22}.

 From the modeling perspective,
 tensor recovery
 covers
 \textit{Low-Rank Tensor Completion} (LRTC) \cite{ zhao2015bayesian22222, long2021bayesian,
 liu2019low,  zheng2021fullyAAAI, wu2022tensor,
 % tong2022scaling,
 % cai2022provable1111,
 %mu2014square22,
  yu2019tensor,        qin2022low, zheng2020tensor44, liu2024revisiting, % wang2023guaranteed,
  feng2023multiplex },  %
   \textit{Robust LRTC} (RLRTC) %
   \cite{goldfarb2014robust, huang2020robust,  li2023robust, liu2024fully,
    hou2025robust, jiang2019robust,lou2019robust,
   wang2020robust,
     song2020robust,yang2022robust },
%%%%%%%%%%%%%%%%%%%%%%%%%%%%%%%%%%%%%%%%%%%%%%%%%%%%%%%%%%%%%%%%%%%%%%%%%
%
%%%%%%%%%%%%%%%%%%%%%%%%%%%%%%%%%%%%%%%%%%%%%%%%%%%%%%%%%%%%%%%%%%%%%%%%%%
%and
 \textit{Tensor Robust Principal Component Analysis} (TRPCA) \cite{lu2019tensor,   zhou2019bayesian11, zhang2020low,
   zheng2020tensor44,
    wang2023guaranteed
   , gao2023tensor, yu2024generalized
   , qin2024tensor},
 and  \textit{Tensor Compressive Sensing} \cite{hou2021robust, wang2022tensor, liu2023tensor, hou2024tensor,  liu2024low11}.
 %
%  have received  widespread  attention recently.
%
%The aforementioned studies
% in terms of
Based on the solutions and techniques employed, %these methods are further subdivided into two branches.
these studies %approaches
%Existing  approaches    for tensor recovery
can be further subdivided into two branches.
 The first branch  is factorization-based methods,
  % (e.g., \cite{zhao2015bayesian22222, zhou2019bayesian11, %zhao2015bayesian, %
% long2021bayesian, liu2019low,
%  zheng2021fullyAAAI, wu2022tensor, %li2021robust,
%  li2023robust,  %, he2023robust
%  tong2022scaling,  %tong2022scaling,
%  cai2022provable1111, liu2024low11
%  }),
  which alternatively update the factors with the predefined initial tensor rank.
  Existing %These
  classical %representative
  studies are %mainly
 % exploited %
 % proposed
 developed
 % by
 % in virtue of
  via
   alternating least squares %\textit{alternating least squares} (ALS)
  \cite{liu2019low,  zheng2021fullyAAAI, wu2022tensor,   li2023robust},
  %\textit{Variational Bayesian Inference} (VBI)
  variational Bayesian inference \cite{zhao2015bayesian22222, zhou2019bayesian11, long2021bayesian},
   gradient-oriented optimization approach \cite{cai2023generalized, tong2022scaling, cai2022provable1111, qin2024guaranteed, liu2024low11}, etc.
  %%%%%%%%%%%%%%%%%%%%%%%%%%%%%%%%%%%%%%%%%%%%%%%%%%%%%%%%%%%%%%%%%%%%%%%%%%%%%%%%%%%%%%%%%%%%%%
  The other important branch   is regularization-based %rank  minimization
   methods,
 %(e.g., \cite{  goldfarb2014robust,  xie2017kronecker,bengua2017efficient,
%  lu2019tensor,   yu2019tensor, huang2020robust,  liu2024fully}),
   which
   % utilizes certain regularization terms to surrogate the tensor rank function.
   %That is to say,
  % In other words,  this type of method
    focuses more on characterizing prior structures
    (e.g., low-rankness and sparsity %  and smoothness
     priors)
   underlying %on
   tensor  data finely via proper
   regularization items. %methods.
   Some typical convex regularizers %regularization strategies
   include
   SNN \cite{goldfarb2014robust} %, % and   square deal \cite{mu2014square22}
    under Tucker framework,
   TRNN \cite{yu2019tensor,huang2020robust} under \textit{tensor ring} (TR) framework,
   FCTN nuclear norm \cite{ liu2024fully} under \textit{Fully-Connected Tensor Network} (FCTN) framework,
   and %UTNN \cite{lu2019tensor,  song2020robust},
   HTNN \cite{ qin2022low}, WSTNN \cite{zheng2020tensor44}, METNN \cite{liu2024revisiting},
   TCTV \cite{wang2023guaranteed}  under \textit{Tensor Singular Value Decomposition} (T-SVD) framework.
To further enhance the recovery performance, a series of nonconvex variants have been subsequently proposed
\cite{ wang2021generalized, yang2022355,  wang2024low2222, zhang2023tensor,
 chen2020robust,qiu2021nonlocal,zhao2022robust,qiu2024robust, %zhao2020nonconvex,
  qin2021robust, qin2022robust ,qin2023nonconvex , gao2023tensor, yu2024generalized
  }.

\begin{figure}[!htbp]
\renewcommand{\arraystretch}{0.0}
\setlength\tabcolsep{1pt}
\centering
\begin{tabular}{c c }
%\centering
%\hline  \\
\includegraphics[width=1.66in, height=1.316in]{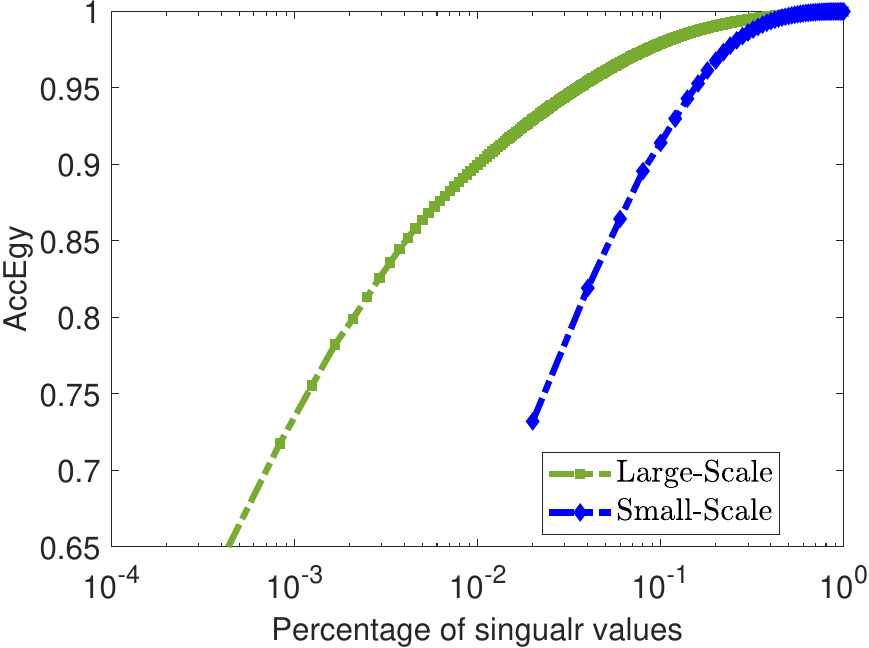}&
\includegraphics[width=1.66in, height=1.316in]{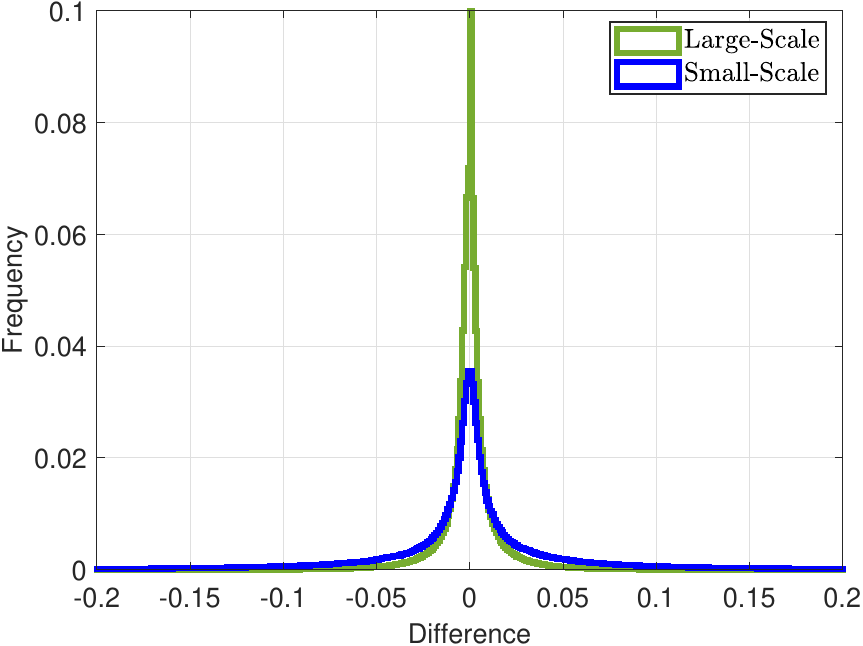} %  hsi_S  smooth_pp1  % smooth-p1 smooth_pp1
%\\ \hline
%
\end{tabular}
%\vspace{-0.15cm}
%\caption{}
\caption{An illustration of the distinctions in $\textbf{L}$ and $\textbf{S}$ prior structures between small-scale and large-scale tensors.
%: Illustrative Examples
%Illustrations of simultaneous $\textbf{L}$ and $\textbf{S}$ prior structures %global low-rankness and local smoothness
%in correlated gradient  tensors.
\textbf{Top:} The AccEgy ($AccEgy = \sum_{i=1}^{k} \sigma_i^2 / \sum_{j=1} \sigma_j^2 $ with $\sigma_i$ denotes the $i$-th tensor singular value) versus the percentage of singular values;
\textbf{Right:}  Frequency histograms  of all corresponding elements of gradient tensor.
 }
\vspace{-0.65248605cm}
\label{INTROls} % \label{fixed-rank-recovery}
\end{figure}
%%%%%%%%%%%%%%%%%%%%%%%%%%%%%%%%%%%%%%%%%%%%%%%%%%%%%%%%%%%%%%%%%%%%%%%%%%%%%%%%%%%%%%%%%%%%%%%%%%%%%%%%%%%%%%%%%%%%%%%%%%%%%%%%%%%%%%%%%%%%
%

%it is increasingly evident that as the scale of tensor data expands, its inherent low - rank and smooth properties become more pronounced.

Although the above %methods based on regularization strategies
regularization-based methods
  yield commendable  performance % outcomes
 in practical applications, %they are  mainly applicable   to small-scale tensor data.
 they demonstrate inefficiency in handling   large tensor data.
 % not be suitable for
%With the rapid advancement of emerging technologies including  artificial intelligence and large model,
%a wide variety of large-scale tensor data has sprung up in numerous fields such as image processing, computer vision, and remote sensing.
%%
In response to the demands of the big data era,
% In response to the demands of the big data era
%%
%Currently,
it is of crucial importance to
address the issue of large-scale tensor recovery
 in a fast and accurate manner.
 %remains fraught with  challenges,  %%%  primarily in terms of modeling and computation.
 Speed is %mainly
 manifested in the computation module, whereas accuracy is %primarily
  embodied in the modeling module.
%Although they have achieved remarkable accomplishments in various fields, they are not efficient in handling large-scale tensor data
However,
 achieving high accuracy and high speed simultaneously remains fraught with  challenges, %is challenging,
 because there is a trade-off between accuracy and speed.

 % However, processing these tensor - structured big data is non - trivial due to their high dimensionality, large scale, and the presence of noise, outliers, and missing values, which necessitates the development of advanced data processing techniques.

\begin{figure}[!htbp]
\renewcommand{\arraystretch}{0.0}
\setlength\tabcolsep{0pt}
\centering
\begin{tabular}{c c }
\centering
\includegraphics[width=1.8in, height=1.4316in]{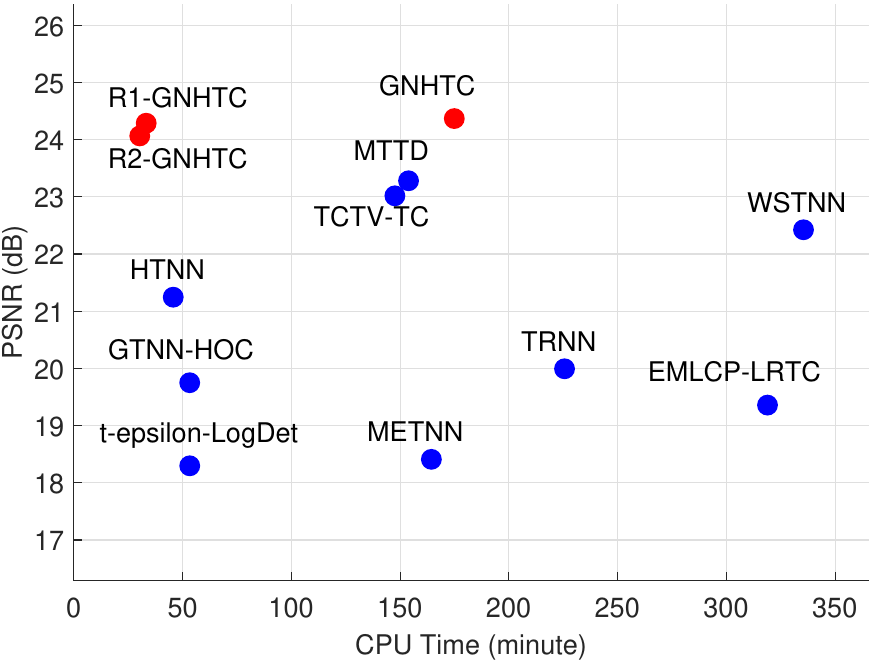}
&
\includegraphics[width=1.7in, height=1.4316in]{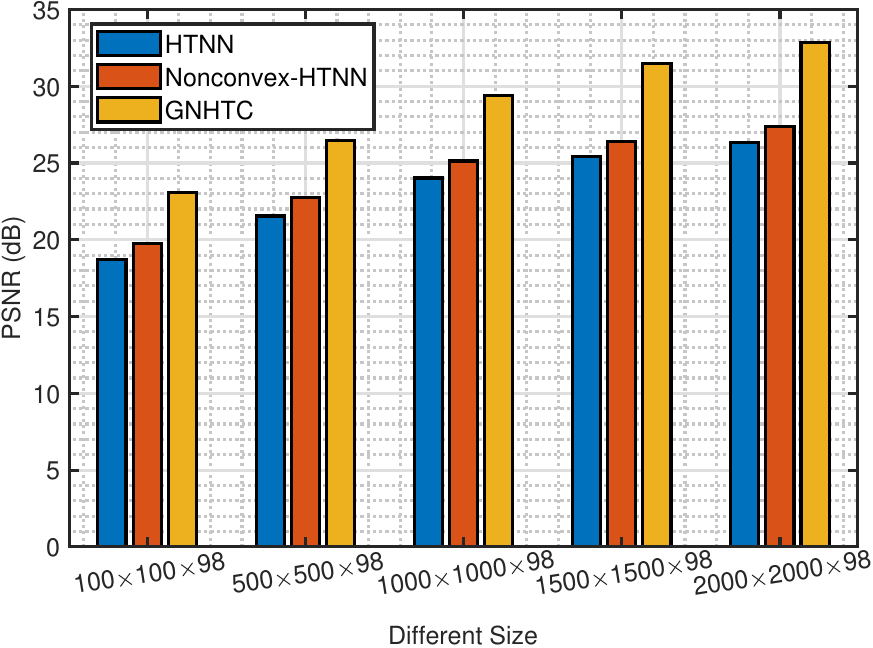}
\end{tabular}
%\vspace{-0.15cm}
%\caption{}
\caption{\textbf{Left:} The   CPU time %comparisons of
 versus % and
  average PSNR
  in LRTC task  (see Table \ref{table-lrtc-MSIMSI} for more details).
The proposed randomized algorithms achieve the lowest computation cost.
\textbf{Right:}
The tensor scale versus
%The  %comparisons of
 average PSNR in LRTC task. %versus
 }
\vspace{-0.76cm}
\label{INTROtimevsPsnr} % \label{fixed-rank-recovery}
\end{figure}

 One of the challenges lies in how to  explore
 effective and concise regularization strategies to reveal %uncover
 the %critical and insightful
 insightful  prior features %structures
 of large-scale high-order tensors, and further investigate reliable and scalable recovery models.
%%%%%%%%%%%%%%%%%%%%%%%%%%%%%%%%%%%%%%%%%%%%%%%%%%%%%%%%%%%%%%%%%%%%%%%%%%%%%%%%%%%%%%%%%%%%%%%%%%%%%%%%%%%%%%%%%%%%%%%%%%%%%%%%%%%
% Thus, how to  tackle the problem of large-scale tensor recovery %remains a challenge.
% remains fraught with a multitude of challenges.
%%%%%%%%%%%%%%%%%%%%%%%%%%%%%%%%%%%%%%%%%%%%%%%%%%%%%%%%%%%%%%%%%%%%%%%%%%%%%%%%%%%%%%%%%%%%%%%%%%%%%%%%%%%%%%%%%%%%%%%%%%%%%%
%On the one hand,
Previous studies
(e.g., \cite{qin2022low, zheng2020tensor44, liu2024revisiting,  wang2023guaranteed,qin2023nonconvex,  feng2023multiplex, zhang2023tensor})
have not managed to cast deep insights upon the prior information of
 large high-order tensors.
Besides, %Existing studies
they are
 %Moreover, they
 still %remain %be
 in a state of unawareness regarding the ramifications
 of tensor scale fluctuations on their %inherent
 structural characteristics.
 %%%%%%%%%%%%%%%%%%%%%%%%%%%%%%%%%%%%%%%%%%%%%%%%%%%%%%%%%%%%%%%%%%%%%%%%%%%%%%%%%%%%%%%%%%%%%%%%%%%%%%%%%%%%
% As illustrated in Figure 1,
%%%%%%%%%%%%%%%%%%%%%%%%%%%%%%%%%%%%%%%%%%%%%%%%%%%%%%%%%%%%%%%%%%%%%%%%%%%%%%%%%%%%%%%%%%%%%%%%%%%%%%%%%%%%
 As illustrated in Figure \ref{INTROls},   we  uncover a  universal
 phenomenon:  with the increase in the scale of tensor data,
  % As the scale of tensor data expands,
  it tends to simultaneously exhibit stronger global low-rankness (\textbf{L}) and local smoothness  (\textbf{S}).
  %%%%%%%%%%%%%%%%%%%%%%%%%%%%%%%%%%%%%%%%%%%%%%%%%%%%%%%%%%%%%%%%%%%%%%%%%%%%%%%%%%%%%%%%%%%%%%%%%%%%%%%%%%%%%%%%%%%%%%%%%%%%%%%%%%%%%%%%%%%
  %
  %%%%%%%%%%%%%%%%%%%%%%%%%%%%%%%%%%%%%%%%%%%%%%%%%%%%%%%%%%%%%%%%%%%%%%%%%%%%%%%%%%%%%%%%%%%%
   %The aforementioned findings %facts
%  inspire us  to investigate   novel generalized  nonconvex regularization strategies
%  that can flexibly  %  profoundly
%  excavate two crucial prior structures  embedded within % the
%  large high-order tensors.
  To profoundly excavate these two crucial prior structures,
  %captures L + S
  % in a concise and simultaneous manner,  % within  the big high-order  tensors,
   we consider
  % investigating
  exploring
    a novel %generalized  nonconvex
    regularization strategy, in which generalized nonconvex constraints are imposed on  % applied to
    % the gradient domain
    the resulting gradient maps
     of the original big %high-order
     tensors.
  %
%    The aforementioned facts inspire us  to investigate  a novel generalized  nonconvex regularization strategy applied
%    to the gradient domain of the original large tensor, which can
%     profoundly excavate two crucial prior structures within  the large tensor.
  %%%%%%%%%%%%%%%%%%%%%%%%%%%%%%%%%%%%%%%%%%%%%%%%%%%%%%%%%%%%%%%%%%%%%%%%%%%%%%%%%%%%%%%%%%%
  %Compared with
  %
%
  Different from  the nonconvex low-rank
  regularization methods that previously acted on the original domain
  \cite{ wang2021generalized, yang2022355,  wang2024low2222, zhang2023tensor,
 chen2020robust,qiu2021nonlocal,zhao2022robust,qiu2024robust, %zhao2020nonconvex,
  qin2021robust, qin2022robust ,qin2023nonconvex, gao2023tensor, yu2024generalized},
   this gradient mapping-based nonconvex paradigm %approach
    can simultaneously
 %  capture
   encode \textbf{L}+\textbf{S} priors in a  concise manner,
    and  demonstrates  %superiority.
  % both
   superiority and effectiveness.
   A supporting example  shown on %the right side of Figure 1 demonstrates that:
  %
  % Correspondingly,
   the right-hand side of
   %in
    Figure \ref{INTROtimevsPsnr}  indicates that
  in the same scenario, as the scale of tensor data expands,
  the performance of tensor recovery %-based
  methods becomes increasingly prominent.
  Moreover, the performance disparity between the proposed nonconvex
   GNHTC method based on the joint \textbf{L} and \textbf{S} priors
  %under the joint \textbf{L} and \textbf{S} prior paradigm
  and
 %
  %%
%  in the same scenario, the performance of tensor recovery methods becomes increasingly evident as the scale of tensor data increases.
%   Moreover, the performance gap between the nonconvex recovery method under the joint \textbf{L} and \textbf{S} prior paradigm and
%   the convex or   nonconvex method  only utilizing the pure \textbf{L} prior widens as the tensor data scale grows.
   %%
   %%%%%%%%%%%%%%%%%%%%%%%%%%%%%%%%%%%%%%%%%%%%%%%%%%%%%%%%%%%%%%%%%%%%%%%%%%%%%%%%%%%%%%%%%%%%%%%%%%%%%%%%%%%%%%%%%%%%%%%%%%%%%%%%%%%%%%%%%%%%%%%%%%%%%%%%
%
 the others  only utilizing the  \textbf{L} prior
   also becomes more
 pronounced.

 % Therefore,
  % Consequently,
  % it is a reasonable modeling approach to consider imposing nonconvex constraints on the resulting gradient tensors of raw large-scale tensor.

  % the right-hand side of Figure \ref{INTROtimevsPsnr}

 From a computational perspective, existing %regularization-based
 tensor recovery   methods generally require to  perform the SVD-based %SVD/T-SVD-based
 (e.g., \cite{goldfarb2014robust, yu2019tensor,huang2020robust,
 qin2022low, zheng2020tensor44, liu2024revisiting,  wang2023guaranteed, feng2023multiplex, %zhang2023tensor,
 wang2021generalized, yang2022355,  wang2024low2222, zhang2023tensor,
 chen2020robust,qiu2021nonlocal,zhao2022robust,qiu2024robust, %zhao2020nonconvex,
  qin2021robust, qin2022robust ,qin2023nonconvex, gao2023tensor, yu2024generalized
 })
or alternating minimization-based
(e.g., \cite{liu2019low,  zheng2021fullyAAAI, wu2022tensor, liu2024fully,  li2023robust})
multiple low-rank approximations.
  %
%which suffers from high computational costs when dealing with large-scale high-order tensor data.
 These studies %will encounter the issue of
 suffer from the curse of dimensionality
 %high computational costs and storage requirements
 when dealing with large-scale high-order tensor data,  thereby leading to high computational costs and storage requirements.
%
%%%%%%%%%%%%%%%%%%%%%%%%%%%%%%%%%%%%%%%%%%%%%%%%%%%%%%%%%%%%%%%%%%%%%%%%%%%%%%%%%%%%%%%%%%%%%%%%%%%%%%%%%%%%%%%%%%%%%%%%%%%%%%%%%%
As illustrated on the left-side of Figure \ref{INTROls}, with the escalation of the scale of tensor data, it typically manifests %displays
a heightened degree   of redundancy, which is tantamount to the low-rank property.
As thus,
%
%Given the aforementioned fact,
%the first research motivation of this paper is to explore effective dimensionality reduction methods for large-scale tensor data,
%
  another challenge lies in how to develop    efficient low-rank %effective
  compression and approximation    methods for %applicable to  %dimension-reduction
large-scale tensors,
% while preserving their % critical
%major structural information,
 with the aim of achieving dimension-reduction
 %avoiding the curse of dimensionality.
  % of large-scale tensors
 and %thereby
  elevating computational efficiency.

%
%%%%%%%%%%%%%%%%%%%%%%%%%%%%%%%%%%%%%%%%%%%%%%%%%%%%%%%%%%%%%%%%%%%%%%%%%%%%%%%%%%%%%%%%%%%%%%%%%%%%%%%%%%%%%%%%%%%%%%%%%%%%%%%%%%%%%%%%%%%%%%%%%

 %%%%%%%%%%%%%%%%%%%%%%%%%%%%%%%%%%%%%%%%%%%%%%%%%%%%%%%%%%%%%%%%%%%%%%%%%%%%%%%%%%%%%%%%%%%%%%%%%%%%%%%%%
 %Driven by the fact %
 %Randomized algorithms
 Randomized sketching approach is a powerful tool for handling optimization problems involving a large amount of data,
 which is known for its low memory and computational complexities
 (e.g., \cite{halko2011finding,  martinsson2020randomized,
martinsson2016randomized1, tropp2017practical1,  %yu2017single,
yu2018efficient1,  duersch2020randomized1, %tropp2019streaming,
feldman2020turning55,
hallman2022block1, demmel2023improved,
%yao2018large111
tropp2023randomized
}).
Driven by these advantages,
 several popular
sketching techniques (e.g., random sampling \cite{fu2020block55, larsen2022practical, zhang2023randomized, tarzanagh2018fast}, random projection \cite{zhang2018randomized,  che2022fast,  %che2020computation33, yu2022practical
che2020computation33,che2021randomized,che2021efficient, che2023randomized,dong2023practical
},
coresets \cite{chhaya2020streaming}) have been employed to design fast randomized low-rank  approximation  algorithms for large-scale tensors.
However,  existing approximation algorithms based on T-SVD essentially perform % can only
dimensionality reduction on the spatial modes \cite{tarzanagh2018fast, zhang2018randomized,  che2022fast,qin2023nonconvex},
 %reduce the spatial dimensions of large tensors,
% which consequently leads to unsatisfactory computational efficiency.
thereby
resulting in suboptimal computational efficiency.
%
% the above-mentioned LRTA algorithms essentially perform a random projection on each frontal slice in the transformed tensor.
 % As a result, there is still room for improvement in their computational efficiency.
%%
To tackle the aforementioned concerns, we consider designing efficient and adaptive Tucker compression methods,
 which executes %random projection
 dimensionality reduction
along each mode of the tensor, as a preprocessing step. Following this preprocessing, the associated deterministic
algorithms under T-SVD framework can be quickly computed via the smaller Tucker core tensor.

\vspace{-0.00cm}
On the basis of the two research contents considered above,
%%%%%%%%%%%%%%%%%%%%%%%%%%%%%%%%%%%%%%%%%%%%%%%%%%%%%%%%%%%%%%%%%%%%%%%%%%%%%%%%%%%%%%%%%%%%%%%%%%%%%%%%%%%%%%%
% modeling  leveraging generalized nonconvex regularizers and the  fixed-accuracy LRTA algorithms
% leveraging full-mode random projection strategy,
%%%%%%%%%%%%%%%%%%%%%%%%%%%%%%%%%%%%%%%%%%%%%%%%%%%%%%%%%%%%%%%%%%%%%%%%%%%%%%%%%%%%%%%%%%%%%%%%%%%%%%%%%%%%%%%%%
we continue to  think about how to develop %efficient and
robust, scalable,    and reliable
 algorithms equipped with convergence guarantee to solve the formulate
tensor recovery  models that leverages  generalized nonconvex regularizers.
 In addition, our consideration also involves
 embedding  the developed  fast randomized compression and approximation   algorithms
 into the main and time-consuming calculations of the proposed recovery  method,
  thereby enhancing its practicality and efficiency of processing large-scale tensor data.
%
%Finally, we apply the proposed algorithms   to various  multidimensional data
% that suffer from   simultaneous elements loss and noise/outliers corruption,
% like  \textit{multispectral images} (MSIs),  \textit{hyperspectral images}  (HSIs),
%\textit{magnetic resonance images} (MRIs),   \textit{color videos} (CVs),
%\textit{multi-temporal remote sensing images} (MRSIs), \textit{light field images} (LFIs).
 %%%
To sum up,  in  this  article, we delve into   an innovative generalized nonconvex  tensor modeling  method,
which  exhibits powerful capabilities in prior characterization of large-scale high-order tensors.
Besides,  novel fast randomized  \textit{low-rank tensor approximation}  (LRTA) algorithms  leveraging full-mode random projection strategy
 are   incorporated to alleviate computational bottlenecks
 encountered when dealing with  large-scale tensor data.
Main contributions of this work are summarized as follows:

%\begin{itemize}
 % \item
1) Firstly,  leveraging the block Krylov subspace approach, block Lanczos bidiagonalization process and random projection technique, we  devise
efficient randomized algorithms for solving fixed-rank and fixed-precision LRTA problems in the Tucker format, respectively.
%
%%%%%%%%%%%%%%%%%%%%%%%%%%%%%%%%%%%%%%%%%%%%%%%%%%%%%%%%%%%%%%%%%%%%%%%%%%%%%
%
 Theoretical bounds on the accuracy of the approximation   error estimate are established.
Compared with the %previous
  deterministic LRTA methods, the proposed one exhibit remarkable advantage in computational efficiency.
% speedup

%\item
2) Secondly, a novel generalized nonconvex  tensor modeling framework is developed
by  investigating new  regularization schemes.
The first scheme  has a strong ability to  encode two insightful and essential
 tensor prior information   simultaneously, i.e., $\textbf{L}$+$\textbf{S}$ priors,
and the second  scheme  can well  enhance the robustness against various structured  noise/outliers,
like entry-wise, slice-wise and tube-wise forms.
%%%%%%%%%%%%%%%%%%%%%%%%%%%%%%%%%%%%%%%%%%%%%%%%%%%%%%%%%%%%%%%%%%%%%%%%%%%%%%%%%%%%%%%%%%%%%%%%%%%%%%%%%%%
%%%%%%%%%%%%%%%%%%%%%%%%%%%%%%%%%%%%%%%%%%%%%%%%%%%%%%%%%%%%%%%%%%%%%%%%%%%%%%%%%%%%%%%%%%%%%%%%%%%%%%%%%%%%%%%%%%%%
Notably,  the proposed framework includes most existing  relevant   approaches as special cases.

%\item
3)
Thirdly, efficient and scalable models, along with their  optimization algorithms with convergence guarantees,
are investigated for several typical tensor recovery tasks  in both  unquantized and quantized scenarios.
The coupling of the proposed randomized LRTA  strategies can significantly  quicken  the computational speed of the developed
 recovery  algorithms, especially for large-scale high-order tensors
 (\textit{see Figure \ref{INTROtimevsPsnr}}).
%(see Section   \ref{experiments}---Subsection \ref{randomized-diss}).
The   practicability,  effectiveness and superiority of the proposed  method
are proven by extensive experiments on  various large-scale multi-dimensional  data.
\textcolor[rgb]{0.00,0.00,0.00}{\textbf{What is particularly noteworthy is that
all theoretical proofs, related algorithms, and more experimental results of this article are provided in the supplementary material.}}

%\section{\textbf{Related Work}}\label{related}

%\subsection{\textcolor[rgb]{0.00,0.00,0.00}{\textbf{An Overview of Randomized LRTA Methods}}}
%\subsubsection{\textcolor[rgb]{0.00,0.00,0.00}{\textbf{ALS-Based LRTA Methods}}}
%\subsubsection{\textcolor[rgb]{0.00,0.00,0.00}{\textbf{SVD-Based LRTA Methods}}}

\vspace{-0.5048cm}
\section{\textbf{notations and preliminaries}}\label{nota}
%\vspace{-0.1cm}
%For brevity,
In this section,  the main notations and preliminaries utilized  in
the whole paper are summarized,
most of which originate form  the literatures \cite{kolda2009tensor, martin2013order, qin2022low,qin2023nonconvex,wang2023guaranteed}.
%\textbf{see supplementary material for details}.

We use $x$, $\textbf{x}$, $\textbf{X}$, and $\bm{\mathcal{X}}$ to denote scalars, vectors, matrices, and tensors, respectively.
%The set of integers $\{1,2, \cdots, n\} $ is denoted as $[n]$.
For a matrix $\bm{{X}} \in \mathbb{R}^{m \times  n}$,
$\bm{I}_{n} \in \mathbb{R} ^{n\times n}$, $\operatorname{trace}(\bm{X})$,
 $ {{\bm{X}}}^{\mit{H}} ({{\bm{X}}}^{\mit{T}})$,
  $ \langle {{\bm{X}}},{{\bm{Y}}} \rangle=\operatorname{trace} ({{\bm{X}}}^{\mit{H}} \cdot {{\bm{Y}}})$ and
   ${\|{{\bm{X}}}\|}_{\star}= {\big(\sum_{i}  \; \big|\sigma_{i}({{\bm{X}}})\big| \big)}$
 denote its
 identity matrix,  trace, conjugate transpose (transpose), inner product and  nuclear   norm, respectively.
 %%%
For an order-$d$ tensor $\boldsymbol{\mathcal{X}} \in \mathbb{R}^{n_1 \times  \cdots \times n_d}$,
$\boldsymbol{\mathcal{X}}_ {i_1,\cdots,i_d}$  denotes
its $(i_1,\cdots,i_d)$-th element,
     ${{\bm{X}}_{(k)}} $ denotes its
      \textbf{mode-$k$ unfolding},
   % ${\|{\boldsymbol{\mathcal{A}}}\|}_{\infty}= \max_{i_{1} \cdots i_d} |{\boldsymbol{\mathcal{A}}_{i_{1} \cdots i_d}}|$ &tensor infinity norm
%    \cr
and
     $
    {\boldsymbol{\mathcal{X}}}^{<j>}:={\boldsymbol{\mathcal{X}}}{(:,:,i_3,\cdots, i_{d})},
    j={\sum_{a=4}^{d} }   {(i_a-1){\Pi}_{b=3}^{a-1}n_b}+i_3$
is called as  its %the matrix
\textbf{face slice}.
Then $ {\operatorname{bdiag}} ({{\boldsymbol{\mathcal{X}}}})$
 is the  \textbf{block diagonal matrix} whose $i$-th block equals to ${\boldsymbol{\mathcal{X}}}^{<i>}$,
 $\forall
%j \in \{1,\cdots,n_3\cdots n_d\}$.
i \in \{1,2,\cdots,n_3\cdots n_d\}$.
%
%%%%%%%%%%%%%%%%%%%%%%%%%%%%%%%%%%%%%%%%%%%%%%%%%%%%%%%%%%%%%%%%%%%%%%%%%%%%%%%%%%%%%%%%%%%%%%%%%%%%%%%%%%%%%%%%%%%%%%%
The \textbf{inner product} of two tensors $\boldsymbol{\mathcal{X}}$ and $\boldsymbol{\mathcal{Y}}$ with the same size is defined as
%the sum of the products of their entries, i.e., $\langle \boldsymbol{\mathcal{X}}, \boldsymbol{\mathcal{Y}} \rangle = \sum_{i_1,i_2,\cdots,i_N}
%\boldsymbol{\mathcal{X}} _{i_1,i_2,\cdots,i_N}    \cdot \boldsymbol{\mathcal{Y}} _{i_1,i_2,\cdots,i_N}$.
$\langle {\boldsymbol{\mathcal{X}}},{\boldsymbol{\mathcal{Y}}} \rangle  = {\sum}_{j=1}^{n_3 n_4 \cdots n_d} \langle {\bm{\mathcal{X}}}^{<j>},{\bm{\mathcal{Y}}}^{<j>} \rangle$.
%%%%%%%%%%%%%%%%%%%%%%%%%%%%%%%%%%%%%%%%%%%%%%%%%%%%%%%%%%%%%%%%%%%%%%%%%%%%%%%%%%%%%%%%%%%%%%%%%%%%%%%%%%%%%%%%%%%%%%%%%
The \textbf{$\ell_1$-norm}, \textbf{Frobenius norm},
 %and
  \textbf{$\operatorname{\textbf{tube}}_1$-norm}
  and \textbf{$\operatorname{\textbf{slice}}_1$-norm}
of $\boldsymbol{\mathcal{X}}$ are defined as
%$\|\boldsymbol{\mathcal{X}}\|_1=\sum_{i_1,i_2,\cdots,i_N}|\boldsymbol{\mathcal{X}}_  {i_1,i_2,\cdots,i_N} |$,
${\|{\boldsymbol{\mathcal{X}}}\|}_{\ell_1}= ({\sum _ {i_{1}\cdots i_{d}}
    }
    |{\boldsymbol{\mathcal{X}}_{i_{1} \cdots i_d}}|)$,
 %$\|\boldsymbol{\mathcal{X}}\|_F=\sqrt{\sum_{i_1,i_2,\cdots,i_N}| \boldsymbol{\mathcal{X}} _{ i_1,i_2,\cdots,i_N} |^2}$
 ${\|{\boldsymbol{\mathcal{X}}}\|}_{\mathnormal{F}}= {(\sum_{i_{1} \cdots i_d} |{\boldsymbol{\mathcal{X}}_{i_{1} \cdots i_d}}|^{2})^\frac{1}{2}}$,
 %and
 $\| {\boldsymbol{\mathcal{X}}}\|_{\operatorname{tube}_1}  = % \sum_{{i_1}=1}^{n_1}  \sum_{{i_2}=1}^{n_2}
\sum_{i_1, i_2}
\|{\boldsymbol{\mathcal{X}}}_{i_1 i_2 : \cdots : }
\|_{ \mathnormal{F}  }$,
$\| {\boldsymbol{\mathcal{X}}}\|_{\operatorname{slice}_1}  = % \sum_{{i_1}=1}^{n_1}  \sum_{{i_2}=1}^{n_2}
\sum_{i_3,\cdots, i_d}
\|{\boldsymbol{\mathcal{X}}} {(:,:, i_3, \cdots , i_d )}
\|_{\mathnormal{F}}$,
%%%
% $\| {\boldsymbol{\mathcal{X}}}\|_{\mathnormal{F}, 1}  =  % \sum_{{i_1}=1}^{n_1}  \sum_{{i_2}=1}^{n_2}
%\sum_{i_1, i_2}
%\|{\boldsymbol{\mathcal{X}}}_{i_1 i_2 : \cdots : }
%\|_{ \mathnormal{F}  }$,
%%%%
respectively.

Let  ${\mathfrak{L}}(\boldsymbol{\mathcal{X}})$ or $\boldsymbol{\mathcal{X}}_{{\mathfrak{L}}}$ represent   the result of invertible linear transforms $\mathfrak{L}$ on $\boldsymbol{\mathcal{X}} \in \mathbb{R}^{n_1\times \cdots \times  n_d}$, i.e.,
 \begin{align}\label{trans}
 \mathfrak{L}(\boldsymbol{\mathcal{X}}) =\boldsymbol{\mathcal{X}} \;{\times}_{3} \;\bm{U}_{n_3} \;{\times}_{4} \;\bm{U}_{n_4} \cdots
    {\times}_{d} \;\bm{U}_{n_d},
\end{align}
 where
  $\boldsymbol{\mathcal{X}} \;{\times}_{j}\; \bm{U}_{n_j}$ denotes
    the mode-$j$ product of  $\boldsymbol{\mathcal{X}}$ with matrix ${\bm{U}}_{n_j}$,
 and
 the
 transform matrices $\bm{U}_{n_i} \in \mathbb{C}^{n_i \times n_i}$ of $\mathfrak{L}$ satisfies:
   \begin{align}\label{orth}
  {{\bm{U}}}_{n_i} \cdot {{\bm{U}}}^{\mit{H}}_{n_i}={{\bm{U}}}^{\mit{H}}_{n_i} \cdot {{\bm{U}}}_{n_i}=\alpha_i {{\bm{I}}}_{n_i},  \forall i\in \{3,\cdots,d\},
   \end{align}
   in which $\alpha_i>0$  is a  constant.
  The inverse operator of $\mathfrak{L}(\boldsymbol{\mathcal{X}})$  is defined as
   $ \mathfrak{L}^{-1} (\boldsymbol{\mathcal{X}}) =\boldsymbol{\mathcal{X}} \;{\times}_{d} \;\bm{U}^{-1}_{n_d} \;{\times}_{{d-1}} \;{{\bm{U}}_{n}}_{d-1}^{-1} \cdots
    {\times}_{3} \;\bm{U}^{-1}_{n_3}$,
    %
    %%%%%%
    and  $\mathfrak{L}^{-1} (\mathfrak{L}(\boldsymbol{\mathcal{X}}))=\boldsymbol{\mathcal{X}}$.

Based on the above concepts,
%we give the definition of high-order t-product.
%we give the definition of
we introduce related
tensor product, tensor decompositions and ranks  utilized %involved
in this paper.

\begin{Definition}\label{def2111}
\textbf{(T-Product \cite{qin2022low})}
Let
${\boldsymbol{\mathcal{X}}} \in \mathbb{R}^{n_1 \times n_2 \times n_3 \times \cdots \times n_d}$
and
${\boldsymbol{\mathcal{Y}}} \in \mathbb{R}^{n_2 \times l \times n_3 \times \cdots \times n_d}$.
Then the invertible linear transforms $\mathfrak{L}$ based t-product is defined as
\begin{align}\label{gene_tproduct}
\boldsymbol{\mathcal{C}}={\boldsymbol{\mathcal{X}}}{*}_{\mathfrak{L}} {\boldsymbol{\mathcal{Y}}}=\mathfrak{L}^{-1}
\big({\mathfrak{L}}(\boldsymbol{\mathcal{X}}) \bigtriangleup
{\mathfrak{L}}(\boldsymbol{\mathcal{Y}})\big),
\end{align}
where
$ \bigtriangleup  $  denotes the face-wise product
($\boldsymbol{\mathcal{P}}=\boldsymbol{\mathcal{M}} \bigtriangleup
     \boldsymbol{\mathcal{N}} $ implies %satisfies
     %$\Longleftrightarrow$
     %${\operatorname{bdiag}} ({{\boldsymbol{\mathcal{C}}}})={\operatorname{bdiag}} ({{\boldsymbol{\mathcal{A}}}})\cdot {\operatorname{bdiag}} ({{\boldsymbol{\mathcal{B}}}})$
    $ {\boldsymbol{\mathcal{P}}}^{<j>}={\boldsymbol{\mathcal{M}}}^{<j>}\cdot{\boldsymbol{\mathcal{N}}}^{<j>}, \;j=1,\cdots,n_3\cdots n_d$).
\end{Definition}

The relevant algebraic basis induced by high-order  t-product  can be found in %the %reference 1
 the literatures \cite{qin2022low,qin2023nonconvex,wang2023guaranteed}, such as
%identity tensor,   orthogonal tensor, f-diagonal tensor
 %Gaussian random tensor
 identity, orthogonality, inverse, transpose, T-SVD factorization, T-QR factorization,   T-SVD rank.

\begin{Definition} \label{theorem11111}
\textbf{(T-SVD  factorization and rank \cite{qin2022low})}
Let $ {\boldsymbol{\mathcal{X}}} \in \mathbb{R}^{n_1  \times \cdots  \times n_d}$,
then it can be factorized as
\begin{equation}
{\boldsymbol{\mathcal{X}}}={\boldsymbol{\mathcal{U}}} {*}_{\mathfrak{L}}   {\boldsymbol{\mathcal{S}}} {*}_{\mathfrak{L}}
{\boldsymbol{\mathcal{V}}}^{\mit{T}} ,
\end{equation}
where ${\boldsymbol{\mathcal{U}}} \in \mathbb{R}^{n_1 \times n_1 \times n_3 \times \cdots \times n_d}$ and
${\boldsymbol{\mathcal{V}}} \in \mathbb{R}^{n_2 \times n_2 \times n_3 \times \cdots \times n_d}$ are  orthogonal, %unitary, %\cite{\cite{\cite{}}} tensors,
 ${\boldsymbol{\mathcal{S}}} \in \mathbb{R}^{n_1 \times n_2 \times n_3 \times \cdots \times n_d}$ is a f-diagonal tensor.
Then, the  T-SVD rank of ${\boldsymbol{\mathcal{X}}}$ is defined as
\begin{align*}
%\textcolor[rgb]{0.00,0.00,1.00}{
 {\operatorname{rank}}_{tsvd}({\boldsymbol{\mathcal{X}}})
&= \sum_{i} {\mathbbm{1}
\big[
{\boldsymbol{\mathcal{S}}}(i,i,:,\cdots,:) \neq \boldsymbol{0}
\big]},
\end{align*}
\textcolor[rgb]{0.00,0.00,0.00}{where %$\mathbbm{1}$ %
%${\sharp}$
$\mathbbm{1}[\cdot]$
denotes the %cardinality of a set.
indicator function.}
\end{Definition}

\begin{Definition} \label{tucker212}
\textbf{(Tucker factorization and rank \cite{kolda2009tensor})}
%For any order-$d$ tensor %
Let
$ {\boldsymbol{\mathcal{X}}} \in \mathbb{R}^{n_1  \times \cdots  \times n_d}$,
then
its Tucker decomposition form is % can be factorized as
\begin{equation}
 \boldsymbol{\mathcal{X}}
 \triangleq
% = %{\bm{\mathcal{A}}}=
 [\bm{\mathcal{C}};  \bm{F}_{1}, \bm{F}_{2},\cdots, \bm{F}_{d}]
 =\boldsymbol{\mathcal{C}} {\times}_{1} \bm{F}_{1}  {\times}_{2} \bm{F}_{2}   \cdots
 %Tucker decomposition, i.e.,
%
 {\times}_{d} \bm{F}_{d},
 \end{equation}
  where $\{\bm{F}_{i}\}_{i=1}^{d} \in \mathbb{R}^{n_i  \times r_i}$
% $ \bm{F}_{i} \in \mathbb{R}^{n_i  \times r_i}$ is the orthogonal %Tucker
% factor matrix along each  mode.
are the factor matrices and can be thought of as the principal components  along different   mode.
  %%%
  $\boldsymbol{\mathcal{C}} \in \mathbb{R}^{r_1  \times \cdots  \times r_d}$ is  the %Tucker
  core tensor and
  %it reflects the interaction between components along %the
 %  different modes,
 its entries show %represent
  the level of interaction between the
different components. % $\{\bm{F}_{i}\}_{i=1}^{d} $.
 The vector $\bm{r}=
 (r_1,  \cdots, r_d)$ is defined as the  Tucker-rank of %the original tensor
 $\bm{\mathcal{X}}$, where $\bm{r}_j=
 {\rm{rank}} (\textbf{X}_{(j)})$, $j=1,\cdots,d$.
 %This decomposition can be abbreviated as  $ $.
\end{Definition}

\begin{proposition} \label{one-bitequ}   \cite{chen2023high}
Let $x$ and $\xi$ be two independent random variables, where $|x| < \beta$ holds almost surely and $\xi$
is uniformly distributed on the interval $[-\theta, \theta]$ with $\gamma   \geq \beta$. Then we have
$$
\mathbb{E} (x) =\mathbb{E} \big(  \theta \cdot \operatorname{sign} (x+\xi ) \big).
$$
\end{proposition}

\vspace{-0.53cm}
%Randomized Techniques Based High-Order Tensor Approximation
\section{\textbf{Fast  Randomized %
 High-Order %Block Lanczos Bidiagonalization Method Based Fixed-Accuracy
Tensor Approximation}}\label{LRTAapproximation}

In this section, fast  and accurate randomized  algorithms are devised
for solving fixed-rank and   fixed-precision LRTA problems, respectively.
%we primarily propose  for solving LRTA problems presented as fixed-precision and fixed-rank issues, respectively.
%%%
Furthermore, we provide the  error bound analysis for these two randomized %LRTA %
algorithms.
 \textbf{\textit{Due to space limitations, the
 complexity analysis and
 %  relevant
  experimental results are included in the supplementary materials.}}

\begin{algorithm}[!htbp]
\setstretch{0.2}
     \caption{
   STHOSVD  Using   Randomized Block Krylov Iteration Method, \textbf{R-STHOSVD-BKI}. %STHOSVD-RBKI
   %\cite{meyer2024unreasonable},
  % $\operatorname{{\textit{RBKI}}} (\bm{{A}},b,  \epsilon, \delta)$. %: Spatial Domain Version..
     }
     \label{STHOSVD-BKI}
      \KwIn{$\bm{\mathcal{X}} %\bm{{X}}
      \in\mathbb{R}^{n_1\times   \cdots \times n_d}$, target rank:
       $\bm{r}=(r_1, \cdots, r_d)$,  processing order:  $\bm{\rho}$, % :\{i_1, i_2, \cdots, i_d\}$.
       block size: $\bm{b}=(b_1,  \cdots, b_d)$,
      Krylov iterations: $\bm{q}=(q_1,  \cdots, q_d)$, $1\leq b_i \leq r_i$, $(q_i+1) \cdot b_i \geq r_i$, $i=1,2, \cdots,d$.
       }
       %%%%%%%%%
      {\color{black}\KwOut{
      %
%$\bm{{Q}}   \in\mathbb{R}^{m \times (q+1) \cdot b}$
%with orthogonal columns.
$  \hat{\bm{\mathcal{X}}}= [\bm{\mathcal{C}};  \bm{F}_{1}, \bm{F}_{2},\cdots, \bm{F}_{d}]
 $.
      }}

Set $\bm{\mathcal{C}}  \leftarrow  \bm{\mathcal{X}} $\;

      \For{$v=1,2,\cdots, \operatorname{\textit{length}} (\bm{\rho})$}
      {

      % $\bm{F}_{\bm{\rho} _{v}} = \operatorname{{\textit{randUBV}}} (\bm{C}_{(\bm{\rho} _{v})},b,  \epsilon, \delta)$\;
Set   $ \bm{A} \leftarrow \bm{C}_{
(\bm{\rho} _{v})
%({v})
}$,
 $n=\operatorname{\textit{size}} (\bm{A} ,2)$\;

  Draw      a Gaussian random matrix  ${\boldsymbol{{G}}}    \in\mathbb{R}^{n \times \bm{b}_{\bm{\rho} _{v}} }$\;

Build the Krylov subspace:

              ${\boldsymbol{{K}}}: = [{\boldsymbol{{A}}}  {\boldsymbol{{G}}}, ({\boldsymbol{{A}}} {\boldsymbol{{A}}}^{\mit{T}})
              {\boldsymbol{{A}}}  {\boldsymbol{{G}}}, \cdots,
              ({\boldsymbol{{A}}} {\boldsymbol{{A}}}^{\mit{T}})^{\bm{q}_{\bm{\rho} _{v}} }
              {\boldsymbol{{A}}}  {\boldsymbol{{G}}}
              ]
              $\;

     Form an orthonormal   basis ${\boldsymbol{{Z}}}$  for $ {\boldsymbol{{K}}}$\;

     Compute   ${\boldsymbol{{M}}}: = {\boldsymbol{{Z}}}^{\mit{T}}  {\boldsymbol{{A}}}  {\boldsymbol{{A}}}^{\mit{T}}
      {\boldsymbol{{Z}}}$\;

      Set $\boldsymbol{U}_{r}$  to  the $r$ top singular vectors of ${\boldsymbol{{M}}}$\;
     %

%Return $\boldsymbol{Q}=  \boldsymbol{Z} \boldsymbol{U}_{r} $\;

Update
  $\bm{F}_{\bm{\rho} _{v}}   \leftarrow    \boldsymbol{Z} \boldsymbol{U}_{r} $,
  $\bm{C}_{(\bm{\rho} _{v})}   \leftarrow  (\bm{F}_{\bm{\rho} _{v}}) ^{\mit{T}}  \bm{C}_{(\bm{\rho} _{v})}$\;
  $  \bm{\mathcal{C}}^{(v)} \leftarrow \bm{C}_{(\bm{\rho} _{v})} $, in tensor format.
}
    \end{algorithm}

\begin{algorithm}[!htbp]
\setstretch{0.2}
     \caption{
   % Block Lanczos bidiagonalization Process.
   Deflated QR, $\operatorname{\textbf{\textit{defl-QR}}} (\bm{{A}}, \delta)$.
     }
     \label{deflated}
      \KwIn{$\bm{{A}}\in\mathbb{R}^{m \times n}$,
     % and an orthogonal tensor
    deflation tolerance:  $\delta $.
       }

%$\bm{\mathcal{U}} ^{(0)} =\textbf{0}; \bm{\mathcal{L}} ^{(1)} =\textbf{0} $\;
Compute the pivoted  QR factorization:  $ {\boldsymbol{{A}}}
 {\boldsymbol{{P}}}   =\tilde {\boldsymbol{{Q}}}   \tilde {\boldsymbol{{R}}}   $\;

Find the largest $s$ such that $|\tilde {\boldsymbol{{R}}} (s,s) | \geq \delta$\;

$ {\boldsymbol{{R}}}=\tilde {\boldsymbol{{R}}} (1:s, :)
 {{\boldsymbol{{P}}}}^{\mit{T}}$\;

 $ {\boldsymbol{{Q}}}=\tilde {\boldsymbol{{Q}}} (:, 1:s) $\;

 %\For{$k=1,2,3, \cdots$}
%      {
%     $  [{\boldsymbol{\mathcal{U}}}^{(m)}, {\boldsymbol{\mathcal{R}}}^{(m)}]=\operatorname{\textit{H-TQR}}( {\boldsymbol{\mathcal{A}}}  {*}_{\mathfrak{L}} {\boldsymbol{\mathcal{V}}}^{(m)}
%     -{\boldsymbol{\mathcal{U}}}^{(m-1)} {*}_{\mathfrak{L}} {\boldsymbol{\mathcal{L}}}^{(m)}
%     , \mathfrak{L})$;\\
%$
% [{\boldsymbol{\mathcal{V}}}^{(m+1)}, ( {\boldsymbol{\mathcal{L}}}^{(m+1)}) ^{\mit{T}}]=\operatorname{\textit{H-TQR}}(
%     {{\boldsymbol{\mathcal{A}}}}^{\mit{T}}   {*}_{\mathfrak{L}}   {\boldsymbol{\mathcal{U}}}^{(m)}-
%{\boldsymbol{\mathcal{V}}}^{(m)} {*}_{\mathfrak{L}} ( {\boldsymbol{\mathcal{R}}}^{(m)}) ^{\mit{T}}
%, \mathfrak{L})
%$.\\
%      }

 {\color{black}\KwOut{\textcolor[rgb]{0.00,0.00,0.00}{
%$\bm{\mathcal{Q}}  \in\mathbb{R}^{n_1\times (q+1)b \times  n_3 \times
%\cdots\times n_d} $,
%$\bm{\mathcal{B}} \in\mathbb{R}^{ (q+1)b \times n_2\times n_3 \times
% \cdots\times n_d} $.
$\bm{{Q}}$,
$\bm{{R}}$, $s$.
}
      }}

    \end{algorithm}

\vspace{-0.5253cm}
\subsection{\textbf{Proposed Randomized LRTA   Algorithms}}

%As we known,  %both THOSVD and STHOSVD
The \textit{truncated higher-order SVD} (THOSVD) \cite{kolda2009tensor} and the \textit{sequentially truncated higher-order SVD} (STHOSVD) \cite{vannieuwenhoven2012new}
algorithms rely directly on SVD when computing the singular vectors of intermediate matrices.
Consequently, they demand significant
%requiring large
memory and high computational complexity when the size of tensors is large.
To alleviate this issue,  Minster et al. \cite{minster2020randomized}
%applied the \textit{randomized SVD} (R-SVD) to the THOSVD and STHOSVD algorithms,
%and then
 designed their corresponding accelerated %randomized
 versions (i.e., R-THOSVD and R-STHOSVD) via the  \textit{randomized SVD} (R-SVD).
Nevertheless,
when the singular spectrum of unfolding matrices of original tensor decays slowly,
basic R-SVD may produce a poor basis in many applications.
Although the R-SVD algorithm using  the power iteration strategy  can  improve the singular value decay rate
\cite{che2020computation33,che2021randomized,che2021efficient, che2023randomized,dong2023practical},
it only uses the result of the last iteration rather than
%and does not use
the results of each iteration.
 So it can not always approach the largest $k$ singular values well in general.
 %%%%%%%%%%%%%%%%%%%%%%%%%%%%%%%%%%%%%%%%%%%%%%%%%%%%%%%%%%%%%%%%%%%%%%%%%%%%%%%%%%%%%
 Motivated by the recently achieved %proposed   %studies
 research findings
  \cite{musco2015randomized, yuan2018superlinear,meyer2024unreasonable},
 this subsection first  considers integrating  the \textit{randomized  block
Krylov iteration} (RBKI) method into STHOSVD algorithm to
accelerate the computation speed.
Please see
Algorithm  \ref{STHOSVD-BKI} for more details.
Note that although STHOSVD has the same worst case error-bound as THOSVD, it is less computationally
complex and requires less storage \cite{vannieuwenhoven2012new, minster2020randomized %, sun2020low,dong2023practical
}.
Therefore, in this paper, only the randomized versions of the STHOSVD algorithm is taken into account.

\begin{algorithm}[!htbp]
\setstretch{0.1}
     \caption{
Adaptive Randomized STHOSVD
%R-STHOSVD
%Randomized T-SVD
Using
%Blocked Bidiagonalization Method.
Block Lanczos Bidiagonalization Process Method, \textbf{AD-RSTHOSVD-BLBP}.  %AD-RSTHOSVD-BLBP  R-STHOSVD-BKI and AD-RSTHOSVD-BLBP
     }
     \label{bb-rsthosvd}
      \KwIn{$\bm{\mathcal{X}}\in\mathbb{R}^{n_1\times \cdots\times n_d}$,
     % transform:   $\mathfrak{L}$,
       relative error:  $\epsilon$,   block size: $b$,   deflation tolerance: $\delta$,
       processing order: $\bm{\rho}$.
       }
       %%%%%%%%%
      {\color{black}\KwOut{ \textcolor[rgb]{0.00,0.00,0.00}{
$  \hat{\bm{\mathcal{X}}}= [\bm{\mathcal{C}};  \bm{F}_{1}, \bm{F}_{2},\cdots, \bm{F}_{d}]
 $.}
      }}

Set $\bm{\mathcal{C}}  \leftarrow  \bm{\mathcal{X}} $\;

      \For{$v=1,2,\cdots, \operatorname{\textit{length}} (\bm{\rho})$}
      {

      % $\bm{F}_{\bm{\rho} _{v}} = \operatorname{{\textit{randUBV}}} (\bm{C}_{(\bm{\rho} _{v})},b,  \epsilon, \delta)$\;
Set   $ \bm{A} \leftarrow \bm{C}_{(\bm{\rho} _{v})}$,  $n=\operatorname{\textit{size}} (\bm{A} ,2)$\;

     Generate  a Gaussian random matrix $\bm{{G}}\in\mathbb{R}^{n \times  b }$\;

       % Compute the results of $\mathfrak{L}$ on $\boldsymbol{\mathcal{A} }$ and $\boldsymbol{\mathcal{G} }$, i.e.,
%       $\mathfrak{L} (\boldsymbol{\mathcal{A}}), \mathfrak{L} (\boldsymbol{\mathcal{G}})$\;

     % \For{$v=1,2,\cdots, n_3\cdots n_d$}
%      {

  ${\boldsymbol{{U}} } ^{\{v\}}=[\;\;];\;
   {\boldsymbol{{B}}}^{\{v\}}=[\;\;];\;
   {\boldsymbol{{V} }}^{\{v\}}=[\;\;]$\;

$
E=
   \| {{\boldsymbol{{A}}}} \|_{\mathnormal{F}} ^{2}$
   \;

     $ %[ \sim,
      {  {\boldsymbol{{V}}}
     } ^{(1)}
     =\operatorname{\textit{qr}}(
     {\boldsymbol{{G}}}, 0)
     $\;

     $ {  {\boldsymbol{{U}}}
     }^{(1)} = \textbf{0} ; \;\;
      {  {\boldsymbol{{L}}}
     } ^{(1)}= \textbf{0} $\;

      $ { {  {\boldsymbol{{V}}}
     }} ^{\{v\}}  = {  {\boldsymbol{{V}}^{(1)}}
     }  ;  \;\;
   {{  {\boldsymbol{{U}}}
     } } ^{\{v\}}=  {  {\boldsymbol{{U}}^{(1)}}
     } $\;

\For{$i=1,2,3, \cdots$}
              {
           $ [ {\boldsymbol{{U}}}^{(i)},
           {\boldsymbol{{R}}}^{(i)}] =
           \operatorname{\textbf{\textit{defl-QR}}}
           \big (    {\boldsymbol{{A}}}
            {\boldsymbol{{V}}}^{(i)}  -
            {\boldsymbol{{U}}}^{(i-1)}
           {\boldsymbol{{L}}}^{(i)} , \delta
         \big  )
            % \mathfrak{L}(\boldsymbol{\mathcal{G}})^{<v>} \big(:,(i-1)b+1:ib \big)
             $\; %{*}_{\mathfrak{L}}

             $\boldsymbol{{U}} ^{\{v\}}=[ \boldsymbol{{U}}^{\{v\}}, {{\boldsymbol{{U}}}^{(i)}}]$\;

             $ %\operatorname{RelErr}=  \operatorname{RelErr}-
             E=E-
   \| { {\boldsymbol{{R}}}^{(i)}} \|_{\mathnormal{F}} ^{2}$
   \;

%$
%\mathfrak{L}\big({\boldsymbol{\mathcal{Y}}}^{(i)}\big)^{<v>}
%=
%-
%%
%\mathfrak{L}(\boldsymbol{\mathcal{Q}})^{<v>} \cdot \mathfrak{L}(\boldsymbol{\mathcal{B}})^{<v>} \cdot
%\mathfrak{L}({\boldsymbol{\mathcal{G}}}^{(i)})^{<v>}
%+
%%
%\mathfrak{L}(\boldsymbol{\mathcal{W}})^{<v>} \big(:,(i-1)b+1:ib \big)
%$
%\;

%$[ {\mathfrak{L}({\boldsymbol{\mathcal{Q}}}^{(i)})}^{<v>},\mathfrak{L}({\boldsymbol{\mathcal{R}}}^{(i)})^{<v>}]
%=\operatorname{qr}\big(\mathfrak{L}({\boldsymbol{\mathcal{Y}}}^{(i)})^{<v>}\big)$\;

$
{ {\boldsymbol{{V}}} }^{(i+1)}=
{{\boldsymbol{{A}}}}  ^{\mit{T}}
{{\boldsymbol{{U}}}^{(i)}}-{ {\boldsymbol{{V}}}}^{(i)}
{ {\boldsymbol{{R}}}^{(i)}}^{\mit{T}}
$\;

$
{{\boldsymbol{{V}}}^{(i+1)}} ={{\boldsymbol{{V}}}^{(i+1)}} -
 {{\boldsymbol{{V}}}} ^{\{v\}}
{ {\boldsymbol{{V}}} ^{\{v\}} } ^{\mit{T}}
{{\boldsymbol{{V}}}^{(i+1)}}
$\;

$[ {{\boldsymbol{{V}}}^{(i+1)}}, {({\boldsymbol{{L}}}^{(i+1)}})^{\mit{T}},s]
= \operatorname{\textbf{\textit{defl-QR}}} \big(
{{\boldsymbol{{V}}}^{(i+1)}}, \delta \big )
$\;

$
 {{\boldsymbol{{V}}}
} ^{\{v\}}=[  {{\boldsymbol{{V}}}
} ^{\{v\}},
{ {\boldsymbol{{V}}}^{(i+1)}}
]
$\;

\If{
$s<b$  %\text{utilize the unblocked randomized technique}
}
{
Draw  a  random standard
Gaussian matrix ${\boldsymbol{{G}}} ^{(i) } \in\mathbb{R}^{n \times (b-s)} $ \;

$
{ {{\boldsymbol{{V}}}^{'}}  ^{(i+1)}} = \operatorname{\textit{qr}}
( {\boldsymbol{{G}}} ^{(i) }    -  {{\boldsymbol{{V}}}  ^{\{v\}}
}
{{ {{\boldsymbol{{V}}}} } ^{\{v\}}
}^
{\mit{T}}
  {\boldsymbol{{G}}} ^{(i) }
  )
$\;

$
 {{\boldsymbol{{V}}}
}^{\{v\}}=[  {{\boldsymbol{{V}}}
} ^{\{v\}},
{{{\boldsymbol{{V}}}^{'}}  ^{(i+1)}}
]
$\;
}
%%% Update  the block bidiagonal matrix ${\bm{B}}^{\{v\}} $  by (\ref{blockB})\;

             $ %\operatorname{RelErr}=  \operatorname{RelErr}-
             E=E-
   \| {  {\boldsymbol{{L}}}^{(i+1)}}\|_{\mathnormal{F}} ^{2}$
   \;

\If{
$ %\operatorname{RelErr}
E <    \epsilon ^2  %\tau ^2
{\|  {\boldsymbol{{A}}}  \|}_{\mathnormal{F}}^{2}$  %\text{utilize the unblocked randomized technique}
}
{
break;
}
% }

 %$[\mathfrak{L}(\boldsymbol{\mathcal{Q}}_{1})^{<v>},\mathfrak{L}(\boldsymbol{\mathcal{R}}_{1})^{<v>}]= \textrm{qr} \big(
% (\mathfrak{L}(\boldsymbol{\mathcal{B}})^{<v>})^{\mit{T}}
% \big) $;\\
%
%$[\mathfrak{L}(\hat{\boldsymbol{\mathcal{U}}})^{<v>},\mathfrak{L}(\boldsymbol{\mathcal{S}})^{<v>},\mathfrak{L}(\hat{\boldsymbol{\mathcal{V}}})^{<v>}]= \textrm{svd} \big(\mathfrak{L}(\boldsymbol{\mathcal{R}}_{1})^{<v>}\big) $;\\
%
%$
%%\hat{\bm{S}}
%\mathfrak{L}(\boldsymbol{\mathcal{V}})^{<v>}
%= \mathfrak{L}(\boldsymbol{\mathcal{Q}}_{1})^{<v>}\cdot \mathfrak{L}(\hat{\boldsymbol{\mathcal{U}}})^{<v>}
%$;\\
%
%$
%%\hat{\bm{S}}
%\mathfrak{L}(\boldsymbol{\mathcal{U}})^{<v>}
%= \mathfrak{L}(\boldsymbol{\mathcal{Q}})^{<v>}\cdot \mathfrak{L}(\hat{\boldsymbol{\mathcal{V}}})^{<v>}
%$;\\
%
%
%

}
      %  \vspace{-0.0cm}

  Update
  $\bm{F}_{\bm{\rho} _{v}}   \leftarrow   \bm{U} ^{\{v\}} $,
  $\bm{C}_{(\bm{\rho} _{v})}   \leftarrow  (\bm{F}_{\bm{\rho} _{v}}) ^{\mit{T}}  \bm{C}_{(\bm{\rho} _{v})}$\;
  $  \bm{\mathcal{C}} ^{(v)}   \leftarrow \bm{C}_{(\bm{\rho} _{v})} $, in tensor format.
}
        \vspace{-0.0mm}
    \end{algorithm}
%%%%%%%%%%%%%%%%%%%%%%%%%%%%%%%%%%%%%%%%%%%%%%%%

%\vspace{-11cm}
For the   proposed fast   STHOSVD   %randomized STHOSVD
algorithm using  RBKI approach,
 % block Krylov iteration method ,
 %
%In the algorithms described in the previous section,
we had to assume prior knowledge of the target rank $(r_1, r_2, \cdots, r_d)$.
% This knowledge may not be available, or may be difficult to estimate, in practice.
%%%%
%%
 Nevertheless,
    %However,
    in practical applications, an estimation of the target %T-SVD
    rank   may not be a tractable task.
    %%%%%%%%%%%%%%%%%%%%%%%%%%%%%%%%%%%%%%%%%%%%%%%%%%%%%%%%%%%%%%%%%%%%%%%%%%%%%%%%%%%%%%%%%%%%%%%%
    %
  %   by devising novel full-mode (or wise-mode) random projection techniques.
%%%%%%%%%%%%%%%%%%%%%%%%%%%%%%%%%%%%%%%%%%%%%%%%%%%%%%%%%%%%%%%%%%%%%%%%%%%%%%%%%%%%%%%%%%%%%%%%%%%%%%%%%%%%%%%%%%%%%
In response to the above issues, in virtue of randomized projection technique and
block Lanczos bidiagonalization process method,
%Thus, in this subsection, we consider %providing %a practical implementation of Algorithm \ref{STHOSVD-BKI} for the fixed-accuracy   LRTA problem.
this subsection sequentially considers
investigating %exploring
an  adaptive  R-STHOSVD algorithm  %randomized STHOSVD  randomized STHOSVD
with prescribed accuracy (i.e., the fixed-accuracy   LRTA %problem
in the Tucker format).
 Wherein, %%%    where
     the desired truncation  rank $(r_1, r_2, \cdots, r_d)$ is not known in advance,
   but we instead want to find %the smallest possible $r$
   a good approximation
  $ \tilde{{{\bm{\mathcal{X}}} } }$
   such that
   $ \|% \hat{\bm{\mathcal{A}}}
{{\bm{\mathcal{X}}} } -
\tilde{{{\bm{\mathcal{X}}} } }
%{\boldsymbol{\mathcal{U}}}{*}_{\mathfrak{L}}{\boldsymbol{\mathcal{B}}}{*}_{\mathfrak{L}}{\boldsymbol{\mathcal{V}}}^{\mit{T}}
  % \in\mathbb{R}^{n_1\times \cdots\times n_d}
 \|_{\mathnormal{F}}  \leq \epsilon  $
   for some tolerance $\epsilon  $.
   %The goal of this paper
   %%%%%%%%%%%%%%%%%%%%%%%%%%%%%%%%%%%%%%%%%%
%
 Extensions of Algorithm \ref{STHOSVD-BKI}
 %(see Algorithm \ref{STHOSVD-BKI})
   to the fixed-accuracy LRTA  problem make use of the fact
   that %tensors $\bm{\mathcal{Q}}$ and $\bm{\mathcal{B}}$
   the Tucker factor matrices
   can be computed incrementally rather than all at once.
   The process can then be terminated once a user-specified error threshold has been reached,
   assuming the error can be efficiently computed or estimated.
   Please see
Algorithm  \ref{bb-rsthosvd} for details.   % Algorithm \ref{STHOSVD-BKI}   Algorithm  \ref{bb-rsthosvd}
%%后面再解释一下算法，
%%%  或者再结合矩阵文献，说一下实现这个算法的几个关键因素

%\vspace{-0.53cm}
%\vspace{-0.64cm}
%Preliminary Knowledge

\vspace{-0.38cm}
\subsection{\textbf{Theoretical Error Bound Analysis}}
% \vspace{-0.1cm}
%
In this subsection,
%we provide the %convergence analysis of the proposed algorithm.
%theoretical error-bound analysis of  Algorithm \ref{STHOSVD-BKI}  and  Algorithm  \ref{bb-rsthosvd}.
%
%%  Theorem  \ref{rbki-tuck} and  Theorem \ref{hallman2022block144444}
we  provide the theoretical error bounds  of
 fixed-rank  LRTA algorithm using small-block   Krylov iteration method (\textit{see Algorithm \ref{STHOSVD-BKI}})  and
 fixed-accuracy LRTA algorithm using block Lanczos bidiagonalization process
 (\textit{see Algorithm  \ref{bb-rsthosvd}})
 under  the Frobenius norm, respectively.
%%
%The main results are given in  Theorem  \ref{rbki-tuck}   and Theorem \ref{hallman2022block144444}
  %Theorems  \ref{rbki-tuck}-\ref{hallman2022block144444}
% below.
%Before that, we present some lemmas.
The detailed proofs of our main  theories (i.e., Theorems \ref{rbki-tuck}-\ref{hallman2022block144444})
can be found in the Supplementary Material.
%

%%%
\begin{Definition}\label{def11}
 %%%%
 Fix block size $b \in [k]$. For each $i \in [k]$, we let ${\mathcal{N}}_{i}  \subset [k]   \backslash \{i\} $
 be the indices of the $b-1$ singular values other than $i$ that minimize $|\frac{\sigma_i({\bm{A}})  -\sigma_j ({\bm{A}})}{\sigma_j({\bm{A}})}|$.
 Then let $g_{\min, b }$ be the $b^{th}$-order gap of ${\bm{A}}$:
 %%%
    \begin{align*}
g_{\min,b}({\bm{A}}) =\min_{i \in [k]}  \min_{j \in [k]   \backslash {\mathcal{N}}_{i}, j \neq i    }
\Big | \frac{\sigma_i({\bm{A}})  -\sigma_j({\bm{A}}) }{\sigma_j({\bm{A}})} \Big|.
\end{align*}
   \end{Definition}
%%%%%%

   \begin{Theorem}\label{rbki-tuck}
   For any %an
   order-$d$  tensor   ${\boldsymbol{\mathcal{X}}} \in \mathbb{R}^{{n_1 \times n_2 \times  \cdots \times n_d}}$ and
   $ \bm{b}_{i}  \leq \bm{r}_{i}$, $i=1,2, \cdots,d$,
   let $g_{\min,\bm{b}_i} ({\bm{X}}_{(i)})$ as in Definition \ref{def11}.
   For any $\varepsilon, \delta \in (0,1)$,
   $q_i =
   \boldsymbol{\mathcal{O}}\big(
   \frac{r_i-b_i}{\sqrt{\varepsilon} }  \log (\frac{1}{g_{\min,b_i}(    {\bm{X}}_{(i)})}  ) + \frac{1} {\sqrt{\varepsilon} }
   \log (\frac{n_i}{ \delta\varepsilon})
   \big)$,
    Algorithm  \ref{STHOSVD-BKI} %%Algorithm  \textbf{R-STHOSVD-BKI}
     with processing order
  $\bm{\rho}:=\{1,2, \cdots, d
  \}$ returns
  the low multilinear rank-$(r_1, r_2, \cdots, r_d)$ approximation of ${\boldsymbol{\mathcal{X}}}$:
 %  Let
    $  \hat{\bm{\mathcal{X}}} %:= [\bm{\mathcal{C}};  \bm{F}_{1}, \bm{F}_{2},\cdots, \bm{F}_{d}]
 =  \boldsymbol{\mathcal{C}} {\times}_{1} \bm{F}_{1}  {\times}_{2} \bm{F}_{2}   \cdots
 {\times}_{d} \bm{F}_{d}$
 % be   by Algorithm  \textbf{R-STHOSVD-BKI} with processing order $\bm{\rho}:=[1,2, \cdots, d]$.   run
  %  for   iterations returns an orthogonal  ${\bm{Q}}  \in \mathbb{R}^{{n \times k}}$
    such that, with probability
   at least $1- \delta$,
 %%%%%%%%%%%%%%%%%%%%%%%%%%%%%%%%%%%%%%%%%%%%%%%%%%%%%%%%%%
 % \cite{vannieuwenhoven2012new} %\cite{qin2022low}
% Then,
  %%%%%%%%%%%%%%%%%%%%
    \begin{equation}   \label{equation11}
%\frac
{\big\| {{\boldsymbol{\mathcal{X}}}}    - \hat{\bm{\mathcal{X}}} \big \|_{\mathnormal{F}} } /  {\sqrt{d}}
\leq
(1+\varepsilon)  \| {{\boldsymbol{\mathcal{X}}}}-
 { \hat{\boldsymbol{\mathcal{X}}}  }_{\operatorname{opt}}
 \|_{\mathnormal{F}},
\end{equation}
where ${ \hat{\boldsymbol{\mathcal{X}}}  }_{\operatorname{opt}}$ %$ \hat{\bm{\mathcal{X}}} ^{(j)}$ denotes  the  $j$-th partial  approximation, defined as
%$ \hat{\bm{\mathcal{X}}} ^{(j)}   =   \bm{\mathcal{C}}^{(j)}    {\times}_{i=1}^{j} \bm{F}_{i} $.
denotes the optimal rank-$\bm{r}$ approximation.
\end{Theorem}

\begin{Definition}
%\textcolor[rgb]{0.00,0.00,0.00}{ }
Suppose that the number of internal iterations of Algorithm  \ref{bb-rsthosvd} %%\textbf{AD-STHOSVD-BLBP} Algorithm
is $K$.
Then, the local loss of orthogonality of factor matrix ${\boldsymbol{{F}}} _{v } := \bm{U} ^{\{v\}}= [\bm{U} ^{(1)} , \cdots,  \bm{U} ^{(K)} ] ,
v=1,2, \cdots, d$  is defined as
\begin{align*}
\varepsilon_{v} = \max \Big \{  \max_{1\leq i \leq K} \big\| ({ \bm{U} ^{(i)}}) ^ {\mit{T}} \cdot {\bm{U} ^{(i)}}
- \bm{I}
\big\|_{2},
\\
 \max_{2 \leq i \leq K} \big\| ({ \bm{U} ^{(i-1)}}) ^ {\mit{T}} \cdot  {\bm{U} ^{(i)}}
\big\|_{2}
  \Big \}.
\end{align*}
The main idea is that we do not require
$ \big\| ({ \bm{U} ^{(i)}} )^ {\mit{T}} \cdot  {\bm{U} ^{(i)}}
- \bm{I}
\big\|_{2}$ to be small. Instead, we need only the milder condition that adjacent blocks be close to orthogonal.
\end{Definition}

\begin{Theorem} %\cite{hallman2022block1}
\label{hallman2022block144444} %\label{rbki-tuck}
Let
$  \hat{\bm{\mathcal{X}}} %:= [\bm{\mathcal{C}};  \bm{F}_{1}, \bm{F}_{2},\cdots, \bm{F}_{d}]
 =  \boldsymbol{\mathcal{C}} {\times}_{1} \bm{F}_{1}  {\times}_{2} \bm{F}_{2}   \cdots
 {\times}_{d} \bm{F}_{d}$,
${\boldsymbol{{U}}} ^{\{j\} },   {\boldsymbol{{B}}} ^{\{j\} }$, and $ {\boldsymbol{{V}}} ^{\{j\} } $
be as produced by %%%\textbf{AD-STHOSVD-BLBP} Algorithm
Algorithm  \ref{bb-rsthosvd}
 with  input
${\boldsymbol{\mathcal{X}}} \in \mathbb{R}^{{n_1 \times  \cdots \times n_d}}$,
deflation tolerance $\delta$
and processing order  $\bm{\rho} = [1,2,\cdots,d] $.
Let
$E_{j}= \big\|  {\boldsymbol{{C}}} _{(j)}^ {(j-1) }
\big
\|_{\mathnormal{F}}^{2}-
\big\|  {\boldsymbol{{B}}} ^{\{j\} }
\big
\|_{\mathnormal{F}}^{2}$,
$\varepsilon_{j}$ denote the local loss of orthogonality of  $ {\boldsymbol{{U}}} ^{\{j\} }$, and
$\varpi_{j}$ be the number of columns removed from ${\boldsymbol{{U}}} ^{\{j\} }$ due to deflation operation.
%Let
Assume that $ {\boldsymbol{{V}}} ^{\{j\} } $ has orthonormal columns. Then
%$$
%\big\|  {\boldsymbol{{C}}} _{(v) }  ^ {(v-1) } -
%  {\boldsymbol{{U}}} ^{\{v\} } {\boldsymbol{{B}}} ^{\{v\} }  ({\boldsymbol{{V}}} ^{\{v\} })^ {\mit{T}}
%\big
%\|_{\mathnormal{F}}^{2}
%=E_{v}
%+
%4 \varepsilon_{v}\big\|  {\boldsymbol{{X}}} _{(v) }
%\big
%\|_{\mathnormal{F}}^{2}
%+
%2 \delta \sqrt{\varpi_{v}} (1+2 \varepsilon_{v})
%\big\|  {\boldsymbol{{X}}} _{(v) }
%\big
%\|_{\mathnormal{F}}.
%$$
\begin{align}   %\label{equation11}
%\big\| {{\boldsymbol{\mathcal{X}}}}    - \hat{\bm{\mathcal{X}}} \big \|_{\mathnormal{F}}^{2}
 \Big\|
{{\boldsymbol{\mathcal{X}}}}    - \hat{\bm{\mathcal{X}}}
\Big\| ^2_{\mathnormal{F}}
& \leq
\sum_{j=1}^{d}  \Big \{
E_{j}
+
(1+ 4 \varepsilon_{j}-\varepsilon_{j}^{2})   \big\|  {\boldsymbol{{X}}} _{(j) }
\big
\|_{\mathnormal{F}}^{2}
 \notag \\ \label{part15}
&
+
 2 \delta \sqrt{\varpi_{j}} (1+3 \varepsilon_{j}  +2 \varepsilon_{j} ^{2} )
\big\|  {\boldsymbol{{X}}} _{(j) }
\big
\|_{\mathnormal{F}}
\Big \}.
\end{align}
\end{Theorem}

\begin{Remark}
  The above theoretical results say that the error bounds for  Algorithm \ref{STHOSVD-BKI} and Algorithm \ref{bb-rsthosvd}
  are all independent of the processing order.
  This means that while some processing orders may lead to more accurate decompositions,
  every processing order has the same worst case error bound.
%
%The above results %
 Theorem \ref{hallman2022block144444} %It
 shows that
 %the error estimate $\sum_{j=1}^{d} E_{j}$ will remain accurate up to terms involving
% the deflation tolerance and the local loss of orthogonality in factor matrix $\{\bm{F}_{j}\}_{j=1}^{d}$.
% The proof makes the
as long as the  local orthogonality is maintained for factor matrix $\{\bm{F}_{j}\}_{j=1}^{d}$
and as long as the number of  deflations   is not too large, we can expect
%the error
 $\sum_{j=1}^{d} E_{j}$ to remain an accurate estimate of the Frobenius norm approximation error, at least when the error tolerance
 is not too small. %${\boldsymbol{{U}}} ^{\{j\} },   {\boldsymbol{{B}}} ^{\{j\} }$, and $ {\boldsymbol{{V}}} ^{\{j\} } $
 Thus, we conclude that even if we use one-side reorthogonalization
 (i.e., only  reorthogonalize $\{ {\boldsymbol{{V}}} ^{\{j\} } \} _{j=1} ^{d}$  regardless of the orthogonality of
 $\{{\boldsymbol{{U}}} ^{\{j\} }\} _{j=1} ^{d}  $), the quantity
  $\sum_{j=1}^{d} E_{j}$
 will remain a reliable estimate of the approximation error.
  Note that the above randomized LRTA algorithms and  corresponding theoretical analysis %error bounds
  can degenerate to
 % the results    existed in  matrix cases.
 existing results  presented in matrix format \cite{yuan2018superlinear,meyer2024unreasonable, hallman2022block1}.
\end{Remark}

\vspace{-0.487653cm}
\section{\textcolor[rgb]{0.00,0.00,0.00}{\textbf{Generalized Nonconvex %Robust High-Order  Tenosor Completion
%Low-Rank Plus Smooth %Tensor Representation
Tensor Modeling}}}  \label{model}  %Robust Tensor Completion

%\vspace{-0.3cm}
%%%%%%%%%%%%%%%%%%%%%%%%%%%%%%%%%%%%%%%%%%%%%%%%%%%%%%%%%%%%%%%%%%%%%%%%%%%%%%%%%%%%%%%%%%%%%%%%%%%%%%
 %\textcolor[rgb]{0.00,0.00,0.00}{This section introduces a generalized nonconvex framework for the RTC problem,
%   organized into  six subsections for detailed elaboration.
% Subsection \ref{nonconvex-regulari} devises two novel unified  nonconvex   regularizers.
%An effective    nonconvex RTC model is proposed in  subsection \ref{Model_Formulation}.
% %
% Based on the randomized LRTA strategies    presented in section  \ref{LRTAapproximation},
% Subsection \ref{gnlstr-oper} proposes   fast solution methods   for  pivotal subproblems.
%  An efficient ADMM algorithm is developed to  solve the formulated model in subsection \ref{optalgadmm}.}
%%
% Subsection \ref{converage} and \ref{timecom}
% provide the convergence analysis and time complexity analysis of the proposed algorithm, respectively.
%%%%%%%%%%%%%%%%%%%%%%%%%%%%%%%%%%%%%%%%%%%%%%%%%%%%%%%%%%%%%%%%%%%%%%%%%%%%%%%%%%%%%%%%%%%%%%%%%%%%%%%%%%%%%%%%%%%

\subsection{\textbf{Generalized
Nonconvex Regularizers}}
\label{nonconvex-regulari}
%Surrogate
%The corresponding proximal operator

\subsubsection{\textbf{Novel  regularizer encoded %low-rankness and smoothness
both L+S priors}}
In many application scenarios, the large-scale %big
 high-order tensors to be estimated are not only with low-rankness but also
possess significant smooth structure.
Existing purely low-rank based methods are not sufficient to adequately meet the accuracy requirement
when facing these simultaneous low-rank and smooth data
\cite{qin2022low, zheng2020tensor44, liu2024revisiting,  %wang2023guaranteed, %feng2023multiplex,
 yang2022355,  wang2024low2222, zhang2023tensor,qin2023nonconvex
%zhang2023tensor
}.
%%%%%%%%%%%%%%%%%%%%%%%%%%%%%%%%%%%%%%%%%%%%%%%%%%%%%%%%%%%%%%%%%%%%%%%%%%%%%%%%%%%%%%%%%%%%%%%%%%
Besides, previous works %studies
on
 simultaneous  $\textbf{L}$+$\textbf{S}$ tensor recovery
 typically encoded the \textbf{L}+\textbf{S} prior features coexisting in tensor data
 as two separate regularization terms (i.e., low-rank term plus total variation term), and then coupled them into various tensor recovery models through additive combination
\cite{chen2018tensor,qiu2021robust,  feng2023multiplex}.
However,
%to  overcome the difficulty of tuning %adjusting balancing
the trade-off  parameter between these two terms % imposed between these two regularizers.
%
 %in the above-mentioned methods,
 is tricky to be selected for achieving or approaching the Pareto optimality.
%%%%%%%%%%%%%%%%%%%%%%%%%%%%%%%%%%%%%%%%%%%%%%%%%%%%%%%%%%%%%%%%%%%%%%%%%%%%%%%%%%%%%%%%%%%%%%%%%%%%%%%%%%%%%%%%%%%%
% \begin{Remark}
 %\textbf{(The Motivation of Our Novel Regularizer)}
 %In the literature \cite{wang2023guaranteed},
 % Wang et al.  proposed a novel  high-order  T-CTV norm,   which incorporates  two insightful tensor  priors (i.e., \textbf{L} and \textbf{S})
% into  a unique regularization term and  ensures that both priors are encoded simultaneously without tuning the corresponding parameters of the two prior regularization terms.
%%%%%%%%%%%%%%%%%%%%%%%%%%%%%%%%%%%%%%%%%%%%%%%%%%%%%%%%%%%%%%%%%%%%%%%%%%%%%%%%%%%%%%%%%%%%%%%%%%%%%%%%%%%%%%%%
Recently,
Wang et al. \cite{wang2023guaranteed} proposed a novel regularization term termed Tensor Correlated Total Variation (T-CTV), which inherently %encodes %both low-rank and smooth
characterizes both
$\textbf{L}$ and $\textbf{S}$ priors of a tensor simultaneously.
%
%Motivated by the fact that   \textbf{L} and \textbf{S} priors %are usually coupled with
%are always  coexist with each other in real visual tensor,
% Wang et al. \cite{wang2023guaranteed} proposed a novel regularization penalty %called  \textit{tensor correlated total variation} (T-CTV)
%(\textit{tensor correlated total variation},  T-CTV) under the  T-SVD framework,
%    %which incorporates  two insightful tensor  priors (i.e., \textbf{L} and \textbf{S}) into  a unique  term.
%which characterizes  %two insightful tensor  priors (i.e., \textbf{L} and \textbf{S}) into  a unique  term.
%both $\textbf{L}$ and $\textbf{S}$ priors of a tensor with a unique  term.
%The underlying mathematical principles indicate that this single regularizer can promote the two priors concurrently.
%%%%%%%%%%%%%%%%%%%%%%%%%%%%%%%%%%%%%%%%%%%%%%%%%%%%%%%%%%%%%%%%%%%%%%%%%%%%%%%%%%%%%%%%%%%%%%%%%%%%%%%%%%%%%%%%%%%%%%%%%%%%%%%%%%
%Besides, unlike existing %related
%researches %on tensor recovery
%adopting  \textbf{L}+\textbf{S} priors \cite{chen2018tensor,qiu2021robust,  feng2023multiplex},
%this study overcomes the difficulty of tuning %adjusting balancing
%the trade-off  parameter imposed between these two regularizers.
%%%%%%%%%%%%%%%%%%%%%%%%%%%%%%%%%%%%%%%%%%%%%%%%%%%%%%%%%%%%%%%%%%%%%%%%%%%%%%%%%%%%%%%%%%%%%%%%%%%%%%%%%%%%%%%%%%%%%%%%%%%%%%%%%%%
Nevertheless, the T-CTV regularizer was established %modeled
by the HTNN  \cite{qin2022low}
%\textit{high-order TNN} (HTNN) \cite{qin2022low}
 in the gradient domain,
 which equally treats all singular components %values
 of each gradient tensor in the transform domain, thus causing
 some  unavoidable biases  in practical application.
% To alleviate this problem,
%%%%%%%%%%%%%%%%%%%%%%%%%%%%%%%%%%%%%%%%%%%%%%%%%%%%%%%%%%%%%%%%%%%%%%%%%%%%%%%%%%%%%%%%%%%%%%%%%%%%%%%%%%%%%%%%%%%%%%%%%%%%%
%In order to better encode the global low-rankness and local smoothness  of a tensor simultaneously,
%  we propose  an unified nonconvex surrogate of T-SVD rank as the %a
%   tighter regularizer, which flexibly incorporates % that
%  %involves
%  a family %class
%   of nonconvex penalty functions.
   %%%%%%%%%%%%%%%%%%%%%%%%%%%%%%%%%%%%%%%%%%%%%%%%%%%%%%%%%%%%%%%%%%%%%%%%%%%%%%%%%%%%%%%%%%%%%%%%%%%%%%%%%%%%%%%
   To  address %model
    simultaneous  $\textbf{L}$+$\textbf{S}$ large-scale %high-order
   tensor recovery problem more effectively and flexibly,
   we propose a unified nonconvex regularization strategy that is tighter for T-SVD rank approximation.
% Our main motivation is that those larger singular values in the gradient tensors should be penalized less, which is quite reasonable. The
%gradient tensor inherits the low-rank property of the original tensor, where the smaller singular values mainly correspond to noise. Therefore, we adopt a reweighting strategy for the gradient tensors to take advantage of the above existing prior information.
%%%%%%%%%%%%%%%%%%%%%%%%%%%%%%%%%%%
%\end{Remark}

%Inspired by the structure of high-order T-CTV
%(see Definition \ref{def10}),
%Inspired by the paradigm of prior characterization presented in Definition \ref{def10},
Below, based on the shortcomings of existing prior characterization paradigms  \cite{chen2018tensor,qiu2021robust,  feng2023multiplex, wang2023guaranteed},
%characterization paradigm
we define the following \textit{Generalized Nonconvex High-Order T-CTV}  (GNHTCTV)
regularizer:
%that encodes the low-rankness and the smoothness of a tensor simultaneously,
%i.e.,
%%%%%%%%%%%%%%%%%%%%%%%%%%%%%%%%%%%%%%%%%%%%%%%%%%%%%%%%%%%%%%%%%%%%%%%%%%%%%%%%%%%%%%%%%%%%%%%%%%%%%%%%%%%%%%%%%%%%%%%%%%%%%%%%%%%
%%%%%%%%%%%%%%%%%%%%%%%%%%%%%%%%%%%%%%%%%%%%%%%%%%%%%%%%%%%%%%%%%%%%%%%%%%%%%%%%%%%%%%%%%%%%%%%%%%%%%%%%%%%%%%%%%%

%%%%%%%%%%%%%%%%%%%%%%%%%%%%%%%%%%%%%%%%%%%%%%%%%%%%%%%%%%%%%%%%%%%%%%%%%%%%%%%%%%%%%%%%%%%%%%%%%%%%%%%%%%%%%%%%%%
\begin{Definition}\label{def10}
\textbf{(Gradient tensor \cite{wang2023guaranteed})}
For ${\boldsymbol{\mathcal{A}}} \in \mathbb{R}^{ n_1\times \cdots \times n_d} $,
its gradient tensor along the $k$-th mode is defined as
%Let $\mathfrak{L}$ be any invertible linear transform in (\ref{trans}) and it satisfies (\ref{orth}),
%${\boldsymbol{\mathcal{S}}}$ be from the middle
%component of
%${\boldsymbol{\mathcal{A}}} = {\boldsymbol{\mathcal{U}}}{{*}_{\mathfrak{L}}}{\boldsymbol{\mathcal{S}}} {*}_{\mathfrak{L}}
% {\boldsymbol{\mathcal{V}}}^{\mit{T}}$.
%%%%%%%%%%%%%%%%%%%%%%%%%%%%%%%%%%%%%%%%%%%%%%%%%%%%%%%%%%%%%%%%%%%%%%%%%%%%%%%%%%%%%%%%
\begin{align}\label{gradient-tensor}
 {\boldsymbol{\mathcal{G}}}_{ k }
&:=
\nabla_{k} ({\boldsymbol{\mathcal{A}}})=
\boldsymbol{\mathcal{A}} \;{\times}_{k} \;\bm{D}_{n_k}, \;\; k=1,2,\cdots, d,
\end{align}
where $\bm{D}_{n_k}$ is a row circulant matrix %. circulant matrix
of $(-1,1, 0, \cdots,0)$, and
$\nabla_{k} $ is defined as the corresponding
difference operator along the $k$-th mode
of tensor ${\boldsymbol{\mathcal{A}}}$.
\end{Definition}

%\begin{Definition}\label{gnhtctv}
%\textbf{(GNHTCTV)}
%Let $\mathfrak{L}$ be any invertible linear transform in (\ref{trans}) and its  transform matrices  satisfy (\ref{orth}).
%%%%%%%%%%%%%%%%%%%
%For any ${\boldsymbol{\mathcal{A}}} \in \mathbb{R}^{ n_1\times n_2 \times n_3  \cdots \times n_d} $,
%%
%then its GNHTCTV  norm is defined as
%%%%%%%%%%%%%%%%%
%%
%\end{Definition}

%%%%%%%%%%%%%%%%%%%%%%%%%%%%%%%%%%%%%%%%%%%%%%%%%%%%%%%%%%%%%%%%%%%%%%%%%%%%%%%%%%%%%%%%%%%%%%%%%%%%%%%%%%%%%%%%%%
\begin{Definition}\label{gnhtctv}
\textbf{(GNHTCTV)}
Let $\mathfrak{L}$ be any invertible linear transform in (\ref{trans}) and
its corresponding
transform matrices  satisfy (\ref{orth}).
%%%%%%%%%%%%%%%%%%
For ${\boldsymbol{\mathcal{A}}} \in \mathbb{R}^{ n_1\times \cdots \times n_d} $,
denote $\Gamma$ as a priori set consisting of directions along which  ${\boldsymbol{\mathcal{A}}} $
equips $\textbf{L}$+$\textbf{S}$ priors,
and ${\boldsymbol{\mathcal{G}}}_{k}$, $k\in \Gamma$ as its correlated gradient tensors.
Then,
the
 GNHTCTV norm
is defined as
%%%%%%%%%%%%%%%%%%%%%%%%%%%%%%%%%%%%%%%%%%%%%%%%%%%%%%%%%%%%%%%%%%%%%%%%%%%%%%%%%%%%%%%%
%%%%%%%%%%%%%%%%%%%%%%%%%%%%%%%%%%%%%%%%%%%%%%%%%%%%%%%%%%%%%%%%%%%%%%%%%%%%%%%%%%%%%%%%
\begin{align*}
%\| {\boldsymbol{\mathcal{A}}}\|_{ \textit{GNHTCTV}}
\Psi( {\boldsymbol{\mathcal{A}}})
&:= \frac{1}{\gamma} \sum_{k \in \Gamma}
 \big\|
 \nabla_{k} ({\boldsymbol{\mathcal{A}}})
 \big\|_
{
\Phi,
\mathfrak{L} } %,
%\notag \\ %\label{GNHTCTV1111}
%&
=
\frac{1}{\gamma} \sum_{k \in \Gamma}
 \Big(
\frac{1}{\rho} \sum_{i=1}^{m} \sum_{j=1}^{n}
\Phi
%\psi
\big(
 \sigma_{ij}^{(k)}
\big)
\Big),
\end{align*}
\textcolor[rgb]{0.00,0.00,0.00}{where  $\gamma := \sharp\{\Gamma\}$ equals to the cardinality of $\Gamma$,}
%where
%${\boldsymbol{\mathcal{Z}}}  \in\mathbb{R}^{n_1\times \cdots\times n_d}$,
$m=\min(n_1,n_2)$, $n= n_3 \cdots n_d$,
 $\rho=\alpha_3\cdots \alpha_d>0$ represents  a constant determined by  $\mathfrak{L}$,
% ${\boldsymbol{\mathcal{G}}}_{k} $  denotes  the gradient tensor along the $k$-th mode of ${\boldsymbol{\mathcal{Z}}}$,
%%%%%%%%%%%%%%%%%%%%%%%%%%%%%%%
${\boldsymbol{\mathcal{S}}}_{k}$  is %from %denotes
the middle
component of
$\nabla_{k}  ({\boldsymbol{\mathcal{A}}} )  = {\boldsymbol{\mathcal{U}}}_{k}  {{*}_{\mathfrak{L}}} {\boldsymbol{\mathcal{S}}}_{k} {*}_{\mathfrak{L}}
 {\boldsymbol{\mathcal{V}}}_{k}^{\mit{T}}$,
 $\sigma_{ij}^{(k)}=  { {\mathfrak{L}}({{{\boldsymbol{\mathcal{S}}}}_{k} })} ^ {<j>}  (i,i)$,
  % $k, \gamma, \Gamma$  are equivalent to the ones appeared in Definition \ref{def10},
  and
$\Phi (\cdot): \mathbb{R}^{+} \rightarrow \mathbb{R}^{+}$  is a generalized
nonconvex function. % satisfying the following assumptions:
\end{Definition}

%\begin{Definition}\label{gnhtctv}
%\textbf{(GNHTCTV)}
%Let $\mathfrak{L}$ be any invertible linear transform in (\ref{trans}) and its  transform matrices  satisfy (\ref{orth}).
%%%%%%%%%%%%%%%%%%%
%For any ${\boldsymbol{\mathcal{A}}} \in \mathbb{R}^{ n_1\times n_2 \times n_3  \cdots \times n_d} $,
%%
%then its GNHTCTV  norm is defined as
%%%%%%%%%%%%%%%%%
%%
%\end{Definition}
%%%%%%%%%%%%%%%%%%%%%%%%%%%%%%%%%%%%%%%%%%%%%%%%%%%%%%%%%%%%%%%%%%%%%%%%%%%%%%%%%%%%%%%%%%%%%%%%%%%%%%%%%%%%%%%%%%%%%%%%%%%%%%%%%%%%%%%
%

\begin{Assumption} \label{assumpt}
The generalized nonconvex function $\Phi (\cdot): \mathbb{R}^{+} \rightarrow \mathbb{R}^{+}$
 %is assumed to
 satisfies the following assumptions:
%\\
\begin{itemize}
     \item
     \textbf{(I)}:
     $\Phi (\cdot)$: $\mathbb{R}^{+} \rightarrow \mathbb{R}^{+}$ is proper, lower semi-continuous
     and symmetric with respect to y-axis;

     \item
      \textbf{(II)}:
      $\Phi (\cdot)$ %: $\mathbb{R}^{+} \rightarrow \mathbb{R}^{+}$ % is nonconvex function.
is concave
%continuous,
and monotonically increasing  on $[0,\infty)$
%,
%and %possibly nonsmooth.
 with %function which satisfies:
 $\Phi(0)=0$.
 %%, {\lim _{x\rightarrow +\infty}} \frac {\Phi (x)} {x} =0  $.
%
   \end{itemize}
   If there is no special explanation, we suppose that Assumption \ref{assumpt} holds throughout the paper.
\end{Assumption}

Many popular nonconvex penalty functions satisfy  the above Assumption \ref{assumpt},
such as
firm function \cite{gao1997waveshrink}, logarithmic (Log) function  \cite{gong2013general},
 $\ell_{q}$ function \cite{marjanovic2012l_q},
 smoothly clipped absolute deviation (SCAD) function \cite{fan2001variable1 },
and  minimax concave penalty  (MCP) function \cite{zhang2010nearly}, capped-$\ell_{q}$ function \cite{li2020matrix, pan2021group}.
% In this way,
Thus,  the novelly proposed   GNHTCTV regularizer %(see Definition \ref{gnhtctv}) %(\ref{unified-regular}),
is %we are
equivalent to employing a family of nonconvex functions
%a family of nonconvex penalty functions is
%employed  to  all  singular values of each  gradient tensor.
onto  the   singular values of all face slices of  each  gradient tensor
%%%%%%%%%%%%%%%%%%%%%%%%%%%%%%%%%%%%%%%%%%%%%%%%%%%%%%%%%%%%%%%%%%%%%%%%%%%%%%%%%%%%%%%%%%
%%%%%%%%%%%%%%%%%%%%%%%%%%%%%%%%%%%%%%%%%%%%%%%%%%%%%%%%%%%%%%%%%%%%%%%%%%%%%%%%%%%%%%%%%%%%%
in the transformed domain.
Many existing regularizers (e.g., \cite{
jiang2019robust, song2020robust,yang2022robust, ng2020patched,
chen2020robust,qiu2021nonlocal,zhao2020nonconvex,zhao2022robust,
zhang2023generalized, qiu2024robust,
%qiu2021nonlocal, qiu2024robust,
%
qin2022low,qin2022robust , qin2021robust, qin2023nonconvex, wang2023guaranteed}) %can be viewed as special cases
can be regarded %considered
 as %special cases %
 specialized instances
of our GNHTCTV. %proposed method.
%From the equation (\ref{unified-regular}), we can find that
For example,
when $\Phi (\cdot)$  is set to the $\ell_1$-norm, the GNHTCTV regularizer  degenerates to the
T-CTV norm proposed
 in \cite{wang2023guaranteed}, i.e.,
\begin{align*}\label{WTSN111}
\| {\boldsymbol{\mathcal{A}}}\|_{ \textit{T-CTV}} %W-TCTV
&:= \frac{1}{\gamma} \sum_{k\in \Gamma}
 \big\|{\boldsymbol{\mathcal{G}}}_{k} \big\|_
{ %\circled{ {\boldsymbol{\mathcal{W}}}  }, %
\circledast,
%\circled{w},
\mathfrak{L} },
\end{align*}
where
$ \|\cdot  \|_
{
\circledast,
\mathfrak{L} }$ denotes the
\textit{high-order TNN} (HTNN) \cite{qin2022low}, i.e.,
$\| {\boldsymbol{\mathcal{A}}}\|_{\circledast,{\mathfrak{L}} }
: = \frac{1}{\rho} \big\|\operatorname{bdiag}(\boldsymbol{\mathcal{A}}_{\mathfrak{L}})\big\|_{\star}$,
where $\rho= \alpha_3 \cdots \alpha_d$.

\subsubsection{\textbf{Novel noise/outliers regularizer}}
Related studies indicate that  the $\ell_{1}$-norm sometimes is %might  be
biased and  statistically suboptimal in %promoting sparsity,
enhancing the robustness against noise/outliers.
Besides,
%the
real-world   tensor data  may also be corrupted by %be damaged by
other structured  noise/outliers (e.g., slice-wise and tube-wise format)  other than the entry-wise form. %distribution,
Owing to %Based on the fact that and
the aforementioned facts,
we hence define
the following generalized  nonconvex noise/outliers regularization penalty: % term:
\begin{equation} \label{unified-regular11}
\Upsilon( {\boldsymbol{\mathcal{E}}})
:=
 %\sum_{{i_1}=1}^{n_1}  \sum_{{i_2}=1}^{n_2} \cdots \sum_{{i_d}=1}^{n_d}
%\phi
\psi
\big(
 %{ {\mathfrak{L}}({{{\boldsymbol{\mathcal{S}}}}_{k} })} ^ {<j>}  (i,i)
h( %\big
{\boldsymbol{\mathcal{E}}} %_{i_1 i_2 \cdots i_d} %\big
)
\big),
\end{equation}
where % $\psi(\cdot)$ has the same properties as $\Phi(\cdot)$.
$\psi (\cdot):\mathbb{R}^{+} \rightarrow \mathbb{R}^{+}$  is %represents   %is
  a generalized
nonconvex function,  which  has the same properties as $\Phi(\cdot)$.
Here, % we  consider
three type of  corrupted  noise/outliers are taken into account: \textbf{1)}
When the tensor ${\boldsymbol{\mathcal{E}}}$ is an entry-wise noise/outlier tensor, and then
$h(\cdot)=\|\cdot\|_{\ell_{1}}$ is defined as an $\ell_{1}$-norm;
% Then, we have
%\begin{equation} \label{unified-regular22}
%%
%\Upsilon( {\boldsymbol{\mathcal{E}}})
%:= \psi (\| {\boldsymbol{\mathcal{E}}}\|_{\ell_{1}} ) =
% \sum_{{i_1}=1}^{n_1}  \sum_{{i_2}=1}^{n_2} \cdots \sum_{{i_d}=1}^{n_d}
%\psi
% %{ {\mathfrak{L}}({{{\boldsymbol{\mathcal{S}}}}_{k} })} ^ {<j>}  (i,i)
%\big( %\big
%|{\boldsymbol{\mathcal{E}}}_{i_1 i_2 \cdots i_d} %\big
%|\big).
%\end{equation}
\textbf{2)}
When %the tensor ${\boldsymbol{\mathcal{E}}}$ is a tube-wise noise/outlier tensor,
the noise/outlier tensor ${\boldsymbol{\mathcal{E}}}$  has  tube-wise structure,  %, %sparsity,
and then
$h(\cdot)=\|\cdot\|_{\operatorname{tube}_1}$ is defined as an $\operatorname{tube}_1$-norm;
% then we have %. Then, we have
%\begin{equation} \label{unified-regular22222}
%%
%\Upsilon( {\boldsymbol{\mathcal{E}}})
%:= \psi (\| {\boldsymbol{\mathcal{E}}}\|_{\operatorname{tube}_1} ) =
% \sum_{{i_1}=1}^{n_1}  \sum_{{i_2}=1}^{n_2} %\cdots \sum_{{i_d}=1}^{n_d}
%\psi
% %{ {\mathfrak{L}}({{{\boldsymbol{\mathcal{S}}}}_{k} })} ^ {<j>}  (i,i)
%\big(% \big
%\|{\boldsymbol{\mathcal{E}}}_{i_1 i_2 : \cdots : } %\big
%\|_{{\mathnormal{F}}} \big
%).
%\end{equation}
%%%
\textbf{3)}
When the %tensor ${\boldsymbol{\mathcal{E}}}$ is a slice-wise noise/outlier tensor,
%noise/outlier tensor ${\boldsymbol{\mathcal{E}}}$  has structured sparsity on the slices, %is a slice-wise ,
 tensor ${\boldsymbol{\mathcal{E}}}$  has structured noise/outlier %sparsity
  on the slices, %is a slice-wise ,
  and then
$h(\cdot)=\|\cdot\|_{\operatorname{slice}_1}$ is defined as an $\operatorname{slice}_1$-norm.
%a Frobenius norm.
%%%%%%%%%%%%%%%%%%%%%%%%%%%%%%%%%%%%%%%%%%%%%%%%%%%%%%%%%%%%%%%%
%Then, % we have
%\begin{equation} \label{unified-regular555555}
%%
%\Upsilon( {\boldsymbol{\mathcal{E}}})
%:=\psi (\| {\boldsymbol{\mathcal{E}}}\|_{\operatorname{slice}_1} ) =
% \sum_{{j}=1}^{n_3 n_4 \cdots n_d}
%\psi
% %{ {\mathfrak{L}}({{{\boldsymbol{\mathcal{S}}}}_{k} })} ^ {<j>}  (i,i)
%\big(% \big
%\|{\boldsymbol{\mathcal{E}}}^{<j>} %\big
%\|_{\mathnormal{F}} \big
%).
%\end{equation}

\vspace{-0.3840503cm}
%\vspace{-0.4cm}
\subsection{\textbf{Fast Generalized Solver for Key Subproblems}}   \label {gnlstr-oper}
%\subsection{\textbf{Generalized Proximity Operator for Sparsity and Low-Rank Inducing Penalties}}   \label {gnlstr-oper}

This subsection %%mainly
 presents the solution method of   two kinds of
 key subproblems  %involved  in  GNRHTC model (\ref{orin_nonconvex}),
 involved in subsequent %tensor recovery
 models,
 i.e.,
%and
\begin{align}\label{E_prox1}
&
\textcolor[rgb]{0.00,0.00,0.00}{
\arg\min_{ \boldsymbol{\mathcal{E}}}
{\lambda} \cdot \Upsilon \big({\boldsymbol{\mathcal{E}}} \big)
 +
\frac
{1}{2}
\|
{\boldsymbol{\mathcal{E}}}-  {\boldsymbol{\mathcal{A}}}
\|^2_{\mathnormal{F}},}
%\end{align}
%%and
%\begin{align}
\\
\label{LS_prox1}
&
\arg\min_{ {\boldsymbol{\mathcal{G}}}}
%\tau \| {\boldsymbol{\mathcal{Z}}}\|_{ { {\boldsymbol{\mathcal{W}}}, {\boldsymbol{\mathcal{S}}}_{p} }}^{p}
%\tau
%\Phi ({\boldsymbol{\mathcal{G}}} )
\tau\cdot
\big\| {\boldsymbol{\mathcal{G}}} \big\|_{\Phi,\mathfrak{L} }
+
\frac{1}{2}\|{\boldsymbol{\mathcal{G}}} -{\boldsymbol{\mathcal{A}}}\|^{2}_{\mathnormal{F}}. %\;\; k\in \Gamma,
\end{align}

\subsubsection{\textbf{Solve the problem (\ref{E_prox1})  via GNHTT %Generalized Nonconvex Thresholding
Operator}}
Before providing the solution to the problem  (\ref{E_prox1}),
we first introduce the proximal  mappings  of the generalized nonconvex functions, %regularization penalty.
which plays a central role in developing highly efficient first-order algorithms.
Specifically, for a nonconvex penalty function $\psi(\cdot)$ satisfying Assumption \ref{assumpt},  its proximity operator is defined as
\begin{equation}\label{equ_gst1}
\operatorname{\textit{Prox}}_ {\psi, \mu} (v)=
\arg \min_{x} \Big \{  \mu\cdot    \psi(x)
 + \frac{1}{2} {(x-v)}^2 \Big \},
\end{equation}
where $\mu>0$ is a penalty parameter.

In the supplementary material, %%%Below,
we summarize the proximity operator %proximal operator
for  several popular nonconvex regularization penalties satisfying Assumption \ref{assumpt},
including firm-thresholding, $\ell_{q}$-thresholding,
capped-$\ell_{q}$ thresholding,
MCP, Log,  and SCAD  penalties.
 %%%
Based on the previous analysis,  the  optimal solution to  the subproblem (\ref{E_prox1})
can be computed by \textit{generalized nonconvex high-order tensor thresholding/shrinkage}  (GNHTT)  operator in an element-wise, tube-wise or slice-wise manner. %operator
%Please see Algorithm \ref{random-wtst} for details.
Specifically, if %when
$h(\cdot)=\|\cdot\|_{\ell_1}$, then we have
  $$ {\hat{{\boldsymbol{\mathcal{E}}}}}_ {i_1 \cdots i_d}
=
\operatorname{\textit{Prox}}_ {\psi, \lambda} ({\boldsymbol{\mathcal{A}}}_ {i_1 \cdots i_d}),
\forall (i_1, \cdots, i_d) \in [n_1] \times \cdots \times [n_d].
$$
If %When
 $h(\cdot)=\|\cdot\|_{\operatorname{tube}_1}$, then we have
%
% $i \in \{1,2,\cdots, n_1\},  j \in \{1,2,\cdots, n_2\} $
 %
$$ {\hat{ {\boldsymbol{\mathcal{E}}} }_{ij :\cdots :} }
  =
 \frac
  %(
  {  {\boldsymbol{\mathcal{A}}}_{ij : \cdots : } } %/
 { { \|{\boldsymbol{\mathcal{A}}}_{ij : \cdots : } \|_{\mathnormal{F}}} } %)
   \cdot
  \operatorname{\textit{Prox}}_ {\psi, \lambda} ({ \|{\boldsymbol{\mathcal{A}}}_{ij : \cdots : } \|_{\mathnormal{F}}}),
  \; \forall (i,j) \in [n_1] \times  [n_2].
  $$
If
$h(\cdot)=\|\cdot\|_{\operatorname{slice}_1  }$, then  we have
%
 %     $k=1,2,\cdots, n_3 \cdots n_d$
    %
$$ {\hat{ {\boldsymbol{\mathcal{E}}} } }^{<k>}
  =
  \frac
 %(
 { {  {\boldsymbol{\mathcal{A}}}}^{<k>}} %/
 { {   \|{\boldsymbol{\mathcal{A}}}^{<k>} \|_{\mathnormal{F}} } } %)
 \cdot
  \operatorname{\textit{Prox}}_ {\psi, \lambda} ({ \|{\boldsymbol{\mathcal{A}}}^{<k> } \|_{\mathnormal{F}}}),
  \forall  k \in [ n_3 \cdots n_d].
  $$

\subsubsection{\textbf{Solve the problem (\ref{LS_prox1})
via %\textit{Generalized Nonconvex High-Order Tensor Singular Value Thresholding}
GNHTSVT Operator}}
Before  providing the solution to the problem (\ref{LS_prox1}), we first introduce the following key definition and theorem.

\begin{Definition}\label{gNtsvt}
\textbf{(GNHTSVT operator)}
Let
${\boldsymbol{\mathcal{A}}}={\boldsymbol{\mathcal{U}}}{*}_{\mathfrak{L}}{\boldsymbol{\mathcal{S}}}{*}_{\mathfrak{L}}
{\boldsymbol{\mathcal{V}}}^{\mit{T}}$
%%%
be the  \text{T-SVD} of
 ${\boldsymbol{\mathcal{A}}} \in \mathbb{R}^{n_1\times  \cdots \times n_d}$, $m=\min(n_1,n_2)$, and $n=n_3 \cdots n_d$.
%%%
For any $\tau>0$, then
the %GNTSVT
\textit{Generalized Nonconvex High-Order Tensor Singular Value Thresholding}  (GNHTSVT)
operator of ${\boldsymbol{\mathcal{A}}}$   is defined as follows
\begin{equation}\label{gntsvt_operator}
{\boldsymbol{\mathcal{D}}}_{\Phi, \tau}({\boldsymbol{\mathcal{A}}}, \mathfrak{L})
={\boldsymbol{\mathcal{U}}}{*}_{\mathfrak{L}}
{\boldsymbol{\mathcal{S}}}_{\Phi, \tau}
{*}_{\mathfrak{L}}{\boldsymbol{\mathcal{V}}}^{\mit{T}},
\end{equation}
where $ {\mathfrak{L}} ( {\boldsymbol{\mathcal{S}}}_{\Phi, \tau}) ^{<j>} (i,i)=
 \operatorname{\textit{Prox}}_ {\Phi, \tau} \big (  {{\mathfrak{L}}(
{\boldsymbol{\mathcal{S}}}}
)^{<j>} (i,i)
\big)
$ for $j=1,2, \cdots,n$, $i=1,2, \cdots, m$,  and
$\operatorname{\textit{Prox}}_ {\Phi, \tau} (\cdot) $ denotes the  proximity operator of nonconvex penalty function
$\Phi(\cdot)$,
which   has the same properties as $\psi(\cdot)$.
\end{Definition}
%%%%%%%%%%%%%%%%%%%%%%%%%%%%%%%

%%%%%%%%%%%%%%%%%%%%%%%%%%%%%%%%%%%%%%%%%%%%%%%%%%%%%%%%%%%%%%%%%%%%%%%%%%%%%%%%%%%%%%%%%%%%%%%%%%%%%%%%%%%%%%%%%%
\begin{Theorem}\label{theorem_wtsvt}
 Let $\mathfrak{L}$ be any invertible linear transform in (\ref{trans}) and
% it satisfies
its  transform matrices satisfy
 (\ref{orth}), $m=\min{(n_1,n_2)}$, $n=n_3 n_4 \cdots n_d $.
For any $\tau>0$ and %${\boldsymbol{\mathcal{Z}}} \in \mathbb{R}^{n_1\times  \cdots \times n_d}$,
${\boldsymbol{\mathcal{A}}}  \in \mathbb{R}^{n_1\times  \cdots \times n_d}$,
if the
nonconvex penalty function
$\Phi(\cdot)$ satisfies Assumption \ref{assumpt},
then GNHTSVT  % GNTSVT
operator (\ref{gntsvt_operator}) obeys
\begin{align}\label{eq_wsvtEE}
{\boldsymbol{\mathcal{D}}}_{\Phi, \tau}({\boldsymbol{\mathcal{A}}}, \mathfrak{L})=
\arg\min_{ {\boldsymbol{\mathcal{G}}}}
\tau \cdot
%\Phi ({\boldsymbol{\mathcal{G}}} )
\big\| {\boldsymbol{\mathcal{G}}} \big\|_{\Phi,\mathfrak{L} }
+
\frac
{1} {2}
\|{\boldsymbol{\mathcal{G}}} -{\boldsymbol{\mathcal{A}}}\|^{2}_{\mathnormal{F}}.
\end{align}
\end{Theorem}

% \begin{proposition}\label{ppppp6666}
%Let ${{\boldsymbol{\mathcal{A}}}}= {{\boldsymbol{\mathcal{U}}}} {*}_{\mathfrak{L}}  {\boldsymbol{\mathcal{B}}} {*}_{\mathfrak{L}}
%{{\boldsymbol{\mathcal{V}}}^{\mit{T}}} \in \mathbb{R}^{n_1 \times n_2 \times \cdots\times n_d}$, where
%${{\boldsymbol{\mathcal{U}}}} \in \mathbb{R}^{n_1 \times k\times  \cdots\times n_d}$ and ${{\boldsymbol{\mathcal{V}}}}
%\in \mathbb{R}^{n_2 \times k\times  \cdots\times n_d}$ are partially orthogonal,
%${{\boldsymbol{\mathcal{B}}}} \in \mathbb{R}^{k \times k \times  \cdots\times n_d}$.
%Then, we have
%\begin{equation*}
%%
%{\boldsymbol{\mathcal{D}}}_{\Phi,\tau}({\boldsymbol{\mathcal{A}}}, \mathfrak{L})={{\boldsymbol{\mathcal{U}}}}
%{*}_{\mathfrak{L}} {\boldsymbol{\mathcal{D}}}_{\Phi,\tau}({\boldsymbol{\mathcal{B}}})
%{*}_{\mathfrak{L}}
%{{\boldsymbol{\mathcal{V}}}^{\mit{T}}}.
%%$$
%\end{equation*}
%%
%\end{proposition}

According to  the above Definition \ref{gNtsvt} and Theorem \ref{theorem_wtsvt},
%Therefore,
the  subproblem (\ref{LS_prox1})   can be efficiently solved in virtue of GNHTSVT operation.
%please see Theorem \ref{theorem_wtsvt} for details.

%%%%%%%%%%%%%%%%%%%%%%%%%%%%%%%%%%%%%%%%%%%%%%%%%%%%%%%%
 \begin{algorithm}[!htbp]
\setstretch{0.0}
     \caption{{\textcolor[rgb]{0.00,0.00,0.00}{Accelerated GNHTSVT operation.
   %
 % $ {\boldsymbol{\mathcal{D}}}_{\Phi, \tau}({\boldsymbol{\mathcal{A}}}, \mathfrak{L})$.
     }}
     }
     \label{random-wtsvt}
      \KwIn{
         $\bm{\mathcal{A}}\in\mathbb{R}^{n_1\times \cdots\times n_d}$, processing order: $\bm{\rho}$,
         transform:  $\mathfrak{L}$, %target T-SVD Rank: $k$,
      % nonconvex
        regularizer: $ \Phi(\cdot)$,
 $\tau> 0$,
 %relative
 %
  target rank:
       $\bm{r}$, %=(r_1, \cdots, r_d)
       block size in Algorithm \ref{STHOSVD-BKI}: $\bm{b}$, % =(b_1, \cdots, b_d)   $\bm{\rho}$, $\bm{b}$, $\bm{q}$,
       Krylov iterations: $\bm{q}$, %=(q_1, \cdots, q_d)
        block size in Algorithm \ref{bb-rsthosvd}: $b$,
        error tolerance:  $\epsilon$,
        deflation tolerance: $\delta$.
 % power iteration: $t$.
 %
% \in\mathbb{R}^{n_1\times   \cdots \times n_d}$,
      %
       }

%%%%%%%%%%%%%%%%%%%%%%%%%%%%%%%%%%%%%%%%%%%%%%%%%%%%%%%

\If{ \text{not utilize the randomized technique}}
{
 Compute the results of $\mathfrak{L}$ on $\boldsymbol{\mathcal{A} }$, i.e., $\mathfrak{L}(\boldsymbol{\mathcal{A}})$\;

{
\For{$v=1,2,\cdots, n_3 n_4 \cdots n_d$}
{

$[{\mathfrak{L}}({\boldsymbol{\mathcal{U}}})^{<v>},{\mathfrak{L}}({\boldsymbol{\mathcal{S}}})^{<v>},{\mathfrak{L}}({\boldsymbol{\mathcal{V}}})^{<v>}]= \textrm{\textit{svd}} \big(   {\mathfrak{L}} ({\boldsymbol{\mathcal{A}}})^{<v>}    \big)
$\;

      $
\tilde{\bm{S}} =  \operatorname{\textit{Prox}}_ {\Phi, \tau} \big [
\operatorname{\textit{diag}} \big(  \mathfrak{L}(\boldsymbol{\mathcal{S}})^{<v>} \big) \big]
$;\\

${  \mathfrak{L}(\boldsymbol{\mathcal{C}})   }^{<v>}=
{\mathfrak{L}(\boldsymbol{\mathcal{U}})}^{<v>}\cdot
\operatorname{\textit{diag}} %\operatorname{Diag}
(\tilde{\bm{S}})
\cdot
{   (\mathfrak{L}(\boldsymbol{\mathcal{V}})^{<v>})   }  ^{\mit{T}}$;\\

}
}
$ {\boldsymbol{\mathcal{D}}}_{\Phi, \tau}({\boldsymbol{\mathcal{A}}},  \mathfrak{L})
%%%
\leftarrow {\mathfrak{L}}^{-1}(
      % {\boldsymbol{\mathcal{U}}}_{\mathnormal{L}}
      {\mathfrak{L}}({\boldsymbol{\mathcal{C}}}))
$.
}

%%%%%%%%%%%%%%%%%%%%%%%%%%%%%%%%%%%%%%%%%%%%%%%%%%%%%%
\ElseIf{\text{utilize the randomized technique}}
{
{

             Obtain %$ {\mathfrak{L}}({\boldsymbol{\mathcal{U}}})^{<v>}$,
             %$ {\mathfrak{L}}({\boldsymbol{\mathcal{B}}})^{<v>}$, and $ {\mathfrak{L}}({\boldsymbol{\mathcal{V}}})^{<v>}$
             $  {\bm{\mathcal{A}}}= [\bm{\mathcal{C}};  \bm{F}_{1}, \bm{F}_{2},\cdots, \bm{F}_{d}]$
             by   Algorithm \ref{STHOSVD-BKI} or Algorithm \ref{bb-rsthosvd}\;

Perform the deterministic GNTSVT operation on Tucker core tensor $\bm{\mathcal{C}}$, i.e.,
$\tilde{{\boldsymbol{\mathcal{C}}}}  =  {\boldsymbol{\mathcal{D}}}_{\Phi, \tau}({\boldsymbol{\mathcal{C}}}, \mathfrak{L})
$\;

%
%              $[\mathfrak{L}(\hat{\boldsymbol{\mathcal{U}}})^{<v>},\mathfrak{L}(\boldsymbol{\mathcal{S}})^{<v>},\mathfrak{L}(\hat{\boldsymbol{\mathcal{V}}})^{<v>}]= \textrm{svd} \big(\mathfrak{L}(\boldsymbol{\mathcal{B}})^{<v>}\big) $;\\

%$
%\%hat{\bm{S}}
%\mathfrak{L}(\boldsymbol{\mathcal{U}})^{<v>}
%= \mathfrak{L}(\boldsymbol{\mathcal{U}})^{<v>}\cdot \mathfrak{L}(\hat{\boldsymbol{\mathcal{U}}})^{<v>}
%$;\\
%
%$
%%\hat{\bm{S}}
%\mathfrak{L}(\boldsymbol{\mathcal{V}})^{<v>}
%= \mathfrak{L}(\boldsymbol{\mathcal{V}})^{<v>}\cdot \mathfrak{L}(\hat{\boldsymbol{\mathcal{V}}})^{<v>}
%$;\\
$ {\boldsymbol{\mathcal{D}}}_{\Phi, \tau}({\boldsymbol{\mathcal{A}}}, \mathfrak{L})=
%\hat {\bm{\mathcal{A}} } = %\hat {\bm{\mathcal{G}} }
 %({\boldsymbol{\mathcal{U}}}{*}_{\mathfrak{L}}{\boldsymbol{\mathcal{S}}}{*}_{\mathfrak{L}}{\boldsymbol{\mathcal{V}}}^{\mit{T}})
 %%%
%%% ({\boldsymbol{\mathcal{D}}}_{\Phi, \tau}({\boldsymbol{\mathcal{C}}}, \mathfrak{L}))
 \tilde{{\boldsymbol{\mathcal{C}}}}
  {{\times}}_{1}  {{\bm{F}}_{1}}
 {{\times}}_{2}  {{\bm{F}}_{2}}  \cdots  {{\times}}_{d}  {{\bm{F}}_{d}}  $. %\\
}

}

    \end{algorithm}

%\vspace{-0.13cm}
\subsubsection{\textbf{Fast Randomized GNHTSVT}}\textcolor[rgb]{0.00,0.00,0.00}{
%From the Definition \ref{gNtsvt} and Theorem \ref{theorem_wtsvt},
%In view of the form of GNTSVT operator,
We can find that
when dealing with large-scale high-order  tensor data,
the major bottleneck of solving the  minimization problem (\ref{LS_prox1})
 is to compute   the GNHTSVT operator  involving time-consuming T-SVD  multiple times.
 %Based on
 To address this issue,
 by the % in virtue of  the %proposed
 %fixed-accuracy LRTA  strategy using the block Lanczos bidiagonalization method,
 R-STHOSVD-BKI and AD-RSTHOSVD-BLBP
 %fixed-rank or fixed-accuracy LRTA methods
  algorithms
  leveraging full-mode random projection, Krylov subspace % block Krylov
  iteration
    and
 block Lanczos bidiagonalization strategies,
 %the fixed-accuracy LRTA leveraging full-mode random projection strategy
% Krylov subspace method, block Lanczos bidiagonalization process approach and random projection strategy
%%%%%%%%%%%%%%%%%%%%%%%%%%%%%%%%%%%%%%%%%%%%%%%%%%%%%%%%%%%%%%%%%%%%%%%%%%%%%%%%
% using the block Lanczos bidiagonalization method (see Algorithm \ref{bb-rsthosvd})
 a fast and  efficient  randomized method %algorithm
 is suggested   to boost   the computational speed
  of GNHTSVT operator  (please see Algorithm    \ref{random-wtsvt} for details)}.
Specifically,
%Apply the full-mode random projection to the original high-order tensor
the adaptive %newly  proposed
randomized Tucker compression algorithm  %leveraging
%the fixed-accuracy LRTA leveraging full-mode random projection strategy
%%%%%%%%%%%%%%%%%%%%%%%%%%%%%%%%%%%%%%%%%%%%%%%%%%%%%%%%%%%%%%%%%%%%%%%%%%%%%%%%
% using the block Lanczos bidiagonalization method (see Algorithm \ref{bb-rsthosvd})
is first applied to the original large-scale high-order tensor. % (\textbf{Line 11,  Algorithm \ref{bb-rsthosvd}}).
Then, the deterministic GNHTSVT operation is employed to the small-scale core tensor generated by randomized %of the
Tucker decomposition. % (\textbf{Line 12,  Algorithm \ref{bb-rsthosvd}}).
Finally,
 %combine the  the result of
 obtain the   approximation tensor by
  back projection  of
 the factor matrices in  Step $1$ and  the result of GNHTSVT operation in  Step $2$.
% (\textbf{Line 13,  Algorithm \ref{bb-rsthosvd}}). %to obtain the   approximation tensor.
Please see Line $11$-$13$ in  Algorithm \ref{random-wtsvt} for details. % bb-rsthosvd
The main idea is utilizing the randomized Tucker factorization in the first step as a preprocessing step
after which the deterministic algorithms can be applied to the smaller Tucker core tensor.

\vspace{-0.35cm}
\section{\textcolor[rgb]{0.00,0.00,0.00}{\textbf{Typical Applications}}}

\subsection{\textbf{Unquantized Tensor Recovery}}

In this subsection, we consider
%applying the proposed %generalized nonconvex modeling and accelerated %randomization computing
                     %randomized LRTA
%modeling and computing %strategies
%strategies to
 two typical nonquantized tensor recovery tasks: 1) Noisy tensor completion; 2) Noise-free tensor completion.

\subsubsection{\textbf{Generalized Nonconvex Model}} \label{Model_Formulation}

Based on the  designed  generalized nonconvex regularizers (see Subsection \ref{nonconvex-regulari}),  a novel   model named
\underline{G}eneralized  \underline{N}onconvex \underline{R}obust
\underline{H}igh-Order
\underline{T}ensor \underline{C}ompletion (GNRHTC)
is proposed in this paper.
%%%%%%%%%%%%%%%%%%%%%%%%%%%%%%%%%%%%%%%%%%%%%%%%%%%%%%%%%%%%%%%%%%%%%%%%%%%%
Mathematically,  the  proposed GNRHTC model  can be formulated as follows:
\begin{equation} \label{orin_nonconvex}
\min_{{\boldsymbol{\mathcal{L}}},{\boldsymbol{\mathcal{E}}}}
 \Psi( {\boldsymbol{\mathcal{L}}})
 + {\lambda}  \Upsilon ({\boldsymbol{\mathcal{E}}}),
\boldsymbol{\bm{P}}_{{{\Omega}}}(%{\boldsymbol{\mathcal{D}}}{*}_{\mathfrak{L}}
{\boldsymbol{\mathcal{L}}}
+{\boldsymbol{\mathcal{E}}})=
\boldsymbol{\bm{P}}_{{{\Omega}}}({\boldsymbol{\mathcal{M}}}),
\end{equation}
where  $\boldsymbol{\bm{P}}_{{{\Omega}}}(\cdot)$ is the projection operator onto the observed index set ${{{\Omega}}}$,
$\Psi({\boldsymbol{\mathcal{L}}})$ represents the regularizer measuring  tensor low-rankness plus smoothness properties concurrently, $\Upsilon({\boldsymbol{\mathcal{E}}})$
  denotes the  noise/outliers  regularization (see Subsection \ref{nonconvex-regulari} for details), and
 $\lambda > 0$  is a  trade-off parameter that balances these two regularization terms. % ${\boldsymbol{\mathcal{M}}}$ is the observed tensor,
%
% \begin{Remark}
 Broadly speaking,  the advantages of GNRHTC model are summarized as follows:

 %\begin{itemize}
% \item
% Note that
 1) The optimization problems formulated in existing works (e.g., \cite{
 qin2022low,wang2023guaranteed,
 jiang2019robust, wang2019robust,wang2020robust,song2020robust,ng2020patched,
 %chen2020robust, zhang2023tensor,
 chen2020robust, qin2022robust , qin2021robust, qin2023nonconvex,
 %zhao2020nonconvex,qiu2021nonlocal,zhang2023generalized, zhang2023generalized77
 zhao2020nonconvex, wang2021generalized, qiu2021nonlocal,zhao2022robust,zhang2023generalized, zhang2023generalized77
%%% yang2022robust11, zhou2019tensor, du2022enhanced, wang2022learning,qin2023generalized,xiao2023robust,liu2023anomaly
 %%%%liu2012robust,liu2016blessing
%%%% candes2011robust,liu2011latent,liu2012robust,yin2015laplacian,liu2016blessing
 })
 can be viewed as special cases of our GNRHTC.
 For example,
 \textbf{1)} Our proposed model (\ref{orin_nonconvex}) is equivalent to the NRTRM model \cite{ qiu2021nonlocal}
 or   GNCM-RTC model \cite{zhang2023generalized},
 when %the dictionary ${\boldsymbol{\mathcal{D}}} $ keeps  an identity tensor,
 the tensor's order is set to $3$ and the local smoothness is not taken into account;
 \textbf{2)} Furthermore, on the basis of the above conditions, if
 an  additional condition is  attached, i.e.,  ${\Omega}$ is the whole observed  set,  then
 our proposed model (\ref{orin_nonconvex}) can reduce to the GNR model \cite{zhang2023generalized77};
 \textbf{3)} %When the nonconvex functions $\psi(\cdot)$, $\Phi(\cdot)$ in the two proposed regularization terms  are linear, % functions,
When the generalized %nonconvex
regularization terms are respectively  set to be the T-CTV norm and $\ell_{1}$-norm, i.e., $\Psi( {\boldsymbol{\mathcal{L}}}) = \| {\boldsymbol{\mathcal{L}}}\|_{ \textit{T-CTV}}$,
 $\Upsilon({\boldsymbol{\mathcal{E}}})= \| {\boldsymbol{\mathcal{E}}}\|_{ 1}$,
  %the dictionary ${\boldsymbol{\mathcal{D}}} $ remains an identity tensor,
 and ${\Omega}$ is the whole indices set, then
 our formulated  model (\ref{orin_nonconvex})  degenerates to the
 T-CTV-TRPCA model \cite{wang2023guaranteed}.
 %There is other
%situation where the theoretical guarantee of matrix completion
%in [67] is the equivalent version of the Theorem 4.1.

  % \item
   2) Compared with the existing
    %convex %subspace approaches
    %tensor-based  low-rank representation
    %robust tensor completion
    RLRTC approaches (\text{e.g.},
    \cite{chen2020robust,qin2022robust,qin2021robust, qin2023nonconvex, zhao2020nonconvex,qiu2021nonlocal,zhao2022robust,zhang2023generalized})
    %\cite{yang2022robust11, zhou2019tensor, du2022enhanced, wang2022learning,qin2023generalized,xiao2023robust,liu2023anomaly}
    %
 %
  %
   within the  T-SVD framework,
   the   GNRHTC  method consider a
    %takes into account a
     more  unified, realistic and  challenging situation.
   % more generalized %,
    %and realistic
   %%%%%%%%%%%%%%%%%%%%%%%%%%%%%%%%%%%%%%%%%%%%%%%%%%%%%
 %%%%%%%%%%  Firstly, the   GNLSTR model      can  process  incomplete and noisy     tensor data  with arbitrary order.
  %Wherein,  two  joint generalized  nonconvex regularizers  enable
%   %
%   the  low-rank plus smooth structure     underlying in the corrupted high-order  tensor  to be well captured,
%   and the robustness against noise/outliers to be well enhanced.
%Firstly,
Specifically,
 the   GNRHTC model   %%%where
%the degenerated tensor is caused by both missing values and noise
can well
 %processing
 process
 the  %%%the acquired %tensors
tensor data  with arbitrary order
%to be processed
%%%are
degraded by  elements loss and noise/outliers corruption.
% more generalized %, comprehensive
    %and realistic
%Secondly,
Besides,  two novel generalized nonconvex  regularizers  $\Psi( \cdot)$ and  $\Upsilon (\cdot)$
are %developed to %designed to
integrated into the  GNRHTC model,
%one of which
the former
%novel generalized nonconvex  regularizer  $\Psi( \cdot)$ is developed to
can synchronously encode %the low-rank plus smooth structure
both \textbf{L} and \textbf{S} priors of a tensor into a unique concise form,
%and another new unified nonconvex regularizer $\Upsilon (\cdot)$ is devised to
and the latter %another one % the other
%can
%be  capable of removing % processing
%remove
is
capable of   processing %resisting %removing
several types of popular structured noise/outliers tensors, like
entries-wise, slice-wise and tube-wise forms.
Existing approaches rarely achieve these goals simultaneously.
% Existing approaches rarely achieve these goals simultaneously.
%% joint
% the union of  two  generalized nonconvex  regularizers  can well reconstruct %capture
%  the  low-rank plus smooth structure %low-rankness plus smoothness
%  underlying in the  tensor data, % synchronously,
%  and can well enhanced the robustness against noise/outliers.
%  %Thirdly,

   %\item

% \end{itemize}

% \end{Remark}

 %traditional model assumes that the underlying data lie in a single subspace
% we consider tensor LR linear representation
% tend to be draw from a union of subspace
 %the model offer stronger clean data recovery capacity by considering more complex data structure.
 % tensor linear representation

% Our work extends the recovery of corrupted data from a single subspace (RPCA) to multiple subspaces.
%Compared to [20], which requires the bases of subspaces to be known for handling the corrupted data from multiple subspaces,
% our method is autonomous, i.e., no extra clean data is required.

%Proximal
%Proximity  Operator and Generalized Nonconvex %Unified
%Singular Value Thresholding
% Proximity Operator for Sparsity and Low-Rank Inducing Penalties
% Generalized Solver for Key Subproblems

\begin{Remark}
When we disregard the noise scenario (i.e.,the trade-off parameter $\lambda $  is set to zero), the proposed model (\ref{orin_nonconvex})  %Model 1
degenerates into the following \textit{generalized nonconvex high-order tensor completion} (GNHTC) model:
\begin{align} %\\
%\label{equ_nonconvexhtc}
& \min_{{\boldsymbol{\mathcal{L}}}, %{\boldsymbol{\mathcal{E}}},
 {\boldsymbol{\mathcal{G}}}_{k}}
\frac{1}{\gamma} \sum_{k\in \Gamma}
 %\Phi ({\boldsymbol{\mathcal{G}}}_{k} )
 \big\| {\boldsymbol{\mathcal{G}}}_{k} \big\|_{\Phi,\mathfrak{L} },
 %+
% {\lambda}  \Upsilon \big(\boldsymbol{\bm{P}}_{{{\Omega}}}({\boldsymbol{\mathcal{E}}})\big),
 %%%%%%%%%%%%%%%%%%%%%%%%%%%%%%%%%%%%%%%%%%%%%%%%%%%%%%%%%%%%%%%%%%%%%%%%%%%%%%%%%%%%%%%%%%%%%%%%%%%%%%%%%
\notag \\ \label{equ_nonconvexhtc}
&
\text{s.t.} \;\;
 {\boldsymbol{\mathcal{G}}}_{k}=\nabla_{k}({\boldsymbol{\mathcal{L}}}),  \;\;
%{\boldsymbol{\mathcal{L}}}
%+{\boldsymbol{\mathcal{E}}}=
%{\boldsymbol{\mathcal{M}}},
\boldsymbol{\bm{P}}_{{{\Omega}}}(%{\boldsymbol{\mathcal{D}}}{*}_{\mathfrak{L}}
{\boldsymbol{\mathcal{L}}}
%+{\boldsymbol{\mathcal{E}}}
)=
\boldsymbol{\bm{P}}_{{{\Omega}}}({\boldsymbol{\mathcal{M}}}).
\end{align}
 Existing  %matrix-based subspace methods,  %liu2012robust,liu2016blessing
   %matrix-based  low-rank representation (MLRR) \ell
   robust %low-rank
   matrix completion (RMC)
   methods (e.g., \cite{%zhou2011godec,
   nie2015joint, lu2015nonconvex,  yao2018large, wen2019robust, zhang2022generalized, WangZhi2025low
   %candes2011robust,liu2011latent,liu2012robust,yin2015laplacian,liu2016blessing
 %  wang2023robust, yao2018large, yao2018large111
   })
   are limited to the case of  matrix data.
   %%%
These efforts requires to reshape the raw data into a large-scale matrix when dealing with high-order tensor
data. % faced in realistic application, like medical images, color videos, traffic data, etc.
Wherein, the matrization operator
destroys the intrinsic structure of the original tensor data, thus
resulting in unsatisfied recovery performance.
%%%%%%%%%%%%%%%%%%%%%%%%%%%%%%%%%%%%%%%%%%
 %  When dealing with  the high-order tensor data  faced in practical applications,  %compared with  %different from
 %  unlike  existing  %matrix-based subspace methods,  %liu2012robust,liu2016blessing
%   matrix-based  low-rank representation (MLRR) methods (e.g., % \cite{candes2011robust,liu2011latent,liu2012robust,yin2015laplacian,liu2016blessing})
 %   that  require to perform  %vectorization or
 %  matrization operator on  tensor data,
 %
%
   On the contrary,
    the  proposed GNHTC and GNRHTC models  (\ref{orin_nonconvex})-
(\ref{equ_nonconvexhtc}) are able to work directly with raw %higher-order matrization
    tensor data without any expansion.
    Thus, it   preserves % retaining  %its internal structural information well
    the intrinsic structure of the original
   tensor data very well,
   and adequately %fully
   characterizes %exploits
   its %spatial %global and local
   spatial-spectral
   correlation.

   %GNLSTR preserves the intrinsic structure of the original
   %tensor data, and fully explores
\end{Remark}

\vspace{-0.3cm}
%\vspace{-0.36cm}
\subsubsection{\textbf{Optimization Algorithm}}\label{optalgadmm}
%\vspace{-0.2cm}

In this subsection, the  ADMM framework  \cite{boyd2011distributed} is adopted to solve the proposed  models (\ref{orin_nonconvex})
and %model
(\ref{equ_nonconvexhtc}).
Given the significant similarities %in the ADMM optimization process between the two models,
 in the ADMM optimization procedures between the two models,
we will
only provide the detailed steps for the former.
The nonconvex model (\ref{orin_nonconvex})  can be equivalently reformulated as follows:
\begin{align}
& \min_{{\boldsymbol{\mathcal{L}}},{\boldsymbol{\mathcal{E}}}, {\boldsymbol{\mathcal{G}}}_{k}}
\frac{1}{\gamma} \sum_{k\in \Gamma}
 %\Phi ({\boldsymbol{\mathcal{G}}}_{k} )
 \big\| {\boldsymbol{\mathcal{G}}}_{k} \big\|_{\Phi,\mathfrak{L} }
 +
 {\lambda}  \Upsilon \big(\boldsymbol{\bm{P}}_{{{\Omega}}}({\boldsymbol{\mathcal{E}}})\big),
\notag \\ \label{equ_nonconvex}
&\text{s.t.} \;\; {\boldsymbol{\mathcal{G}}}_{k}=\nabla_{k}({\boldsymbol{\mathcal{L}}}), \;\;
{\boldsymbol{\mathcal{L}}}
+{\boldsymbol{\mathcal{E}}}=
{\boldsymbol{\mathcal{M}}},
\end{align}

The  augmented Lagrangian function of (\ref{equ_nonconvex}) is
\begin{align}
&
\mathcal{F}(
{\boldsymbol{\mathcal{L}}},  \{ {\boldsymbol{\mathcal{G}}}_{k} , k \in \Gamma\},
 {\boldsymbol{\mathcal{E}}},
\{ \Lambda_{k} , k \in \Gamma\},
{\boldsymbol{\mathcal{Y}}})
 =
 \sum_{k\in \Gamma} \Big (
\frac{1}{\gamma}
 %\Phi ({\boldsymbol{\mathcal{G}}}_{k} )
 \big\| {\boldsymbol{\mathcal{G}}}_{k} \big\|_{\Phi,\mathfrak{L} }
 +
 \notag \\
 &
 \langle   \Lambda_{k} ,  \nabla_{k}({\boldsymbol{\mathcal{L}}}) - {\boldsymbol{\mathcal{G}}}_{k}\rangle +
\frac {\mu}{2}
\| \nabla_{k}({\boldsymbol{\mathcal{L}}}) - {\boldsymbol{\mathcal{G}}}_{k}\|^{2}_{{{\mathnormal{F}}}}\Big )
+ {\lambda}  \Upsilon \big(\boldsymbol{\bm{P}}_{{{\Omega}}}({\boldsymbol{\mathcal{E}}})\big)+
 \notag \\ \label {lag-tttt}
&\langle {\boldsymbol{\mathcal{Y}}}, %{{\boldsymbol{\mathcal{L}}}+{\boldsymbol{\mathcal{E}}}-\boldsymbol{\mathcal{M}}}
{\boldsymbol{\mathcal{M}}} -  %{\boldsymbol{\mathcal{D}}}{*}_{\mathfrak{L}}
{\boldsymbol{\mathcal{L}}}
-{\boldsymbol{\mathcal{E}}}
\rangle +
\frac {\mu}{2}
\| % {\boldsymbol{\mathcal{L}}}+{\boldsymbol{\mathcal{E}}} -{\boldsymbol{\mathcal{M}}}
{\boldsymbol{\mathcal{M}}} -  %{\boldsymbol{\mathcal{D}}}{*}_{\mathfrak{L}}
{\boldsymbol{\mathcal{L}}} -{\boldsymbol{\mathcal{E}}}
\|^{2}_{{{\mathnormal{F}}}},
\end{align}
where  $\mu$ is the regularization  parameter, and
$\{ \Lambda_{k} , k \in \Gamma\}$,  ${\boldsymbol{\mathcal{Y}}}$ are Lagrange multipliers.
It can be further expressed
as
\begin{align}
&
\mathcal{F}(
{\boldsymbol{\mathcal{L}}},  \{ {\boldsymbol{\mathcal{G}}}_{k} , k \in \Gamma\},
{\boldsymbol{\mathcal{E}}},
\{ \Lambda_{k} , k \in \Gamma\},
 {\boldsymbol{\mathcal{Y}}})
 %&
 =
{\lambda}  \Upsilon \big(\boldsymbol{\bm{P}}_{{{\Omega}}}({\boldsymbol{\mathcal{E}}})\big)+
 \notag \\
 &
 \sum_{k\in \Gamma} \Big(
 \frac
 {1} {\gamma}
  %\Phi ({\boldsymbol{\mathcal{G}}}_{k} )
  \big\| {\boldsymbol{\mathcal{G}}}_{k} \big\|_{\Phi,\mathfrak{L} }
  +
  {\mu}/ {2}
 \|    \nabla_{k}({\boldsymbol{\mathcal{L}}}) - {\boldsymbol{\mathcal{G}}}_{k}+
 {  \Lambda_{k} }  / {\mu}
 \|^{2}_{{{\mathnormal{F}}}} \Big ) +
 \notag \\ \label {lag-tttt11RRR}
&\frac {\mu}{2}
\| % {\boldsymbol{\mathcal{L}}}+{\boldsymbol{\mathcal{E}}} -{\boldsymbol{\mathcal{M}}}
{\boldsymbol{\mathcal{M}}} -  %{\boldsymbol{\mathcal{D}}}{*}_{\mathfrak{L}}
{\boldsymbol{\mathcal{L}}}
-{\boldsymbol{\mathcal{E}}}
+ %\frac
{{\boldsymbol{\mathcal{Y}}}}  /{\mu}
\|^{2}_{{{\mathnormal{F}}}} +C,
\end{align}
%%%%%%%%%%%%%%%%%%%%%%%%%%%%%%%%%%%%%%%%%%%%%%%%%%%%%%%%%%%%%%%%%%%%%%%%%%%%%%%%%%%%%%%%%%%%%%%%%%%%%%%%%%%%%%%%%%%%%%%%%%%%%%%%%%%%%%%%%%%
%
where  $C$ is only the multipliers dependent squared items. Below, we show how to solve the subproblems for each involved variable.

\noindent {\textbf{Update ${ {\boldsymbol{\mathcal{L}}}^{\{t+1\}}}$
(low-rank component)}}
Taking the derivative in (\ref {lag-tttt11RRR}) with respect to $\boldsymbol{\mathcal{L}}$,
it gets the following linear system:
\begin{align}
&
\Big (  {\boldsymbol{\mathcal{I}}}
+\sum_{k\in \Gamma}  \nabla_{k}^{\mit{T}}  \nabla_{k} \Big) (\boldsymbol{\mathcal{L}})
=
 %{{\boldsymbol{\mathcal{D}}}}^{\mit{T}} {*}_{\mathfrak{L}}
({\boldsymbol{\mathcal{M}}} %-  {\boldsymbol{\mathcal{D}}}{*}_{\mathfrak{L}} {\boldsymbol{\mathcal{Z}}}
-{\boldsymbol{\mathcal{E}}}^{\{t\}}   + \frac
 {  {{\boldsymbol{\mathcal{Y}}}}^{\{t\}} }   {\mu^{\{t\}}})
 +
\sum_{k\in \Gamma}
\notag \\ \label {coef-tensor}
&
\nabla_{k}^{\mit{T}} \big({{\boldsymbol{\mathcal{G}}}}_{k}^{\{t\}}- %\frac
{  \Lambda_{k}^{\{t\}} } /  {\mu^{\{t\}}} \big),
\end{align}
where $\nabla_{k}^{\mit{T}}(\cdot)$ denotes the  transpose operator of $\nabla_{k}(\cdot) $.
Note that the difference operation on tensors has been proved to be linear via %t-product %
%tensor-tensor product
high-order t-product
\cite{wang2023guaranteed}.
Referring to the literature \cite{wang2008new},
we can apply multi-dimensional FFT, which diagonalizes $\nabla_{k}(\cdot)$'s corresponding difference tensors
$  {{\boldsymbol{\mathcal{D}}}}_{k}$, enabling to efficiently obtain the optimal solution of (\ref {coef-tensor})
in virtue of convolution theorem of Fourier transforms, i.e.,
\begin{align} \label{gnrhtc-LLLLLLwwwww}
{\boldsymbol{\mathcal{L}}}^{\{t+1\}}
&=\mathscr{F}^{-1}
\bigg(
\frac
{
\mathscr{F} ({\boldsymbol{\mathcal{M}}}- {\boldsymbol{\mathcal{E}}}^{\{t\}} + \frac{{\boldsymbol{\mathcal{Y}}}^{\{t\}}}  {\mu^{\{t\}}})
+  {\boldsymbol{\mathcal{T}}}
}
{\textbf{1} + \sum_{k \in \Gamma}  \mathscr{F}( {{\boldsymbol{\mathcal{D}}}}_{k})^{\mit{H}}  \odot \mathscr{F}( {{\boldsymbol{\mathcal{D}}}}_{k})    }
\bigg),
\end{align}
where
%%%%%%%%%%
${\boldsymbol{\mathcal{T}}}=  \sum_{k \in \Gamma }   \mathscr{F}( {{\boldsymbol{\mathcal{D}}}}_{k})^{\mit{H}}
\odot \mathscr{F}(
{\boldsymbol{\mathcal{G}}}_{k}^{\{t\}}
- \frac {  \Lambda_{k}^{\{t\}} }   {\mu^{\{t\}}}
)
$,
$\textbf{1} $ is a tensor with all elements as $1$,
$ \odot$ is componentwise multiplication, and the division  is componentwise as well.

\noindent {\textbf{Update ${ {\boldsymbol{\mathcal{G}}}_{k}^{\{t+1\}}}, k \in \Gamma$ (gradient  component)}}
For each $k \in \Gamma$, extracting all items containing ${\boldsymbol{\mathcal{G}}}_{k}$ from
(\ref {lag-tttt}), we can get that
%The  optimization subproblem  (\ref{L_hat}) concerning  ${\boldsymbol{\mathcal{L}}^{\psi+1}}$ can  be written as
\begin{align}\label{L_prox}
\textcolor[rgb]{0.00,0.00,0.00}{
%\arg
\min_{{\boldsymbol{\mathcal{G}}}_{k}}
\frac{1}{\gamma} %\sum_{k\in \Gamma}
 %%\Phi ({\boldsymbol{\mathcal{G}}}_{k} )
  \big\| {\boldsymbol{\mathcal{G}}}_{k} \big\|_{\Phi,\mathfrak{L} }
+
\frac {\mu^{\{t\}}} {2}
\big\|
\nabla_{k}({\boldsymbol{\mathcal{L}}}^{\{t+1\}}) - {\boldsymbol{\mathcal{G}}}_{k}
+
\frac {  \Lambda_{k}^{\{t\}} }   {\mu^{\{t\}}}
\big\|^2_{\mathnormal{F}}.}
\end{align}
%%%%%%%%%%%%%%%%%%%%%%%%%%%%%%%%%%%%%%%%%%%%%%%%%%%%%%%%%%%%%%%%%%%%%%%%%%%%%%%%
This key subproblem is analogous to  the form of  the minimization problem  (\ref{LS_prox1}),
and its  close-form solution can be efficiently obtained by utilizing  the fast randomized  GNHTSVT operation. Please see  \ref{gnlstr-oper} for details.

\noindent
\textcolor[rgb]{0.00,0.00,0.00}{
{\textbf{Update ${\boldsymbol{\mathcal{E}}^{\{t+1\}}}$ (noise/outliers component)}}}
\textcolor[rgb]{0.00,0.00,0.00}{
The optimization subproblem %(\ref{E_hat})
with respect to $\boldsymbol{\mathcal{E}}$ %
can  be written as}
\begin{align*}
\textcolor[rgb]{0.00,0.00,0.00}{
\min_{
\boldsymbol{\mathcal{E}}}
{\lambda}  \Upsilon \big(\boldsymbol{\bm{P}}_{{{\Omega}}}({\boldsymbol{\mathcal{E}}})\big)
 +
\frac {\mu^{\{t\}}}{2}
\big\|
{\boldsymbol{\mathcal{E}}}-  \big({\boldsymbol{\mathcal{M}}}-
%{\boldsymbol{\mathcal{D}}}{*}_{\mathfrak{L}}
 {\boldsymbol{\mathcal{L}}}^{\{t+1\}}
+
%\frac{{\boldsymbol{\mathcal{Y}}}^{k}}  {\beta^k}
\frac{{\boldsymbol{\mathcal{Y}}}^{\{t\}}}  {\mu^{\{t\}}}\big)
\big\|^2_{\mathnormal{F}}.}
\end{align*}
%%%%%%%%%%%%%%%%%%%%%%%%%%%%%%%%%%%%%%%%%%%%%%%%%%%%%%%%%%%%%%%%%%%%%%%%%%%%%%%%%%%%%%%%%%%%%%%%%%%%%%
%
%%%%%%%%%%%%%%%%%%%%%%%%%%%%%%%%%%%%%%%%%%%%%%%%%%%%%%%%%%%%%%%%%%%%%%%%%%%%%%%%%%%%%%%%%%%%%%%%%%%%%
\textcolor[rgb]{0.00,0.00,0.00}{
Let ${\boldsymbol{\mathcal{H}}}^{\{t\}}={\boldsymbol{\mathcal{M}}}-
%{\boldsymbol{\mathcal{D}}}{*}_{\mathfrak{L}}
{\boldsymbol{\mathcal{L}}}^{\{t+1\}}+
%\frac{{\boldsymbol{\mathcal{Y}}}^{k}}  {\beta^k}
\frac{{\boldsymbol{\mathcal{Y}}}^{\{t\}}}  {\mu^{\{t\}}}$.
The above problem  can be solved by the following two subproblems with respect to $\boldsymbol{\bm{P}}_{{{\Omega}}}({\boldsymbol{\mathcal{E}}}^{\{t+1\}})$ and $\boldsymbol{\bm{P}}_{{{\Omega}}_{\bot}}({\boldsymbol{\mathcal{E}}}^{\{t+1\}})$, respectively.}

%
  %%%%%%%%%%%% algorithm4%%%%%%%%%%%%%%%%%%%%%%%%%%%%%%%%%%%%%%%%%%%%%%%%
\begin{algorithm}[!htbp]
\setstretch{0.3} %设置具有指定弹力的橡皮长度（原行宽的1.35 倍）
\caption{
%Solve the proposed model (\ref{orin_nonconvex}) by ADMM.
%GNLSTR for RHTC
 Solve GNRHTC model (\ref{orin_nonconvex}) via ADMM.
}\label{algorithm1}

 \KwIn{
$\boldsymbol{\bm{P}}_{{{\Omega}}}({\boldsymbol{\mathcal{M}}}) \in \mathbb{R}^{n_1 \times \cdots \times n_d}$,
$\mathfrak{L}$, $ \Phi(\cdot),  \psi(\cdot)$, $h(\cdot)$,  $\Gamma$, % as a priori set
 $\lambda$, %$\tau> 0$,
 $\bm{\rho}$, $\bm{b}$, $\bm{q}$,
 $\epsilon$, $b$, $\delta$.
 % target T-SVD Rank: $k$,
%        %
%  block size: $b$,
%  power iteration: $t$,
%  $0<p,q<1$,
%  $c_1, c_2, {\epsilon}_1, {\epsilon}_2$.
  }

 \textbf{Initialize:} $ { {\boldsymbol{\mathcal{G}}}_{k}^{\{0\}}}=  %{\boldsymbol{\mathcal{L}}}^{0}=
 {\boldsymbol{\mathcal{E}}}^{\{0\}}
 =
 \Lambda_{k} ^{\{0\}}=
 {\boldsymbol{\mathcal{Y}}}^{\{0\}}
 =
 \boldsymbol{0}$,
 $\vartheta$,
%$\beta^0=10^{-3}$,
$\mu^{\{0\}}$,
%$\beta^{\max}=10^{8}$,
$\mu^{\max}$,
$\varpi$,
$t=0$\;

\While{\text{not converged}}
{

Update ${ {\boldsymbol{\mathcal{L}}}^{\{t+1\}}}$  by computing (\ref {coef-tensor})
;\\
 \textcolor[rgb]{0.00,0.00,0.00}{Update $ { {\boldsymbol{\mathcal{G}}}_{k}^{\{t+1\}}}$ by computing
 (\ref{L_prox}) for each $k \in  \Gamma$;} \\

  \textcolor[rgb]{0.00,0.00,0.00}{Update $\boldsymbol{\bm{P}}_{{{\Omega}}}({\boldsymbol{\mathcal{E}}}^{\{t+1\}})$ by computing (\ref{E_prox_ome});}\\

     \textcolor[rgb]{0.00,0.00,0.00}{Update $\boldsymbol{\bm{P}}_{{{\Omega}}_{\bot}}({\boldsymbol{\mathcal{E}}}^{\{t+1\}})$ by computing (\ref{E_prox_ome1});}\\

 \textcolor[rgb]{0.00,0.00,0.00}{Update   $\Lambda_{k} ^{\{t+1\}}, {\boldsymbol{\mathcal{Y}}}^{\{t+1\}}, \mu^{\{t+1\}}$
 by computing
 (\ref{lagrange11})-(\ref{lagrange22});}\\  %\label{lagrange11}

 Check the convergence condition
%\begin{align*}
%&\textcolor[rgb]{0.00,0.00,0.00}{\|{\boldsymbol{\mathcal{L}}}^{\{t+1\}}-{\boldsymbol{\mathcal{L}}}^{\{t\}}\|{_{\infty}}
% \leq \varpi,
%%\\
%\;
%%&
%\|{\boldsymbol{\mathcal{E}}}^{\{t+1\}}-{\boldsymbol{\mathcal{E}}}^{\{t\}}\|{_{\infty}}  \leq \varpi,}\\
%&\textcolor[rgb]{0.00,0.00,0.00}{\|
%% \boldsymbol{\boldsymbol{\mathcal{L}}}^{\psi+1}+ {\boldsymbol{\mathcal{E}}}^{\psi+1}- {\boldsymbol{\mathcal{M}}}
%{\boldsymbol{\mathcal{M}}} - % {\boldsymbol{\mathcal{D}}}{*}_{\mathfrak{L}}
% {\boldsymbol{\mathcal{L}}}^{\{t+1\}} -{\boldsymbol{\mathcal{E}}}^{\{t+1\}}
%\|{_{\infty}}
%   \leq \varpi.}
%%
%\end{align*}
by computing (\ref{convercon}).
}
{\color{black}\KwOut{
  ${\boldsymbol{\mathcal{L}}}
  \in \mathbb{R}^{n_1 \times \cdots \times n_d}$.
   %and ${\boldsymbol{\mathcal{E}}}  \in \mathbb{R}^{n_1 \times \cdots \times n_d}$.
  }}
\end{algorithm}
%%%%%%%%%%%%%%%%%%%%%%%%%%%%%%%%%%%%%%%%

\noindent
\textcolor[rgb]{0.00,0.00,0.00}{
 \textbf{
 Regarding $\boldsymbol{\bm{P}}_{{{\Omega}}}({\boldsymbol{\mathcal{E}}}^{\{t+1\}})$:}}
\textcolor[rgb]{0.00,0.00,0.00}{the optimization subproblem  with respect to $\boldsymbol{\bm{P}}_{{{\Omega}}}({\boldsymbol{\mathcal{E}}}^{\{t+1\}})$
 is formulated as following}
\begin{align}\label{E_prox_ome}
\textcolor[rgb]{0.00,0.00,0.00}{
\min_{
\boldsymbol{\bm{P}}_{{{\Omega}}}(\boldsymbol{\mathcal{E}})
} %\lambda \|\boldsymbol{\mathcal{E}}\|_{q}^{q}
%\lambda\|\boldsymbol{\bm{P}}_{{{\Omega}}}({\boldsymbol{\mathcal{E}}})\|_{{{\boldsymbol{\mathcal{W}}}_{2}}^{\psi},\ell_q}^{q}
{\lambda}  \Upsilon \big(\boldsymbol{\bm{P}}_{{{\Omega}}}({\boldsymbol{\mathcal{E}}})\big)
 +
%\frac
\frac{\mu^{\{t\}}}{2}
%{\beta^k/2}
\big\| \boldsymbol{\bm{P}}_{{{\Omega}}}(
{\boldsymbol{\mathcal{E}}}-{\boldsymbol{\mathcal{H}}}^{\{t\}}
)
\big\|^2_{\mathnormal{F}}.}
\end{align}
\noindent
This  subproblem is equivalent to the form of the minimization problem (\ref{E_prox1}),
and its close-form solution can be obtained by utilizing the %WTST operation.
 generalized nonconvex shrinkage operator.  Please see  \ref {gnlstr-oper} for details.

%\item
\noindent
\textcolor[rgb]{0.00,0.00,0.00}{
  \textbf{%(II)
   Regarding
$\boldsymbol{\bm{P}}_{{{\Omega}}_{\bot}}({\boldsymbol{\mathcal{E}}}^{\{t+1\}})$:}}
\textcolor[rgb]{0.00,0.00,0.00}{the optimization subproblem
 with respect to $\boldsymbol{\bm{P}}_{{{\Omega}}_{\bot}}({\boldsymbol{\mathcal{E}}}^{\{t+1\}})$
 is formulated as following}
 \begin{align}\label{E_prox_ome1}
 \textcolor[rgb]{0.00,0.00,0.00}{
\boldsymbol{\mathrm{P}}_{{{\Omega}}_{\bot}}({\boldsymbol{\mathcal{E}}}^{\{t+1\}})=
\min_{
{\boldsymbol{\bm{P}}}_{{{\Omega}}_{\bot}} (\boldsymbol{\mathcal{E}})
}
\frac{\mu^{\{t\}}}{2}
%{\beta^\psi/2}
\big\| \boldsymbol{\bm{P}}_{{{\Omega}}_{\bot}}(
{\boldsymbol{\mathcal{E}}}-{\boldsymbol{\mathcal{H}}}^{\{t\}}
)
\big\|^2_{\mathnormal{F}}.}
\end{align}
\noindent
The  closed-form solution for  subproblem
(\ref{E_prox_ome1})  can be obtained through the standard least square regression method.

\noindent
{\textbf{Update $\Lambda_{k} ^{\{t+1\}}, k \in \Gamma $,   ${\boldsymbol{\mathcal{Y}}}^{\{t+1\}}$,  and
$\mu^{\{t+1\}}$  (Lagrange multipliers and %step size
 penalty parameter)}}
Based on the rule of ADMM framework,
the lagrange multipliers are updated by the following equations:
\begin{align}\label{lagrange11}
{\boldsymbol{\mathcal{Y}}}^{\{t+1\}}
&=
{\boldsymbol{\mathcal{Y}}}^{\{t\}}+
\mu^{\{t\}}
({\boldsymbol{\mathcal{M}}}-  %{\boldsymbol{\mathcal{D}}}{*}_{\mathfrak{L}}
{\boldsymbol{\mathcal{L}}}^ {\{t+1\}}-{\boldsymbol{\mathcal{E}}}^{\{t+1\}}),
\\
{\Lambda_{k}}^{\{t+1\}} &={\Lambda}_{k}^{\{t\}}+
\mu^{\{t\}} \big(
\nabla_{k}({\boldsymbol{\mathcal{L}}}^{\{t+1\}})- {\boldsymbol{\mathcal{G}}}^{\{t+1\}}_{k}
%\nabla_{k}({\boldsymbol{\mathcal{Z}}}^{t+1})
 \big ),
\\ \label{lagrange22}
\mu^{\{t+1\}}
&= \min \left( \mu^{\operatorname{max}},\vartheta \mu^{\{t\}} \right),
\end{align}
\noindent where $\vartheta$ stands for the growth rate. %   is usually set to $1.15$. %a control constant.
%It has been demonstrated that the update scheme of $\vartheta$  can accelerate the convergence of ADMM algorithm.

\noindent
\textbf{Check the convergence condition:} The convergence condition is defined as follows:
\begin{equation} \label{convercon}
%\text{Error}=
\max
\left\{
 \begin{aligned} %
&
\textcolor[rgb]{0.00,0.00,0.00}{\|{\boldsymbol{\mathcal{L}}}^{\{t+1\}}-{\boldsymbol{\mathcal{L}}}^{\{t\}}\|{_{\infty}} },
\|{\boldsymbol{\mathcal{E}}}^{\{t+1\}}-{\boldsymbol{\mathcal{E}}}^{\{t\}}\|{_{\infty}}
\\
&
\|
\nabla_{k}  {\boldsymbol{\mathcal{L}}}^{\{t+1\}}- {\boldsymbol{\mathcal{G}}}^{\{t+1\}}_{k}
\|_{{\infty}} ,
\| {\boldsymbol{\mathcal{G}}}_{k} ^{\{t+1\}}-{\boldsymbol{\mathcal{G}}}_{k} ^{\{t\}}\|_{{\infty}}
\\
&
\|
% \boldsymbol{\boldsymbol{\mathcal{L}}}^{\psi+1}+ {\boldsymbol{\mathcal{E}}}^{\psi+1}- {\boldsymbol{\mathcal{M}}}
{\boldsymbol{\mathcal{M}}} - % {\boldsymbol{\mathcal{D}}}{*}_{\mathfrak{L}}
 {\boldsymbol{\mathcal{L}}}^{\{t+1\}} -{\boldsymbol{\mathcal{E}}}^{\{t+1\}}
\|_{{\infty}} %%\\
\end{aligned}
\right\}
%\\
\leq  \varpi.
\end{equation}

The whole ADMM optimization scheme is summarized in Algorithm \ref{algorithm1}.
%It is worth
 Noting that the time complexity analysis of the proposed algorithms is provided in the supplementary material.

%\vspace{-0.14cm}

\subsubsection{\textbf{Convergence Analysis}} \label{converage}

In this subsection, we provide the convergence analysis of the proposed algorithm.
% The main results are given in
%Theorem \ref{conver1111} and  Theorem \ref{conver2222}  below. %Before  that, we present some lemmas.
The detailed proofs of relevant theories and lemmas can be found in the Supplementary Material.

\begin{Lemma}\label{yy1}
The sequences
%{Lk }, {Ek }, {Mk }
%$\{{\boldsymbol{\mathcal{L}}}^{k+1}\}$, $\{{\boldsymbol{\mathcal{E}}}^{k+1}\}$ and
$\big \{{\boldsymbol{\mathcal{Y}}}^{\{t\}} \big \}$ and  $\big \{ \Lambda_{k} ^{\{t\}}, k \in \Gamma \big \}$
generated by \textcolor[rgb]{0.00,0.00,0.00}{\text{Algorithm}} \ref{algorithm1}
are bounded.
\end{Lemma}

%
%\begin{Lemma}\label{eee222}
%The sequences
%%
%%$\{{\boldsymbol{\mathcal{L}}}^{k}\}$ and $\{{\boldsymbol{\mathcal{E}}}^{k}\}$
%%
%$ \big \{     {\boldsymbol{\mathcal{L}}}^{\{t\}},  % {\boldsymbol{\mathcal{L}}},
%\{ {\boldsymbol{\mathcal{G}}}_{k}^{\{t\}}, k \in \Gamma\},
%%\{ \Lambda_{k} ^{\{t\}}, k \in \Gamma\},
% {\boldsymbol{\mathcal{E}}}^{\{t\}}  \big \}$
%generated by \textcolor[rgb]{0.00,0.00,0.00}{\text{Algorithm}} \ref{algorithm1}
% are bounded.
%\end{Lemma}

\begin{Lemma}\label{yy15559999999}
Suppose that the   sequences
$\big \{{\boldsymbol{\mathcal{Y}}}^{\{t\}} \big\}$ and $\big\{ \Lambda_{k} ^{\{t\}}, k \in \Gamma \big \}$
generated by \textcolor[rgb]{0.00,0.00,0.00}{\text{Algorithm}}  \ref{algorithm1} %\textbf{GNRHTC}
are   bounded,
then
the sequence
$ \big \{     {\boldsymbol{\mathcal{L}}}^{\{t\}} ,   %\big\}$,  % {\boldsymbol{\mathcal{L}}},
%$\big
\{ {\boldsymbol{\mathcal{G}}}_{k}^{\{t\}}, k \in \Gamma  \},
   {\boldsymbol{\mathcal{E}}}^{\{t\}} \big\}$
 %
  % \{ \Lambda_{k} ^{\{t\}}, k \in \Gamma\}, {\boldsymbol{\mathcal{Y}}}^{\{t\}}
 %  \Big \}$
%generated by \textcolor[rgb]{0.00,0.00,0.00}{\text{Algorithm}} \ref{algorithm1}
 is bounded.
\end{Lemma}

\begin{Theorem}\label{conver2222} %\label{conver}
\textcolor[rgb]{0.00,0.00,0.00}{
Suppose that
the sequence
%{Lk }, {Ek }, {Mk }
%$\{{\boldsymbol{\mathcal{L}}}^{k+1}\}$, $\{{\boldsymbol{\mathcal{E}}}^{k+1}\}$ and
$\big \{{\boldsymbol{\mathcal{Y}}}^{\{t\}} ,    \Lambda_{k} ^{\{t\}}, k \in \Gamma \big \}$
%,
%generated by \textcolor[rgb]{0.00,0.00,0.00}{\text{Algorithm}} \ref{algorithm1}
generated by Algorithm \ref{algorithm1}
is bounded.
Then, the  sequences $\{{\boldsymbol{\mathcal{L}}}^{\{t+1\}}\}$,
$\{ {\boldsymbol{\mathcal{G}}}_{k}^{\{t+1\}}, k \in \Gamma\}$, and
$\{{\boldsymbol{\mathcal{E}}}^{\{t+1\}}\}$ % and $\{{\boldsymbol{\mathcal{Y}}}^{\{t+1\}}\}$
satisfy:}
\begin{align*}
& \textcolor[rgb]{0.00,0.00,0.00}{ 1)\lim_{t \rightarrow \infty}{
\|{\boldsymbol{\mathcal{M}}}-{\boldsymbol{\mathcal{L}}}^{\{t+1\}}-{\boldsymbol{\mathcal{E}}}^{\{t+1\}}\|{_{\mathnormal{F}}}}=0;}
 \\
& 2) \lim_{t  \rightarrow \infty}
{\|
\nabla_{k}({\boldsymbol{\mathcal{L}}}^{\{t+1\}})- {\boldsymbol{\mathcal{G}}}^{\{t+1\}}_{k}
\|{_{\mathnormal{F}}}
} =0, \;\; k \in \Gamma;
\\
&
\textcolor[rgb]{0.00,0.00,0.00}
{3)\lim_{t   \rightarrow \infty} {\|{\boldsymbol{\mathcal{L}}}^{\{t+1\}}-{\boldsymbol{\mathcal{L}}}^{\{t\}}\|{_{\mathnormal{F}}}}
%=0;
=
}
%\\
%&
%2)
\lim_{t  \rightarrow \infty} { \|{\boldsymbol{\mathcal{E}}}^{\{t+1\}}-{\boldsymbol{\mathcal{E}}}^{\{t\}}\| {_{\mathnormal{F}}}=0;}
\\
& \textcolor[rgb]{0.00,0.00,0.00}{ 4)\lim_{t \rightarrow \infty}{
\| {\boldsymbol{\mathcal{G}}}_{k} ^{\{t+1\}}-{\boldsymbol{\mathcal{G}}}_{k} ^{\{t\}}\|{_{\mathnormal{F}}}}=0;
\;\; k \in \Gamma.}
%\\
%
\end{align*}
\end{Theorem}

\begin{Theorem} \label{conver1111}
Let
$ \big \{
{\boldsymbol{\mathcal{L}}}^{\{t\}},  % {\boldsymbol{\mathcal{L}}},
\{ {\boldsymbol{\mathcal{G}}}_{k}^{\{t\}}, k \in \Gamma\},
 {\boldsymbol{\mathcal{E}}}^{\{t\}} ,
\{ \Lambda_{k} ^{\{t\}}, k \in \Gamma\},
  {\boldsymbol{\mathcal{Y}}}^{\{t\}}  \big \}$ be  a sequence
generated by \textcolor[rgb]{0.00,0.00,0.00}{\text{Algorithm}} \ref{algorithm1}.
Suppose that
the sequences
%{Lk }, {Ek }, {Mk }
%$\{{\boldsymbol{\mathcal{L}}}^{k+1}\}$, $\{{\boldsymbol{\mathcal{E}}}^{k+1}\}$ and
$\big \{{\boldsymbol{\mathcal{Y}}}^{\{t\}} \big\}$ and  $\big\{  \Lambda_{k} ^{\{t\}}, k \in \Gamma \big \}$
%,
%generated by \textcolor[rgb]{0.00,0.00,0.00}{\text{Algorithm}} \ref{algorithm1}
are bounded.
Then,  any accumulation point of %the
the sequence
$ \big \{     {\boldsymbol{\mathcal{L}}}^{\{t\}},  % {\boldsymbol{\mathcal{L}}},
\{ {\boldsymbol{\mathcal{G}}}_{k}^{\{t\}}, k \in \Gamma\},   {\boldsymbol{\mathcal{E}}}^{\{t\}} ,
\{ \Lambda_{k} ^{\{t\}}, k \in \Gamma\},
 {\boldsymbol{\mathcal{Y}}}^{\{t\}}  \big \}$
 is a \textit{Karush-Kuhn-Tucker} (KKT) point of the optimization problem (\ref{equ_nonconvex}).
\end{Theorem}

\vspace{-0.5700605cm}

\subsection{\textbf{Quantized Tensor Recovery}}
%This subsection
In this subsection, we
consider  %applying the proposed modeling and computing strategies to % one %
two
 typical quantized
tensor recovery task, i.e., \textit{One-Bit High-Order Tensor Completion} (OBHTC) and
 \textit{One-Bit Robust High-Order Tensor Completion} (OBRHTC).
% 1) Noisy tensor completion; 2) Noise-
%free tensor completion.

\subsubsection{\textbf{Generalized Nonconvex Model}}
%In this subsection, we
%This subsection
%
A %a %proposes
novel %nonconvex
OBHTC model is investigate  from  uniformly dithered binary %one-bit
 observations.
Specifically, let $y_k$ ($\forall k \in [m]$) be the $k$-th noisy observation from the low-rank tensor
${\boldsymbol{\mathcal{L}}} ^{*}  \in\mathbb{R}^{n_1\times \cdots\times n_d}$ via
$ %$
y_k  =  \langle{\boldsymbol{\mathcal{P}}}_{k},{\boldsymbol{\mathcal{L}}} ^{*}  \rangle + \epsilon _k %, %\xi_k,
$,  %$
where ${\boldsymbol{\mathcal{P}}}_{k}$ is the sampler that uniformly %distributed on the
and randomly extracts one entry of ${\boldsymbol{\mathcal{L}}} ^{*}$, and $\epsilon _k$ denotes additive noise independent of
${\boldsymbol{\mathcal{P}}}_{k}$. %Furthermore,
Then,  we propose to quantize $y_k$ to
$$
q_k=\operatorname{sign} (y_k + \xi_k),
$$
where all dithers $\xi_k,    k \in \{1,2, \cdots, m\}$ are uniformly distributed on the interval $[-\theta, \theta] $ with some $\theta>0$.
Based on Proposition  \ref{one-bitequ},  we can utilized $\theta q_k$   as a substitute for the full observation $q_k$.
Consequently, we propose the following regularized OBHTC model:
\begin{align} \label{1bittensor}
&  \hat {\boldsymbol{\mathcal{L}}}=
  \arg \min _    {   \|  \boldsymbol{\mathcal{L}} \|_{\infty} \leq \alpha %  \boldsymbol{\mathcal{L}}
   }
\frac{1} {2m} \sum_{k=1} ^{m} \big(
 \langle{\boldsymbol{\mathcal{P}}}_{k},{\boldsymbol{\mathcal{L}}}  \rangle - \theta q_k \big)^2
%\Big )
+ \lambda
%\frac{1}{\gamma} \sum_{k\in \Gamma}
%    %\Phi ({\boldsymbol{\mathcal{G}}}_{k} )
 %\big\| {\boldsymbol{\mathcal{G}}}_{k} \big\|_{\Phi,\mathfrak{L} },
 \Psi( {\boldsymbol{\mathcal{L}}}),
% \notag \\ \label{equ_nonconvexhtc_1btc}
%&
%\text{s.t.} \;\;
% {\boldsymbol{\mathcal{G}}}_{k}=\nabla_{k}({\boldsymbol{\mathcal{L}}}), \|  \boldsymbol{\mathcal{L}} \|_{\infty} <0 \;\;
\end{align}
where
$\lambda $ is the regularization parameter,
%\|  \boldsymbol{\mathcal{L}} \|_{\infty} <0
$\Psi( \cdot)$
%Where the regular term 1
is formulated  according to %Formula
Definition \ref{gnhtctv}, $m$ denotes the number of observations, and the assumption
$\| \boldsymbol{\mathcal{L}} \|_{\infty} \leq \alpha$ ($\alpha>0$) helps make the recovery %of $\boldsymbol{\mathcal{L}}$
well-posed by preventing excessive ``spikiness".

%
%
%
  %%%%%%%%%%%% algorithm4%%%%%%%%%%%%%%%%%%%%%%%%%%%%%%%%%%%%%%%%%%%%%%%%
\begin{algorithm}[!htbp]
\setstretch{0.0} %设置具有指定弹力的橡皮长度（原行宽的1.35 倍）
\caption{
%Solve the proposed model (\ref{orin_nonconvex}) by ADMM.
%GNLSTR for RHTC
 Solve % GNRHTC model (\ref{orin_nonconvex})
 GNOBHTC model (\ref{1bittensor})
 via ADMM.
}\label{algorithm2222100000}

 \KwIn{
One-bit observations: $\{q_k\}_{\forall k \in [m]} $,
% $\boldsymbol{\bm{P}}_{{{\Omega}}}({\boldsymbol{\mathcal{M}}}) \in \mathbb{R}^{n_1 \times \cdots \times n_d}$,
uniform samplers: ${\boldsymbol{\mathcal{P}}}_{k} (\forall k \in [m]) $,
$\mathfrak{L}$, $ \Phi(\cdot)$,  %\psi(\cdot)$, $h(\cdot)$,
$\Gamma$, $\gamma := \sharp\{\Gamma\}$, % as a priori set
 $\lambda$, $\theta$, $\alpha$,  %$\tau> 0$,
 $\bm{\rho}$, $\bm{b}$, $\bm{q}$,
 $\epsilon$, $b$, $\delta$.
 % target T-SVD Rank: $k$,
%        %
%  block size: $b$,
%  power iteration: $t$,
%  $0<p,q<1$,
%  $c_1, c_2, {\epsilon}_1, {\epsilon}_2$.
  }

 \textbf{Initialize:} $ { {\boldsymbol{\mathcal{G}}}_{i}^{\{0\}}}=  %{\boldsymbol{\mathcal{L}}}^{0}=
 {\boldsymbol{\mathcal{Z}}}^{\{0\}}
 =
 \Lambda_{i} ^{\{0\}}=
 {\boldsymbol{\mathcal{Y}}}^{\{0\}}
 =
 \boldsymbol{0}$,
 $\vartheta=1.05$,
%$\beta^0=10^{-3}$,
$\mu^{\{0\}}=10^{-6}$,
%$\beta^{\max}=10^{8}$,
$\mu^{\max}=10^{3}$,
$\varpi$,
$t=0$\;

\While{\text{not converged}}
{

Update ${ {\boldsymbol{\mathcal{L}}}^{\{t+1\}}}$  by computing (\ref{1bit-L});
\\

Update ${ {\boldsymbol{\mathcal{Z}}}^{\{t+1\}}}$  by computing (\ref{1bit-Z});
\\

 \textcolor[rgb]{0.00,0.00,0.00}{Update $ { {\boldsymbol{\mathcal{G}}}_{i}^{\{t+1\}}}$ by computing}
 (\ref{1bit-g}) for each $i \in  \Gamma$;
  \\

 \textcolor[rgb]{0.00,0.00,0.00}{Update   $\Lambda_{i} ^{\{t+1\}}, {\boldsymbol{\mathcal{Y}}}^{\{t+1\}}, \mu^{\{t+1\}}$
 by %computing
 (\ref{1bit-Y})-(\ref{1bit-MU})};
 \\  %\label{lagrange11}

 Check the convergence condition
%\begin{align*}
%&\textcolor[rgb]{0.00,0.00,0.00}{\|{\boldsymbol{\mathcal{L}}}^{\{t+1\}}-{\boldsymbol{\mathcal{L}}}^{\{t\}}\|{_{\infty}}
% \leq \varpi,
%%\\
%\;
%%&
%\|{\boldsymbol{\mathcal{E}}}^{\{t+1\}}-{\boldsymbol{\mathcal{E}}}^{\{t\}}\|{_{\infty}}  \leq \varpi,}\\
%&\textcolor[rgb]{0.00,0.00,0.00}{\|
%% \boldsymbol{\boldsymbol{\mathcal{L}}}^{\psi+1}+ {\boldsymbol{\mathcal{E}}}^{\psi+1}- {\boldsymbol{\mathcal{M}}}
%{\boldsymbol{\mathcal{M}}} - % {\boldsymbol{\mathcal{D}}}{*}_{\mathfrak{L}}
% {\boldsymbol{\mathcal{L}}}^{\{t+1\}} -{\boldsymbol{\mathcal{E}}}^{\{t+1\}}
%\|{_{\infty}}
%   \leq \varpi.}
%%
%\end{align*}
by computing % (\ref{convercon}).
$$
\| {{\boldsymbol{\mathcal{L}}} }  ^{\{\nu+1\}}-
%{\boldsymbol{\mathcal{E}}}^{\{\nu\}}
 {{\boldsymbol{\mathcal{L}}} }   ^{\{\nu\}}
\|{_{\mathnormal{F}}} /
\|
%{\boldsymbol{\mathcal{E}}}^{\{\nu\}}
 {{\boldsymbol{\mathcal{L}}} }   ^{\{\nu\}}
\|{_{\mathnormal{F}}}
 \leq \varpi;
$$
}
{\color{black}\KwOut{
  $\hat{{\boldsymbol{\mathcal{L}}}}
  \in \mathbb{R}^{n_1 \times \cdots \times n_d}$.
   %and ${\boldsymbol{\mathcal{E}}}  \in \mathbb{R}^{n_1 \times \cdots \times n_d}$.
  }}
\end{algorithm}
%
%%%%%%%%%%%%%%%%%%%%%%%%%%%%%%%%%%%%%%%%%%%%%%%%%%%%%%%%%%%%%%%%%%%%%%%%%%%%

To handle the incomplete %missing
high-order tensor damaged by both Gaussian noise and sparse noise simultaneously, we can extend the Model (\ref{1bittensor}) to the following model:
\begin{align}  %
%&
& \hat {\boldsymbol{\mathcal{L}}}=
  \arg \min _    {   \|  \boldsymbol{\mathcal{L}} \|_{\infty} \leq \alpha %  \boldsymbol{\mathcal{L}}
   }
\frac{1} {2m} \sum_{k=1} ^{m} \big(
 \langle{\boldsymbol{\mathcal{P}}}_{k},{\boldsymbol{\mathcal{L}}} + {\boldsymbol{\mathcal{S}}}  \rangle - \theta \hat{q}_k \big)^2
%\Big )
+ \lambda_1
%\frac{1}{\gamma} \sum_{k\in \Gamma}
%    %\Phi ({\boldsymbol{\mathcal{G}}}_{k} )
 %\big\| {\boldsymbol{\mathcal{G}}}_{k} \big\|_{\Phi,\mathfrak{L} },
 \Psi( {\boldsymbol{\mathcal{L}}})
\notag \\  \label{1bittensor-gu-sp}  %
&  +
 {\lambda}_2  \Upsilon ({\boldsymbol{\mathcal{S}}}),
% \notag \\ \label{equ_nonconvexhtc_1btc}
%&
%\text{s.t.} \;\;
% {\boldsymbol{\mathcal{G}}}_{k}=\nabla_{k}({\boldsymbol{\mathcal{L}}}), \|  \boldsymbol{\mathcal{L}} \|_{\infty} <0 \;\;
\end{align}
%%%%%%%%%%%%%%%%%%%%%%%%%%%%%%%%%%%%%%%%%%%%%%%%%%%%%%%%%%%%%%%%%%%%%%%%%%%%%%%
%
where
$\lambda_1 $ and $\lambda_2 $ are  the regularization parameter, ${\boldsymbol{\mathcal{L}}}$ is an underlying tensor,
${\boldsymbol{\mathcal{S}}}$ represents sparse noise, and
$$
\hat{q}_k=\operatorname{sign} ( \langle{\boldsymbol{\mathcal{P}}}_{k},{\boldsymbol{\mathcal{L}}} + {\boldsymbol{\mathcal{S}}}   \rangle + \epsilon _k + \xi_k),
$$
in which
the definitions of ${\boldsymbol{\mathcal{P}}}_{k}$,  $ \epsilon _k$, and $\xi_k$ %other symbols
are similar to those of %the symbols in
Model (\ref{1bittensor}).
In this paper, % the model (\ref{1bittensor})
Model (\ref{1bittensor})  and  Model (\ref{1bittensor-gu-sp})
are  named as \textit{generalized nonconvex %1BHTC
OBHTC} (GNOBHTC) and \textit{generalized nonconvex OBRHTC}
(GNOBRHTC), respectively.
% Model (\ref{1bittensor})  and Model (\ref{1bittensor-gu-sp})

\subsubsection{\textbf{Optimization Algorithm}}
To facilitate problem-solving, we start by introducing %an auxiliary tensor
auxiliary variables to equivalently reshape model (\ref{1bittensor}) as
\begin{align}
%&  \hat {\boldsymbol{\mathcal{L}}}=
  %\arg
 &  \min _    {  \boldsymbol{\mathcal{L}}, {\boldsymbol{\mathcal{Z}}},  %\|  \boldsymbol{\mathcal{L}} \|_{\infty} <\alpha %  \boldsymbol{\mathcal{L}}
  {\boldsymbol{\mathcal{G}}}_{i} }
\frac{1} {2m} \sum_{k=1} ^{m} ( %\big(
 \langle{\boldsymbol{\mathcal{P}}}_{k},{\boldsymbol{\mathcal{L}}}  \rangle - \theta q_k %\big)
 )^2
%\Big )
+
\frac{\lambda}{\gamma} \sum_{i \in \Gamma}
    %\Phi ({\boldsymbol{\mathcal{G}}}_{k} )
% \big
 \| {\boldsymbol{\mathcal{G}}}_{i} %\big
 \|_{\Phi,\mathfrak{L} }
% \notag \\
% \textit
 + \operatorname{{I}} _{ \{
  \|  \boldsymbol{\mathcal{L}} \|_{\infty} \leq \alpha
 \} },
 \notag \\  \label{1bittensorqqqq} %\label{equ_nonconvex}
&\text{s.t.} \;\; {\boldsymbol{\mathcal{Z}}}={\boldsymbol{\mathcal{L}}},
 {\boldsymbol{\mathcal{G}}}_{i}=\nabla_{i}({\boldsymbol{\mathcal{Z}}}), i \in \Gamma, \;\;
% \Psi( {\boldsymbol{\mathcal{L}}}),
% \notag \\ \label{equ_nonconvexhtc_1btc}
%&
%\text{s.t.} \;\;
% {\boldsymbol{\mathcal{G}}}_{k}=\nabla_{k}({\boldsymbol{\mathcal{L}}}), \|  \boldsymbol{\mathcal{L}} \|_{\infty} <0 \;\;
\end{align}
where the indicator function
 $\operatorname{I} _{ \{
  \|  \boldsymbol{\mathcal{L}} \|_{\infty} \leq \alpha
 \} }$ is such that: 1) $\operatorname{I} _{ \{
  \|  \boldsymbol{\mathcal{L}} \|_{\infty} \leq \alpha
 \} }=0$, if $\|  \boldsymbol{\mathcal{L}} \|_{\infty} \leq \alpha$;
 2) $\operatorname{I} _{ \{
  \|  \boldsymbol{\mathcal{L}} \|_{\infty} > \alpha
 \} }=+ \infty $, if $\|  \boldsymbol{\mathcal{L}} \|_{\infty} > \alpha$.
The ADMM framework is utilized  to optimize formula (\ref{1bittensorqqqq}),
 %The ADMM iteration system %framework
% with respect to  GN1BHTC model
% The ADMM framework is utilized  to optimize the above problem,
 and the iteration  %entire optimization
 procedure is briefly summarized %as follows: %
 in Algorithm \ref{algorithm2222100000}. Specifically,
%  Due to space constraints, we put the detailed optimization steps of all subproblems in the supplementary materials.
\begin{align}
{ {\boldsymbol{\mathcal{L}}}^{\{t+1\}}}
& =
\boldsymbol{{\mathit{P}}}_{  \| \boldsymbol{\mathcal{L}}    \|_{\infty} \leq \alpha  }
[
(m {\mu^{\{t\}}}   { {\boldsymbol{\mathcal{Z}}}^{\{t\}}}+\boldsymbol{\mathcal{J}} _{1} -m {\boldsymbol{\mathcal{Y}}}^{\{t\}} )
 %\\
 \notag
  \\  \label{1bit-L}
&  \oslash (m {\mu^{\{t\}}} \textbf{1}
+ \boldsymbol{\mathcal{J}} _{2}
)
],
%
%
%$$
%\notag
\\ \label{1bit-Z} % \label{1bit-L}  \label{1bit-Z}
{\boldsymbol{\mathcal{Z}}}^{\{t+1\}}
& =\mathscr{F}^{-1}
\bigg(
\frac
{
\mathscr{F} ( %{\boldsymbol{\mathcal{M}}}-
{\boldsymbol{\mathcal{L}}}^{\{t\}} + \frac{{\boldsymbol{\mathcal{Y}}}^{\{t\}}}  {\mu^{\{t\}}})
+  {\boldsymbol{\mathcal{T}}}
}
{\textbf{1} + \sum_{i \in \Gamma}  \mathscr{F}( {{\boldsymbol{\mathcal{D}}}}_{i})^{\mit{H}}  \odot \mathscr{F}( {{\boldsymbol{\mathcal{D}}}}_{i})    }
\bigg),
%$$
\\ \label{1bit-g} %
{ {\boldsymbol{\mathcal{G}}}_{i}^{\{t+1\}}} & = {\boldsymbol{\mathcal{D}}}_{\Phi, \frac{\lambda }{ \gamma {\mu^{\{t\}}} } }( %{\boldsymbol{\mathcal{A}}},
 \nabla_{i}({\boldsymbol{\mathcal{Z}}}^{\{t+1\}}) %- {\boldsymbol{\mathcal{G}}}_{i}
+
\frac {  \Lambda_{i}^{\{t\}} }   {\mu^{\{t\}}},
 \mathfrak{L}),
\\ \label{1bit-Y}
%\begin{align*} %
%
{\boldsymbol{\mathcal{Y}}}^{\{t+1\}}
&   =
{\boldsymbol{\mathcal{Y}}}^{\{t\}}+
\mu^{\{t\}}
(%{\boldsymbol{\mathcal{M}}}-  %{\boldsymbol{\mathcal{D}}}{*}_{\mathfrak{L}}
{\boldsymbol{\mathcal{L}}}^ {\{t+1\}}-{\boldsymbol{\mathcal{Z}}}^{\{t+1\}}),
\\
{\Lambda_{i}}^{\{t+1\}} & ={\Lambda}_{i}^{\{t\}}+
\mu^{\{t\}} \big(
\nabla_{i}({\boldsymbol{\mathcal{Z}}}^{\{t+1\}})- {\boldsymbol{\mathcal{G}}}^{\{t+1\}}_{i}
%\nabla_{k}({\boldsymbol{\mathcal{Z}}}^{t+1})
 \big ),
\\ \label{1bit-MU}
\mu^{\{t+1\}}
& = \min \left( \mu^{\operatorname{max}},\vartheta \mu^{\{t\}} \right),
%\end{align*}
\end{align}
%%%%%%%%%%%%%%%%%%%%%%%%%%%%%%%%%%%%%%%%%%%%%%%%%%%%%%%%%%%%%%%%%%%%%%%%%%%%%%%
%
%
 \begin{table}[!htbp]
  \caption{
  Experimental datasets used for evaluation and their detailed information.
  }
  \label{exp-datasets}

  \centering
\scriptsize
\footnotesize
\renewcommand{\arraystretch}{0.1}
\setlength\tabcolsep{0.5pt}

\begin{tabular}{c l c }
    \Xhline{1pt}
   % \hline
    %\hline

      \multicolumn{1}{c}{Data-Type} &  \multicolumn{1}{c}{Data-Name} &\multicolumn{1}{c}{Data-Dimension}  \\

     \Xhline{1pt}
\hline
     \hline

  %  \hline
  %   \hline

     \multirow{4}{*}
     {  \tabincell{c} {Order-$3$ HSIs\\ %hyperspectral images\\
       \href{{https://engineering.purdue.edu/~biehl/MultiSpec/hyperspectral.html}} {[Link1]}
     \href{{https://www.ehu.eus/ccwintco/index.php?title=Hyperspectral_Remote_Sensing_Scenes}} {[Link2]}
     \href{{http://szu-hsilab.com/szu-tree-dataset/}} {[Link3]}
     % http://szu-hsilab.com/szu-tree-dataset/
     % %https://engineering.purdue.edu/~biehl/MultiSpec/hyperspectral.html
     }}
     & HSI $1$: KSC  &$512\times 614 \times 176$
\\

 \qquad
     & HSI $2$: Pavia
     &$1096 \times 715 \times 102$
\\

%\qquad
%     & HSI $3$: PaviaU  %\href{{https://www.ehu.eus/ccwintco/index.php?title=Hyperspectral_Remote_Sensing_Scenes}} {(Link)}
%%are preprocessed with the size of $300 \times 300 \times 60$,
%     &$610 \times 340 \times 103$
%\\
%
%\qquad
%     &
%HSI $4$: DCMall  %\href{{https://www.ehu.eus/ccwintco/index.php?title=Hyperspectral_Remote_Sensing_Scenes}} {(Link)}
%     &$1280 \times 307 \times 191$
%\\

\qquad
     &
HSI $3$: SZUTreeHSI-R1  %\href{{https://www.ehu.eus/ccwintco/index.php?title=Hyperspectral_Remote_Sensing_Scenes}} {(Link)}
     &$1000  \times 1000  \times 98 %80
     $
\\

\qquad
     &
HSI $4$: SZUTreeHSI-R2  %\href{{https://www.ehu.eus/ccwintco/index.php?title=Hyperspectral_Remote_Sensing_Scenes}} {(Link)}
     &$1000  \times 1000  \times 98 %80
     $
\\

\hline

   %  \hline

      \multirow{5}{*}
     {  \tabincell{c} {Order-$3$ MRIs
     \\
       \href{{https://www.kaggle.com/c/second-annual-data-science-bowl/data}} {[Link4]} }}
        & MRI $1$: 844-4ch6      &$500 \times 651 \times 30$

\\

 \qquad
     & MRI $2$: 1138-ch8
     &  $559 \times 531 \times 30$

\\

\qquad
     &
MRI $3$: 849-sax38
     &$528 \times 704 \times 30$
\\

\qquad
  & MRI $4$: 958-sax11 &$608 \times 456 \times 30$

\\

\qquad
     &
MRI $5$: 958-4ch10
     &$407 \times 608 \times 30$
\\

\hline

 \multirow{5}{*}
     {  \tabincell{c} {Order-$3$ MSIs  % 10https://www.cs.columbia.edu/CAVE/databases/multispectral/
     \\
       \href{{https://www.cs.columbia.edu/CAVE/databases/multispectral/}} {[Link5]} }}
        & MSI $1$: Cloth     &$512 \times 512 \times 31$

\\

 \qquad
     & MSI $2$: Thread-Spools
     &  $512 \times 512 \times 31$

\\

\qquad
     &
MSI $3$: Jelly-Beans
     &$512 \times 512 \times 31$
\\
% , ,  and  datasets,

\qquad
  & MSI $4$: Feathers &$512 \times 512 \times 31$

\\

%\qquad
%     &
%MRI $5$: 958-4ch10
%     &$407 \times 608 \times 30$
%\\

\hline

     %\hline

      \multirow{6}{*}
     {  \tabincell{c} {Order-$4$ CVs %\\ color videos
     \\
       \href{{https://media.xiph.org/video/derf/}} {[Link6]} }}
     & CV $1$:
    Akiyo   &$288 \times 352 \times 3 \times 300$
\\

 \qquad
    & 	
CV $2$: Container   &$288 \times 352 \times 3 \times 300$
\\

\qquad
    & 	
CV $3$: Silent
     &$288 \times 352 \times 3 \times 300$
\\

\qquad
     &
     CV $4$: Waterfall
     &$288 \times 352 \times 3 \times 260$
\\

\qquad
     &
CV $5$: %hall-monitor   &$288 \times 352 \times 3 \times 300$
News &$288 \times 352 \times 3 \times 300$
\\
\qquad
     &
CV $6$:  Galleon &$288 \times 352 \times 3 \times 360$
\\  %galleon &$288 \times 352 \times 3 \times 360$

 \hline

      \multirow{4}{*}
     {  \tabincell{c} {Order-$4$ MRSIs %\\ color videos
     \\
       \href{{https://take5.theia.cnes.fr/atdistrib/take5/client/\#/home}} {[Link7]} %}}
        \href{{https://theia.cnes.fr/atdistrib/rocket/\#/home}} {[Link8]} }}
     & MRSI $1$:  SPOT-5  &   $1000  \times  1000  \times    4 \times    13$
\\

\qquad & MRSI $2$:  Landsat-7 &  $1000 \times   1000    \times  6  \times   11$\\
\qquad & MRSI $3$:  T22LGN &  $1000 \times   1000   \times   4  \times    7$ \\
\qquad & MRSI $4$:  T29RMM & $1000 \times   1000   \times   4    \times  6$ \\

 \hline
 \multirow{6}{*}
     {  \tabincell{c} {Order-$5$ LFIs %\\ color videos
     \\
        \href{{https://www.irisa.fr/temics/demos/IllumDatasetLF/index.html}} {[Link9]} }}
     & LFI $1$:  Mini   &  $200\times 300 \times 3\times 15\times15$\\
 \qquad&LFI $2$: Framed  &$200\times 300 \times 3\times 15\times15$\\
  \qquad&LFI $3$:  Bee1 &$200\times 300 \times 3\times 15\times15$\\
     \qquad&LFI $4$: Bench &$200\times 300 \times 3\times 15\times15$\\
      \qquad&LFI $5$: Duck  &$200\times 300 \times 3\times 15\times15$\\
       \qquad&LFI $6$: Trees2 &$200\times 300 \times 3\times 15\times15$\\

\hline
     \hline
     \Xhline{1pt}

\end{tabular}
\vspace{-0.6464cm}
\end{table}
%%%%%%%%%%%%%%%%%%%%%%%%%%%%%%%%%%%%%%%%
\noindent where
$\boldsymbol{{\mathit{P}}}_{  \| \boldsymbol{\mathcal{L}}    \|_{\infty} \leq \alpha  } (\cdot)$ stands for
projection operator,
$ \oslash$ is denotes  elementwise division,
the %form of
 definition of ${\boldsymbol{\mathcal{T}}}$  is the same as the formula  (\ref{gnrhtc-LLLLLLwwwww}),
 $ {\boldsymbol{\mathcal{D}}}_{\Phi, \tau}(\cdot, \cdot)$ denotes
 the GNHTSVT operation (see Algorithm \ref{random-wtsvt}).
\textbf{Due to space constraints of this paper,
%we put the %detailed optimization step of all subproblems
%detailed  sub-problems optimization, %  steps of all  sub-problems,
%time complexity,  and
%convergence analysis of Algorithm \ref{algorithm2222100000}
we have placed the detailed ADMM optimization algorithms
for solving Model (\ref{1bittensor})  and Model (\ref{1bittensor-gu-sp}),
along with their corresponding time complexity analysis and convergence analysis %, in the supplementary materials.
in the supplementary materials.}

\vspace{-0.2309cm}

\section{\textbf{EXPERIMENTAL RESULTS}}\label{experiments}

In this section, we perform extensive experiments on both
synthetic and   real-world tensor data  to substantiate the superiority and effectiveness of the proposed approach.
\textcolor[rgb]{0.00,0.00,0.00}{Due to the page limitations of this paper,
partial synthetic and real  experiments are provided in the supplementary material.
All the experiments  are run  on the following two platforms:
\textbf{1)} Windows 10 and Matlab (R2016a) with an Intel(R) Xeon(R) Gold-5122 3.60GHz CPU and 192GB memory;
\textbf{2)} Windows 10 and Matlab (R2022b) with an Intel(R) Xeon(R) Gold-6230 2.10GHz CPU and 128GB memory.}

\textcolor[rgb]{0.00,0.00,0.00}{
\textbf{Experimental Datasets:}
In this section, we choose several %type four major
types of  real  tensorial data to verify the superiority and effectiveness of the proposed method over  compared algorithms.
  Experimental datasets used for evaluation and their detailed information
  are summarized in Table
   \ref{exp-datasets},
  which cover \textit{multispectral images} (MSIs),  \textit{hyperspectral images}  (HSIs),
\textit{magnetic resonance images} (MRIs),   \textit{color videos} (CVs),
\textit{multi-temporal remote sensing images} (MRSIs), and \textit{light field images} (LFIs).}
%%%%%%%%%%%%%%%%%%%%%%%%%%%%%%%%%%%
%
   In our experiments, we   normalize the gray-scale  value of the  tested tensors to the interval  $[0, 1]$.
For unquantized  scenarios, the observed noisy tensor is constructed as follows:
 the %random-valued
 impulse noise with ratio $\operatorname{\textit{NR}}$ is uniformly and randomly  added to %nr  %each frontal slice of
 the ground-truth   tensor,  %sr
 and then we   sample ($\textit{SR} \cdot \prod_{i=1}^{d} n_i$) pixels from the noisy tensor to form the observed tensor
 $\boldsymbol{\bm{P}}_{{{\Omega}}}({\boldsymbol{\mathcal{M}}})$ at random.
For quantized scenarios, the sampling rate is defined as
 $
 \operatorname{SR}= {m} / {\prod_{i=1}^{d} n_i}
 $,   where $m$ denotes the number of samples.
 %The
% one-bit observations $\{q_k\}_{\forall k \in [m]} $ are generated by Formula
The  prequantization random noises $\{\epsilon _k\}   _{\forall k \in [m]}  $
are generated from a mean-zero Gaussian distribution with standard deviation $\sigma >0$.

 \textbf{Evaluation Metrics:}
The \textit{Peak Signal-to-Noise Ratio} (PSNR), the \textit{structural similarity} (SSIM),
the \textit{relative square error} (RSE),
and the CPU time are employed to evaluate the recovery performance.
Generally, better recovery performances are reflected by higher PSNR and SSIM values and lower RSE values.
%
% The parameters in compared methods are manually adjusted to the optimal performance, which refers to the discussion in their articles.
%Meanwhile, the best and the second-best results are highlighted by bold and underline, respect
The best and the second-best results are %highlighted by bold and underline,
 highlighted in red and blue,
respectively.
% The best and the second-best results are highlighted in red and blue, respectively.

\textbf{Parameters Setting:}  Unless otherwise stated, all parameters involved in
the competing methods were optimally assigned or selected
as suggested in the reference papers. % (\textit{Please refer to the supplementary materials for details}).
%
%%%%%%%%%%%%%%%%%%%%%%%%%%%%%%%%%%%%%%%%%%%%%%%%%%%%%%%%%%%%%%%%%%%%%%%%%%%%%%%%%%%%%%%%%%%%%%%%%%%%%%%%%%%%%%%%%%%%%%%%%%%%%%%%%%%%%%%%%%%
%
% Below, we provide parameters for the proposed method
% Regarding the parameter settings for the proposed algorithm:
{\textit{In the supplementary material,  %Below,
we provide the parameter settings %for the proposed algorithm:
regarding the  proposed and compared methods.}}

\vspace{-0.3cm}
\subsection{\textbf{%Comparative
Experiments 1: Noise-Free Tensor Completion in Unquantized Tensor Recovery}}
 \textbf{Competing Algorithms:}
In this subsection, we compare the proposed GNHTC method and its
two randomized versions
with a
number of state-of-the-art LRTC %rank relaxed low rank matrix completion
 methods, including %APG [31], WNNM [6], IRNN-Sp [4],
%IRNN-SCAD [4], FaNCL-LSP [9], FaNCL-CNN [9], MSS [55],
%niAPG [37], CNNM3 [56], TRNM [26], TNNR-IDD [51], and
%NCARL [2].
TRNN \cite{yu2019tensor},
HTNN \cite{qin2022low},
METNN \cite{liu2024revisiting},
MTTD \cite{feng2023multiplex},
WSTNN \cite{zheng2020tensor44},
EMLCP-LRTC \cite{zhang2023tensor},
TCTV-TC  \cite{wang2023guaranteed},
GTNN-HOC \cite{wang2024low2222},
and
t-$\epsilon$-LogDet \cite{ yang2022355}.
In our randomized  versions, one is named  R1-GNHTC, integrating the % that integrates %fuses
fixed-rank LRTC strategy,
while the other is called  R2-GNHTC,  incorporating the %that incorporates
fixed-accuracy LRTC strategy.
%(namely,  R-GNHTC1 fused fixed-rank LRTC strategy and
%R-GNHTC2 fused fixed-accuracy LRTC strategy)

%SMTTD \cite{feng2023multiplex},   t-$\epsilon$-LogDet \cite{ yang2022355},  and .

 \textbf{Results and Analysis:}
%
%The PSNR values and  running time  output by various LRTC  methods for different color videos.
 % Quantitative evaluation %
 Quantitative evaluation results
  of the proposed and compared %various
  LRTC  methods on MSIs, MRSIs, and CVs are presented in
 Table  \ref{table-lrtc-MSIMSI}. % including PSNR, SSIM, RSE and  CPU Time (Second).
These experimental
results %across three types of tensor %multidimensional  data
consistently demonstrate that, especially % particularly
 under extremely low sampling rates, the  proposed %deterministic
 GNHTC method  far surpasses
% significantly outperforms
 existing high-order LRTC % tensor completion
 methods,  such as  HTNN, METNN,
MTTD,
WSTNN, and
EMLCP-LRTC, which  purely couple global low-rankness,
 in terms of the PSNR, SSIM, and RSE metrics.
Besides, % In addition,
  our deterministic  GNHTC algorithm  achieves a PSNR gain of approximately $1$dB compared to the convex TCTV-TC method.
  %%%%%%%%%%%%%%%%%%%%%%%%%%%%%%%%%%%%%%%%%%%%%%
  From the perspective of computational efficiency, as the scale of tensor data increases, %as the data scale increases,
  the %time cost
  gap
  of CPU Time (Second)
  between the two proposed randomized  algorithms and other deterministic algorithms widens.
  Notably, when benchmarked against  %compared to
  competitive approaches such as METNN, MTTD, WSTNN, EMLCP-LRTC, and TCTV-TC,
  our randomized  algorithms on MRSIs recovery demonstrate  a speedup %ranging from
 of  $5$$\sim$$9$ times  ($3$$\sim$$7$ times on CVs).
 The corresponding %visual comparisons %are provided in Figure 1.
visual quality comparisons %are %presented/shown in Figure 1.
are provided in Figure \ref{fig-lrtc-msimsimsi}.
This %which
indicates that when other comparison algorithms fail, the proposed method can still restore the general appearance.
Since the evaluation metrics %numerical results
obtained by  %R$1$-GNRHTC and R$2$-GNRHTC %
two randomized  methods
are very similar, we only present the visual results %outputs %
of %R$1$-GNHTC/
R$1$-GNHTC  % R1-GNHTC
% fixed-rank scheme-based
algorithm.

%%%%%%%%%%%%%%%%%%%%%%%%%%%%%%%%%%%%%%%%%%%%%%%%%%%%%%%%%%%%%%%%%%%%%%*%%%
\begin{table*} %[tp]
\renewcommand{\arraystretch}{0.65}
\setlength\tabcolsep{1.53pt}
  \centering
  \caption{
  %The PSNR values and  running time  output by various LRTC  methods for different color videos.
  Quantitative evaluation %PSNR, SSIM, RSE and  CPU Time (Second)
  of the proposed and compared %various
  LRTC  methods on MSIs, MRSIs, and CVs.
  % on %HSIs and
   %MSI datasets.
   %order-$3$ MSIs and order-$4$ CVs.
  }
  \vspace{-0.15cm}
  \label{table-lrtc-MSIMSI}
  \scriptsize
 % \tiny
% \footnotesize
  \begin{threeparttable}
   % \begin{tabular}{c | c| c | c| c|c|c|c| c|c|c|c| c|c|c|c| c|c}
     \begin{tabular}{cccc cccc cccc  c  c}
   % \hline
    \Xhline{1pt}
    %\multirow{2}{*}
   \tabincell{c}   {SR}&
   % \multirow{2}{*}
   \tabincell{c} {Evaluation\\ Metric}&
   %%%%%%%%%%%%%%%%%%%%%%%%%%%%%%%%%%%%%%%%%%%%%%%%%%%%%%%%%%%%%%%%%%%

%%%%%%%%%%%%%%%%%%%%%%%%%%%%%%%%%%%%%%%%%%%%%%%%%%%%%%%%%%%
\tabincell{c}{ TRNN \\  \cite{yu2019tensor} }& %
\tabincell{c} {HTNN \\ \cite{qin2022low} }& %
   \tabincell{c}{METNN \\  \cite{liu2024revisiting} }& %
% yang2022robust, liu2024fully, lou2019robust
\tabincell{c}{MTTD   \\ \cite{feng2023multiplex} }& %
 \tabincell{c}{WSTNN \\ \cite{zheng2020tensor44} }&
%%%%%%%%%%%%%%%%%%%%%%%%%%%%%%%%%%%%%%%%%%%%%%%%%%%%%%%%%%%%%%%%%%%%%%%%

\tabincell{c}{EMLCP \\ -LRTC\cite{zhang2023tensor}}&
 \tabincell{c} {GTNN-HOC \\ \cite{wang2024low2222}}&
 \tabincell{c} {t-$\epsilon$-LogDet \\ \cite{ yang2022355}}&
 \tabincell{c}{TCTV-TC  \\ \cite{wang2023guaranteed}}
 %\cite{wang2023guaranteed}
 & \textbf{{GNHTC}}& %\textit
 \textbf{{R1-GNHTC}}&  \textbf{{R2-GNHTC}}
   % PSNR&Time (s)&PSNR&Time (s)&PSNR&Time (s)&PSNR&Time (s)&PSNR&Time (s)
   % &PSNR&Time (s)&PSNR&Time (s)&PSNR&Time (s)

    \cr

   % \hline
    %\hline
    \Xhline{1pt}
\multicolumn{14}{c}{\textbf{\textit{Results on Third-Order  MSIs}}}\\
\hline
     \hline
   \multirow{4}{*}{    \tabincell{c}   {SR=$1\%$ }  } &MPSNR
 &14.2487&20.5380&18.4703&21.0184&18.9146&23.4133&20.5633&20.5657&{{27.4766}}& %\textcolor[rgb]{0,0,1}{\textbf{27.2920}}&27.3478
 \textcolor[rgb]{1,0,0}{\textbf{28.1947}} &	\textcolor[rgb]{0,0,1}{\textbf{28.1181}} &	28.0127

 \cr
    %% highlighted in red and blue,
    % \qquad	&MPSNR	& & & & & & & & & & & & & & & & \cr
    %%
   \qquad	&MSSIM	 &0.3385&0.5129&0.5707&0.5400&0.4903&0.7278&0.4927&0.4926&{{0.8075}}&
   %\textcolor[rgb]{1,0,0}{\textbf{0.8268}}&0.7944
   \textcolor[rgb]{1,0,0}{\textbf{0.8363}} &	\textcolor[rgb]{0,0,1}{\textbf{0.8243}} 	& 0.8209
 \cr

   \qquad	&MRSE	& 0.9168&0.4460&0.5865&0.4183&0.5727&0.3477&0.4362&0.4474&{{0.2078}}&
   %0.2135&\textcolor[rgb]{0,0,1}{\textbf{0.2106}}
   \textcolor[rgb]{1,0,0}{\textbf{0.1932}} &	\textcolor[rgb]{0,0,1}{\textbf{0.1946}} &	0.1967
\cr
    \qquad	&MTime	&720.859&\textcolor[rgb]{0,0,1}{\textbf{397.523}}&2030.789&755.391&1670.466&1906.961&465.174&
    \textcolor[rgb]{1,0,0}{\textbf{383.452}}&1444.838 & %1508.467&521.034
    1542.876 &	535.696 &	628.513

    \cr

   \hline
   \multirow{4}{*}{    \tabincell{c}   {SR=$3\%$ }  } &MPSNR
  & 17.0734&26.3468&25.8454&26.7355&29.4583&29.9045&26.4128&27.2031&{{31.4842}}&
  %\textcolor[rgb]{1,0,0}{\textbf{31.9741}}&30.9668
  \textcolor[rgb]{1,0,0}{\textbf{32.4408}} &	\textcolor[rgb]{0,0,1}{\textbf{32.0220}} &	31.6791
\cr
    %%
    % \qquad	&MPSNR	& & & & & & & & & & & & & & & & \cr
    %%
   \qquad	&MSSIM	&0.4950&0.7903&0.8244&0.8066&0.9093&0.9096&0.7930&0.8026&{{0.9099}}&
   %\textcolor[rgb]{1,0,0}{\textbf{0.9312}}&0.8795
   \textcolor[rgb]{1,0,0}{\textbf{0.9316}} &	\textcolor[rgb]{0,0,1}{\textbf{0.9101}} &	0.8983
\cr
   \qquad	&MRSE	&0.6506&0.2370&0.2606&0.2217&0.1595&0.1678&0.2299&0.2163&{{0.1327}}&
   %\textcolor[rgb]{1,0,0}{\textbf{0.1276}}&0.1422
   \textcolor[rgb]{1,0,0}{\textbf{0.1195}} &	\textcolor[rgb]{0,0,1}{\textbf{0.1254}} &	0.1314
\cr
    \qquad	&MTime	&723.259&\textcolor[rgb]{0,0,1}{\textbf{350.539}}&2017.946&713.037&1930.939&1909.037&453.553&\textcolor[rgb]{1,0,0}{\textbf{349.175}}&1422.979&
    %1506.717&523.082
    1544.423 &	534.529 &	626.512
 \cr

   \hline
 %%  \cr
  \multirow{4}{*}{    \tabincell{c}   {SR=$5\%$ }  } &MPSNR
 & 19.8601&28.7757&31.2017&28.9041&32.6543&33.4464&29.7611&30.3920&{{33.6687}}&
 %\textcolor[rgb]{1,0,0}{\textbf{34.4788}}&32.9570
 \textcolor[rgb]{1,0,0}{\textbf{34.6742}} &	\textcolor[rgb]{0,0,1}{\textbf{34.0703}} &	33.3479
\cr
    %%
    % \qquad	&MPSNR	& & & & & & & & & & & & & & & & \cr
    %%

   \qquad	&MSSIM	&0.5826&0.8680&0.9275&0.8751&\textcolor[rgb]{0,0,1}{\textbf{0.9555}}&0.9501&0.8787&0.8890&0.9420&
   %\textcolor[rgb]{1,0,0}{\textbf{0.9602}}&0.9119
   \textcolor[rgb]{1,0,0}{\textbf{0.9586}}	&0.9395 &	0.9200
\cr
   \qquad	&MRSE	 &0.4760&0.1801&0.1389&0.1756&0.1116&0.1188&0.1595&0.1519&{{0.1045}}&
   %\textcolor[rgb]{1,0,0}{\textbf{0.0957}}&0.1147
   \textcolor[rgb]{1,0,0}{\textbf{0.0929}} &	\textcolor[rgb]{0,0,1}{\textbf{0.0999}} 	& 0.1100
 \cr
    \qquad	&MTime	& 726.038&\textcolor[rgb]{1,0,0}{\textbf{335.946}}&2032.377&676.609&1869.232&1908.997&445.278&\textcolor[rgb]{0,0,1}{\textbf{345.258}}&1404.638&
    %1507.796&587.589

    1540.097 &	597.850 &	690.531
 \cr

%
%
%25.5483&32.4790&36.9903&32.5179&37.4408&41.1821&34.2262&34.6778&37.1307&38.0371&36.0476
%0.7607&0.9378&0.9802&0.9469&0.9846&0.9926&0.9494&0.9538&0.9722&0.9819&0.9510
%0.2645&0.1197&0.0710&0.1173&0.0664&0.0431&0.0983&0.0949&0.0713&0.0641&0.0816
%728.7550&320.6770&2028.5129&655.1494&684.5235&1688.5675&441.2913&349.2009&1336.9600&1506.7176&697.7519

   \hline
    \hline
   % \Xhline{1pt}
\multicolumn{14}{c}{\textbf{\textit{Results on Fourth-Order  %CVs
MRSIs}}}\\
\hline
     \hline

      \multirow{4}{*}{    \tabincell{c}   {SR=$0.5\%$ }  } &MPSNR
 & 17.2533&19.8339&15.7720&20.9478&20.8181&18.9815&18.3208&16.3231&21.5348&\textcolor[rgb]{0,0,1}{\textbf{22.7191}}& {{22.4954}}
 & \textcolor[rgb]{1,0,0}{\textbf{22.7820}}\cr  %

   \qquad	&MSSIM	 & 0.3910&0.4132&0.3557&0.4658&0.4701&0.4772&0.3637&0.3250&0.5359&\textcolor[rgb]{1,0,0}{\textbf{0.5905}}&{{0.5695}}
   & \textcolor[rgb]{0,0,1}{\textbf{0.5771}} \cr

   \qquad	&MRSE	&
0.5510&0.3644&0.6399&0.3268&0.3491&0.3974&0.4082&0.5451&0.3161&\textcolor[rgb]{0,0,1}{\textbf{0.2631}}&{{0.2660}}
& \textcolor[rgb]{1,0,0}{\textbf{0.2586}}\cr
    \qquad	&MTime	& 13633.161&{{2852.105}}&9833.043&9106.050&20063.516&19070.785&3183.932&3343.147&9196.801&10506.549&\textcolor[rgb]{1,0,0}{\textbf{1744.389}}
  & \textcolor[rgb]{0,0,1}{\textbf{1870.817}}  \cr

    \hline
     \multirow{4}{*}{    \tabincell{c}   {SR=$1\%$ }  } &MPSNR
 &   19.9934&21.2452&18.4103&23.0171&22.4238&19.3586&19.7498&18.2973&23.2805&\textcolor[rgb]{1,0,0}{\textbf{24.3680}}&{{24.0644}}
 & \textcolor[rgb]{0,0,1}{\textbf{24.2869}}\cr

   \qquad	&MSSIM	 &  0.4437&0.4974&0.4292&0.5802&0.5626&0.5358&0.4675&0.4250&0.6388&\textcolor[rgb]{1,0,0}{\textbf{0.6828}}&{{0.6727}}
   &\textcolor[rgb]{0,0,1}{\textbf{0.6773}} \cr

   \qquad	&MRSE	& 0.3853&0.3156&0.4714&0.2620&0.2873&0.3758&0.3509&0.4306&0.2564&\textcolor[rgb]{0,0,1}{\textbf{0.2235}}&{{0.2244}}  & \textcolor[rgb]{1,0,0}{\textbf{0.2194}}\cr
    \qquad	&MTime	& 13536.008&{{2739.258}}&9861.887&8855.057&20129.556&19136.206&3193.774&3194.831&9231.643&10493.834&\textcolor[rgb]{1,0,0}{\textbf{1815.473}} & \textcolor[rgb]{0,0,1}{\textbf{1992.958}} \cr
  \hline
    \hline
   % \Xhline{1pt}
\multicolumn{14}{c}{\textbf{\textit{Results on Fourth-Order %MRSIs %Results on Fifth-Order  LFIs
CVs}}}\\
\hline
     \hline

      \multirow{4}{*}{    \tabincell{c}   {SR=$0.5\%$ }  } &MPSNR
 &12.8418&23.8049&10.6131&24.2577&20.4145&18.7434&22.8838&11.4957&\textcolor[rgb]{0,0,1}{\textbf{26.2302}}&\textcolor[rgb]{1,0,0}{\textbf{26.9312}}&24.7963
 & 25.3219
 \cr

   \qquad	&MSSIM	 &  0.4538&0.6208&0.3436&0.6917&0.6270&0.3322&0.4877&0.1129&\textcolor[rgb]{0,0,1}{\textbf{0.7332}}&\textcolor[rgb]{1,0,0}{\textbf{0.7479}}&0.6766
   & 0.6829\cr

   \qquad	&MRSE	& 0.5543&0.1588&0.7079&0.1520&0.2268&0.2721&0.1747&0.6254&\textcolor[rgb]{0,0,1}{\textbf{0.1187}}&\textcolor[rgb]{1,0,0}{\textbf{0.1101}}&0.1399
   & 0.1314\cr
    \qquad	&MTime	& 8884.572&{{5589.675}}&13830.999&16389.214&30441.946&31394.932&5838.133&6139.929&11029.124&12837.645&\textcolor[rgb]{1,0,0}{\textbf{3002.928}}
    & \textcolor[rgb]{0,0,1}{\textbf{3481.961}} \cr

    \hline
     \multirow{4}{*}{    \tabincell{c}   {SR=$1\%$ }  } &MPSNR
 &17.9342&25.9737&12.5734&26.0504&26.2277&25.9325&24.7281&13.9930&\textcolor[rgb]{0,0,1}{\textbf{28.3663}}&\textcolor[rgb]{1,0,0}{\textbf{28.7626}}&26.0067
 &26.3973
 \cr

   \qquad	&MSSIM	 & 0.5492&0.6971&0.4150&0.7449&0.7526&0.6601&0.5665&0.1717&\textcolor[rgb]{0,0,1}{\textbf{0.7929}}&\textcolor[rgb]{1,0,0}{\textbf{0.7994}}&0.7087
   & 0.7155\cr

   \qquad	&MRSE	& 0.3138&0.1254&0.5735&0.1246&0.1232&0.1275&0.1420&0.4703&\textcolor[rgb]{0,0,1}{\textbf{0.0941}}&\textcolor[rgb]{1,0,0}{\textbf{0.0906}}&0.1219
   & 0.1164\cr
    \qquad	&MTime	& 8900.368&{{5528.504}}&13871.996&16361.459&30502.989&31443.264&5802.668&6135.755&11073.196&12629.969&\textcolor[rgb]{1,0,0}{\textbf{3293.247}}
    & \textcolor[rgb]{0,0,1}{\textbf{4008.781}}\cr

     \hline \hline

      \Xhline{1pt}
    \end{tabular}
    \end{threeparttable}
    \vspace{-0.3cm}
\end{table*}

%%%%%%%%%%%%%%%%%%%%%%%%%%%%%%%%%%%%%%%%%%%%%%%%%%%%%%%%%%%%%%%%%%%%%%%%%%%%%%%%%%%%%%%%%%%%%%%%%%%%%%%%%%%%%%%%%%%%%%%%%%%%%%%%%%%%%%%%%%%%%%%%%
\begin{figure*}[!htbp]
\renewcommand{\arraystretch}{0.7}
\setlength\tabcolsep{0.43pt}
\centering
\begin{tabular}{ccc  ccc ccc ccc  }%cc ccc  ccc c cc
\centering

\includegraphics[width=0.586in, height=0.73in]{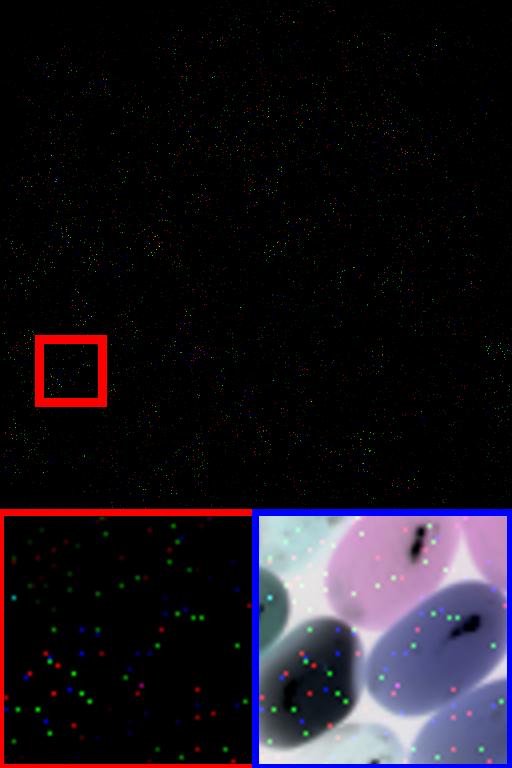}
&
\includegraphics[width=0.586in, height=0.73in]{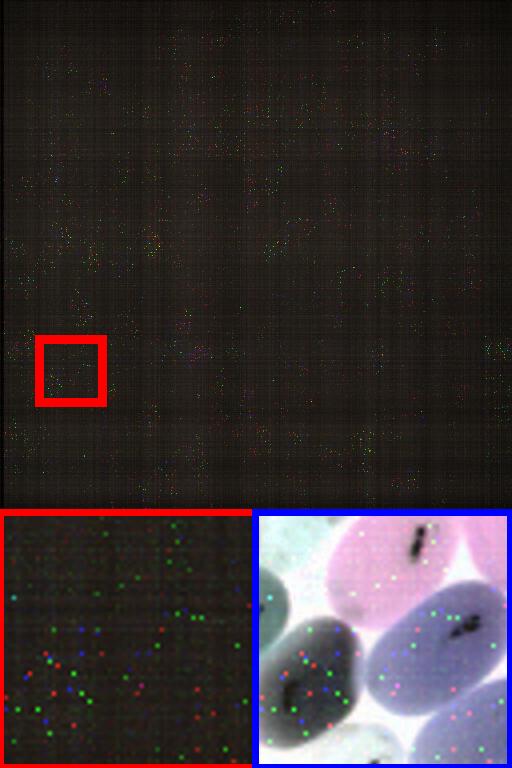}
&
\includegraphics[width=0.586in, height=0.73in]{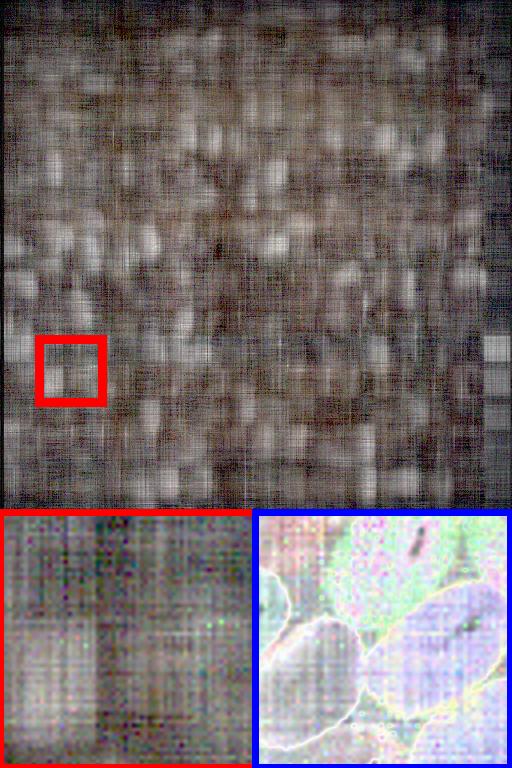}
&
\includegraphics[width=0.586in, height=0.73in]{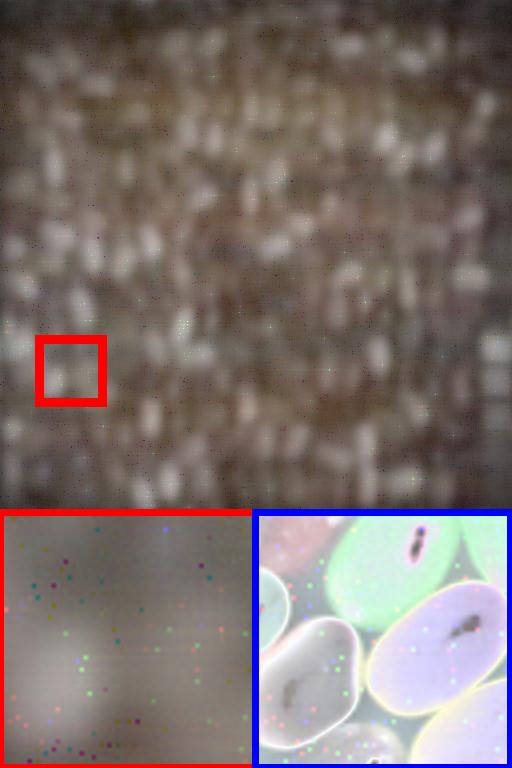}
&
\includegraphics[width=0.586in, height=0.73in]{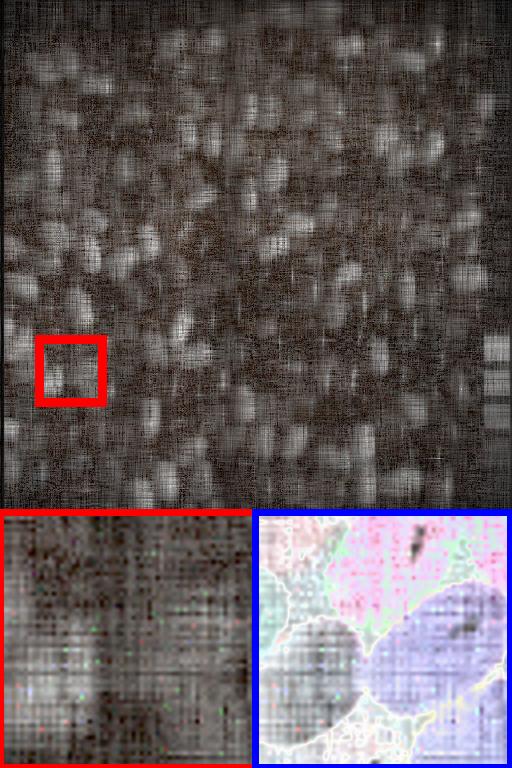}
&
\includegraphics[width=0.586in, height=0.73in]{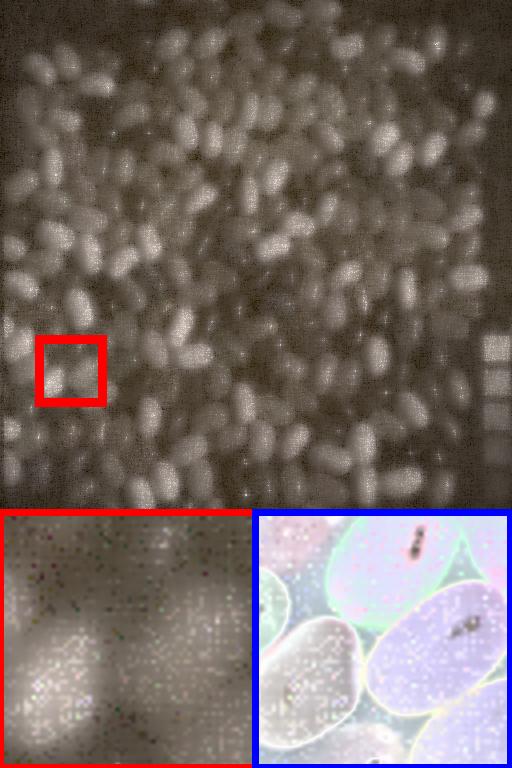}
&
\includegraphics[width=0.586in, height=0.73in]{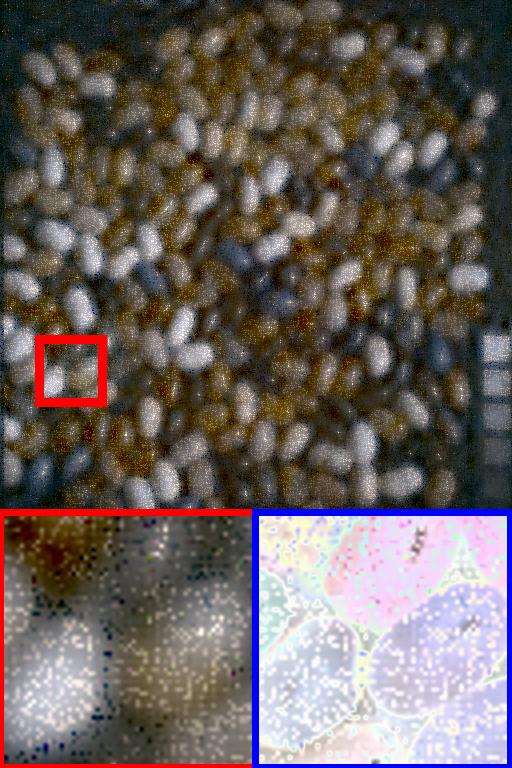}
&
\includegraphics[width=0.586in, height=0.73in]{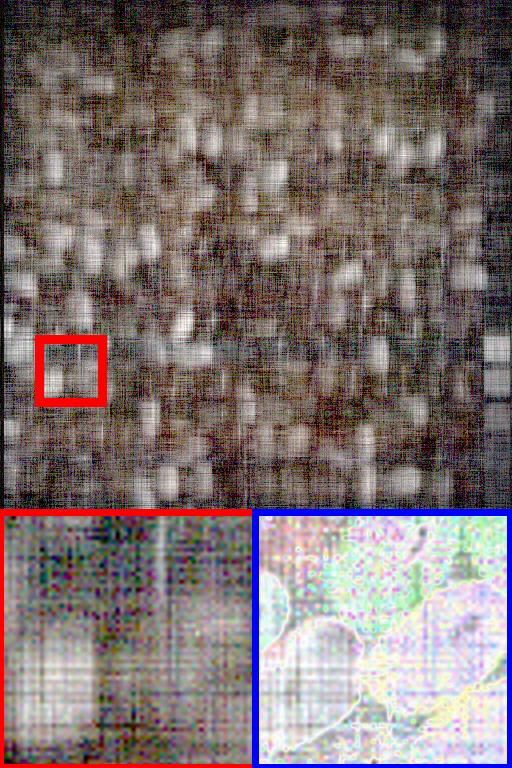}
&
\includegraphics[width=0.586in, height=0.73in]{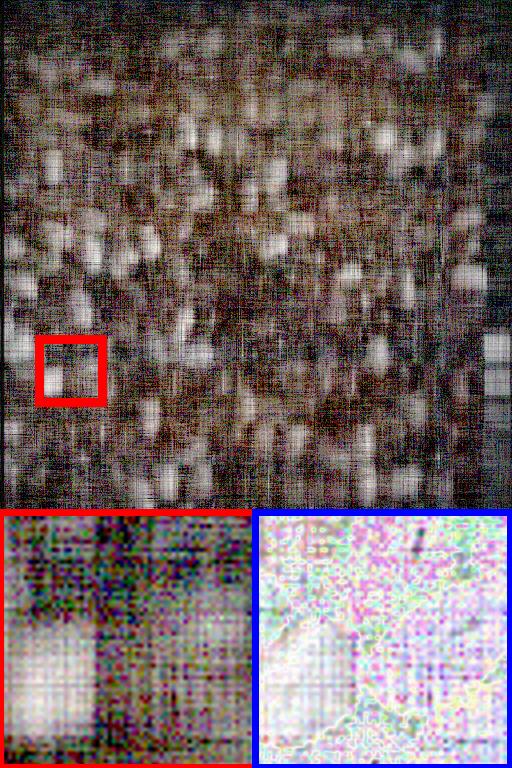}
&
\includegraphics[width=0.586in, height=0.73in]{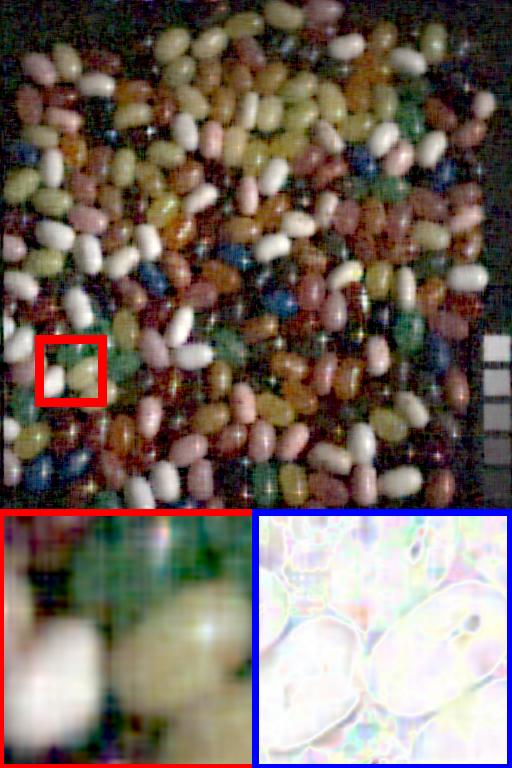}
&
\includegraphics[width=0.586in, height=0.73in]{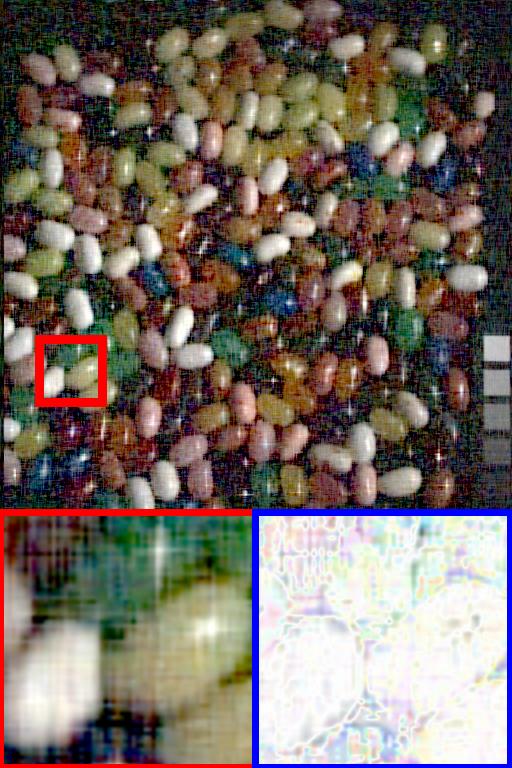}
&
\includegraphics[width=0.586in, height=0.73in]{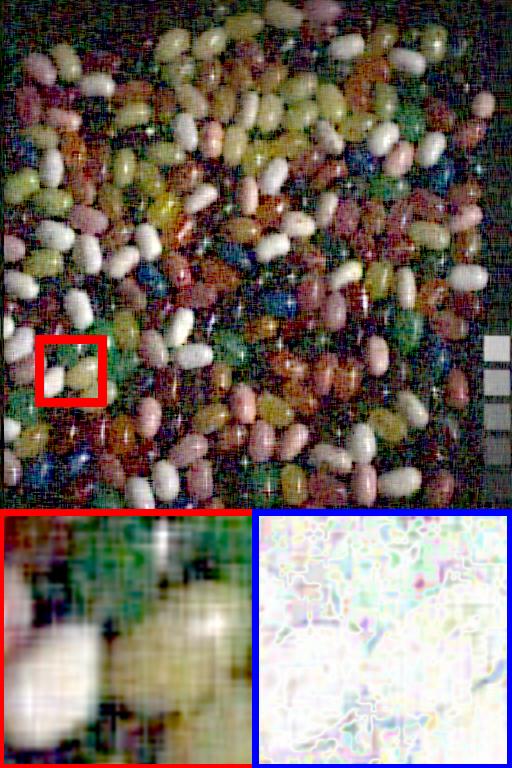} \\ \toprule

%\includegraphics[width=0.586in, height=0.73in]{msi-visu1-001/s12}
%&
%\includegraphics[width=0.586in, height=0.73in]{msi-visu1-001/s1}
%&
%\includegraphics[width=0.586in, height=0.73in]{msi-visu1-001/s2}
%&
%\includegraphics[width=0.586in, height=0.73in]{msi-visu1-001/s3}
%&
%\includegraphics[width=0.586in, height=0.73in]{msi-visu1-001/s4}
%&
%\includegraphics[width=0.586in, height=0.73in]{msi-visu1-001/s5}
%&
%\includegraphics[width=0.586in, height=0.73in]{msi-visu1-001/s6}
%&
%\includegraphics[width=0.586in, height=0.73in]{msi-visu1-001/s7}
%&
%\includegraphics[width=0.586in, height=0.73in]{msi-visu1-001/s8}
%&
%\includegraphics[width=0.586in, height=0.73in]{msi-visu1-001/s9}
%&
%\includegraphics[width=0.586in, height=0.73in]{msi-visu1-001/s10}
%&
%\includegraphics[width=0.586in, height=0.73in]{msi-visu1-001/s11}
%\\
% \toprule
 %\\
\includegraphics[width=0.586in, height=0.73in]{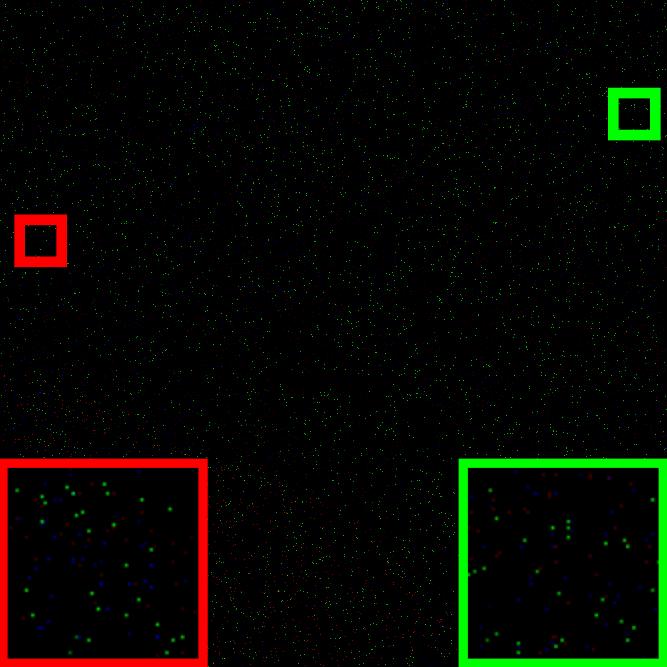}
&
\includegraphics[width=0.586in, height=0.73in]{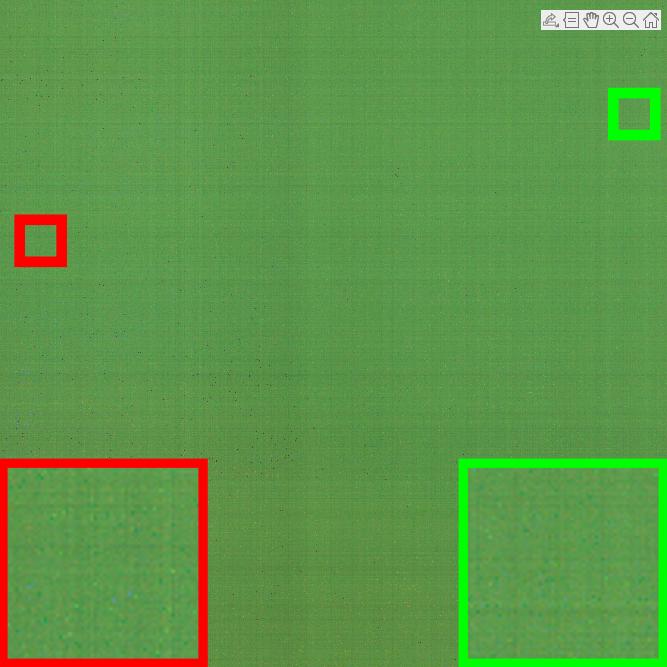}
&
\includegraphics[width=0.586in, height=0.73in]{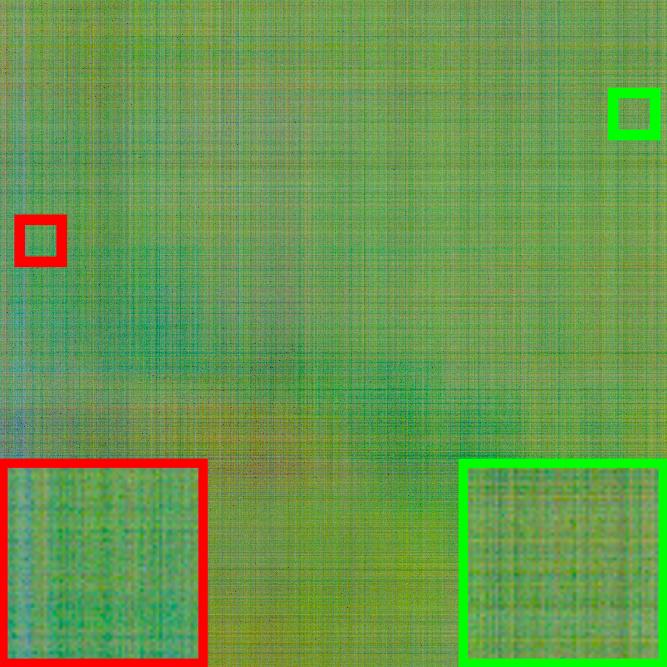}
&
\includegraphics[width=0.586in, height=0.73in]{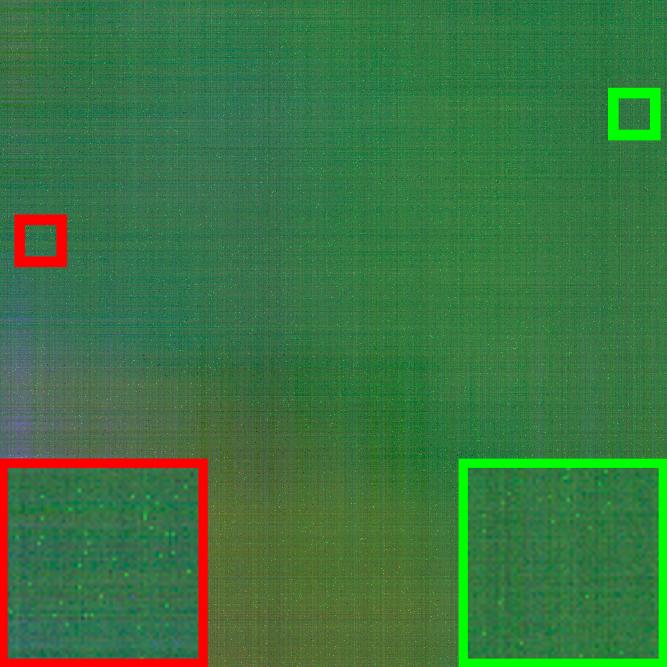}
&
\includegraphics[width=0.586in, height=0.73in]{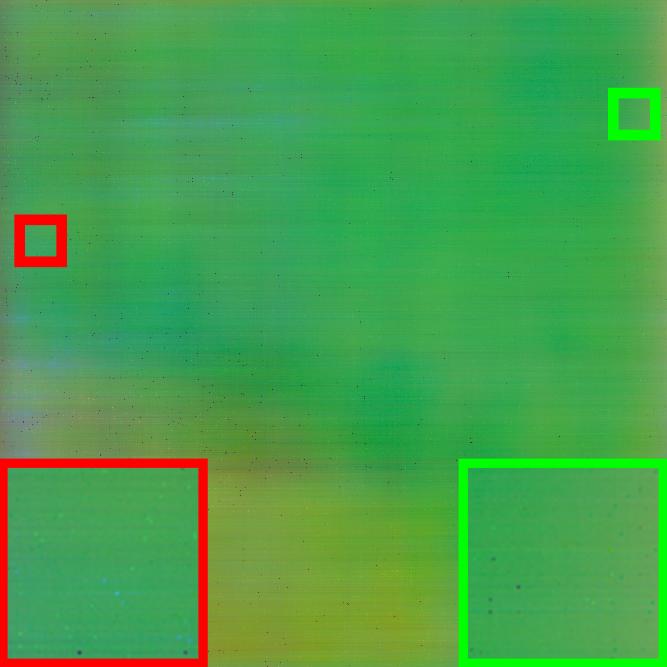}
&
\includegraphics[width=0.586in, height=0.73in]{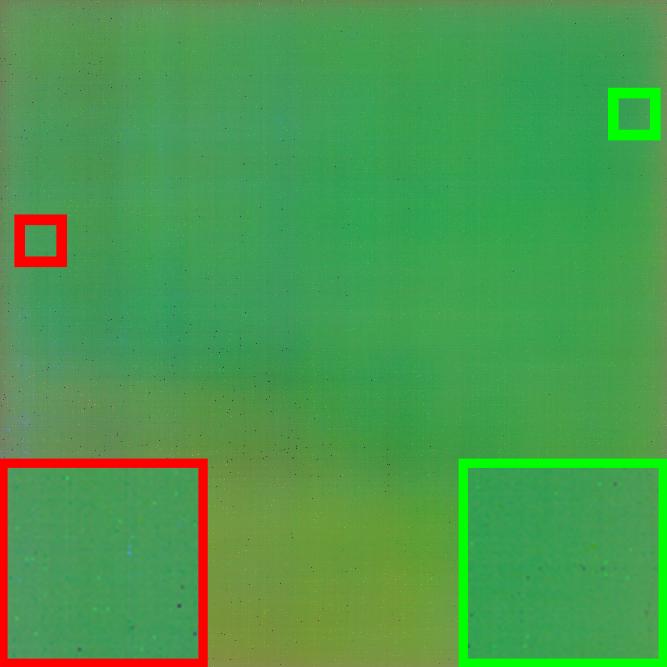}
&
\includegraphics[width=0.586in, height=0.73in]{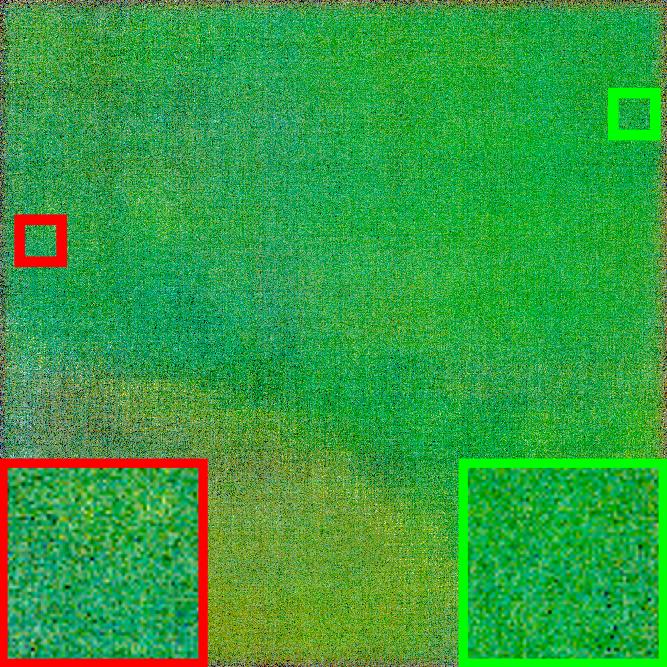}
&
\includegraphics[width=0.586in, height=0.73in]{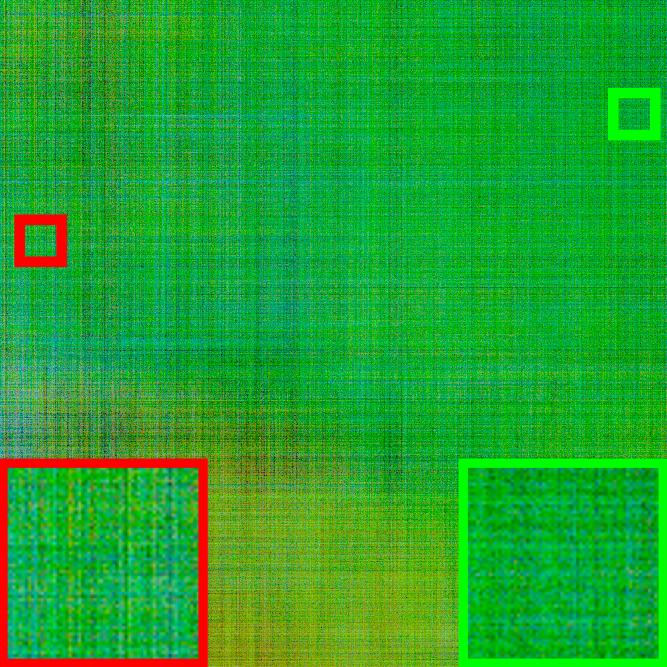}
&
\includegraphics[width=0.586in, height=0.73in]{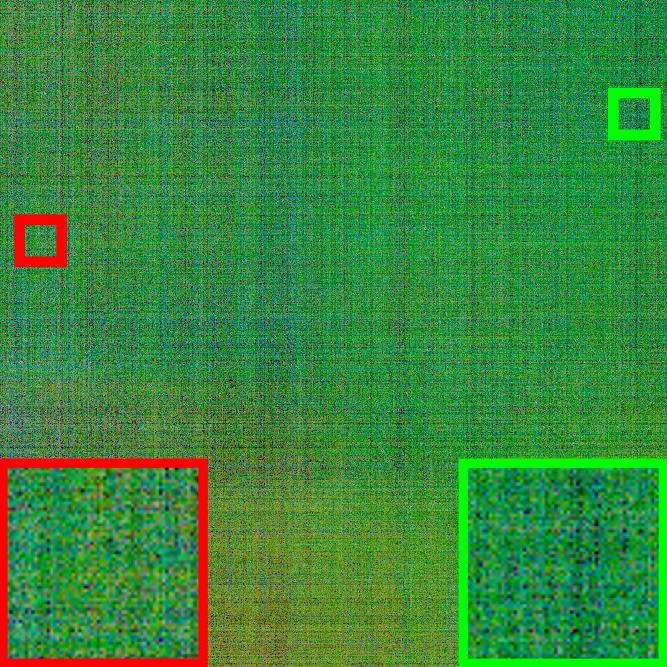}
&
\includegraphics[width=0.586in, height=0.73in]{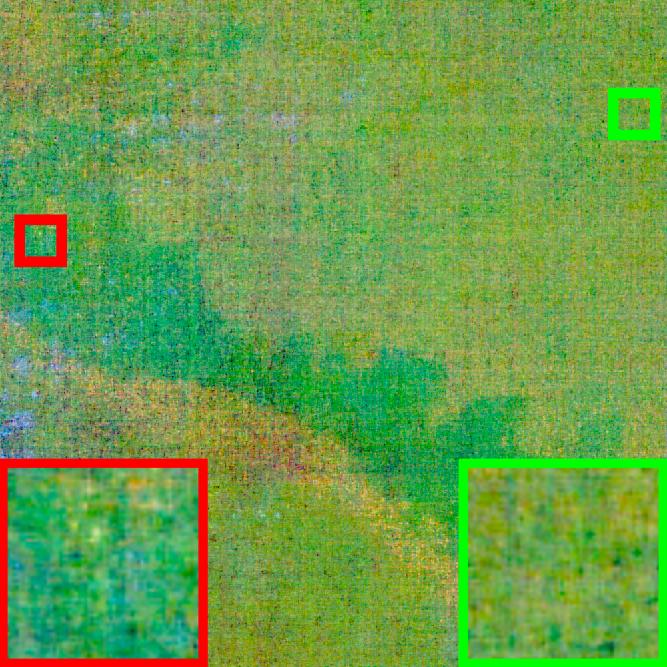}
&
\includegraphics[width=0.586in, height=0.73in]{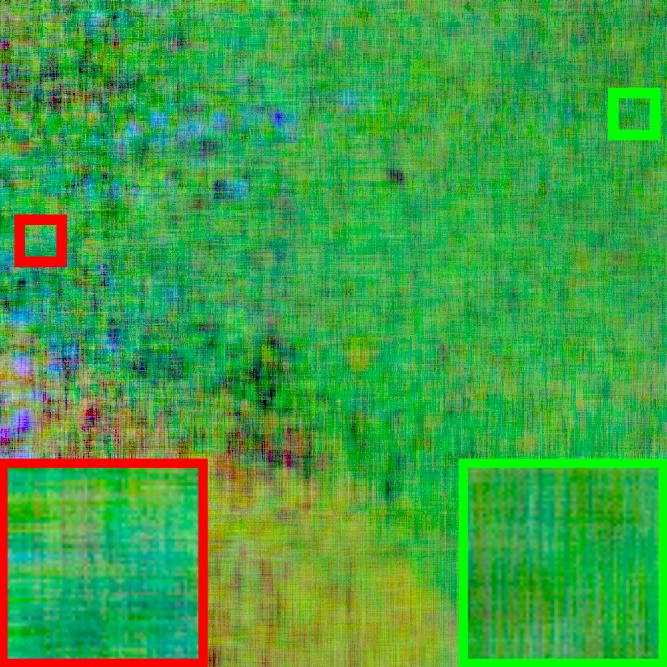}
&
\includegraphics[width=0.586in, height=0.73in]{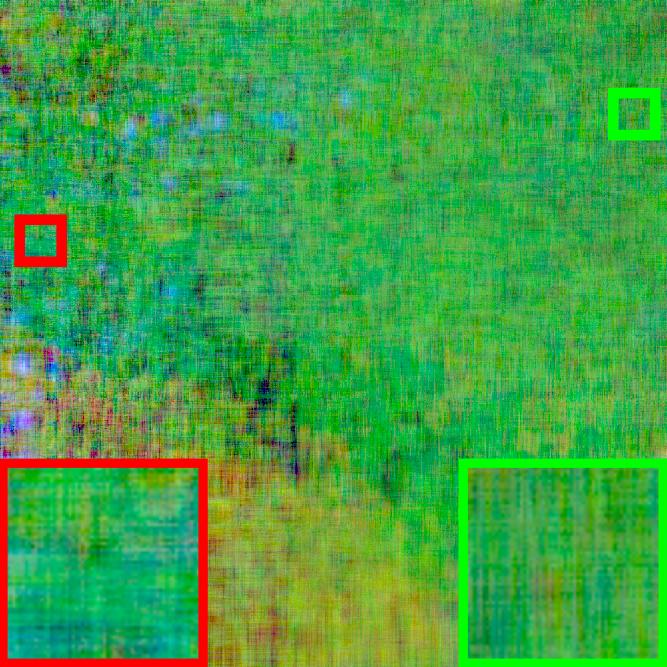}
\\  \toprule
\includegraphics[width=0.586in, height=0.73in]{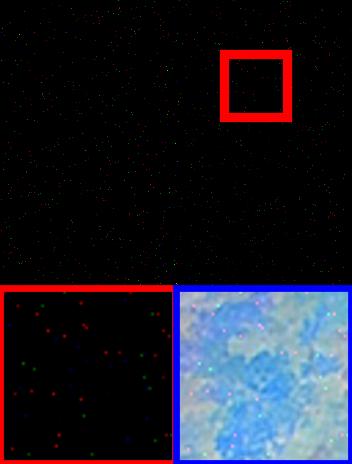}
&
\includegraphics[width=0.586in, height=0.73in]{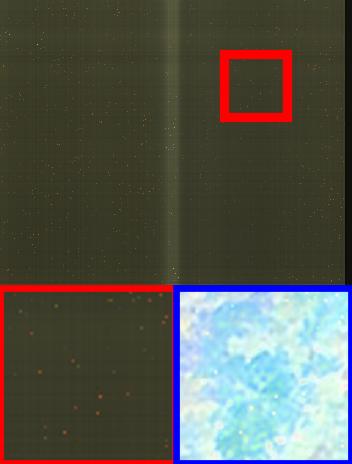}
&
\includegraphics[width=0.586in, height=0.73in]{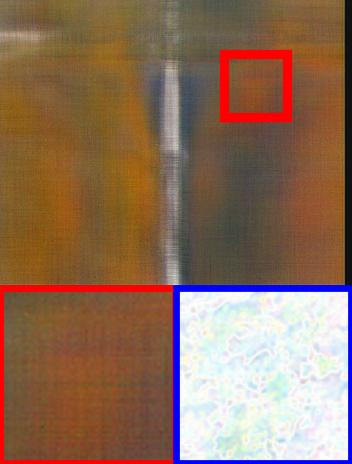}
&
\includegraphics[width=0.586in, height=0.73in]{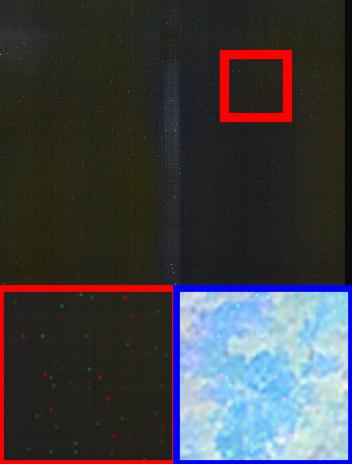}
&
\includegraphics[width=0.586in, height=0.73in]{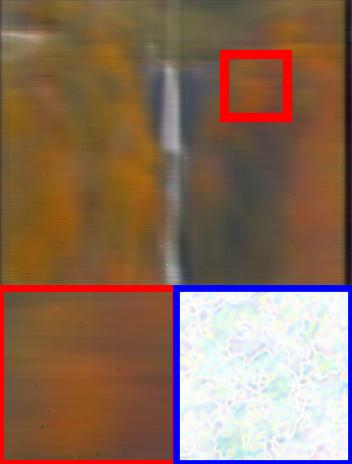}
&
\includegraphics[width=0.586in, height=0.73in]{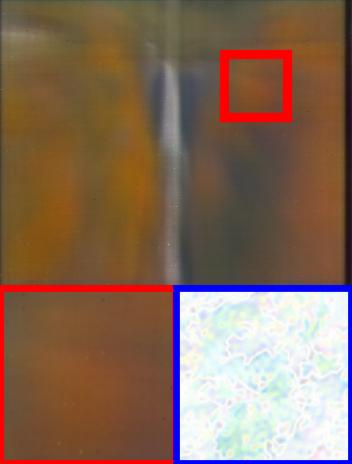}
&
\includegraphics[width=0.586in, height=0.73in]{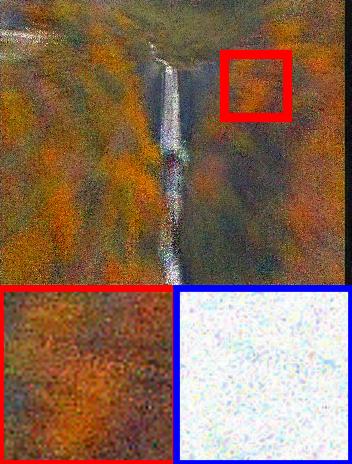}
&
\includegraphics[width=0.586in, height=0.73in]{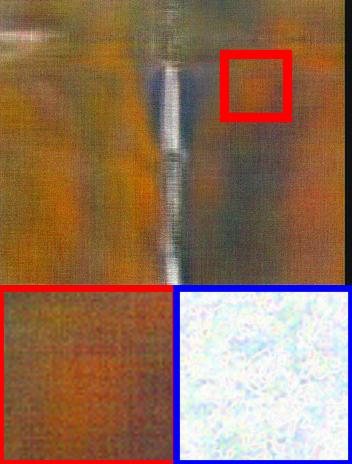}
&
\includegraphics[width=0.586in, height=0.73in]{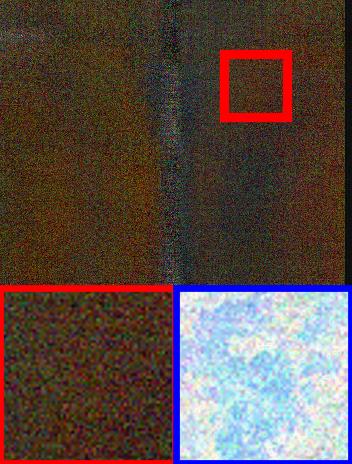}
&
\includegraphics[width=0.586in, height=0.73in]{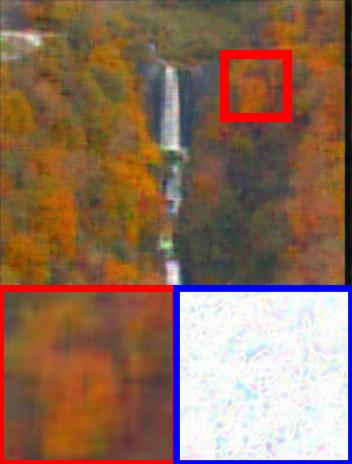}
&
\includegraphics[width=0.586in, height=0.73in]{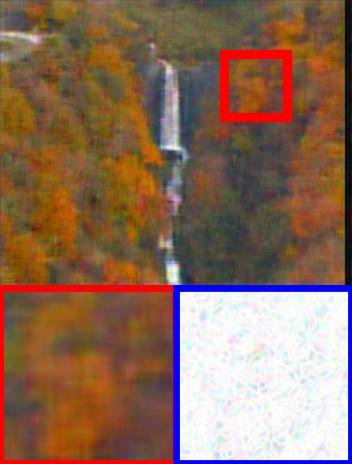}
&
\includegraphics[width=0.586in, height=0.73in]{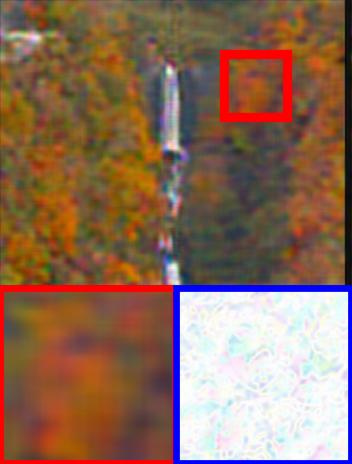}
\\  \toprule

(a)   &
  (b)  & (c) &
(d) & (e)
 &(f)& (g) &
 (h) &
 (i)
 &  (j)
  &
 (k) &(l)
\end{tabular}
% \vspace{-0.4cm}
\vspace{-0.15cm}
\caption{
Visual comparison of various LRTC methods on %for the recovery.
 MSIs (top, SR=$1\%$), MRIs (middle, SR=$1\%$), and CVs (bottom, SR=$0.5\%$).
%inpainting.
%From top to bottom, %the parameter pair $(SR, NR)$ are
%the sampling ratios are all  set to be  $1\%$.
%(0.1, 0.5), (0.05, 0.5) and (0.01, 0.5), respectively.
%Top row: the (5, 1)-th frame of Landsat-7. Middle row: the (2, 6)-th frame of
%SPOT-5. Bottom row: the (3, 5)-th frame of T22LGN.
From left to right:
(a) Observed,
(b)  TRNN, % \cite{yu2019tensor},
 (c) HTNN, % \cite{qin2022low},
 (d) METNN,  % \cite{feng2023multiplex},
  (e) MTTD, % \cite{liu2024revisiting},
   (f) WSTNN, % \cite{zheng2020tensor44},
    (g) EMLCP-LRTC, % \cite{zhang2023tensor},
    (h) GTNN-HOC, % \cite{wang2024low2222},
     (i) t-$\epsilon$-LogDet, % \cite{ yang2022355}, %-Fast,
  (j)  TCTV-TC, %  \cite{wang2023guaranteed},
   (k) GNHTC,
    (l) R1-GNHTC.}
\vspace{-0.59cm}
\label{fig-lrtc-msimsimsi}
\end{figure*}

\vspace{-0.3cm}
% unquantized and quantized tensor completion,

\subsection{\textbf{%Comparative
Experiments 2:
 Noisy Tensor Completion in
%Noisy Tensor Completion
Unquantized %and quantized
Tensor Recovery %---
}}

%{\textbf{%Comparative
%A-1: Noisy Tensor Completion
%Unquantized %and quantized
%Tensor Recovery
%}}

   \textbf{Competing Algorithms:}
   We %mainly %apply the proposed method (HWTSN+w‘q) and its two accelerated versions to several real-world applications, and also
   compare the proposed method (GNRHTC) and its  two % randomized
   accelerated versions %named R-GNRHTC
with several  RLRTC approaches:
%%%%%%%%%%%%%%%%%%%%%%%%%%%%%%%%%%%%%%%%%%%%%%%%%%%%%%%%%%%
TRNN  \cite{huang2020robust},
%%%%%%%%%%%%%%%%%%%%%%%%%%%%%%%%%%%%%%%%%%%%%%%%%%%%%%%%%%%
TTNN  \cite{song2020robust},   TSPK \cite{lou2019robust},  TTLRR \cite{yang2022robust},
LNOP \cite{chen2020robust}, NRTRM \cite{qiu2021nonlocal},
%%%%%%%%%%%%%%%%%%%%%%%%%%%%%%%%%%%%%%%%%%%%%%%%%%%%%%%%%%%%%%%%%%%%%%%%
HWTNN \cite{qin2021robust}, HWTSN \cite{qin2023nonconvex},  randomized HWTSN (R-HWTSN) %-Fast
\cite{qin2023nonconvex}, and TCTV-RTC \cite{wang2023guaranteed}.
%
%%%%%%%%%%%%%%%%%%%%%%%%%%%%%%%%%%%%%%%%%%%%%%%%%%%%%%%%%%%%%%%%%%%%%%%%%%%%%%%%%%%%%%%%%%%%%%%%%%%%%%%%%%%%%%%%%%%%%%%%%%%%%%%%%%
It is noteworthy that the T-CTV regularizer was solely applied in developing high-order LRTC and TRPCA models in reference \cite{wang2023guaranteed}. %In this paper, to
To highlight the advantages of non-convex regularization over convex regularization within the context of a %jointly coupled
joint $\textbf{L}$+$\textbf{S}$ priors %representation
strategy, we additionally include % included %devise %extend
the baseline TCTV-RTC model, which can be viewed as a degenerate version of model
%
 %11 a=b=L1.
(\ref{orin_nonconvex}) when  $\Phi(\cdot) = \psi(\cdot)=\ell_{1}$.

In our proposed randomized  algorithms,
the versions coupled with fixed-rank and fixed-accuracy LRTA strategies are named %R$1$-GNHTC/
R$1$-GNRHTC  and % R$2$-GNHTC/
R$2$-GNRHTC, respectively.
Since the evaluation metrics %numerical results
obtained by  %R$1$-GNRHTC and R$2$-GNRHTC %
two randomized  methods
are very similar, we only present the visual results %outputs %
of %R$1$-GNHTC/
R$1$-GNRHTC % fixed-rank scheme-based
algorithm.

\subsubsection{\textbf{Results and Analysis %Results
on Third-Order HSIs and MRIs datasets}}\label{hsi-mri}

In term of quantitative evaluation indicators (e.g., average PSNR, SSIM, RSE and CPU Time),
Table  \ref{table-mri-hsi} % and \ref{table-hsihsi}
  reports the restoration results
%\label{table-hsihsi} %\label{table-mri}
 of various RLRTC methods on MRIs datasets and HSIs datasets. %, respectively.
From these results, we can observe that:
%1) Compared with other compared algorithms (some of which only utilize global low-rankness),
%T-CTV and GNRHTC, which simultaneously incorporate global low-rank and local smoothness priors, achieve better recovery performance;
 %
\textbf{1)} In comparison to other compared algorithms, some of which rely solely on global low-rankness, TCTV-RTC %T-CTV
and GNRHTC exhibit superior recovery performance by integrating both global low-rank and local smoothness priors;
 %
 %
 %The non-convex GNRHTC method surpasses the convex T-CTV method by approximately 1-2 dB in PSNR, highlighting the superiority of non-convex approaches in high-order tensor modeling;
 %
 \textbf{2)} The proposed %nonconvex
 GNRHTC method gains approximately $1$-$2$ dB in PSNR over the convex TCTV-RTC %T-CTV
 method,
 demonstrating the advantage of nonconvex  regularization scheme in high-order tensor modeling;
 \textbf{3)} For the proposed RLRTC method, while incurring a slight loss in accuracy, the version embedding randomized  techniques significantly enhances computational efficiency compared to its deterministic counterpart.
 In other words, the randomized GNRHTC is approximately $3$-$4$ times faster than the deterministic version across various sampling rates and impulse noise ratios.
 Some inpainting examples concerning MRI and HSI datasets are depicted in Figure \ref{fig-medical-mri} and  \ref{fig_hsi}, respectively.
 Compared to other popular %competitors
 peers, the proposed method is able to recover %restore
 %clearer and
 more distinct detail and contour information, which is especially evident in low-sampling scenarios (e.g., $1\%$ and $5\%$).

%%%%%%%%%%%%%%%%%%%%%%%%%%%%%%%%%%%%%%%%%%%%%%%%%%%%%%%%%%%%%%%%%%%%%%*%%%
\begin{table*} %[tp]
\renewcommand{\arraystretch}{0.7}
\setlength\tabcolsep{3pt}
  \centering
  \caption{
  %The PSNR values and  running time  output by various LRTC  methods for different color videos.
  Quantitative evaluation %results %PSNR, SSIM, RSE and  CPU Time (Second)
  PSNR, SSIM, RSE and  CPU Time (Second)
  of the proposed and compared %various
  RLRTC  methods on third-order
  HSIs
  and
   MRI datasets.
  }
  \vspace{-0.15cm}
  \label{table-mri-hsi}
  \scriptsize
 % \tiny
% \footnotesize
  \begin{threeparttable}
   % \begin{tabular}{c | c| c | c| c|c|c|c| c|c|c|c| c|c|c|c| c|c}
     \begin{tabular}{cccc cccc cccc  c c}
   % \hline
    \Xhline{1pt}
    %\multirow{2}{*}
   \tabincell{c}   {SR %\\
   $\textbf{\&}$ % \\
   NR}&
   % \multirow{2}{*}
   \tabincell{c} {Evaluation\\ Metric}&
   %%%%%%%%%%%%%%%%%%%%%%%%%%%%%%%%%%%%%%%%%%%%%%%%%%%%%%%%%%%%%%%%%%%
%%%%%%%%%%%%%%%%%%%%%%%%%%%%%%%%%%%%%%%%%%%%%%%%%%%%%%%%%%%
\tabincell{c}{TTNN\\ \cite{song2020robust}}&
\tabincell{c} {TSPK \\\cite{lou2019robust}}&
   \tabincell{c}{TTLRR \\ \cite{yang2022robust}}&
% yang2022robust, liu2024fully, lou2019robust
\tabincell{c}{LNOP\\ \cite{chen2020robust}}&
 \tabincell{c}{NRTRM\\ \cite{qiu2021nonlocal}}&
%%%%%%%%%%%%%%%%%%%%%%%%%%%%%%%%%%%%%%%%%%%%%%%%%%%%%%%%%%%%%%%%%%%%%%%%
\tabincell{c}{HWTNN  \\\cite{qin2021robust}}&
 \tabincell{c} {HWTSN\\\cite{qin2023nonconvex}}&
 \tabincell{c} {R-HWTSN  \\\cite{qin2023nonconvex}}&
 \tabincell{c}{TCTV-RTC  \\\cite{wang2023guaranteed}}
 %\cite{wang2023guaranteed}
 & \textbf{GNRHTC}&
 \textbf{R1-GNRHTC} & \textbf{R2-GNRHTC}
   % PSNR&Time (s)&PSNR&Time (s)&PSNR&Time (s)&PSNR&Time (s)&PSNR&Time (s)
   % &PSNR&Time (s)&PSNR&Time (s)&PSNR&Time (s)

    \cr

  % % \hline
%    %\hline
%    \Xhline{1pt}
%%\multicolumn{13}{c}{Results on Third-Order  HSIs}\\
%\hline
%     \hline
% third-order

 \Xhline{1pt}
\multicolumn{14}{c}{\textbf{\textit{Results on Third-Order  MRIs}}}\\
\hline
     \hline

  \multirow{4}{*}{    \tabincell{c}   {SR=$0.1$  \\ $\textbf{\&}$ \\NR= $0.5$}  } &MPSNR
 & 23.8409& 22.7573& 24.4542& 23.8835& 24.6133& 25.9725& 26.9158& 27.0206& 30.4559& \textcolor[rgb]{1,0,0}{\textbf{31.5027}}& \textcolor[rgb]{0,0,1}{\textbf{31.2989}} & 30.9439
   \cr
    %%
    % \qquad	&MPSNR	& & & & & & & & & & & & & & & & \cr
    %%

   \qquad	&MSSIM	&0.7115&0.6461&0.5349&0.5235&0.7222&0.6353&0.7327&0.7615&{{0.8972}}&\textcolor[rgb]{0,0,1}{\textbf{0.9039}}&0.8967&
    \textcolor[rgb]{1,0,0}{\textbf{0.9222}}
    \cr
   \qquad	&MRSE	 &0.2305&0.2619&0.2112&0.2340&0.2116&0.1860&0.1683&0.1658&0.1076&\textcolor[rgb]{1,0,0}{\textbf{0.0954}}&\textcolor[rgb]{0,0,1}{\textbf{0.0978}} & 0.1019
   \cr
    \qquad	&MTime	&\textcolor[rgb]{0,0,1}{\textbf{255.368}}&420.435&509.757&324.997&323.459&366.837&471.374&\textcolor[rgb]{1,0,0}{\textbf{174.369}} %257.8133
    &851.556&991.191&257.378  & 287.212 \cr

     \hline
     \multirow{4}{*}{    \tabincell{c}   {SR=$0.05$  \\ $\textbf{\&}$ \\NR=$0.5$}  } &MPSNR
  & 20.4528& 19.8246& 20.4787& 19.8633& 20.4842& 22.6056& 22.7329& 22.5098& 26.5574& \textcolor[rgb]{1,0,0}{\textbf{28.0649}}& \textcolor[rgb]{0,0,1}{\textbf{27.9266}}  & 27.8158
    \cr
    %%
    % \qquad	&MPSNR	& & & & & & & & & & & & & & & & \cr
    %%
   \qquad	&MSSIM	&0.5547&0.4519&0.3377&0.2569&0.6274&0.5204&0.5750&0.5757&0.8152&\textcolor[rgb]{0,0,1}{\textbf{0.8362}}&{{0.8167}}
   & \textcolor[rgb]{1,0,0}{\textbf{0.8478}}
   \cr
   \qquad	&MRSE	&0.3379&0.3621&0.3322&0.3623&0.3370&0.2634&0.2668&0.2736&0.1669&\textcolor[rgb]{1,0,0}{\textbf{0.1410}}&\textcolor[rgb]{0,0,1}{\textbf{0.1434}} &
    0.1455
   \cr
    \qquad	&MTime	&278.546&420.158&522.046&320.049&323.467&364.056&452.823& \textcolor[rgb]{1,0,0}{\textbf{164.333}} %236.5717
    &831.944&988.753&\textcolor[rgb]{0,0,1}{\textbf{238.458}} & 269.629\cr

  % \hline
%     \hline

%\hline
 %    \hline

\hline
    \hline
   % \Xhline{1pt}
\multicolumn{14}{c}{\textbf{\textit{Results on Third-Order  %CVs
HSIs}}}\\
\hline
     \hline

     \multirow{4}{*}{    \tabincell{c}   {SR=$0.05$  \\ $\textbf{\&}$ \\NR=$0.5$}  } &MPSNR
  &    22.6243&22.1360&23.3060&22.8259&22.5062&22.9506&       24.5271   &23.4906&\textcolor[rgb]{0,0,1}{\textbf{27.9127}}& \textcolor[rgb]{1,0,0}{\textbf{28.7180}} %27.7180
  &26.7984       &    26.3515 \cr
   \qquad	&MSSIM	& 0.5534&0.4655&0.4767&0.4402&0.6070&0.5504&0.5651   &0.5645&\textcolor[rgb]{0,0,1}{\textbf{0.7752}}& \textcolor[rgb]{1,0,0}{\textbf{0.8166}} %0.7666
   &0.7089 & 0.6960
   \cr
   \qquad	&MRSE	& 0.4057&0.4278&0.3724&0.3962&0.4122&0.3934& 0.3271    &0.3693&\textcolor[rgb]{0,0,1}{\textbf{0.2216}}& \textcolor[rgb]{1,0,0}{\textbf{0.2068}} %0.2268
   &
   0.2540 & 0.2674 \cr
    \qquad	&MTime	& 3554.179&6127.126&6956.448&4962.299&4841.568&4022.126&4987.924&\textcolor[rgb]{1,0,0}{\textbf{1998.454}}&8635.938&10778.794&
    \textcolor[rgb]{0,0,1}{\textbf{2735.866}} & 2990.031\cr

     \hline
     \multirow{4}{*}{    \tabincell{c}   {SR=$0.01$  \\ $\textbf{\&}$ \\NR=$0.5$}  } &MPSNR
  &  19.4161&18.7400&18.3855&17.1151&19.1376&17.1177&17.3052&17.2905&{{22.9375}}&  \textcolor[rgb]{1,0,0}{\textbf{23.1347}} &     22.5751  &
  \textcolor[rgb]{0,0,1}{\textbf{23.0700}}
  \cr
   \qquad	&MSSIM	& 0.4237&0.3595&0.2508&0.4554&0.4230&0.3011&0.2752&0.2740&{{0.5295}}& \textcolor[rgb]{1,0,0}{\textbf{0.5456}} &  0.4101&
   \textcolor[rgb]{0,0,1}{\textbf{0.5443}}
   \cr
   \qquad	&MRSE & 0.5829&0.6303&0.6563&0.7597&0.6019&0.7600&0.7439&0.7450&{{0.3895}}&\textcolor[rgb]{1,0,0}{\textbf{0.3835}}   &0.4056	&
   \textcolor[rgb]{0,0,1}{\textbf{0.3864}} \cr
    \qquad	&MTime	&
%4608.7448
3608.7448
&5997.314&7336.936&4641.734&4631.646&3939.969&4675.494&\textcolor[rgb]{1,0,0}{\textbf{1683.462}}&9034.219&10656.386&\textcolor[rgb]{0,0,1}{\textbf{2676.158}}  & % 2614.831 %
2844.831
\cr

   %\hline
   \hline
     \hline

    %  \Xhline{1pt}

      \Xhline{1pt}
    \end{tabular}
    \end{threeparttable}
    \vspace{-0.3cm}
\end{table*}

%%%%%%%%%%%%%%%%%%%%%%%%%%%%%%%%%%%%%%%%%%%%%%%%%%%%%%%%%%%%%%%%%%%%%%%%%%%%%%%%%%%%%%%%%%%%%%%%%%%%%%%%%%%%%%%%%%%%%%%%%%%%%%%%%%%%%%%%%%%%%%%%%
\begin{figure*} [!htbp]
\renewcommand{\arraystretch}{0.7}
\setlength\tabcolsep{0.43pt}
\centering
\begin{tabular}{ccc  ccc ccc ccc  }%cc ccc  ccc c cc
\centering
\includegraphics[width=0.586012in, height=0.81in]{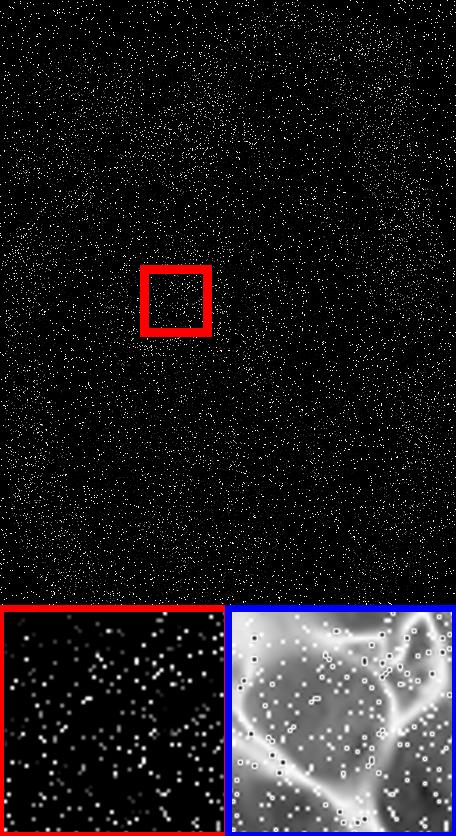}
&
\includegraphics[width=0.586012in, height=0.81in]{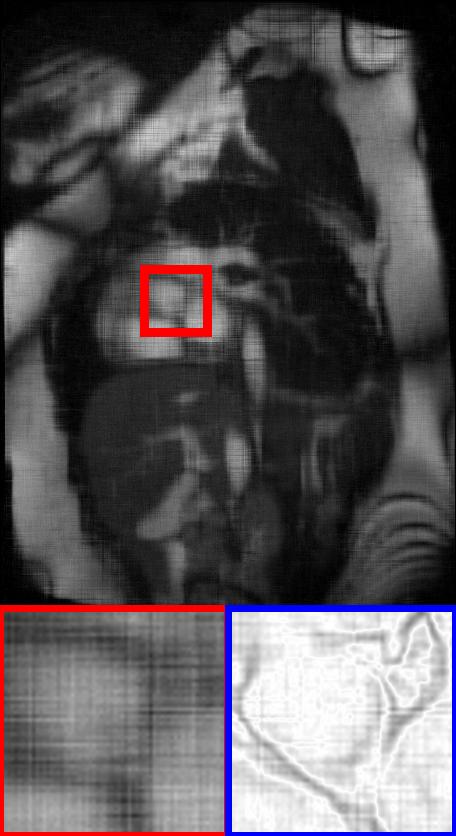}
&
\includegraphics[width=0.586012in, height=0.81in]{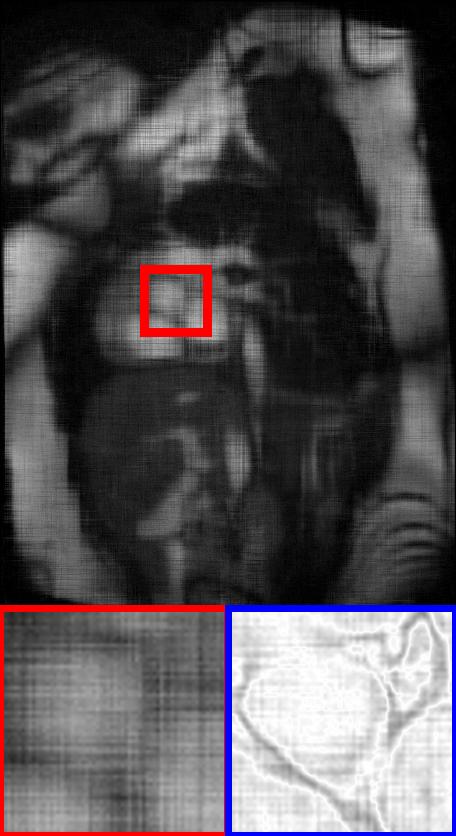}
&
\includegraphics[width=0.586012in, height=0.81in]{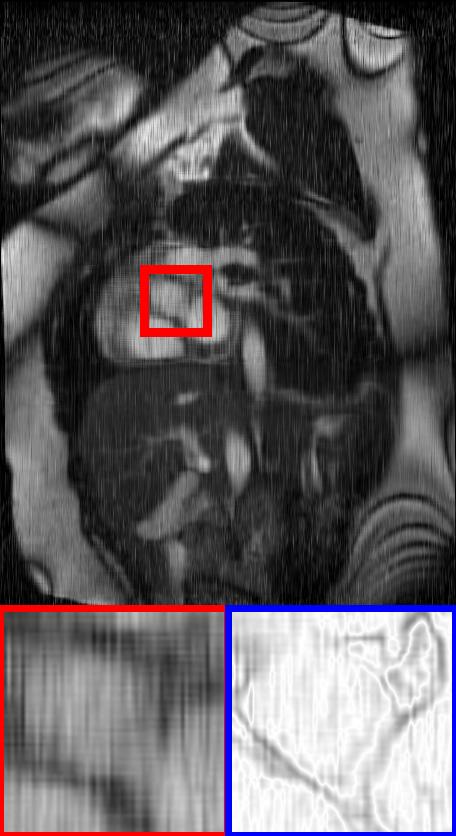}
&
\includegraphics[width=0.586012in, height=0.81in]{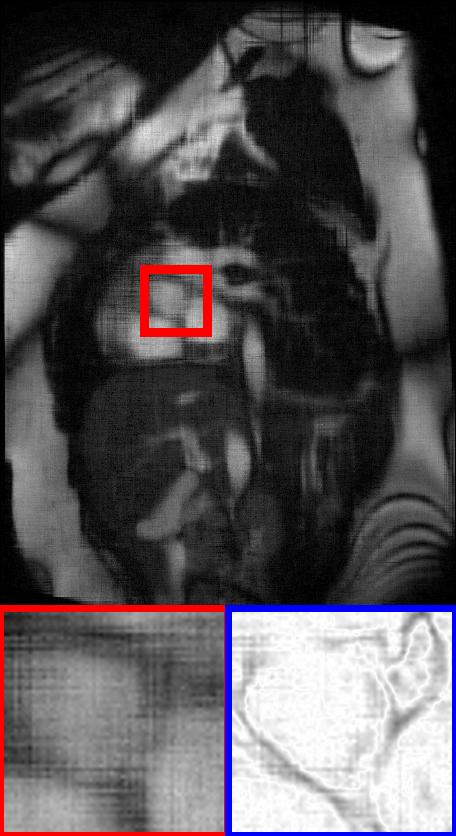}
&
\includegraphics[width=0.586012in, height=0.81in]{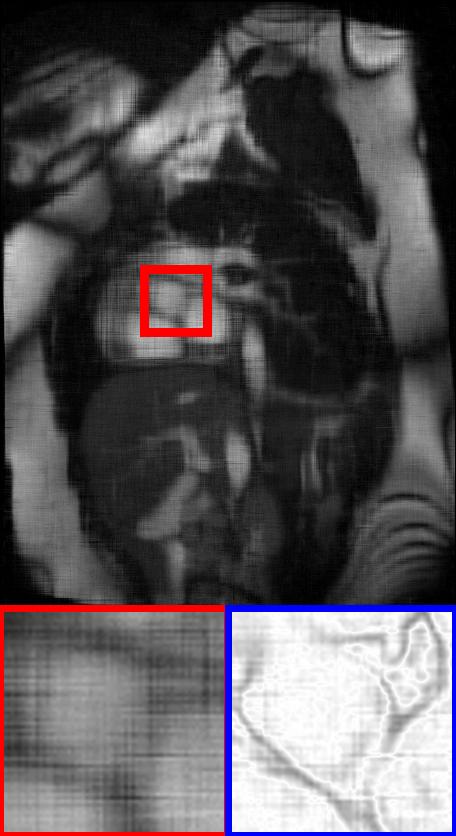}
&
\includegraphics[width=0.586012in, height=0.81in]{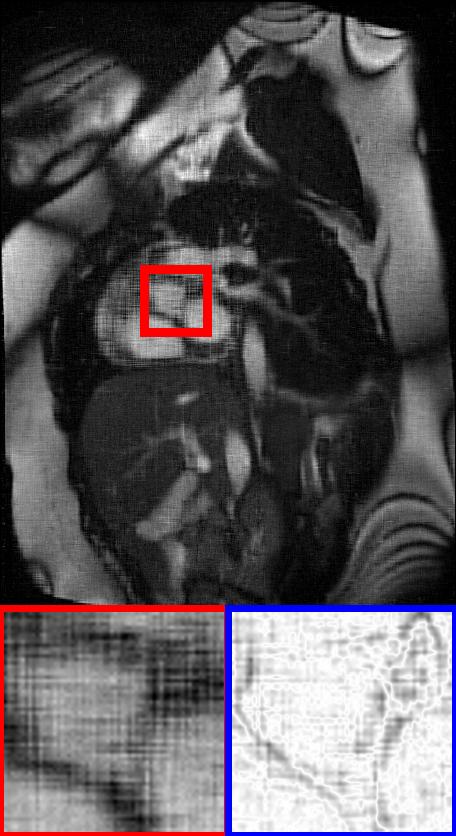}
&
\includegraphics[width=0.586012in, height=0.81in]{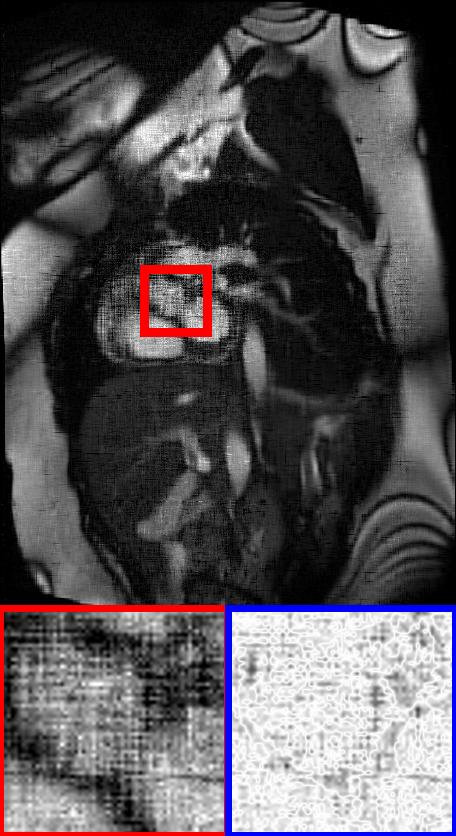}
&
\includegraphics[width=0.586012in, height=0.81in]{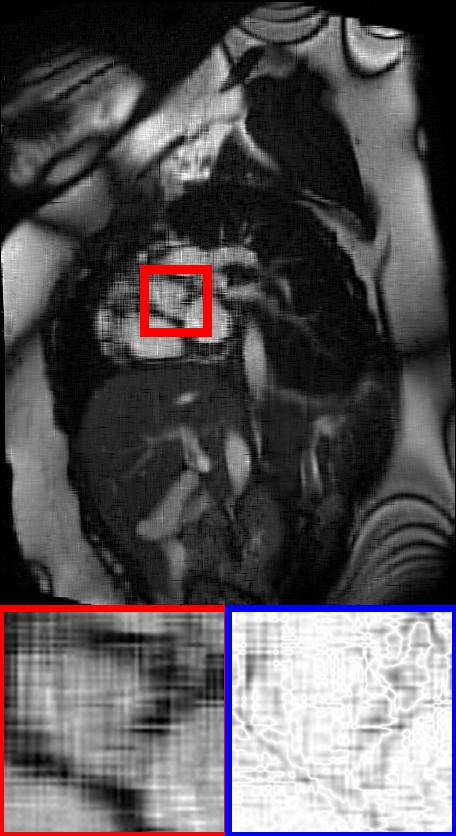}
&
\includegraphics[width=0.586012in, height=0.81in]{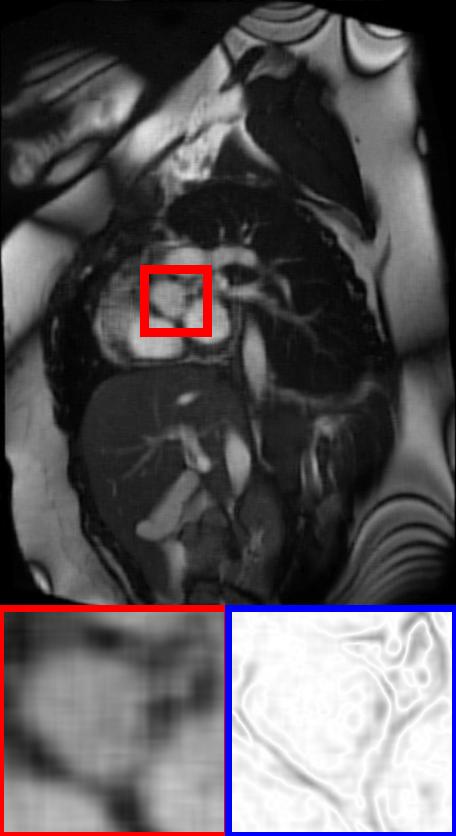}
&
\includegraphics[width=0.586012in, height=0.81in]{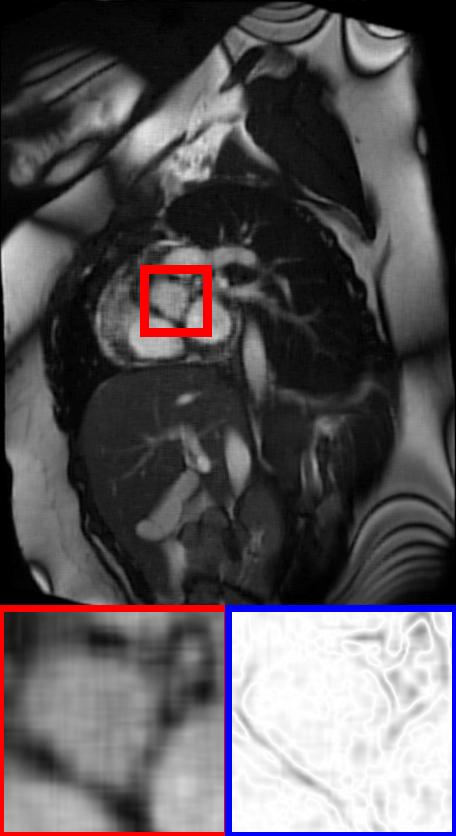}
&
\includegraphics[width=0.586012in, height=0.81in]{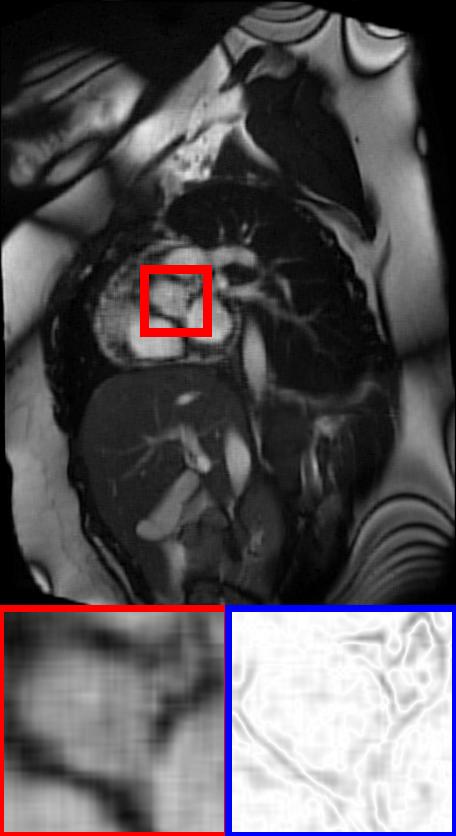}
\\

\includegraphics[width=0.586012in, height=0.81in]{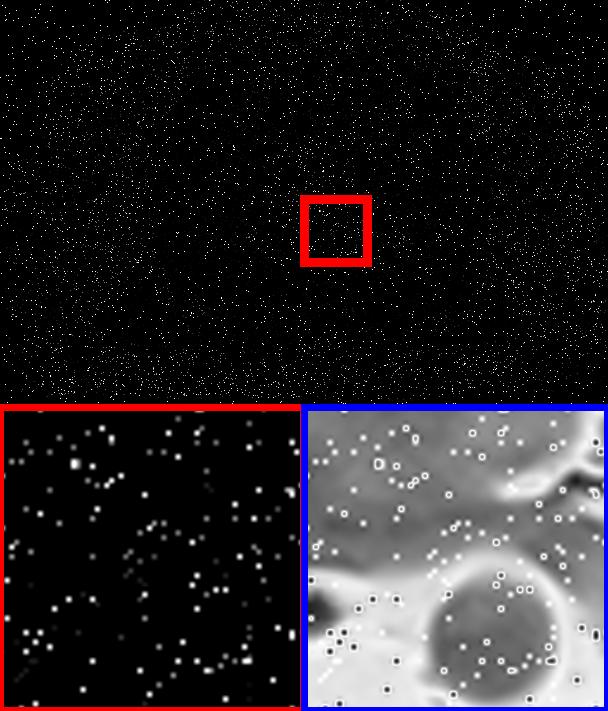}
&
\includegraphics[width=0.586012in, height=0.81in]{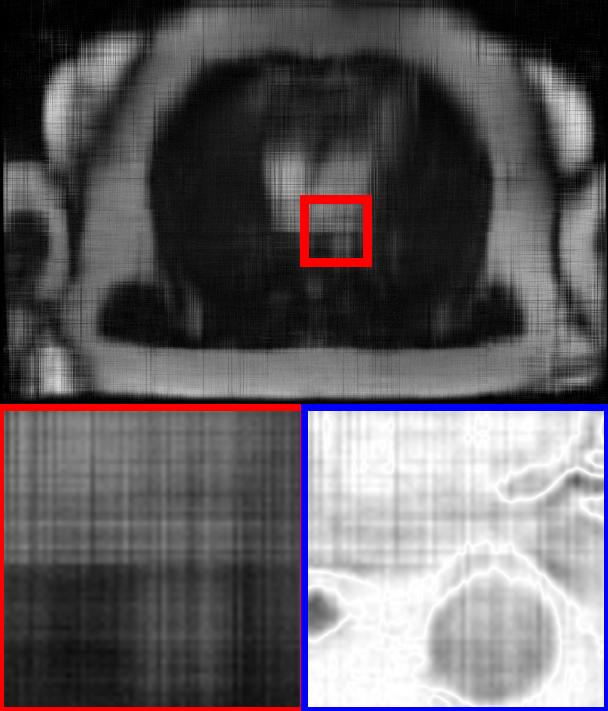}
&
\includegraphics[width=0.586012in, height=0.81in]{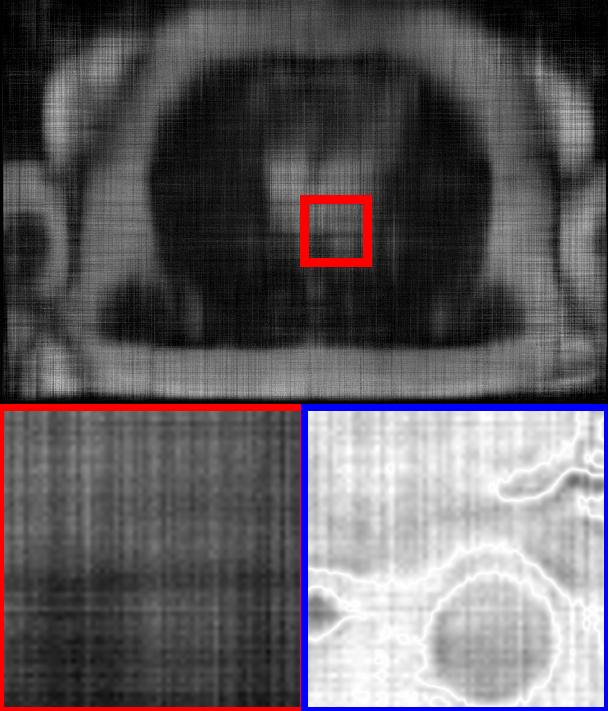}
&
\includegraphics[width=0.586012in, height=0.81in]{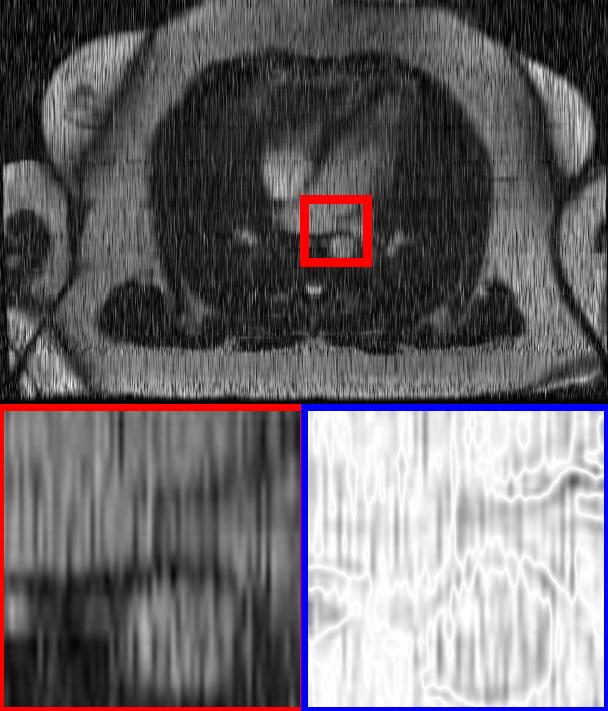}
&
\includegraphics[width=0.586012in, height=0.81in]{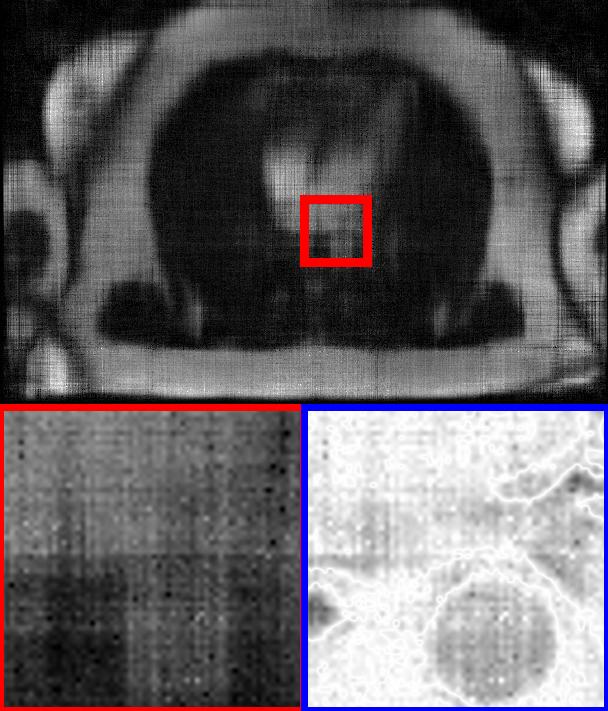}
&
\includegraphics[width=0.586012in, height=0.81in]{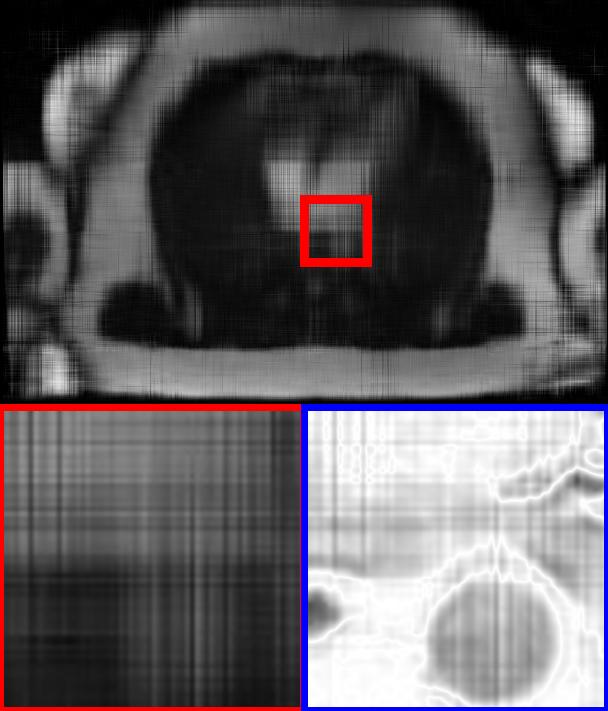}
&
\includegraphics[width=0.586012in, height=0.81in]{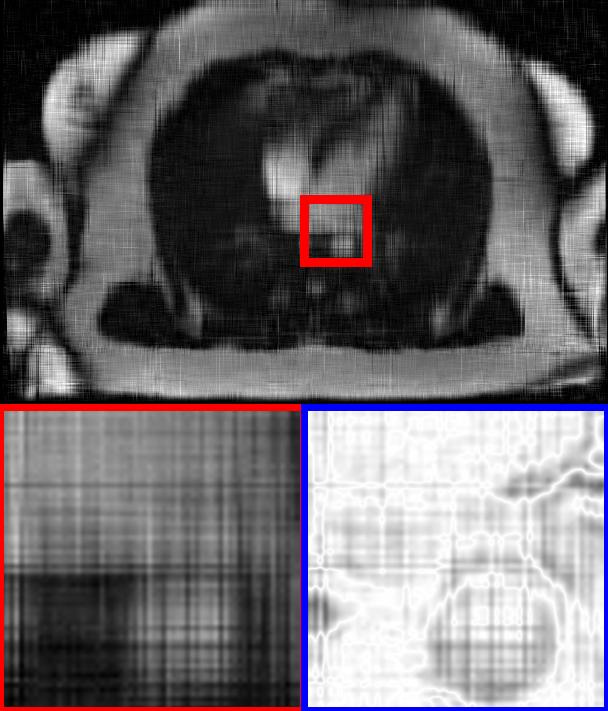}
&
\includegraphics[width=0.586012in, height=0.81in]{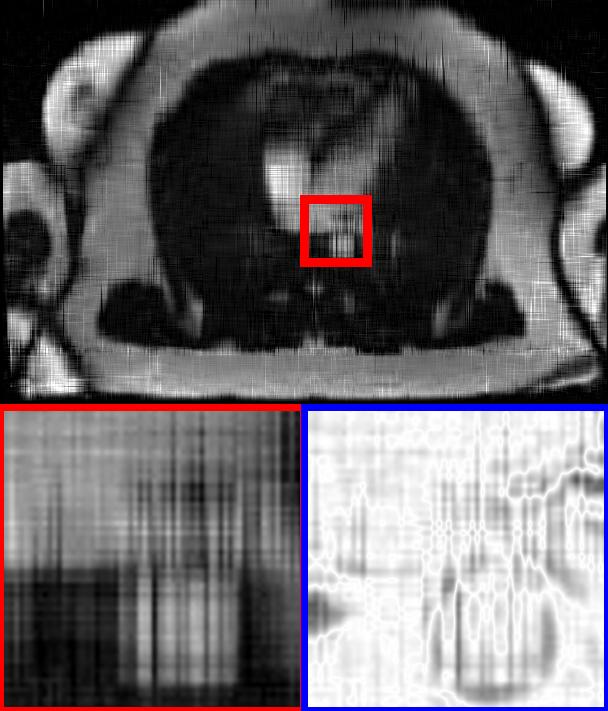}
&
\includegraphics[width=0.586012in, height=0.81in]{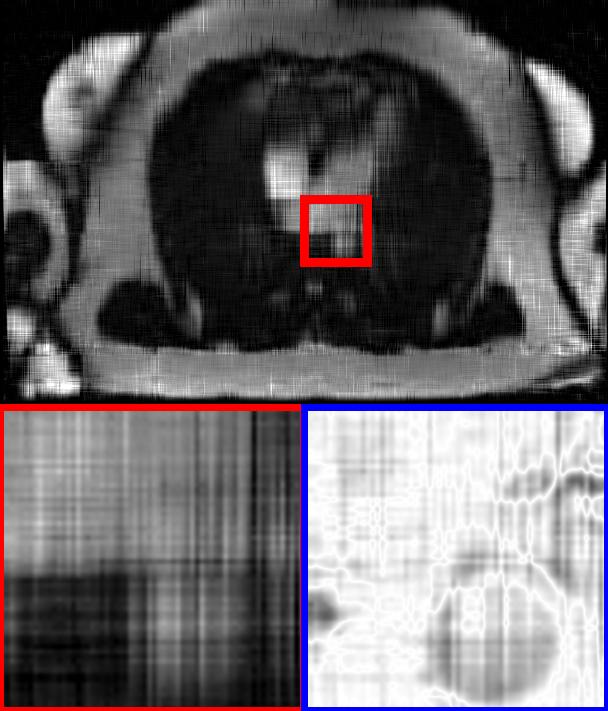}
&
\includegraphics[width=0.586012in, height=0.81in]{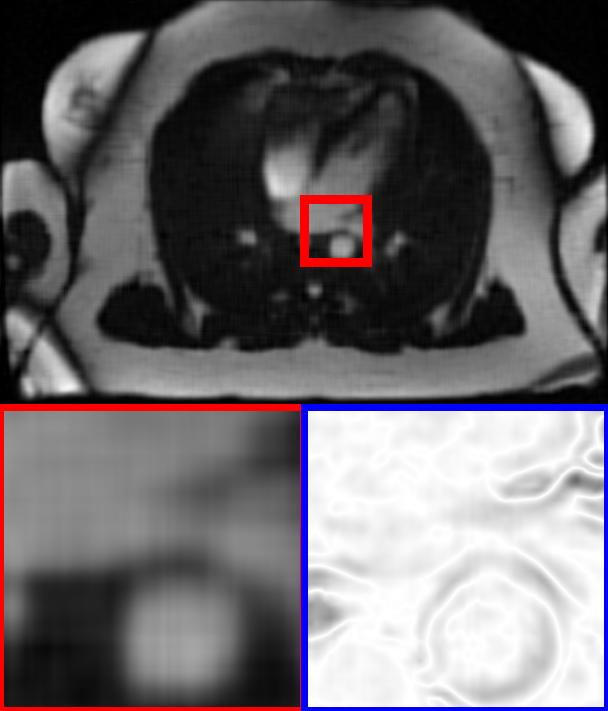}
&
\includegraphics[width=0.586012in, height=0.81in]{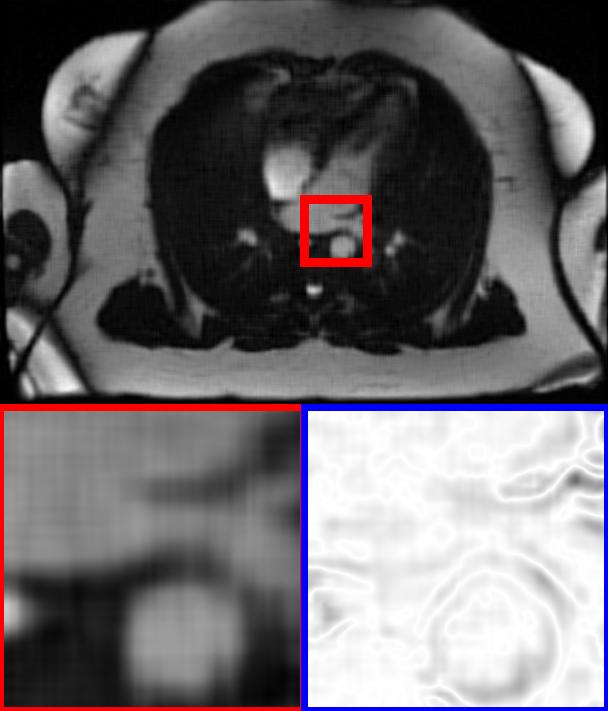}
&
\includegraphics[width=0.586012in, height=0.81in]{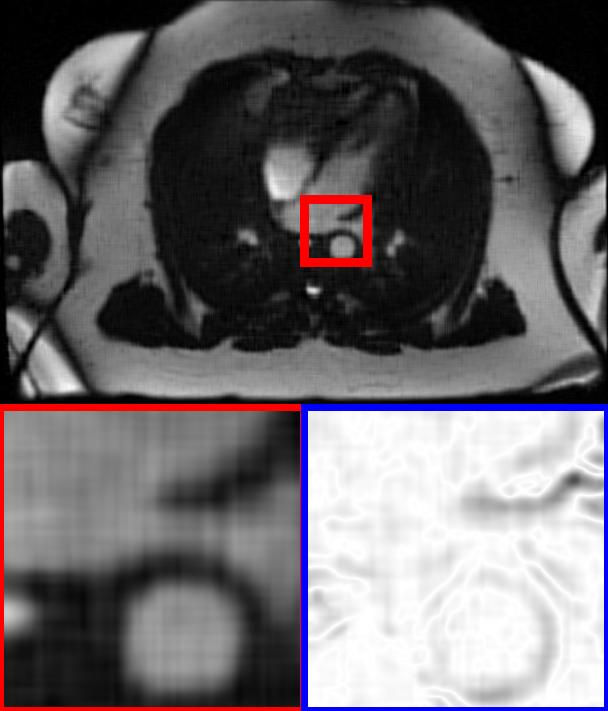}
\\
%TTNN
%[23]
%TSPK
%[20]
%TTLRR
%[24]
%LNOP
%[27]
%NRTRM
%[28]
%HWTNN
%[33]
%HWTSN
%[35]
%HWTSN-Fast
%[35]
%TCTV
%[50]
%GNRHTC R-GNRHTC

%{{{(a) \scriptsize{TTNN}}}}  &
%  {(b) \scriptsize{TSPK}}  & {(c) \scriptsize{TTLRR}}
%&{(d) \scriptsize{LNOP}} & {(e) \scriptsize{NRTRM}}
% &{(f) \scriptsize{HWTNN}}& {(g) \scriptsize{HWTSN}}&
% {(h) \scriptsize{HWTSN-Fast}}&
% {(i) \scriptsize{TTNN}}
% &  {(j) \scriptsize{TTNN}}
%  &
% {(k) \scriptsize{TTNN}} &(l) \scriptsize{TTNN}
(a)   &
  (b)  & (c) &
(d) & (e)
 &(f)& (g) &
 (h) &
 (i)
 &  (j)
  &
 (k) &(l)
\end{tabular}
\vspace{-0.15cm}
\caption{
Visual comparison of various RLRTC methods for  MRI datasets inpainting. From top to bottom, the parameter pair $(SR, NR)$ are
%(0.1, 0.5),
(0.1, 0.5) and (0.05, 0.5), respectively.
%Top row: the (5, 1)-th frame of Landsat-7. Middle row: the (2, 6)-th frame of
%SPOT-5. Bottom row: the (3, 5)-th frame of T22LGN.
From left to right: (a) Observed,
(b)  TTNN, (c) TSPK, (d) TTLRR, (e) LNOP, (f) NRTRM, (g) HWTNN,  (h) HWTSN,  (i) R-HWTSN,  (j)  TCTV-RTC, (k) GNRHTC, (l) R1-GNRHTC.}
\vspace{-0.4cm}
\label{fig-medical-mri} %\label{fig_hsi}
\end{figure*}

\begin{figure*}[!htbp]
\renewcommand{\arraystretch}{0.7}
\setlength\tabcolsep{0.43pt}
\centering
\begin{tabular}{ccc  ccc ccc ccc  }%cc ccc  ccc c cc
\centering

%
%\includegraphics[width=0.586in, height=0.658in]{hsi-visu1-tree2-15/s12} % hsi-visu1-tree2-15
%&
%\includegraphics[width=0.586in, height=0.658in]{hsi-visu1-tree2-15/s1}
%&
%\includegraphics[width=0.586in, height=0.658in]{hsi-visu1-tree2-15/s2}
%&
%\includegraphics[width=0.586in, height=0.658in]{hsi-visu1-tree2-15/s3}
%&
%\includegraphics[width=0.586in, height=0.658in]{hsi-visu1-tree2-15/s4}
%&
%\includegraphics[width=0.586in, height=0.658in]{hsi-visu1-tree2-15/s5}
%&
%\includegraphics[width=0.586in, height=0.658in]{hsi-visu1-tree2-15/s6}
%&
%\includegraphics[width=0.586in, height=0.658in]{hsi-visu1-tree2-15/s7}
%&
%\includegraphics[width=0.586in, height=0.658in]{hsi-visu1-tree2-15/s8}
%&
%\includegraphics[width=0.586in, height=0.658in]{hsi-visu1-tree2-15/s9}
%&
%\includegraphics[width=0.586in, height=0.658in]{hsi-visu1-tree2-15/s10}
%&
%\includegraphics[width=0.586in, height=0.658in]{hsi-visu1-tree2-15/s11} \\

\includegraphics[width=0.586in, height=0.81in]{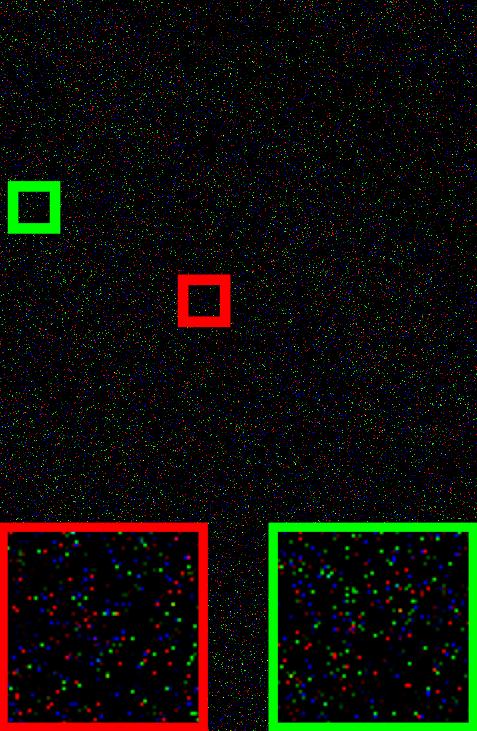}
&
\includegraphics[width=0.586in, height=0.81in]{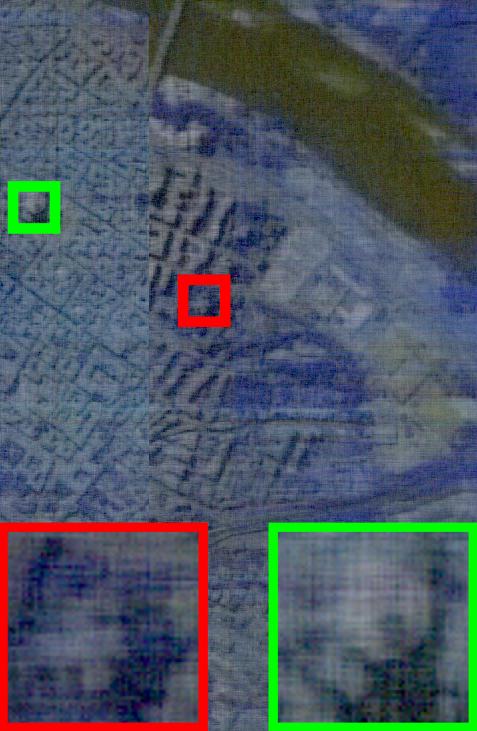}
&
\includegraphics[width=0.586in, height=0.81in]{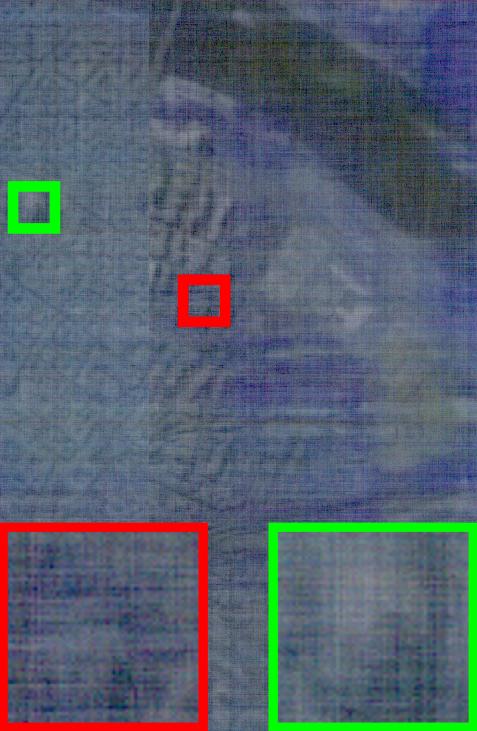}
&
\includegraphics[width=0.586in, height=0.81in]{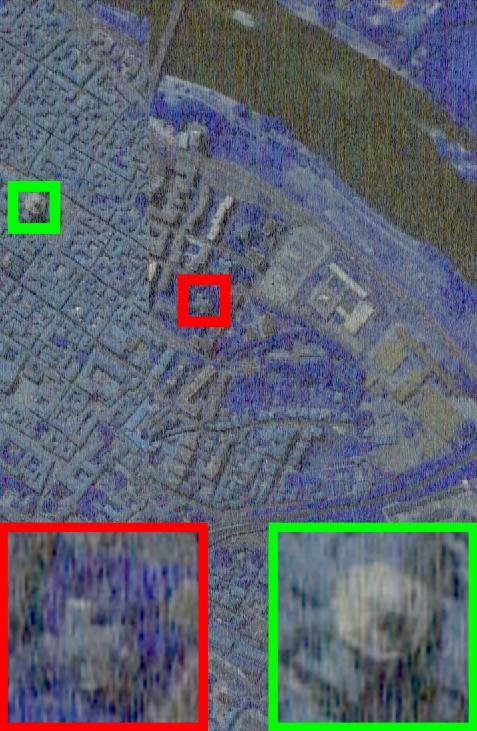}
&
\includegraphics[width=0.586in, height=0.81in]{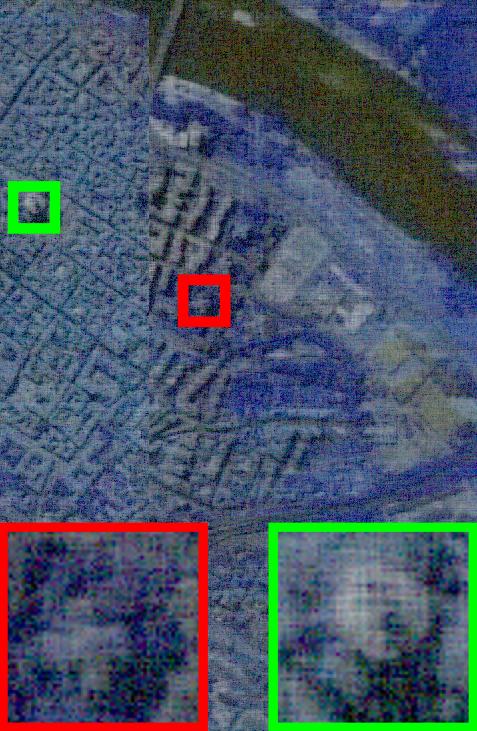}
&
\includegraphics[width=0.586in, height=0.81in]{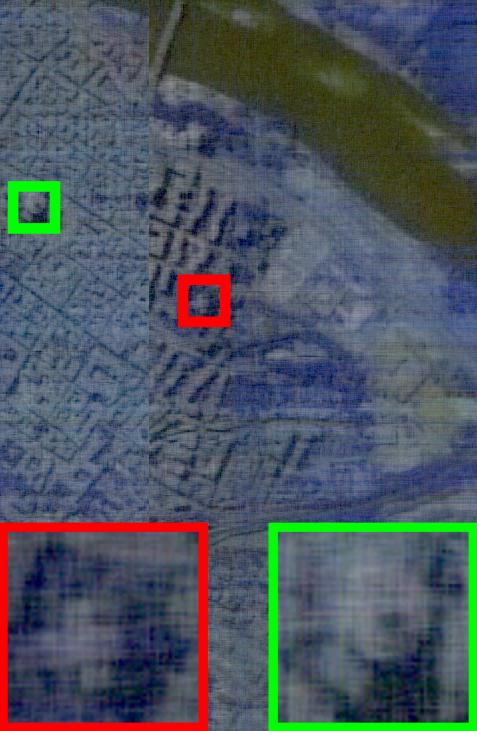}
&
\includegraphics[width=0.586in, height=0.81in]{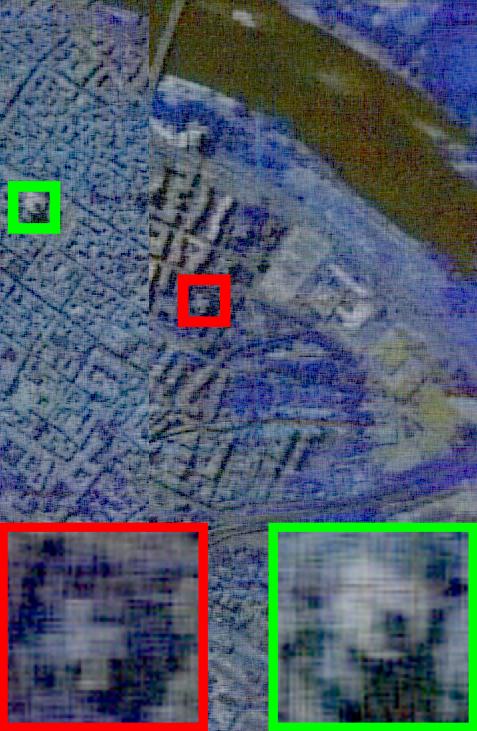}
&
\includegraphics[width=0.586in, height=0.81in]{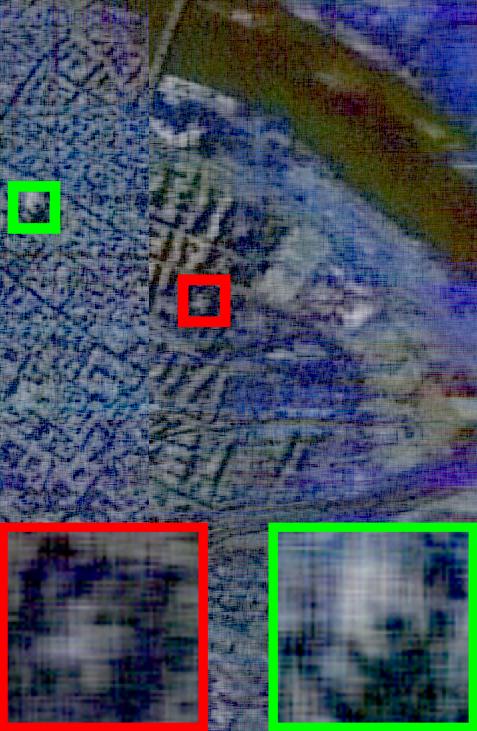}
&
\includegraphics[width=0.586in, height=0.81in]{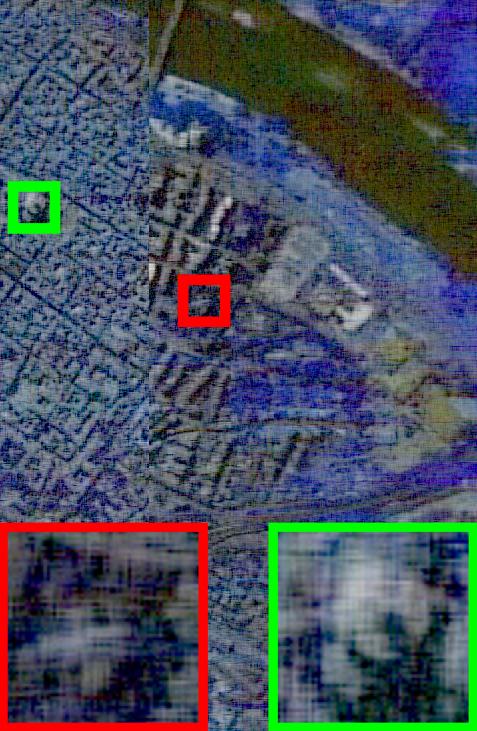}
&
\includegraphics[width=0.586in, height=0.81in]{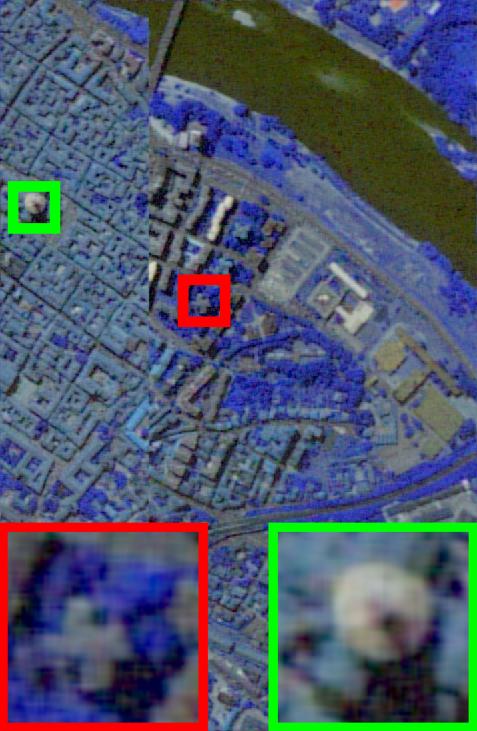}

&
\includegraphics[width=0.586in, height=0.81in]{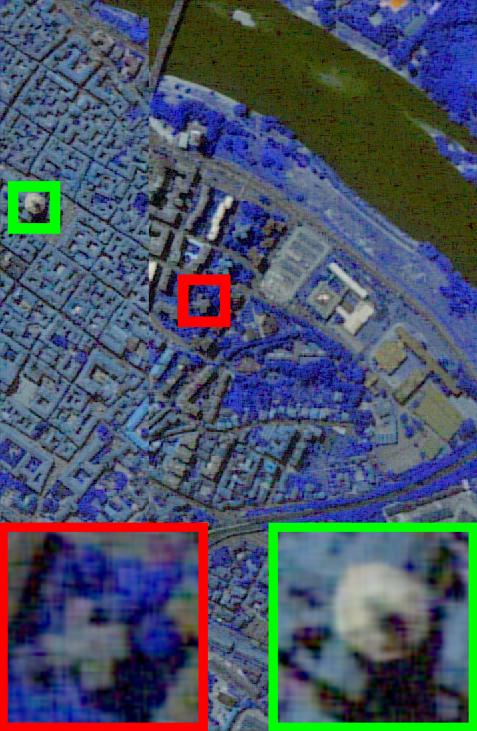}

&
\includegraphics[width=0.586in, height=0.81in]{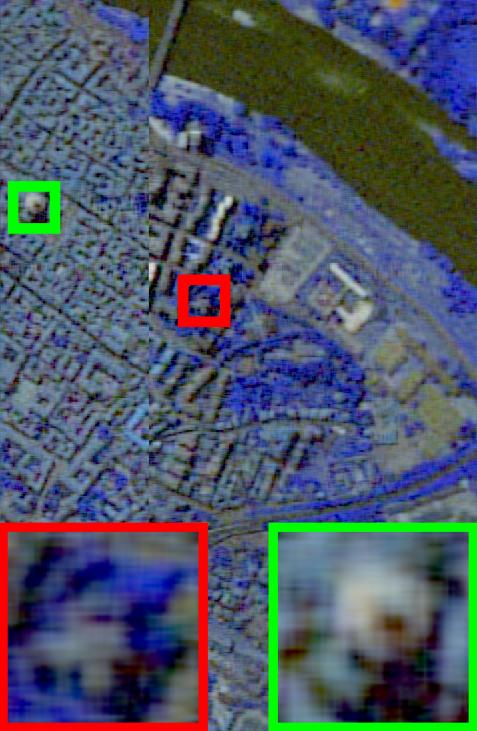}  %hsi-visu1-pavia-055
\\
%
%\includegraphics[width=0.586012in, height=0.81in]{hsi-visu1-tree1-015/s12}
%&
%\includegraphics[width=0.586012in, height=0.81in]{hsi-visu1-tree1-015/s1}
%&
%\includegraphics[width=0.586012in, height=0.81in]{hsi-visu1-tree1-015/s2}
%&
%\includegraphics[width=0.586012in, height=0.81in]{hsi-visu1-tree1-015/s3}
%&
%\includegraphics[width=0.586012in, height=0.81in]{hsi-visu1-tree1-015/s4}
%&
%\includegraphics[width=0.586012in, height=0.81in]{hsi-visu1-tree1-015/s5}
%&
%\includegraphics[width=0.586012in, height=0.81in]{hsi-visu1-tree1-015/s6}
%&
%\includegraphics[width=0.586012in, height=0.81in]{hsi-visu1-tree1-015/s7}
%&
%\includegraphics[width=0.586012in, height=0.81in]{hsi-visu1-tree1-015/s8}
%&
%\includegraphics[width=0.586012in, height=0.81in]{hsi-visu1-tree1-015/s9}
%
%&
%\includegraphics[width=0.586012in, height=0.81in]{hsi-visu1-tree1-015/s10}
%
%&
%\includegraphics[width=0.586012in, height=0.81in]{hsi-visu1-tree1-015/s11}
% \\
 \includegraphics[width=0.586in, height=0.658in]{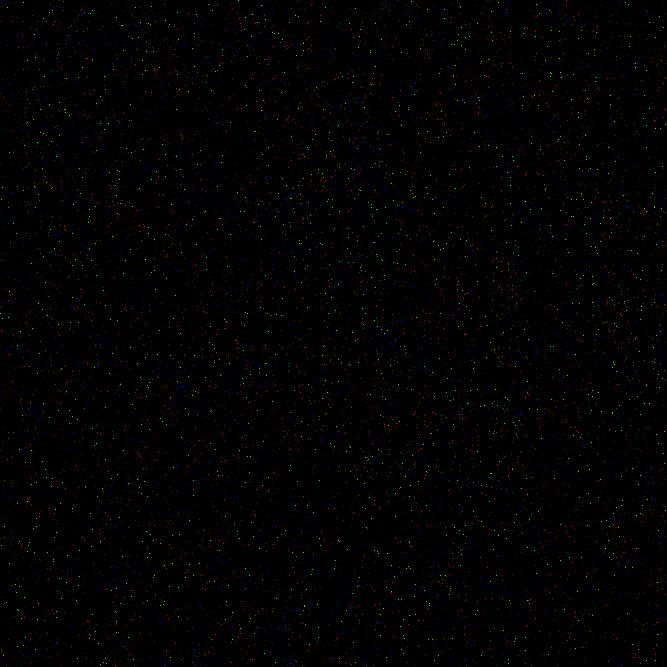}
&
\includegraphics[width=0.586in, height=0.658in]{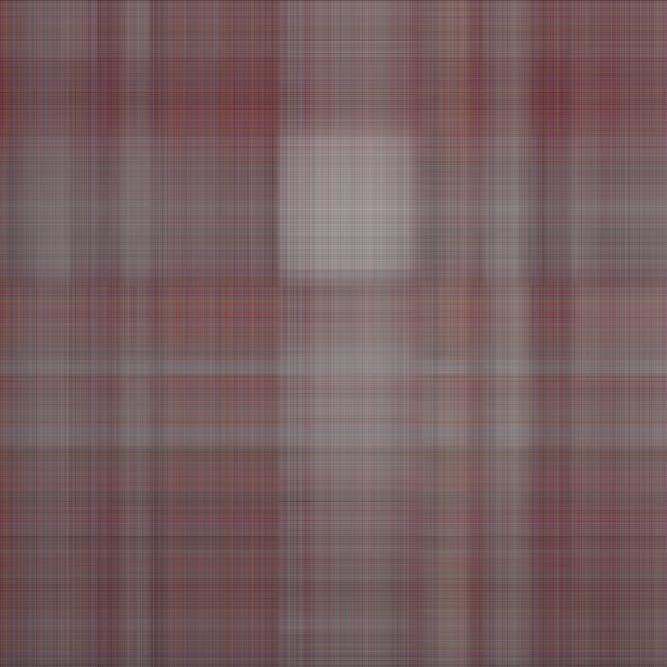}
&
\includegraphics[width=0.586in, height=0.658in]{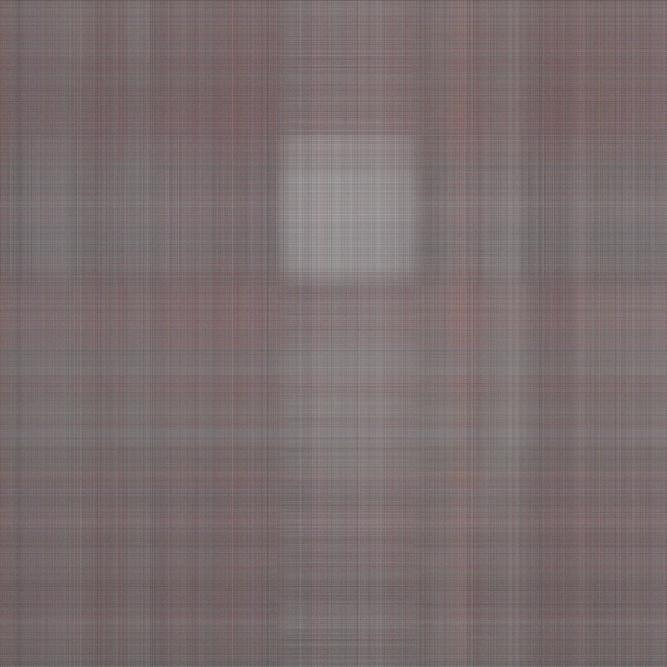}
&
\includegraphics[width=0.586in, height=0.658in]{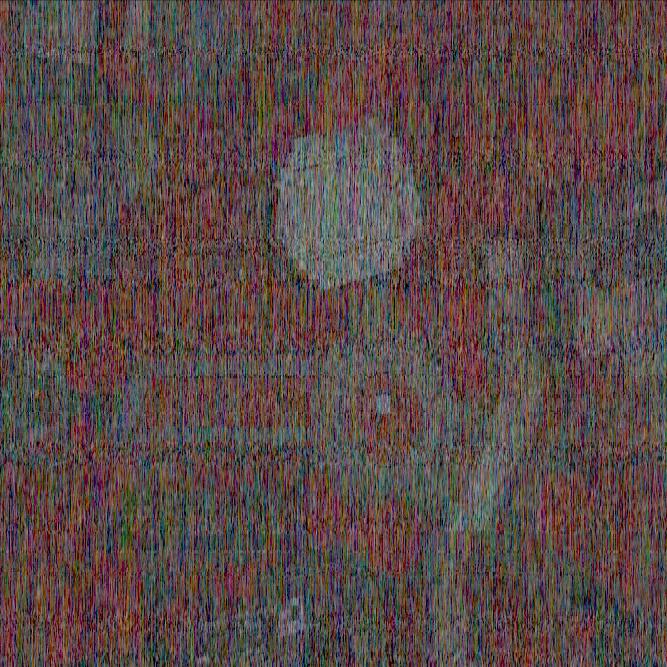}
&
\includegraphics[width=0.586in, height=0.658in]{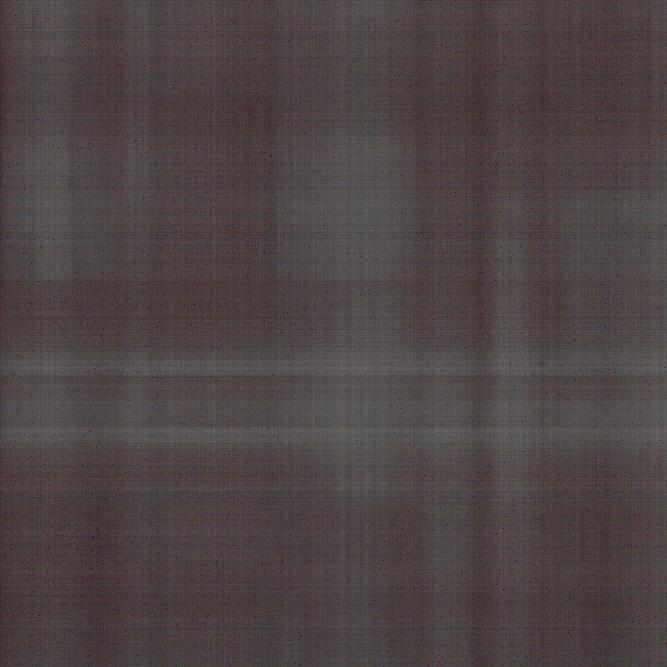}
&
\includegraphics[width=0.586in, height=0.658in]{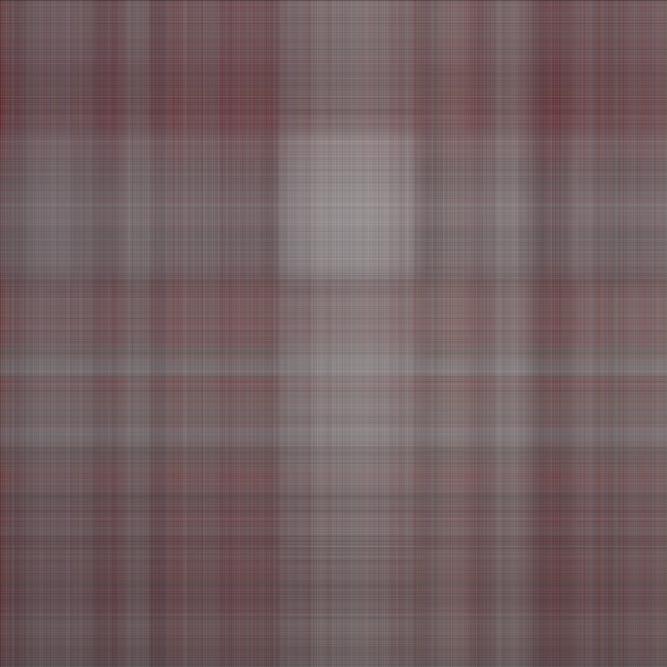}
&
\includegraphics[width=0.586in, height=0.658in]{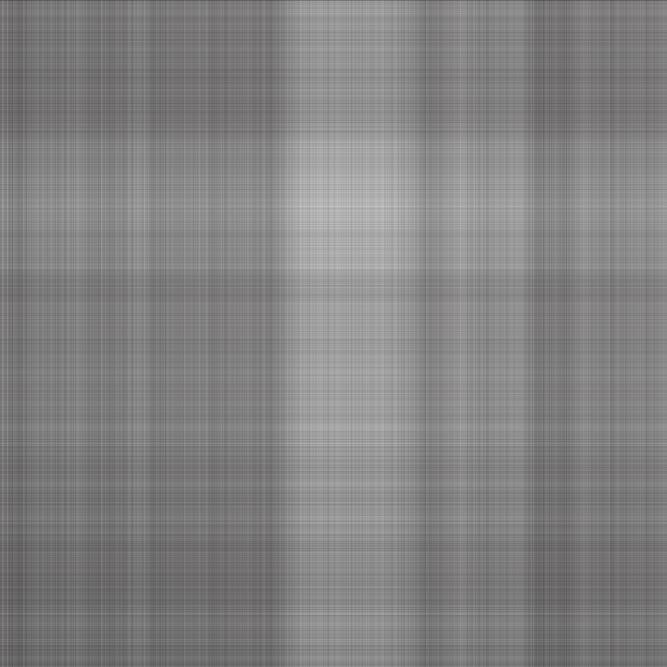}
&
\includegraphics[width=0.586in, height=0.658in]{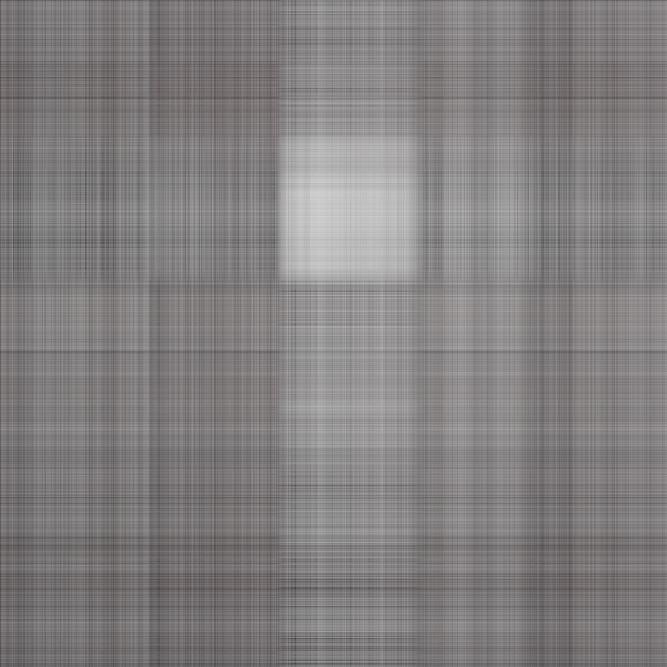}
&
\includegraphics[width=0.586in, height=0.658in]{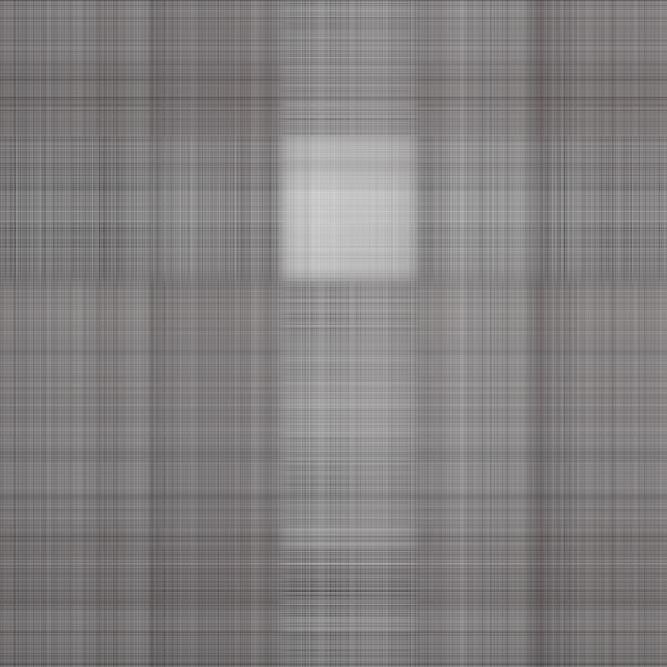}
&
\includegraphics[width=0.586in, height=0.658in]{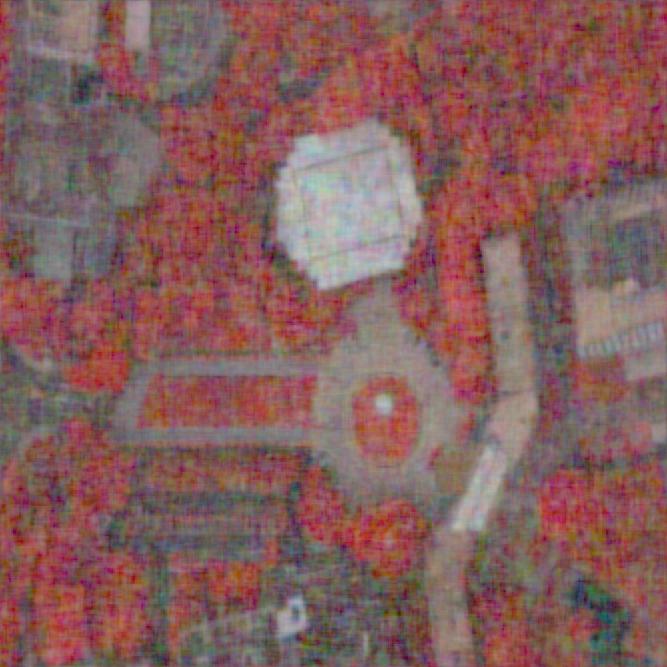}

&
\includegraphics[width=0.586in, height=0.658in]{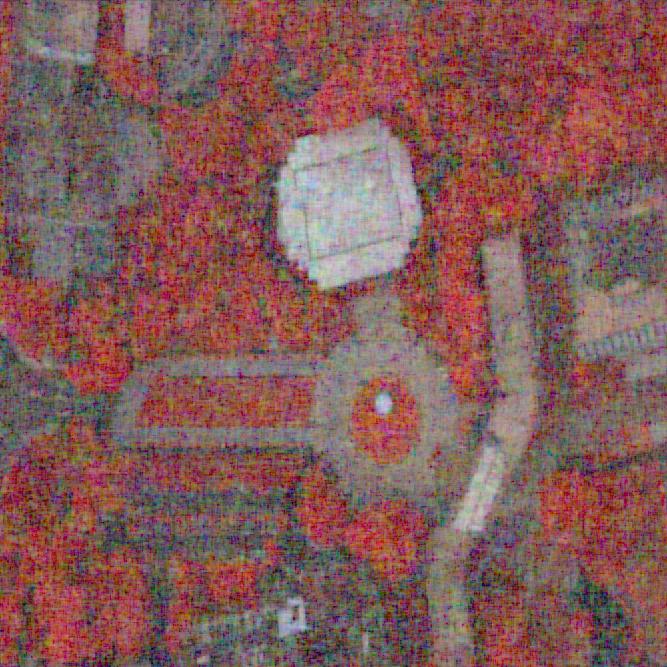}

&
\includegraphics[width=0.586in, height=0.658in]{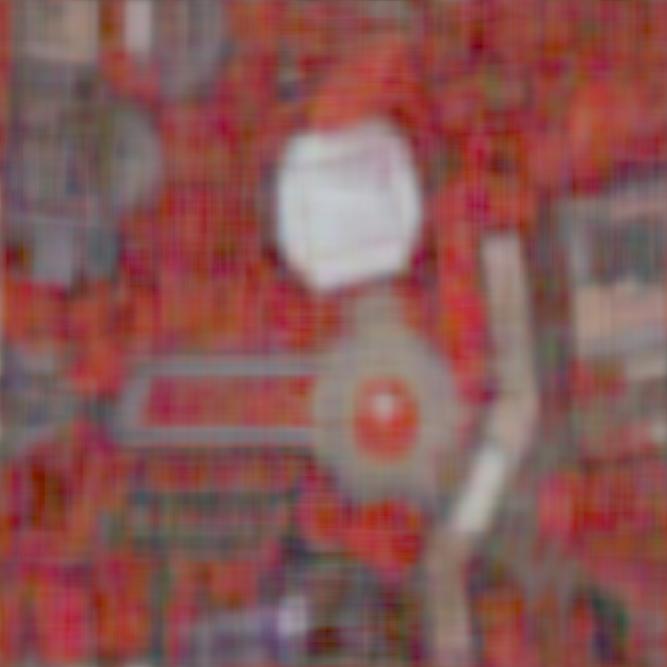} \\

(a)   &
  (b)  & (c) &
(d) & (e)
 &(f)& (g) &
 (h) &
 (i)
 &  (j)
  &
 (k) &(l)
\end{tabular}
% \vspace{-0.4cm}
\vspace{-0.15cm}
\caption{
Visual comparison of various RLRTC methods for  HSI datasets recovery.
%inpainting.
From top to bottom, the parameter pair $(SR, NR)$ are
%(0.1, 0.5),
(0.05, 0.5) and (0.01, 0.5), respectively.
%Top row: the (5, 1)-th frame of Landsat-7. Middle row: the (2, 6)-th frame of
%SPOT-5. Bottom row: the (3, 5)-th frame of T22LGN.
From left to right: (a) Observed,
(b)  TTNN, (c) TSPK, (d) TTLRR, (e) LNOP, (f) NRTRM, (g) HWTNN,  (h) HWTSN,  (i) R-HWTSN, %-Fast,
  (j)  TCTV-RTC, (k) GNRHTC, (l) R-GNRHTC.}
\vspace{-0.4cm}
\label{fig_hsi}
\end{figure*}

%%%%%%%%%%%%%%%%%%%%%%%%%%%%%%%%%%%%%%%%%%%%%%%%%%%%%%%%%%%%%%%%%%%%%%*%%%
\begin{table*}[tp]
\renewcommand{\arraystretch}{0.695}
\setlength\tabcolsep{2.5pt}
  \centering
  \caption{Quantitative evaluation PSNR, SSIM, RSE, CPU Time (Second) of the proposed and compared %various RTC
  RLRTC
   methods
  on fourth-order MRSIs and  CVs, and fifth-order  LFIs.
  }
  \vspace{-0.15cm}
  \label{table-mrsi-CV-LFI}
  \scriptsize
 % \tiny
  \begin{threeparttable}
     \begin{tabular}{cccc cccc cccc cccc cc  c}
   % \hline
    \Xhline{1pt}
   \tabincell{c}   {SR %\\
    $\textbf{\&}$ %\\
   NR}&
   \tabincell{c} {Evaluation\\ Metric}&
%%%%%%%%%%%%%%%%%%%%%%%%%%%%%%%%%%%%%%%%%%%%
\tabincell{c}{TRNN  \\    \cite{huang2020robust}}&
\tabincell{c}{TTNN  \\    \cite{song2020robust}}&
\tabincell{c} {TSPK \\    \cite{lou2019robust}}&
   \tabincell{c}{TTLRR  \\ \cite{yang2022robust}}&
% yang2022robust, liu2024fully, lou2019robust
\tabincell{c}{LNOP \\  \cite{chen2020robust}}&
% T-CTV \cite{wang2023guaranteed}
 \tabincell{c}{NRTRM\\  \cite{qiu2021nonlocal}}&
%%%%%%%%%%%%%%%%%%%%%%%%%%%%%%%%%%%%%%%%%%%%%%%%%%%%%%%%%%%%%%%%%%%%%%%%
\tabincell{c}{HWTNN  \\    \cite{qin2021robust}}&
 \tabincell{c} {HWTSN  \\  \cite{qin2023nonconvex}}&
 \tabincell{c} {R-HWTSN %-Fast
  \\  \cite{qin2023nonconvex}}&
  \tabincell{c} {TCTV-RTC \\ \cite{wang2023guaranteed}  }
 & \textbf{GNRHTC}&
 \textbf{R1-GNRHTC}& \textbf{R2-GNRHTC}

    \cr
    \Xhline{1pt}
    % \hline
    %\hline
   % \Xhline{1pt}
\multicolumn{15}{c}{\textbf{\textit{Results on Fourth-Order MRSIs}}}\\
\hline
     \hline

     \multirow{4}{*}{    \tabincell{c}   {SR=0.1 \\ $\textbf{\&}$ \\NR=1/3}  } &MPSNR
  &   20.8668&23.5098&23.5289&23.9632&23.8890&23.6531&23.9005&24.0602&23.9741&26.0960&{{26.3349}}&\textcolor[rgb]{1,0,0}{\textbf{26.3656}} & \textcolor[rgb]{0,0,1}{\textbf{26.3597}}  \cr
   \qquad	&MSSIM	&0.4904&0.5281&0.5283&0.4840&0.4909&0.5482&0.4931&0.5064&0.5002&\textcolor[rgb]{1,0,0}{\textbf{0.6553}}&\textcolor[rgb]{0,0,1}{\textbf{0.6496}}&0.6202 & 0.6203  \cr
   \qquad	&MRSE	& 0.3387&0.2711&0.2687&0.2398&0.2518&0.2684&0.2518&0.2473&0.2476&0.2171&{{0.2092}}&\textcolor[rgb]{1,0,0}{\textbf{0.1839}} & \textcolor[rgb]{0,0,1}{\textbf{0.1842}} \cr
    \qquad	&MTime	&   3942.46&1719.65&3176.77&3550.68&2213.47&2113.38&1994.28&2654.98&\textcolor[rgb]{1,0,0}{\textbf{1194.43}}&4108.24&5428.79&\textcolor[rgb]{0,0,1}{\textbf{1340.78}} &1708.39 \cr

     \hline
     \multirow{4}{*}{    \tabincell{c}   {SR=0.1  \\ $\textbf{\&}$ \\NR=0.5}  } &MPSNR
  &  19.7638&22.1445&22.0155&21.3846&21.8048&22.5362&22.3005&22.1249&22.0964&24.3981&{{24.5911}}&\textcolor[rgb]{0,0,1}{\textbf{24.9018}} & \textcolor[rgb]{1,0,0}{\textbf{24.9179}} \cr
   \qquad	&MSSIM	& 0.4724&0.4208&0.4306&0.3155&0.4141&0.4767&0.4042&0.3994&0.4069&\textcolor[rgb]{0,0,1}{\textbf{0.5804}}&\textcolor[rgb]{1,0,0}{\textbf{0.5851}}&0.5555 & 0.5588\cr
   \qquad	&MRSE	& 0.3802&0.3121&0.3176&0.2884&0.3380&0.3051&0.3003&0.2963&0.2990&0.2644&{{0.2627}}&\textcolor[rgb]{0,0,1}{\textbf{0.2214}}& \textcolor[rgb]{1,0,0}{\textbf{0.2212}} \cr
    \qquad	&MTime	& 3967.29&1731.32&3207.05&3482.18&2268.15&2118.69&1982.67&2636.47&\textcolor[rgb]{1,0,0}{\textbf{1203.59}}&4010.51&5398.73&
    \textcolor[rgb]{0,0,1}{\textbf{1347.37}} & 1675.08 \cr

  \hline
    \hline
   % \Xhline{1pt}
\multicolumn{15}{c}{\textbf{\textit{Results on Fourth-Order CVs}}}\\
\hline
     \hline

     \multirow{4}{*}{    \tabincell{c}   {SR=0.1 \\ $\textbf{\&}$ \\NR=1/3}  } &MPSNR
  & 21.6719&27.2999&27.3469&28.4580&28.4289&27.9123&28.9986&\textcolor[rgb]{0,0,1}{\textbf{30.1071}}&27.5967&29.3515&\textcolor[rgb]{1,0,0}{\textbf{31.0519}}&28.6433  & 28.3286 \cr

   \qquad	&MSSIM	& 0.6877&0.8505&\textcolor[rgb]{0,0,1}{\textbf{0.8983}}&0.8397&0.8301&0.8626&0.8644&0.8833&0.8112&0.8589&\textcolor[rgb]{1,0,0}{\textbf{0.9123}}&0.8159
   & 0.8110 \cr
   \qquad	&MRSE	& 0.1850&0.0977&0.0908&0.0859&0.0858&0.0912&0.0811&\textcolor[rgb]{0,0,1}{\textbf{0.0725}}&0.0950&0.0785&\textcolor[rgb]{1,0,0}{\textbf{0.0627}}&0.0859 & 0.0889\cr
    \qquad	&MTime	&  %3048&2252&3756&5999&2930&2530&2769&3413&1874&4653&6200&1755
    3048.46&2251.99&3755.75&5999.45&2930.37&2530.27&2768.61&3413.39&\textcolor[rgb]{0,0,1}{\textbf{1874.28}}&4653.09&6200.28&\textcolor[rgb]{1,0,0}{\textbf{1755.31}}
   & 2160.53 %2360.53
   \cr

    \hline

     \multirow{4}{*}{    \tabincell{c}   {SR=0.1 \\ $\textbf{\&}$ \\NR=0.5}} &MPSNR
  &  19.0174&25.7810&26.0552&26.7229&25.7733&26.4971&26.8800&\textcolor[rgb]{0,0,1}{\textbf{27.9984}}&26.0465&26.3012&\textcolor[rgb]{1,0,0}{\textbf{28.8414}}&26.8368
 & 26.2677 \cr

   \qquad	&MSSIM	& 0.5966&0.8114&\textcolor[rgb]{1,0,0}{\textbf{0.8651}}&0.7248&0.7673&0.8265&0.8133&0.8315&0.7708&0.7737&\textcolor[rgb]{0,0,1}{\textbf{0.8534}}&0.7708 & 0.7534\cr
   \qquad	&MRSE	& 0.2503&0.1157&0.1049&0.1056&0.1161&0.1071&0.1029&\textcolor[rgb]{0,0,1}{\textbf{0.0916}}&0.1122&0.1107&\textcolor[rgb]{1,0,0}{\textbf{0.0804}}&0.1053
   & 0.1124\cr
    \qquad	&MTime	& %2921&2393&3662&5923&2982&2569&2727&3390&1872&4461&6257&1576
    2920.95&2393.36&3662.44&5922.98&2982.35&2569.06&2726.56&3390.48&\textcolor[rgb]{0,0,1}{\textbf{1871.82}}&4461.77&6257.18&\textcolor[rgb]{1,0,0}{\textbf{1576.33}}
    & 1732.92\cr

 %  \hline
%     \hline
%
%\hline
%     \hline
\hline
    \hline
   % \Xhline{1pt}
\multicolumn{15}{c}{\textbf{\textit{Results on Fifth-Order %Third-Order  %CVs
LFIs}}}\\
\hline
     \hline

     \multirow{4}{*}{    \tabincell{c}   {SR=0.05 \\ $\textbf{\&}$ \\NR=1/3}  } &MPSNR
  &19.5675&27.7763&25.4831&28.7039&25.8129&28.5711&28.3114&  \textcolor[rgb]{0,0,1}{\textbf{30.1991}} &28.6890&29.9660& \textcolor[rgb]{1,0,0}{\textbf{30.7448}}  &27.3391  & 26.7011 \cr
   \qquad	&MSSIM	& 0.6088&0.8842&0.8359&0.8869&0.7319&\textcolor[rgb]{0,0,1}{\textbf{0.9076}}&0.8794&    0.8624  &0.8730&0.9032& \textcolor[rgb]{1,0,0}{\textbf{0.9239}}   &0.8416 & 0.8294 \cr
   \qquad	&MRSE	& 0.3299&0.1295&0.1677&0.1154&0.1606&0.1184&0.1227&   \textcolor[rgb]{0,0,1}{\textbf{0.1003}}  &0.1187&0.1029&   \textcolor[rgb]{1,0,0}{\textbf{0.0940}}   &0.1393 & 0.1492 \cr
    \qquad	&MTime	& %3132&1765&3294&3933&2601&2079&2142&2325&1327&2932&3630&1012

    3132.39&1765.27&3293.65&3932.82&2600.87&2078.55&2142.27&2325.07&{{1326.56}}&2931.57&3630.29&\textcolor[rgb]{1,0,0}{\textbf{1011.59}}
&\textcolor[rgb]{0,0,1}{\textbf{1200.67}}
     \cr

    \hline
     \multirow{4}{*}{    \tabincell{c}   {SR=0.05 \\ $\textbf{\&}$ \\NR=0.5}} &MPSNR
  & 17.0493&24.5345&22.7260&    25.4637 &19.7810&25.1724&25.3275&26.8663&26.3289&
 25.1870  &   \textcolor[rgb]{1,0,0}{\textbf{27.8194}}  &
     \textcolor[rgb]{0,0,1}{\textbf{27.2358}}   &   26.9874  \cr
    %%
    % \qquad	&MPSNR	& & & & & & & & & & & & & & & & \cr
    %%
   \qquad	&MSSIM	& 0.5350&0.8069&0.7037&     0.7735 &0.5272&0.8275&0.7595&\textcolor[rgb]{0,0,1}{\textbf{0.8332}}&
   0.8076&
  0.7837 & \textcolor[rgb]{1,0,0}{\textbf{0.8668}}  &
    0.7941 &      0.7859 \cr
   \qquad	&MRSE	& 0.4365&0.1874&0.2293&  0.1715     &0.3243&0.1740&0.1715&0.1461&0.1553
   &0.1771 & \textcolor[rgb]{1,0,0}{\textbf{0.1277}}
    & \textcolor[rgb]{0,0,1}{\textbf{0.1398}} &   0.1434  \cr
    \qquad	&MTime	& %3169&1838&3219&3781&2646&2086&2131&2334&1320&2851&3532&997

    3168.99&1838.05&3219.03&3780.72&2645.66&2085.77&2131.19&2334.08&{{1319.79}}&2851.29&3531.73&\textcolor[rgb]{1,0,0}{\textbf{997.29}}
&\textcolor[rgb]{0,0,1}{\textbf{1174.51}}
     \cr

   \hline
     \hline

      \Xhline{1pt}
    \end{tabular}
    \end{threeparttable}
    \vspace{-0.3cm}
\end{table*}

%%%%%%%%%%%%%%%%%%%%%%%%%%%%%%%%%%%%%%%%%%%%%%%%%%%%%%%%%%%%%%%%%%%%%%%%%%%%%%%%%%%%%%%%%%%%%%%%%%%%%%%%%%%%%%%%%%%%%%%%%%%%%%%%%%%%%%%%%%%%%%%%%
\begin{figure*}[!htbp]
\renewcommand{\arraystretch}{0.7}
\setlength\tabcolsep{0.43pt}
\centering
\begin{tabular}{ccc  ccc ccc ccc c }%cc ccc  ccc c cc
\centering
\includegraphics[width=0.54in, height=0.581in]{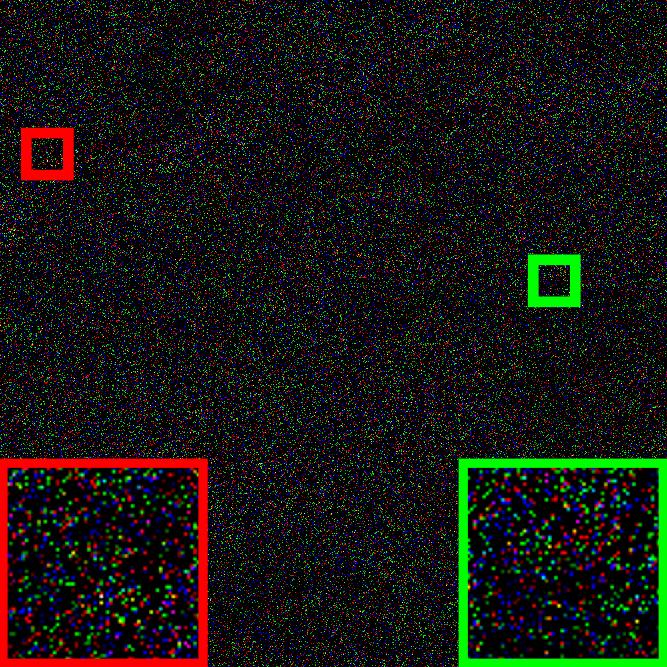} %
&
\includegraphics[width=0.54in, height=0.581in]{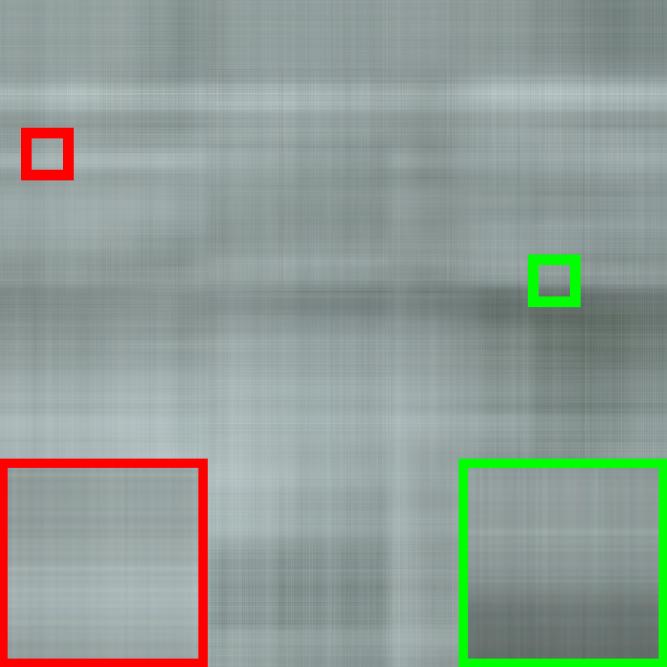}
&
\includegraphics[width=0.54in, height=0.581in]{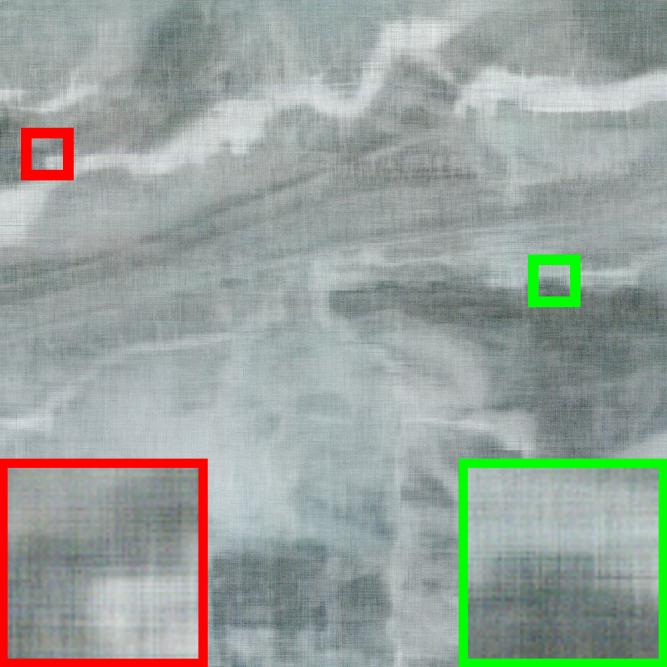}
&
\includegraphics[width=0.54in, height=0.581in]{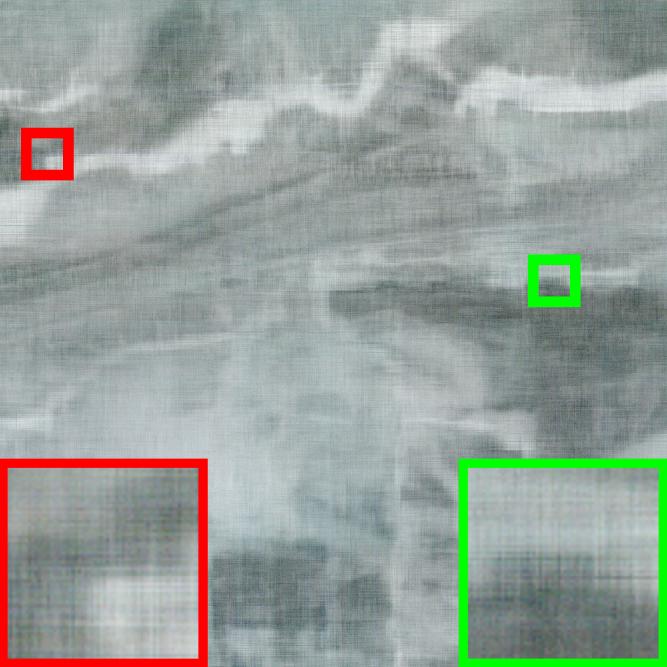}
&
\includegraphics[width=0.54in, height=0.581in]{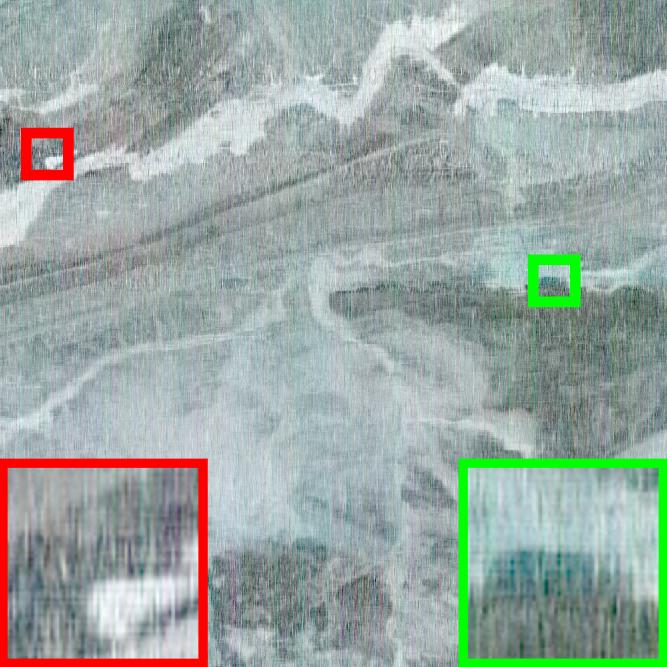}
&
\includegraphics[width=0.54in, height=0.581in]{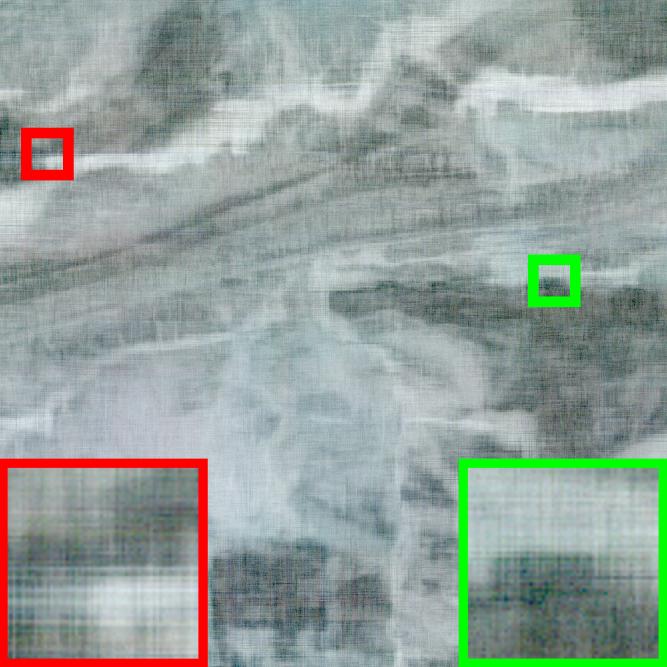}
&
\includegraphics[width=0.54in, height=0.581in]{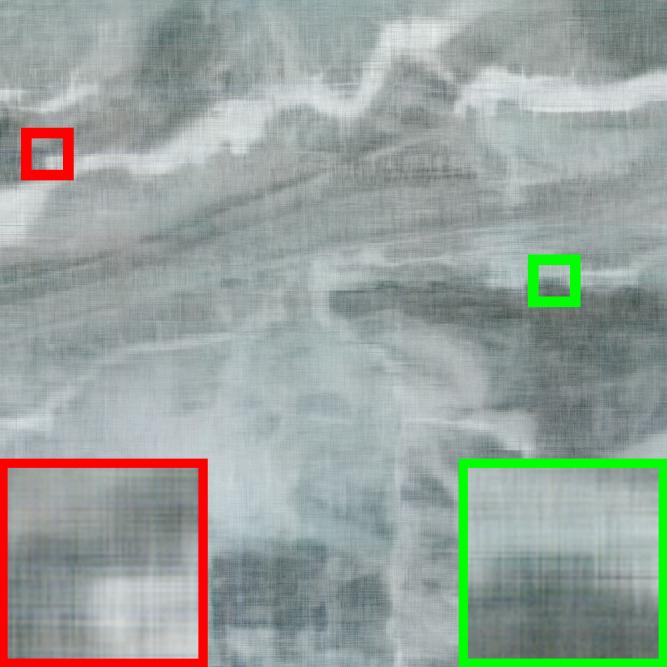}
&
\includegraphics[width=0.54in, height=0.581in]{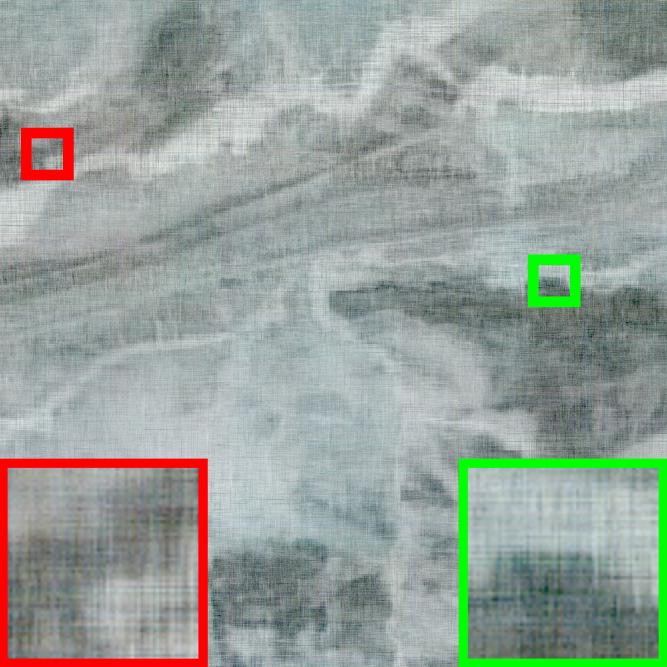}
&
\includegraphics[width=0.54in, height=0.581in]{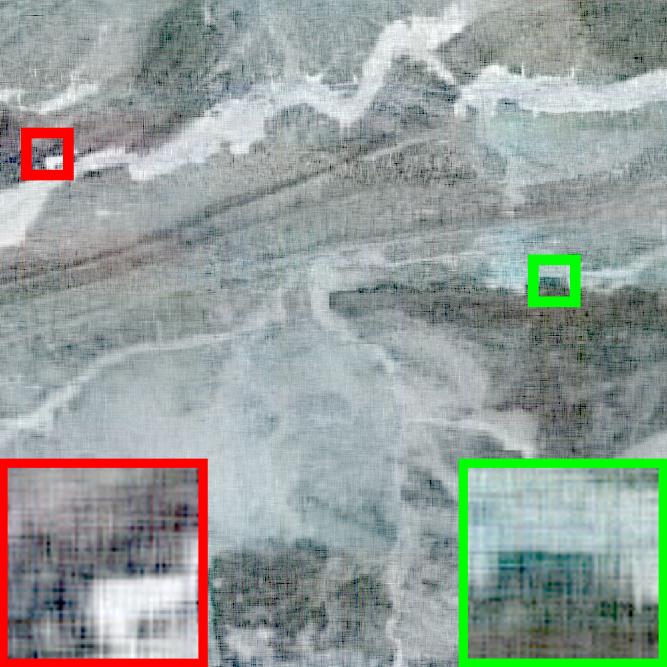}
&
\includegraphics[width=0.54in, height=0.581in]{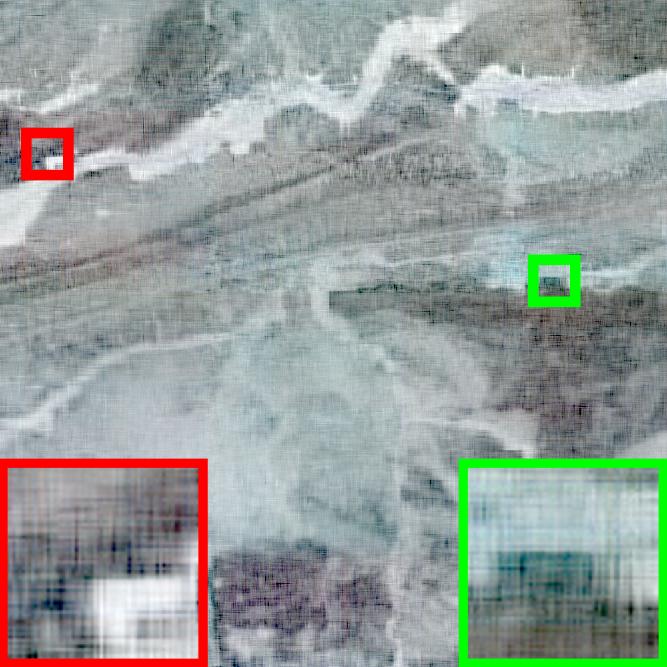}
&
\includegraphics[width=0.54in, height=0.581in]{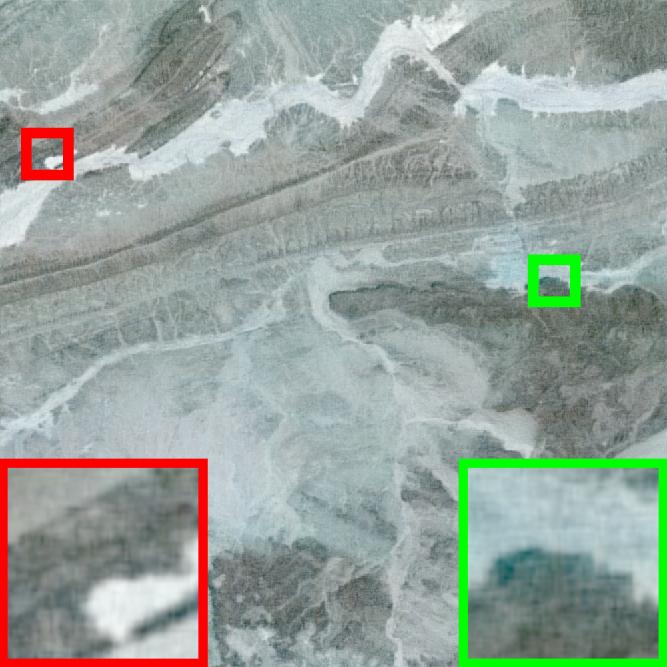}
&
\includegraphics[width=0.54in, height=0.581in]{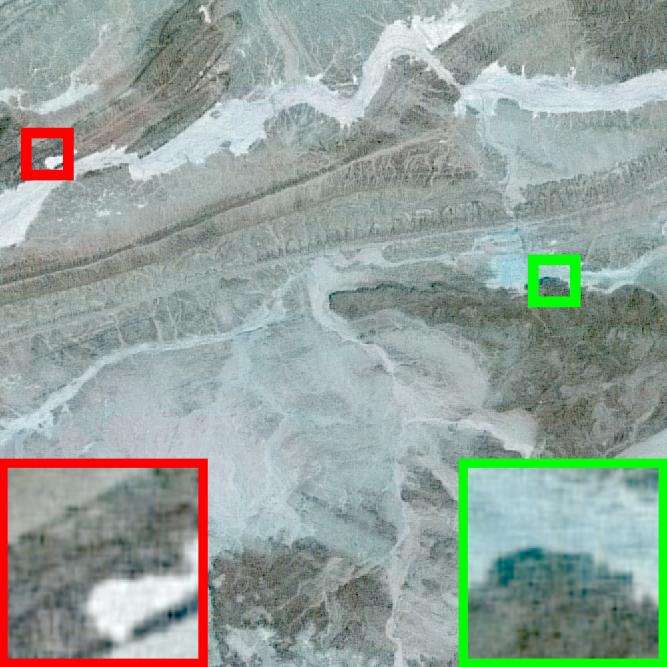}
&
\includegraphics[width=0.54in, height=0.581in]{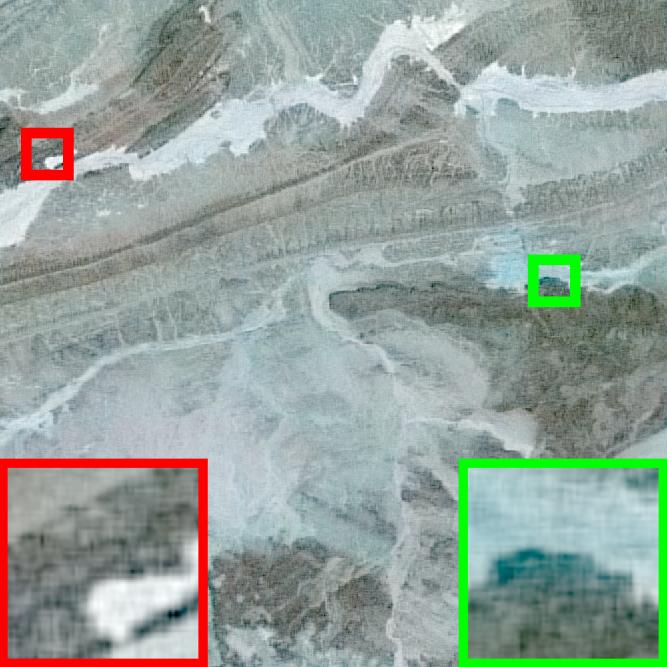}
\\
\includegraphics[width=0.54in, height=0.581in]{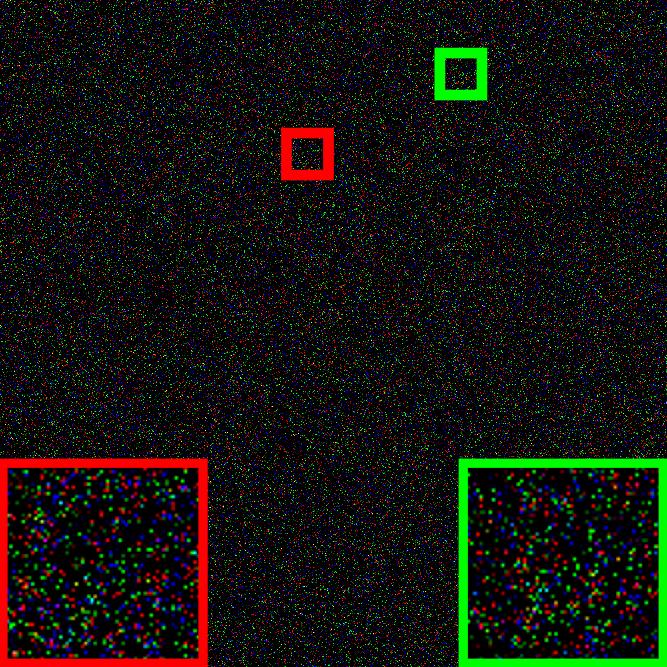} %
&
\includegraphics[width=0.54in, height=0.581in]{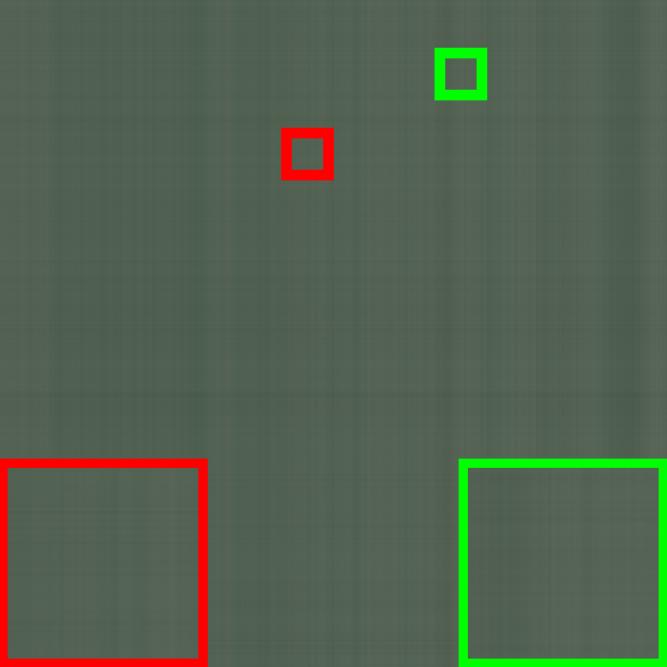}
&
\includegraphics[width=0.54in, height=0.581in]{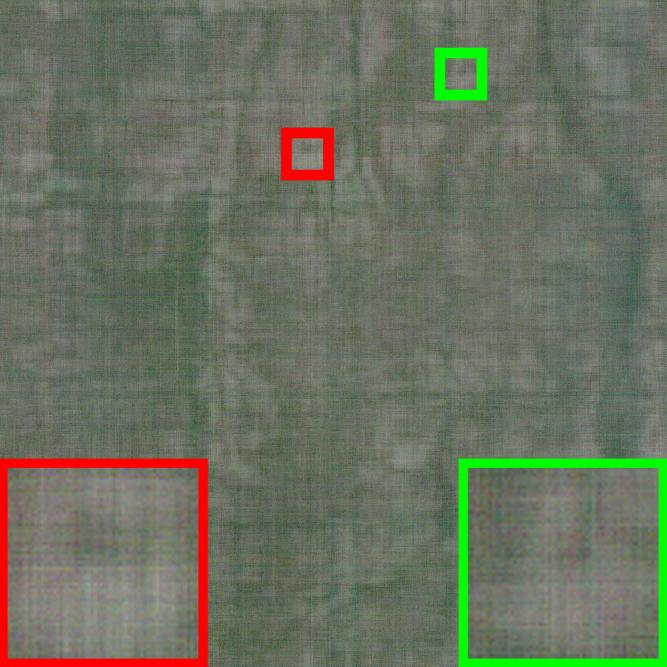}
&
\includegraphics[width=0.54in, height=0.581in]{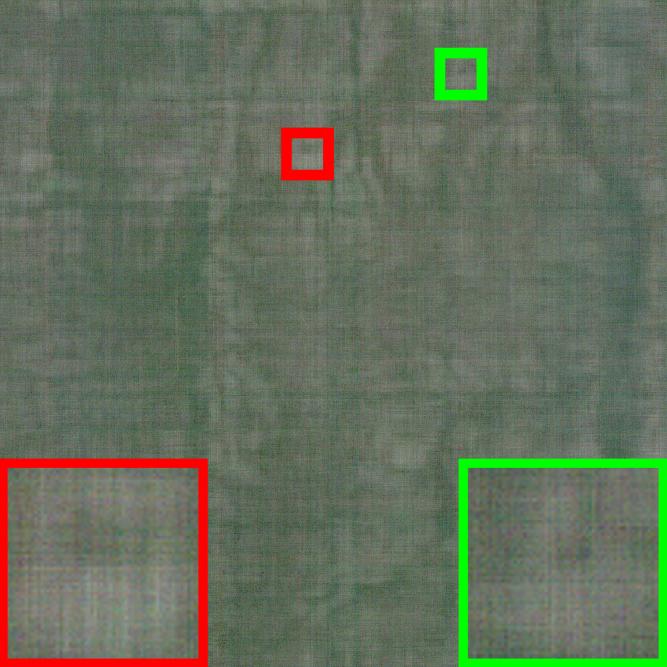}
&
\includegraphics[width=0.54in, height=0.581in]{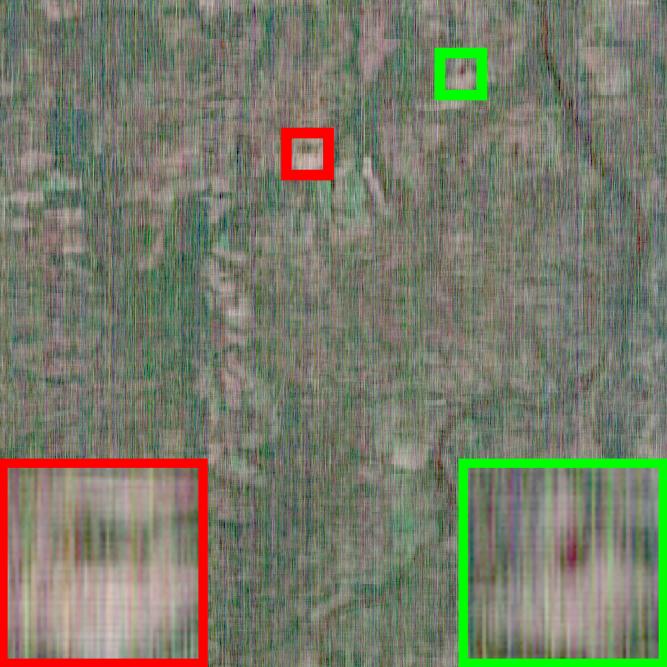}
&
\includegraphics[width=0.54in, height=0.581in]{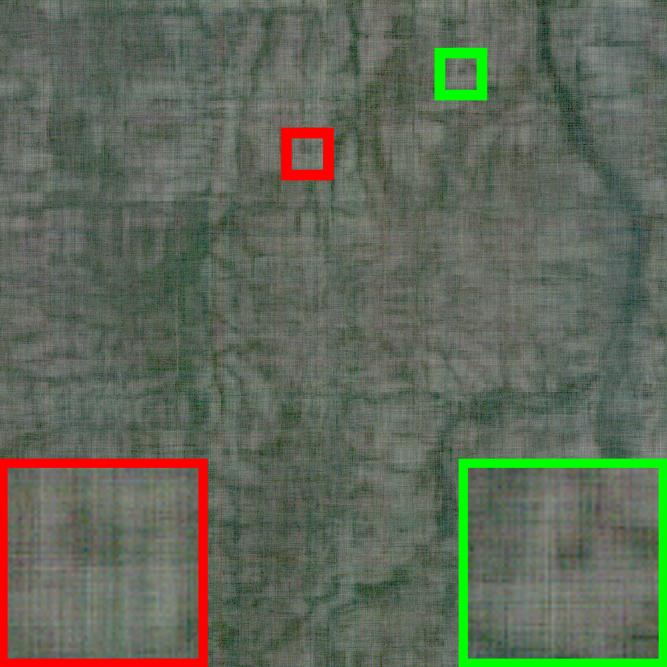}
&
\includegraphics[width=0.54in, height=0.581in]{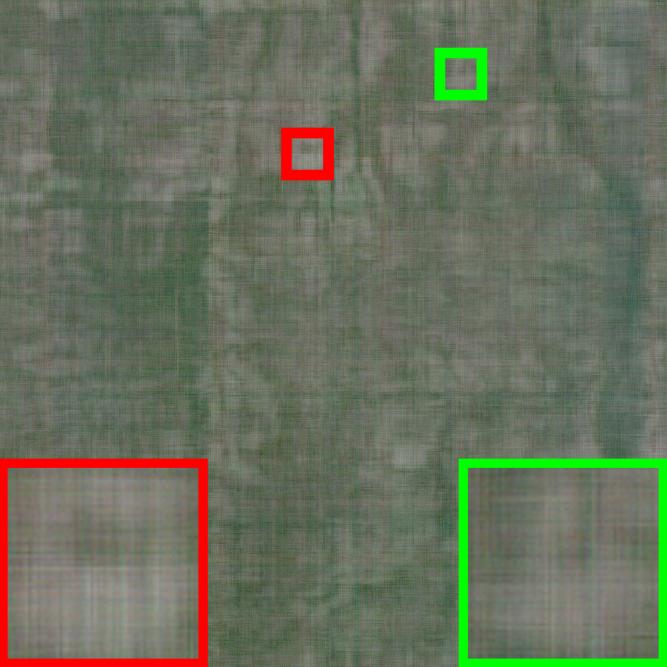}
&
\includegraphics[width=0.54in, height=0.581in]{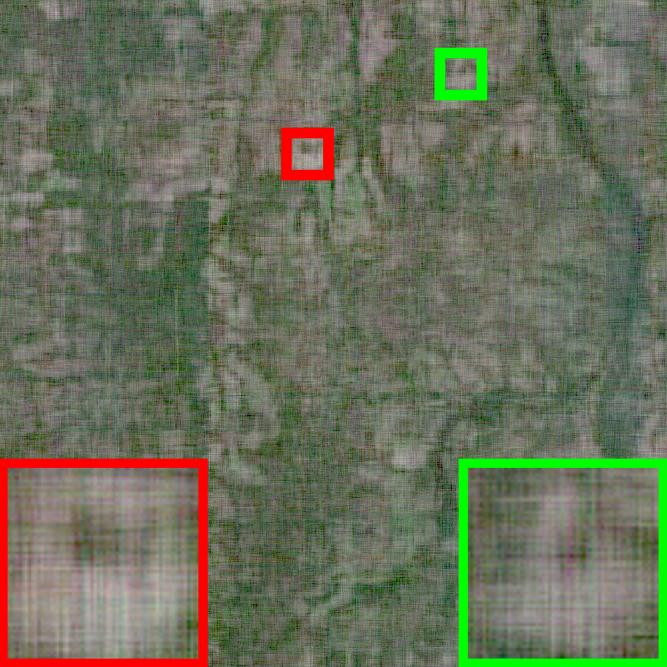}
&
\includegraphics[width=0.54in, height=0.581in]{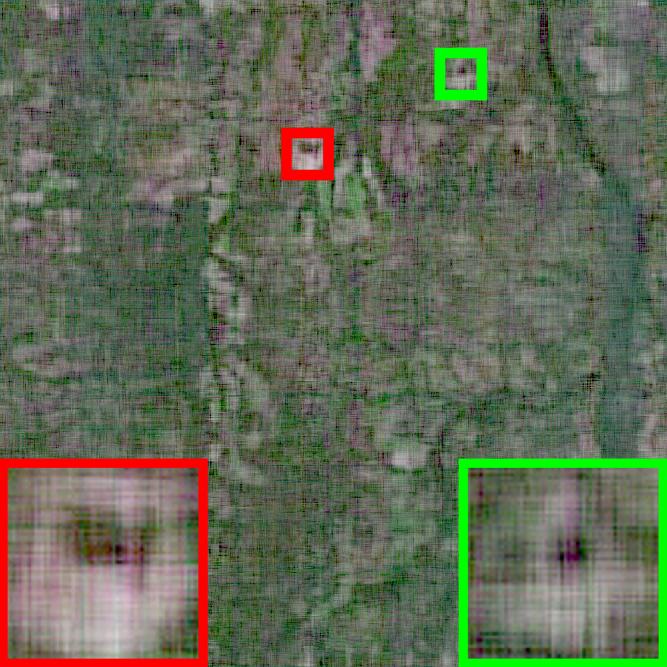}
&
\includegraphics[width=0.54in, height=0.581in]{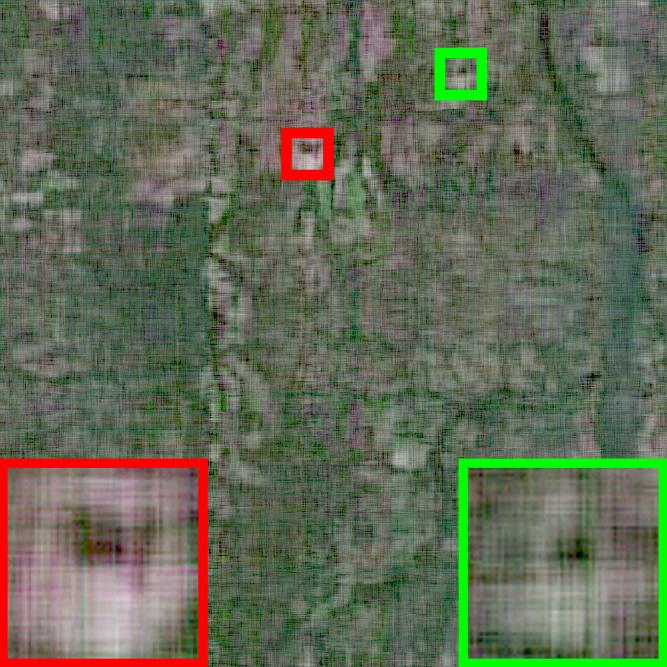}
&
\includegraphics[width=0.54in, height=0.581in]{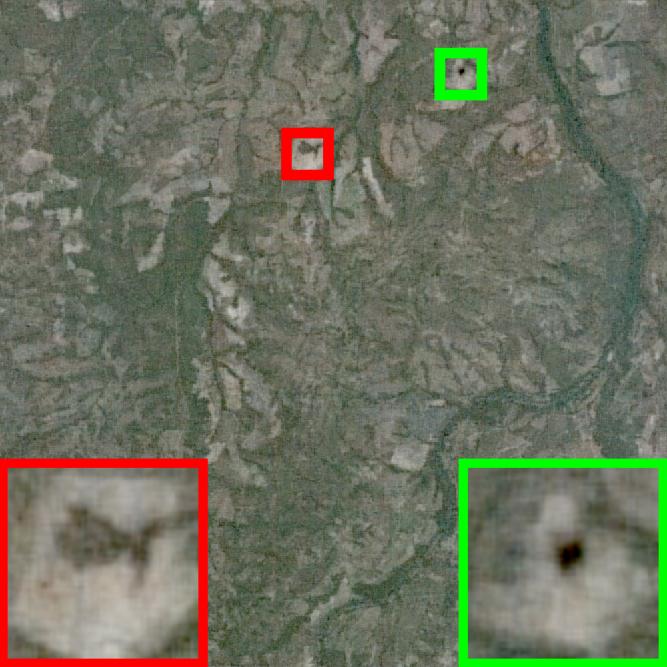}
&
\includegraphics[width=0.54in, height=0.581in]{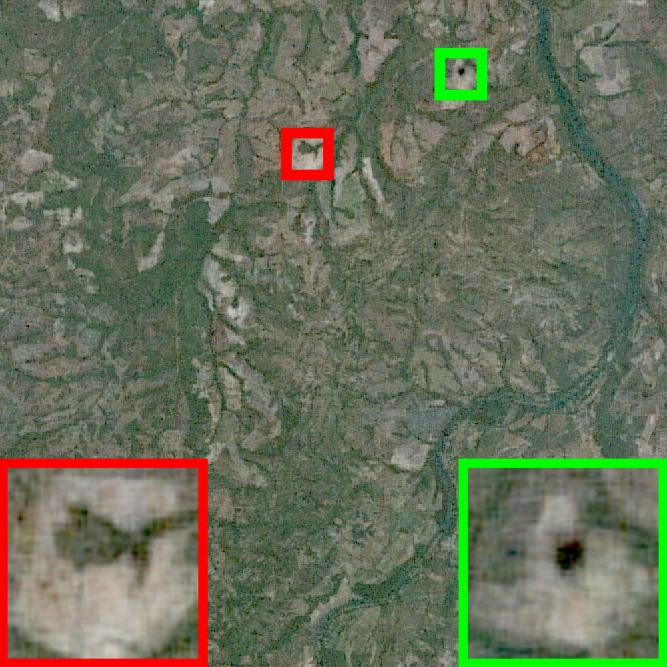}
&
\includegraphics[width=0.54in, height=0.581in]{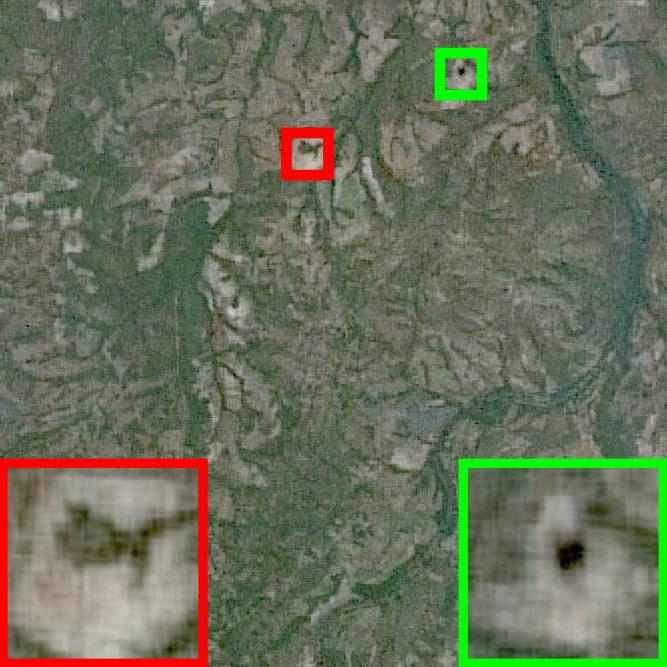}
\\

\includegraphics[width=0.54in, height=0.581in]{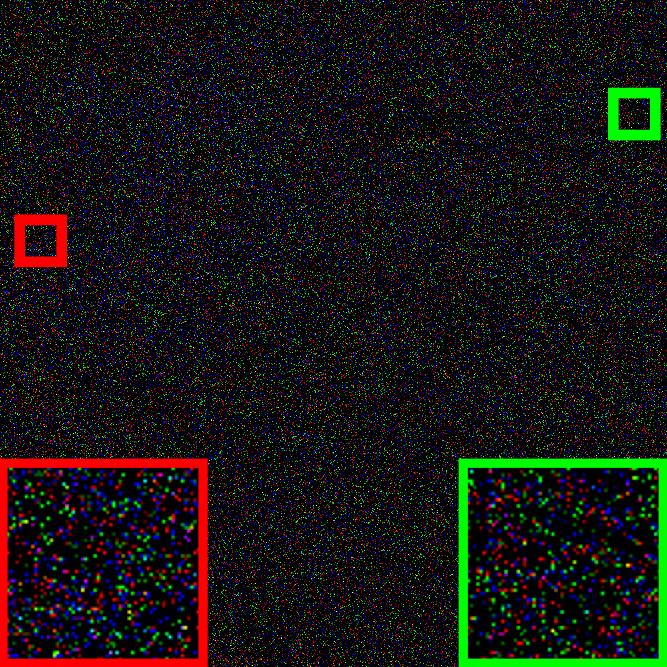} %
&
\includegraphics[width=0.54in, height=0.581in]{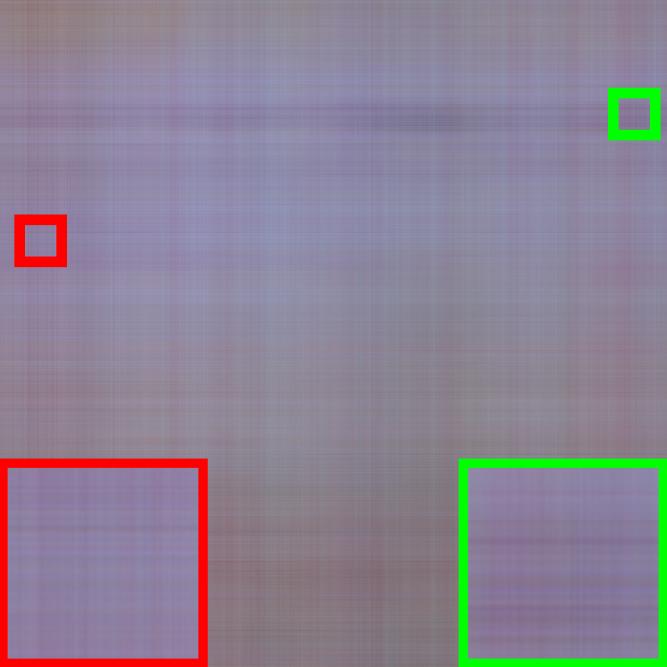}
&
\includegraphics[width=0.54in, height=0.581in]{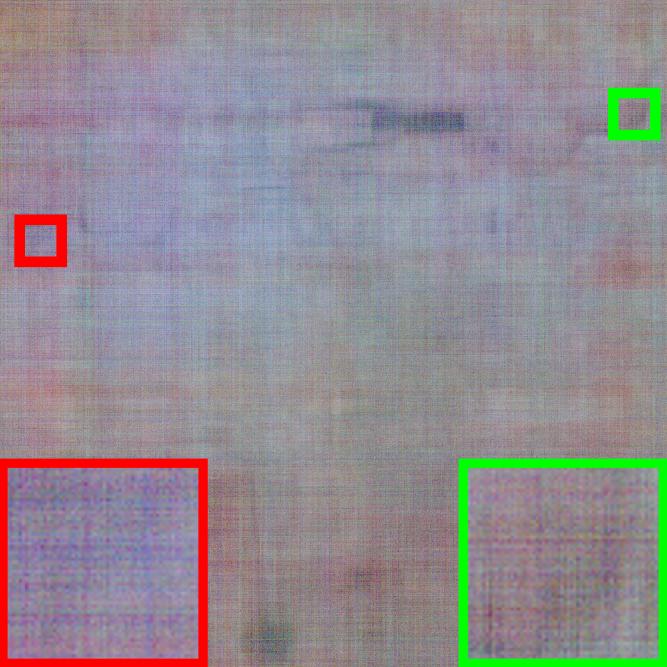}
&
\includegraphics[width=0.54in, height=0.581in]{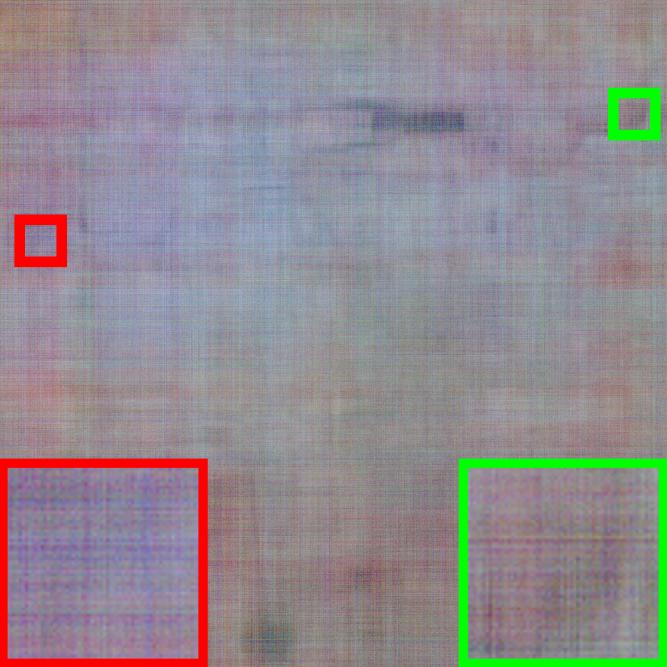}
&
\includegraphics[width=0.54in, height=0.581in]{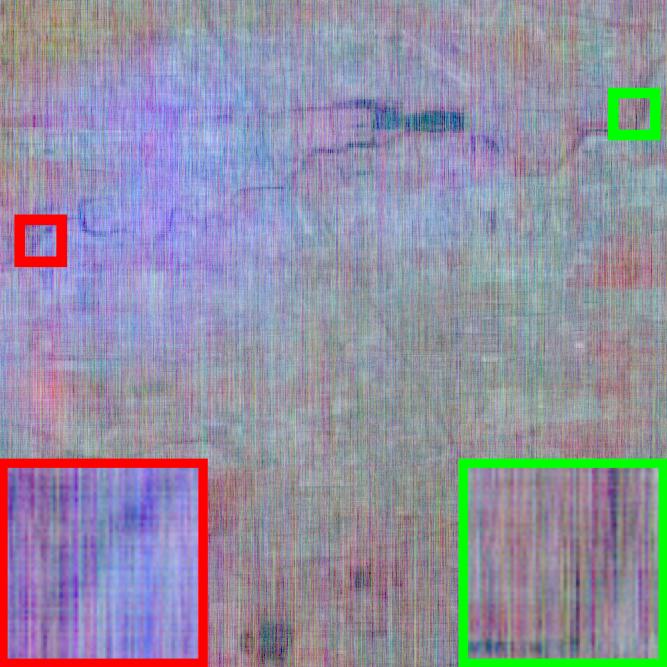}
&
\includegraphics[width=0.54in, height=0.581in]{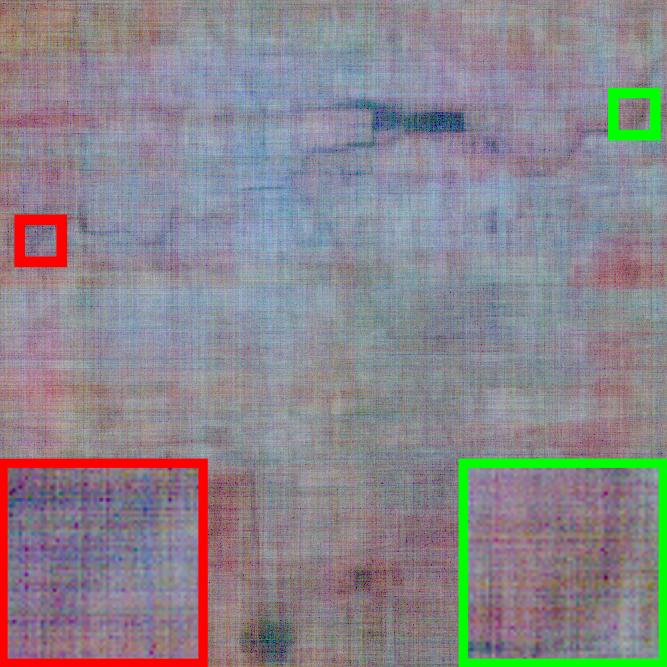}
&
\includegraphics[width=0.54in, height=0.581in]{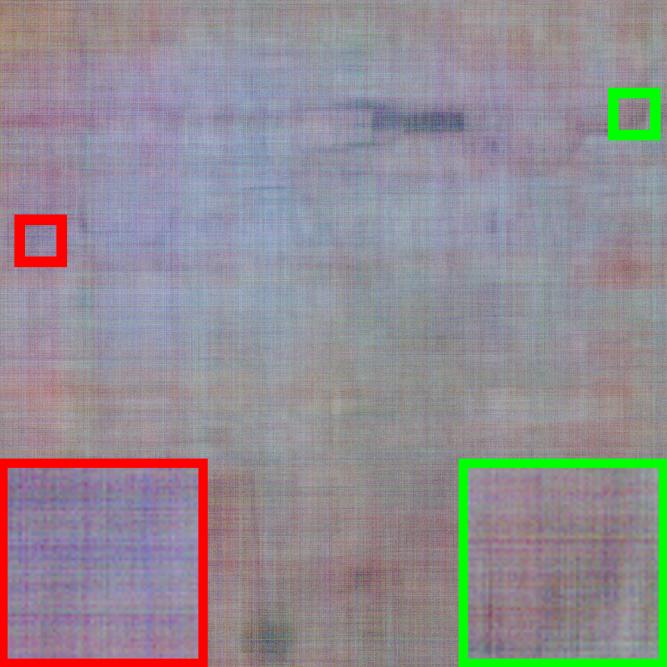}  %mrsi-visul-spot15
&
\includegraphics[width=0.54in, height=0.581in]{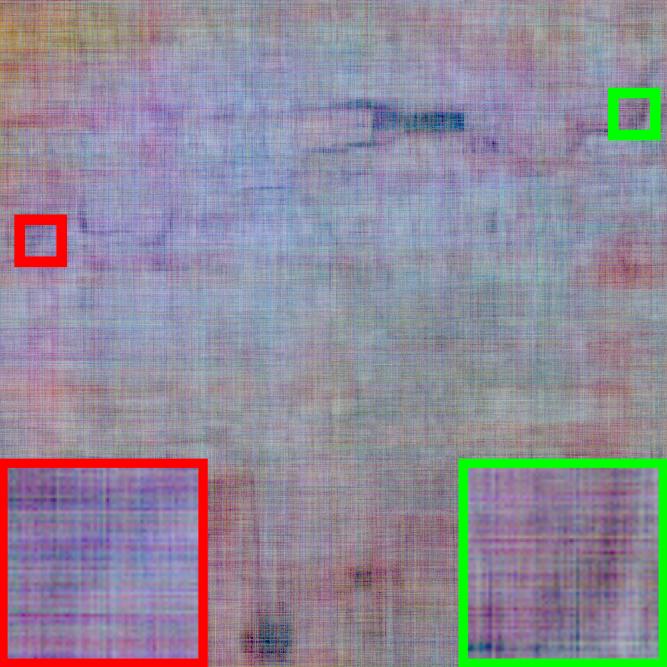}
&
\includegraphics[width=0.54in, height=0.581in]{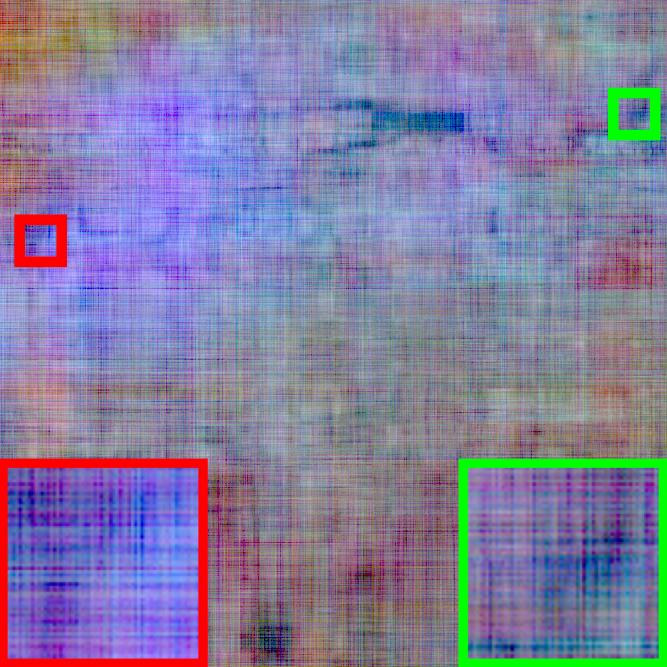}
&
\includegraphics[width=0.54in, height=0.581in]{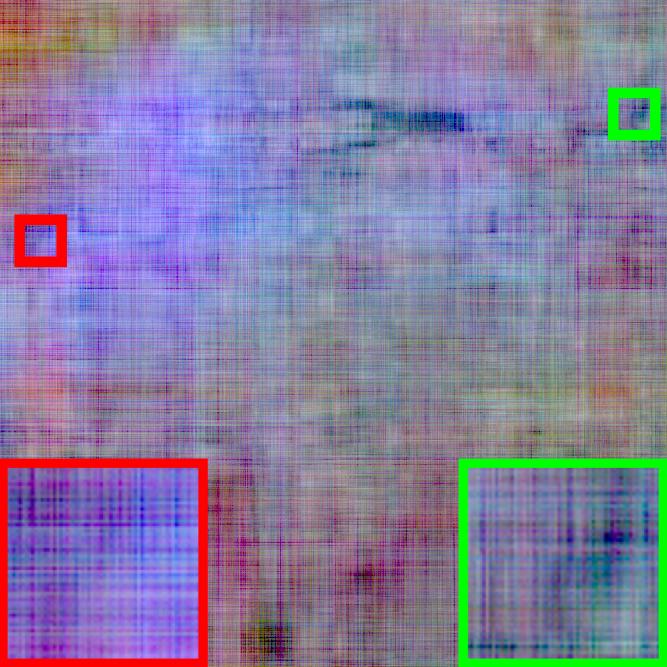}
&
\includegraphics[width=0.54in, height=0.581in]{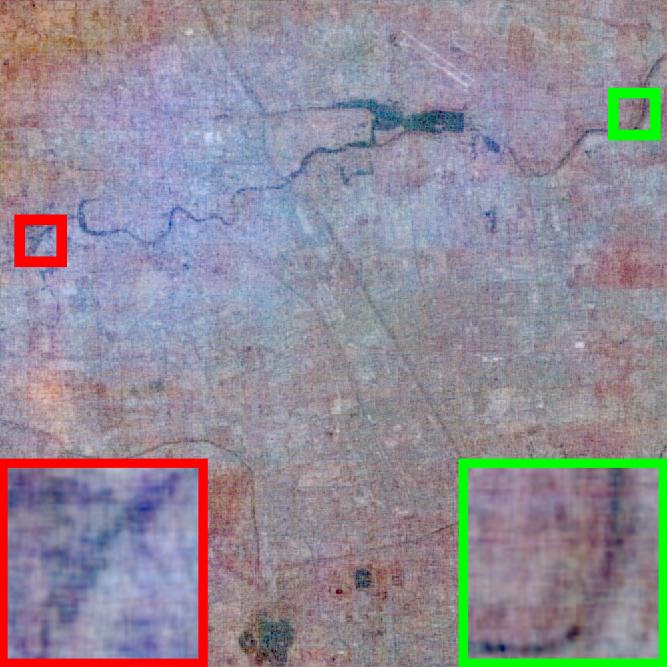}
&
\includegraphics[width=0.54in, height=0.581in]{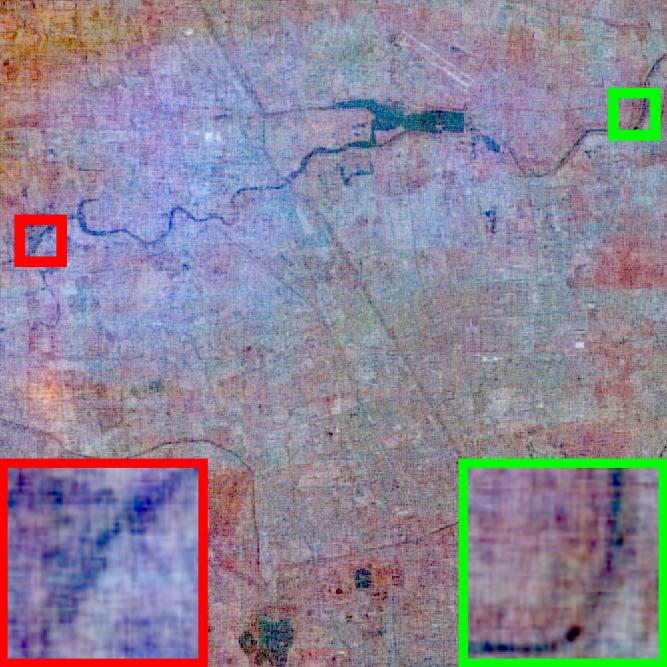}
&
\includegraphics[width=0.54in, height=0.581in]{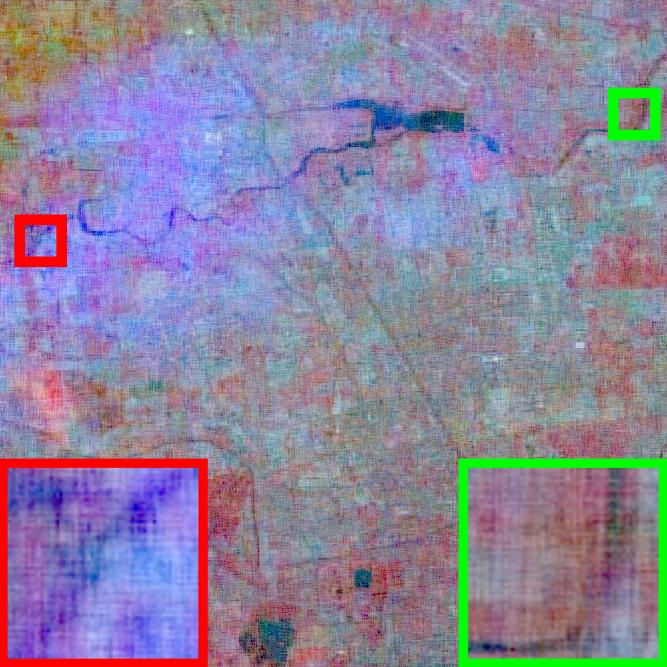}
\\

(a)   &
  (b)  & (c) &
(d) & (e)
 &(f)& (g) &
 (h) &
 (i)
 &  (j)
  &
 (k) &(l)&(m)
\end{tabular}
\vspace{-0.15cm}
\caption{
Visual comparison of various RLRTC methods for  MRSIs inpainting under $(SR, NR)= (0.1, 0.5)$.
 %From top to bottom, the parameter pair $(SR, NR)$ are
%(0.1, 0.5),
%(0.1, 0.5) and (0.05, 0.5), respectively.
%Top row: the (5, 1)-th frame of Landsat-7. Middle row: the (2, 6)-th frame of
%SPOT-5. Bottom row: the (3, 5)-th frame of T22LGN.
From left to right: (a) Observed, (b) TRNN,
(c)  TTNN, (d) TSPK, (e) TTLRR, (f) LNOP, (g) NRTRM, (h) HWTNN,  (i) HWTSN,  (j) R-HWTSN, %HWTSN-Fast,
 (k)  TCTV-RTC, (l) GNRHTC, (m) R-GNRHTC.}
\vspace{-0.5cm}
\label{fig_MRSI}
\end{figure*}

%\vspace{-0.15cm}

%%%%%%%%%%%%%%%%%%%%%%%%%%%%%%%%%%%%%%%%%%%%%%%%%%%%%%%%%%%%%%%%%%%%%%%%%%%%%%%%%%%%%%%%%%%%%%%%%%%%%%%%%%%%%%%%%%%%%%%%%%%%%%%%%%%%%%%%%%%%%%%%%
\begin{figure*}[!htbp]
\renewcommand{\arraystretch}{0.7}
\setlength\tabcolsep{0.43pt}
\centering
\begin{tabular}{ccc  ccc ccc ccc c }%cc ccc  ccc c cc
\centering
\includegraphics[width=0.5286012in, height=0.81in]{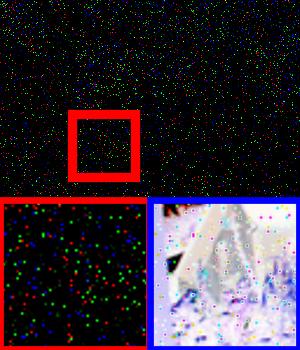}
&
\includegraphics[width=0.5286012in, height=0.81in]{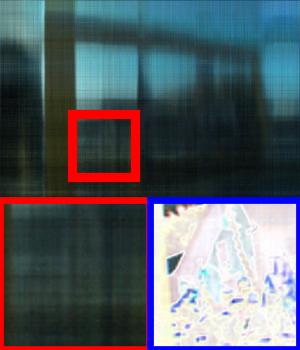}
&
\includegraphics[width=0.5286012in, height=0.81in]{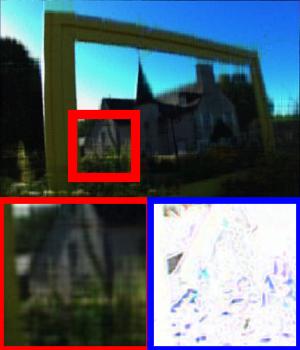}
&
\includegraphics[width=0.5286012in, height=0.81in]{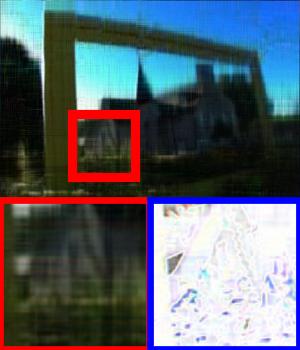}
&
\includegraphics[width=0.5286012in, height=0.81in]{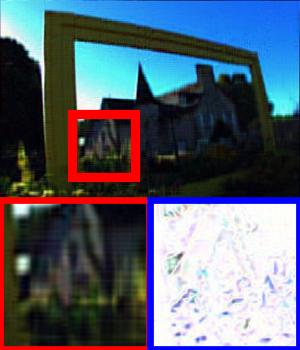} %s4 --->s12
&
\includegraphics[width=0.5286012in, height=0.81in]{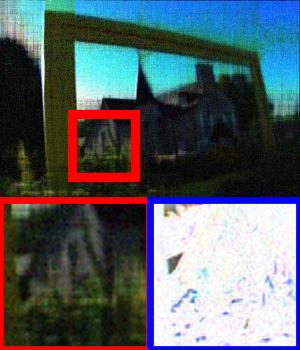}
&
\includegraphics[width=0.5286012in, height=0.81in]{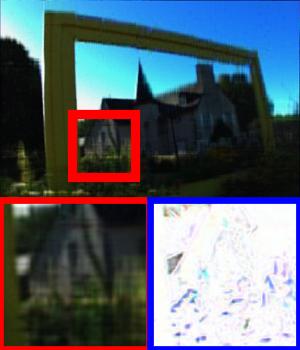}
&
\includegraphics[width=0.5286012in, height=0.81in]{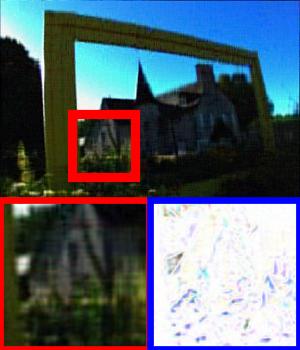}
&
\includegraphics[width=0.5286012in, height=0.81in]{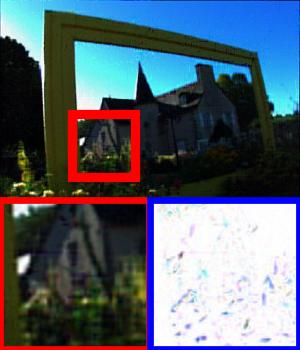}
&
\includegraphics[width=0.5286012in, height=0.81in]{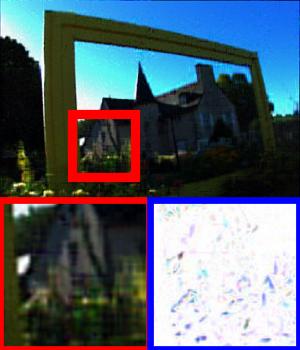}
&
\includegraphics[width=0.5286012in, height=0.81in]{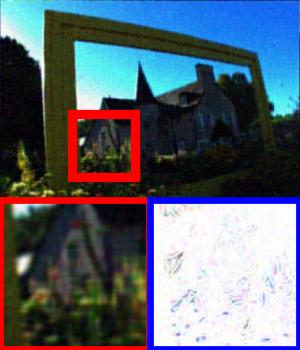}
&
\includegraphics[width=0.5286012in, height=0.81in]{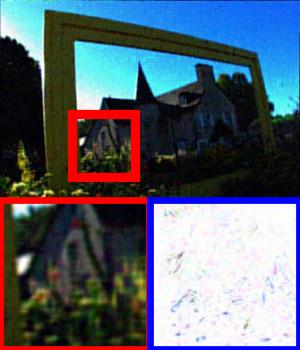}
&
\includegraphics[width=0.5286012in, height=0.81in]{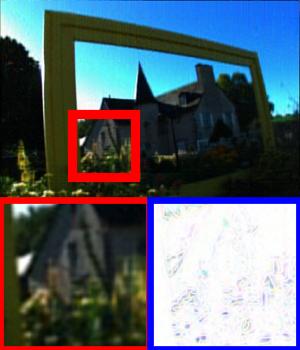}  %s12----->s4
\\

%\includegraphics[width=0.5286012in, height=0.81in]{cvcv-visu1-new13/s13}
%&
%\includegraphics[width=0.5286012in, height=0.81in]{cvcv-visu1-new13/s1}
%&
%\includegraphics[width=0.5286012in, height=0.81in]{cvcv-visu1-new13/s2}
%&
%\includegraphics[width=0.5286012in, height=0.81in]{cvcv-visu1-new13/s3}
%&
%\includegraphics[width=0.5286012in, height=0.81in]{cvcv-visu1-new13/s4}
%&
%\includegraphics[width=0.5286012in, height=0.81in]{cvcv-visu1-new13/s5}
%&
%\includegraphics[width=0.5286012in, height=0.81in]{cvcv-visu1-new13/s6}
%&
%\includegraphics[width=0.5286012in, height=0.81in]{cvcv-visu1-new13/s7}
%&
%\includegraphics[width=0.5286012in, height=0.81in]{cvcv-visu1-new13/s8}
%&
%\includegraphics[width=0.5286012in, height=0.81in]{cvcv-visu1-new13/s9}
%&
%\includegraphics[width=0.5286012in, height=0.81in]{cvcv-visu1-new13/s10}
%&
%\includegraphics[width=0.5286012in, height=0.81in]{cvcv-visu1-new13/s11}
%&
%\includegraphics[width=0.5286012in, height=0.81in]{cvcv-visu1-new13/s12}
%\\

\includegraphics[width=0.5286012in, height=0.81in]{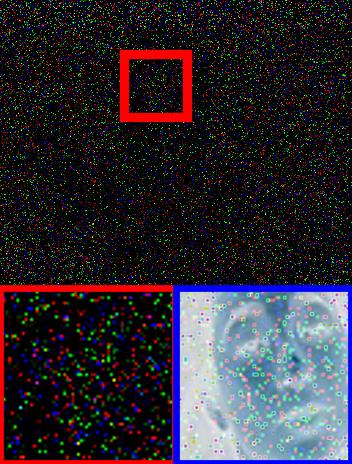}
&
\includegraphics[width=0.5286012in, height=0.81in]{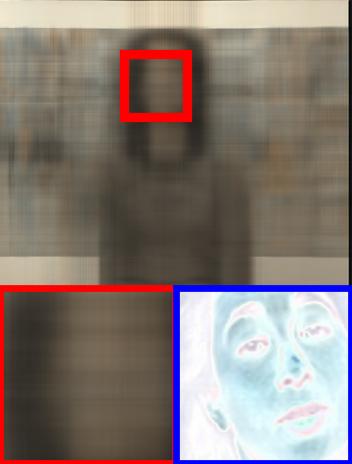}
&
\includegraphics[width=0.5286012in, height=0.81in]{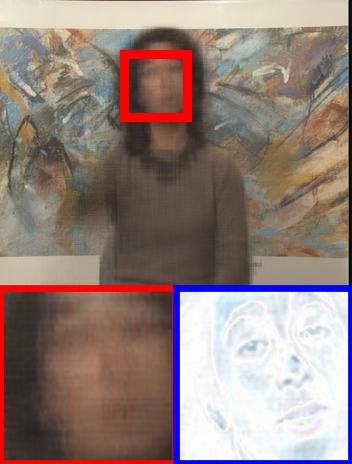}
&
\includegraphics[width=0.5286012in, height=0.81in]{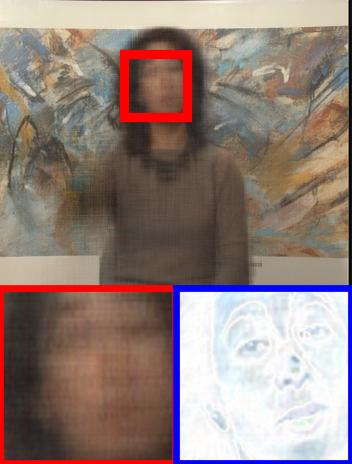}
&
\includegraphics[width=0.5286012in, height=0.81in]{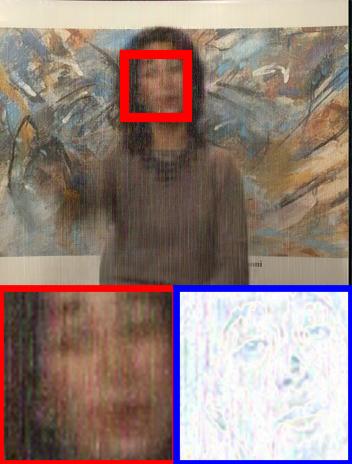} %%%
&
\includegraphics[width=0.5286012in, height=0.81in]{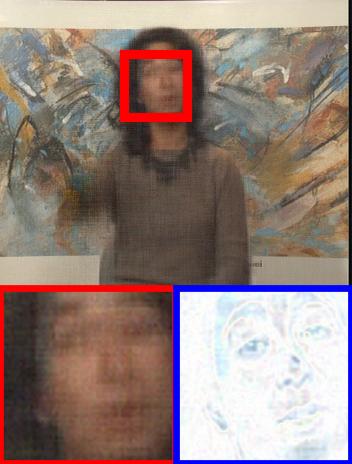}
&
\includegraphics[width=0.5286012in, height=0.81in]{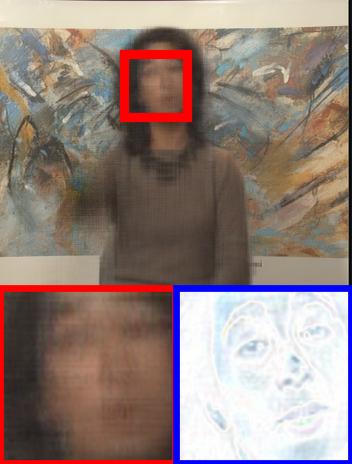}
&
\includegraphics[width=0.5286012in, height=0.81in]{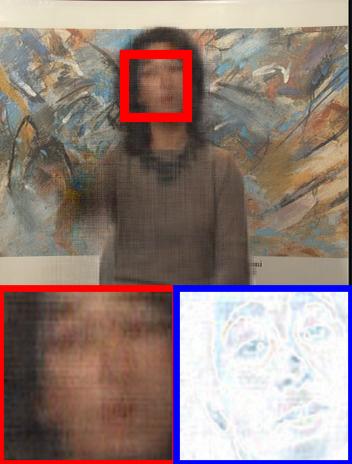}
&
\includegraphics[width=0.5286012in, height=0.81in]{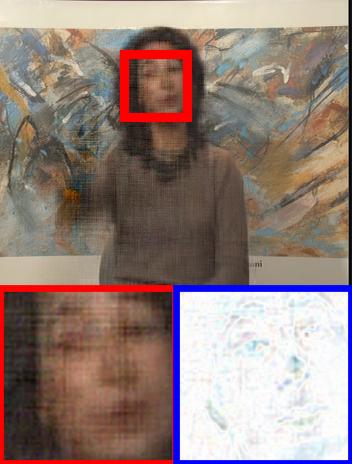}
&
\includegraphics[width=0.5286012in, height=0.81in]{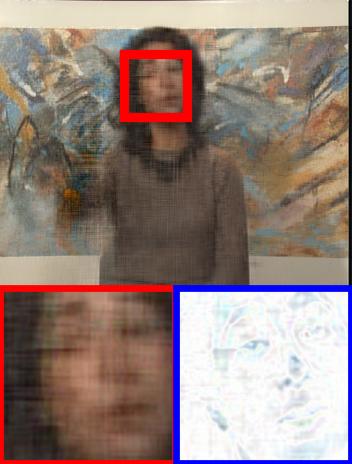}
&
\includegraphics[width=0.5286012in, height=0.81in]{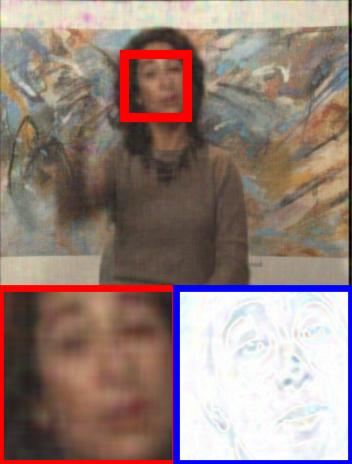}
&
\includegraphics[width=0.5286012in, height=0.81in]{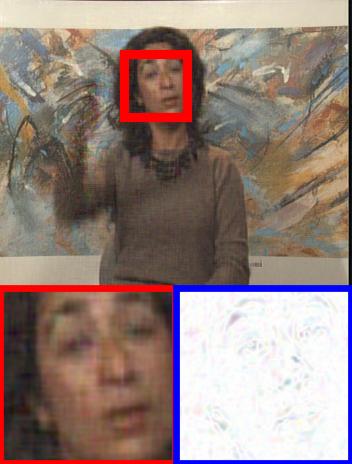}
&
\includegraphics[width=0.5286012in, height=0.81in]{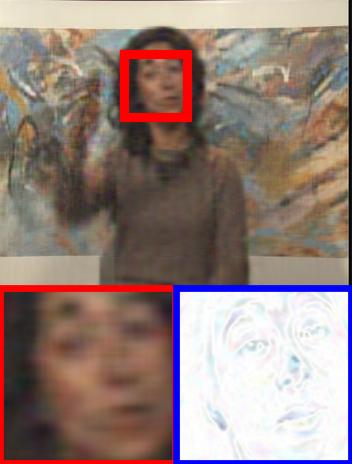}
\\

(a)   &
  (b)  & (c) &
(d) & (e)
 &(f)& (g) &
 (h) &
 (i)
 &  (j)
  &
 (k) &(l)&(m)
\end{tabular}
\vspace{-0.15cm}
\caption{
Visual comparison of various RLRTC methods for  CVs/LFIs inpainting under %$(SR, NR)= (0.1, 1/3)$
$(SR, NR)= (0.05, 0.5)$
 and  $(SR, NR)= (0.1, 0.5)$, respectively.
 %From top to bottom, the parameter pair $(SR, NR)$ are
%(0.1, 0.5),
%(0.1, 0.5) and (0.05, 0.5), respectively.
%Top row: the (5, 1)-th frame of Landsat-7. Middle row: the (2, 6)-th frame of
%SPOT-5. Bottom row: the (3, 5)-th frame of T22LGN.
From left to right: (a) Observed, (b) TRNN,
(c)  TTNN, (d) TSPK, (e) TTLRR, (f) LNOP, (g) NRTRM, (h) HWTNN,  (i) HWTSN,  (j) R-HWTSN,  (k)  TCTV-RTC, (l) GNRHTC, (m) R-GNRHTC.}
\vspace{-0.2cm}
\label{fig-cvcv} % \label{fig-lfilfi}
\end{figure*}

%
%%%%%%%%%%%%%%%%%%%%%%%%%%%%%%%%%%%%%%%%%%%%%%%%%%%%%%%%%%%%%%%%%%%%%%%%%%%%%%%%%%%%%%%%%%%%%%%%%%%%%%%%%%%%%%%%%%%%%%%%%%%%%%%%%%%%%%%%%%%%%%%%%
\begin{figure*}[!htbp]
\renewcommand{\arraystretch}{0.5}
\setlength\tabcolsep{10pt}
\centering
\begin{tabular}{ccc }
\centering

\includegraphics[width=1.976in, height=1.573in]{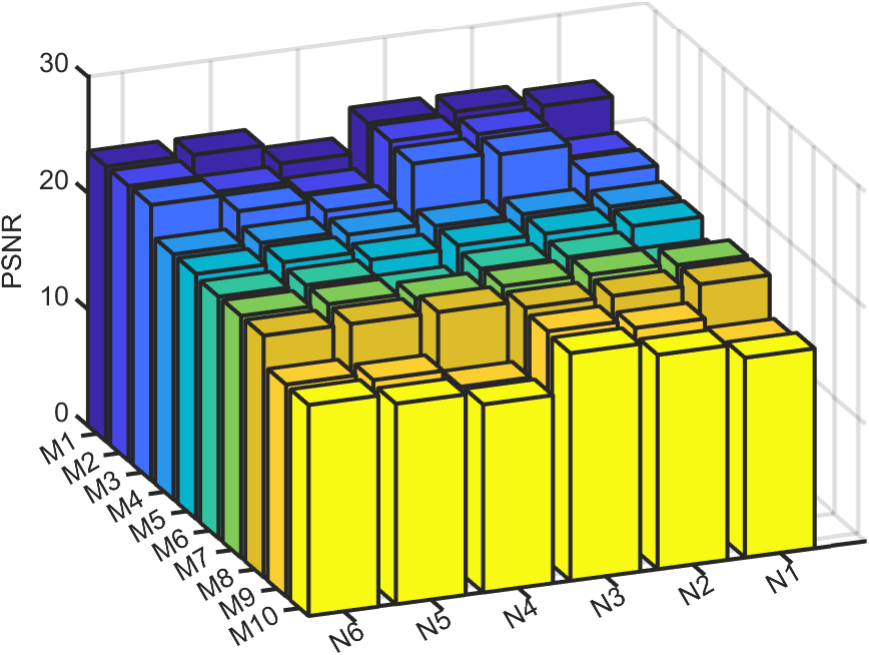}&
\includegraphics[width=1.976in, height=1.573in]{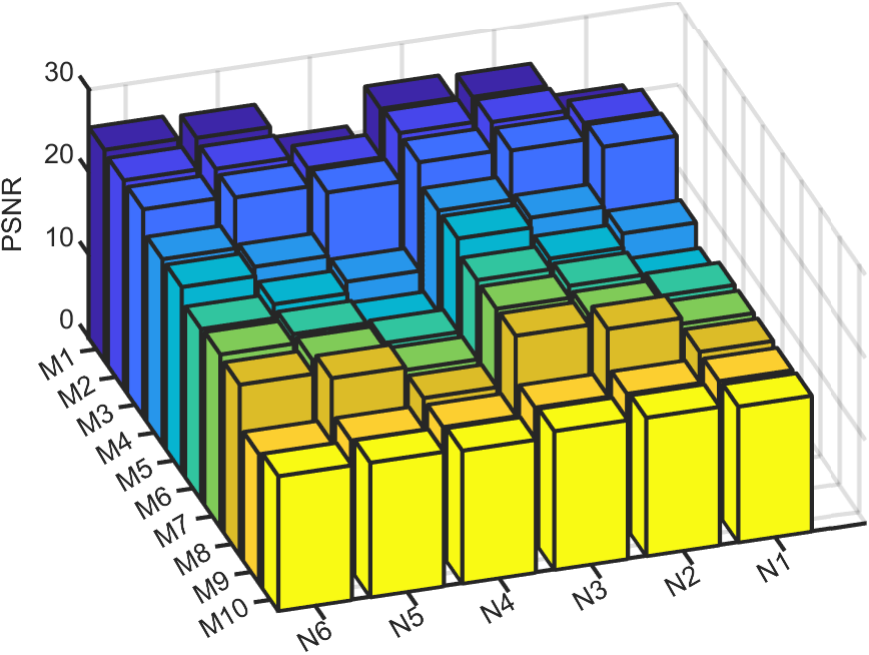}&
\includegraphics[width=1.976in, height=1.573in]{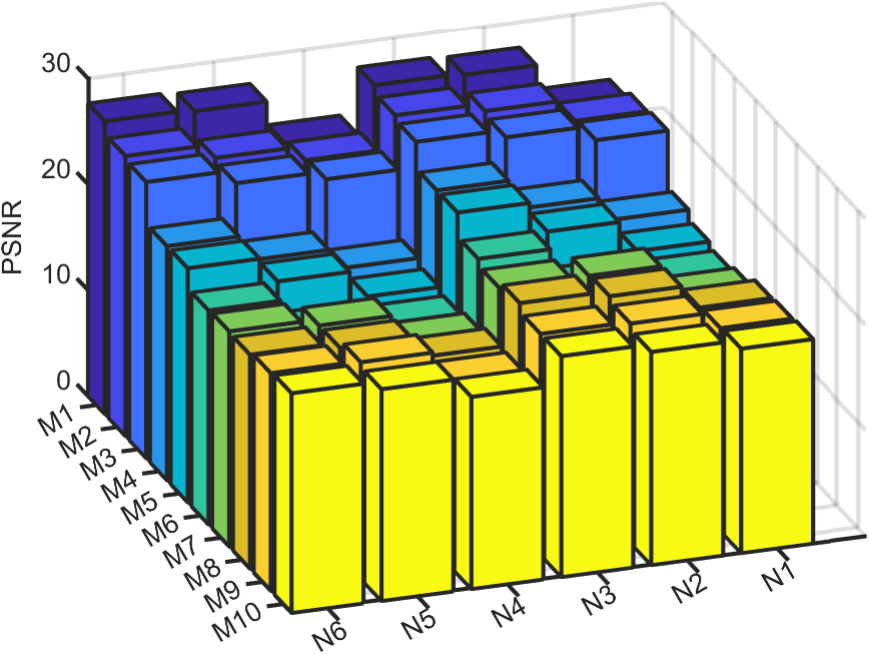}
\\
\includegraphics[width=1.976in, height=1.4573in]{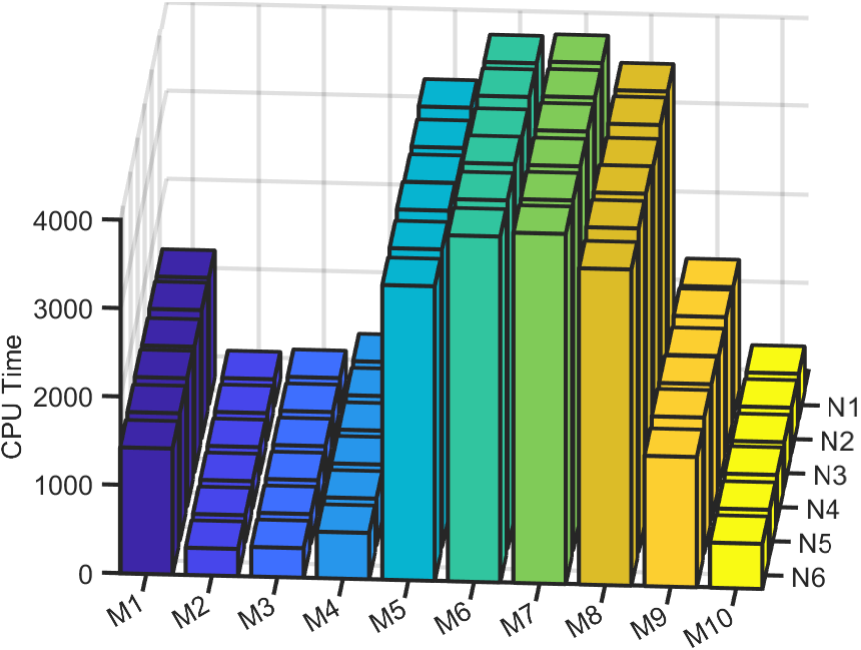}&
\includegraphics[width=1.976in, height=1.4573in]{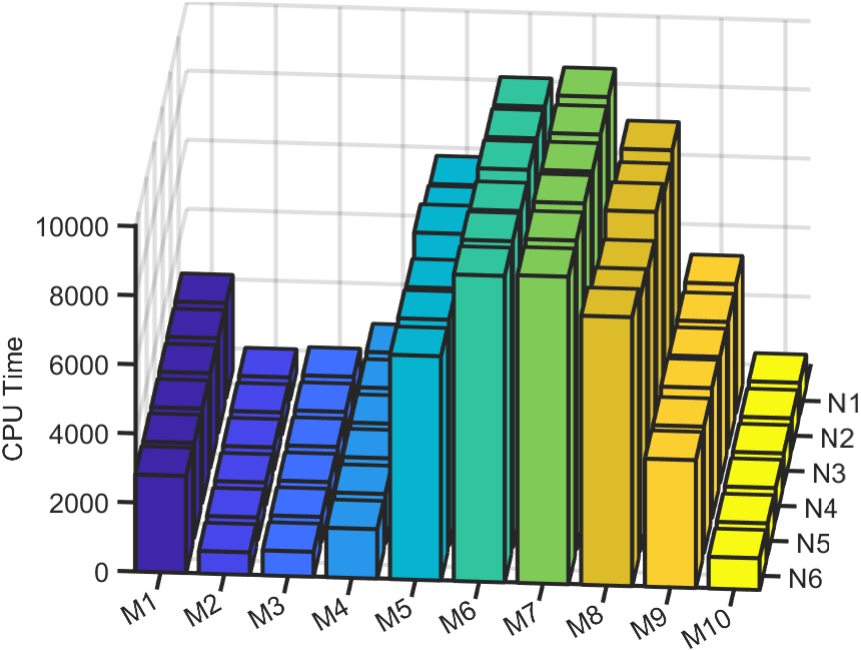}&
\includegraphics[width=1.976in, height=1.4573in]{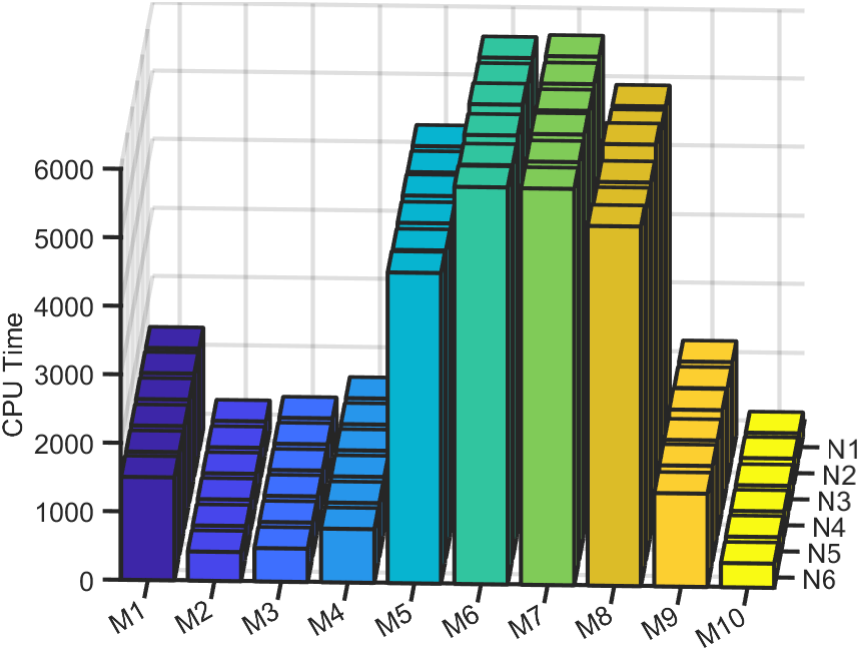}
\\
\footnotesize{{{(a)}} Fourth-order MRSIs}  &
  \footnotesize{(b) Fourth-order CVs}  & \footnotesize{(c) Fifth-order LFIs}

\end{tabular}
\vspace{-0.15cm}
\caption{
The   recovery  performance
of the  proposed and compared OBTC
%GNRHTC  algorithm
under
 various  sampling rates  and Gaussian noise levels. % in which
 \textbf{M1:}  GNOBHTC, \textbf{M2:} R1-GNOBHTC, \textbf{M3:} R2-GNOBHTC,
 \textbf{M4:} 1BRTC \cite{hou2025robust}, \textbf{M5:} MLE-TNN \cite{hou2021one},
  \textbf{M6:} MLE-Max \cite{ghadermarzy2019learning}, \textbf{M7:} MLE-Factor \cite{ghadermarzy2019learning},
 \textbf{M8:} MLE-T2M \cite{aidini20181}, \textbf{M9:}  MLE-Mat \cite{davenport20141}, \textbf{M10:}
 L2-Mat \cite{chen2023high}.
 \textbf{N1:} $\operatorname{SR}=0.1, \sigma=0$,
  \textbf{N2:} $\operatorname{SR}=0.3, \sigma=0$,
   \textbf{N3:} $\operatorname{SR}=0.5, \sigma=0$,
    \textbf{N4:} $\operatorname{SR}=0.1, \sigma=0.2$,
     \textbf{N5:} $\operatorname{SR}=0.3, \sigma=0.2$,
      \textbf{N6:} $\operatorname{SR}=0.5, \sigma=0.2$.
   %
   %    approaches, including  MLE-Mat ,  L2-Mat ,   MLE-TNN ,
 % MLE-T2M ,   MLE-Max , MLE-Factor ,  and  $1$BRTC .
%%%%%%%%%%%%%%%%%%%%%%%%%%%%%%%%%%%%%%%%%%%
%
}
\vspace{-0.2cm}
\label{fig-nonconvex-hsi11111111}
\end{figure*}

%%%%%%%%%%%%%%%%%%%%%%%%%%%%%%%%%%%%%%%%%%%%%%%%%%%%%%%%%%%%%%%%%%%%%%%%%%%%%%%%%%%%%%%%%%%%%%%%%%%%%%%%%%%%%%%%%%%%%%%%%%%%%%%%%%%%%%%%%%%%%%%%%
\begin{figure*}[!htbp]
\renewcommand{\arraystretch}{0.7}
\setlength\tabcolsep{0.43pt}
\centering
\begin{tabular}{ccc  ccc  ccc  ccc   }%cc ccc  ccc c cc
\centering
%\\
%\toprule
% %\toprule
%\\
%(a)   &
%  (b)  & (c) &
%(d) & (e)
% &(f)\\
\includegraphics[width=0.586in, height=0.658in]{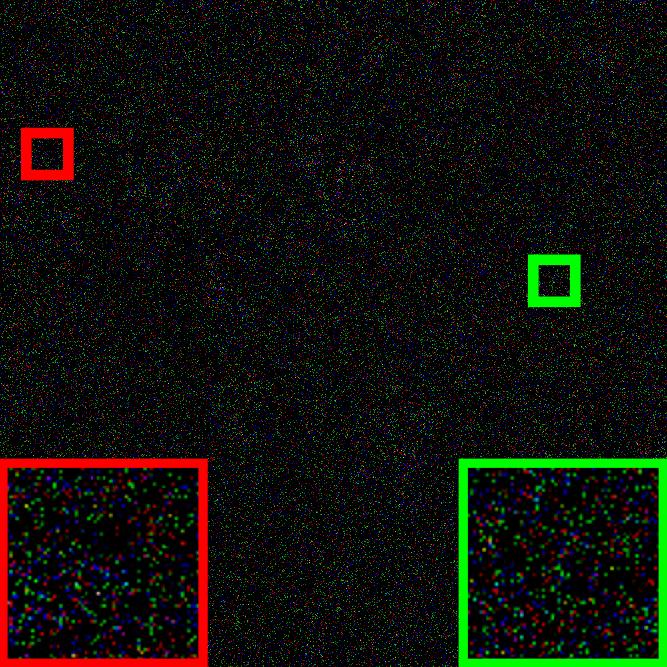}
&
\includegraphics[width=0.586in, height=0.658in]{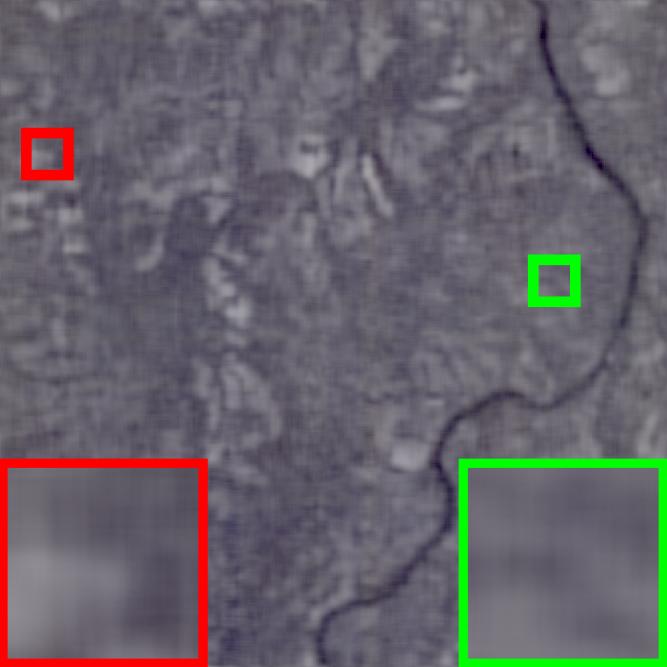}
&
\includegraphics[width=0.586in, height=0.658in]{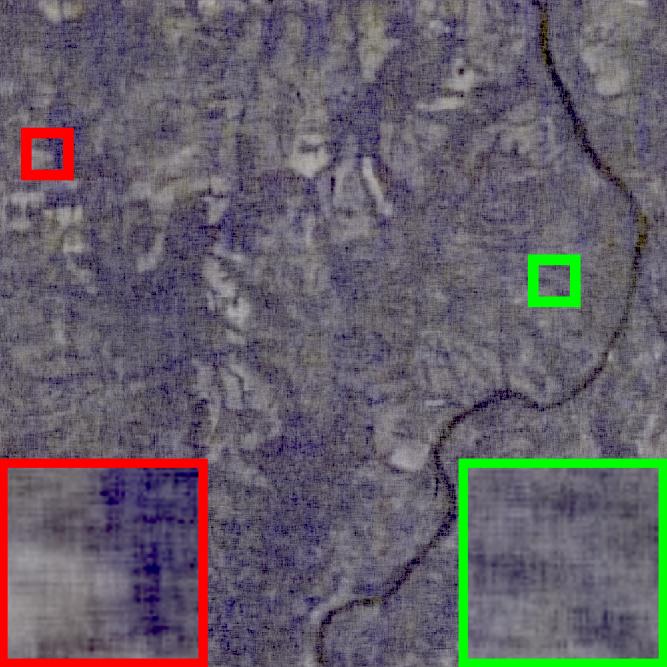}
&
\includegraphics[width=0.586in, height=0.658in]{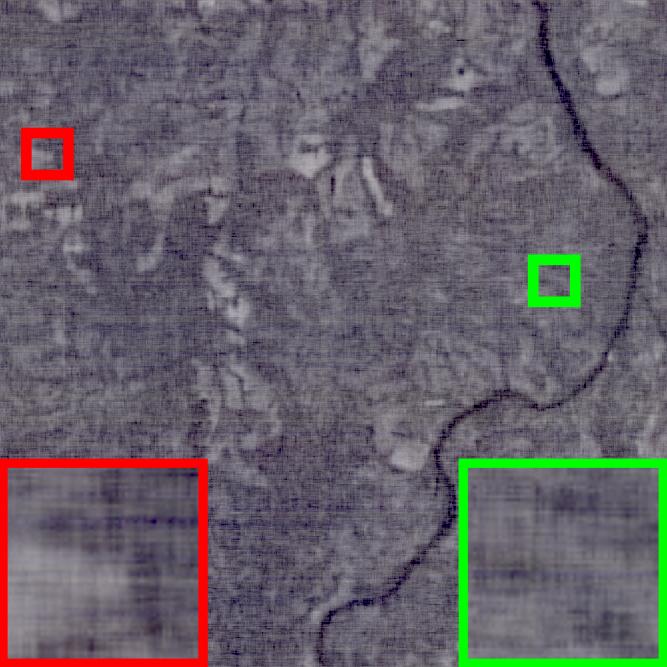}
&
\includegraphics[width=0.586in, height=0.658in]{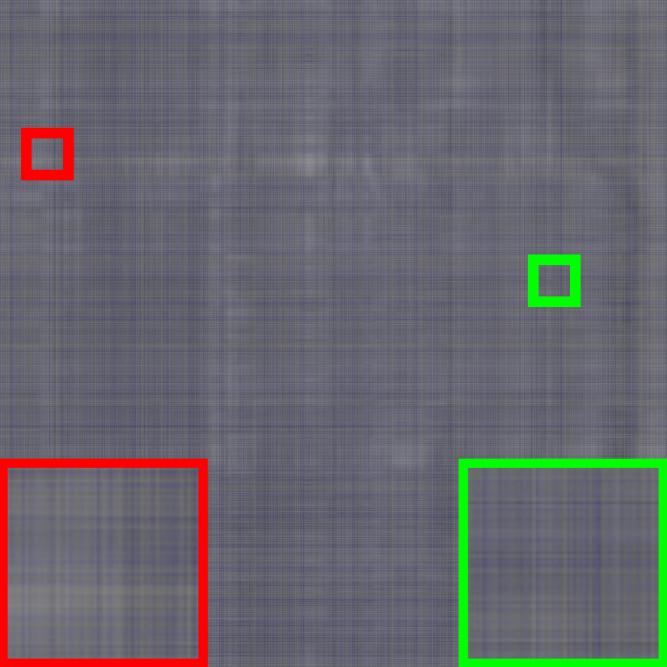}
&
\includegraphics[width=0.586in, height=0.658in]{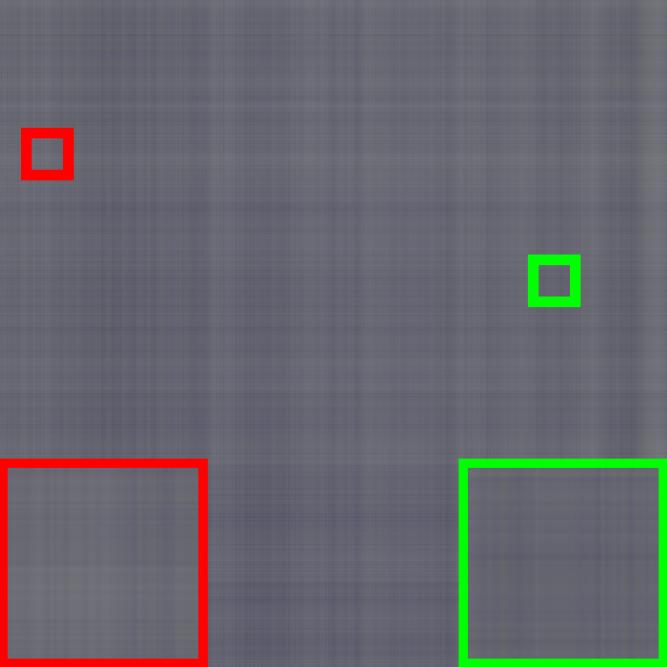}
&
%\\
%(g) &
% (h) &
% (i)
% &  (j)
%  &
% (k) &(l)\\
%%
\includegraphics[width=0.586in, height=0.658in]{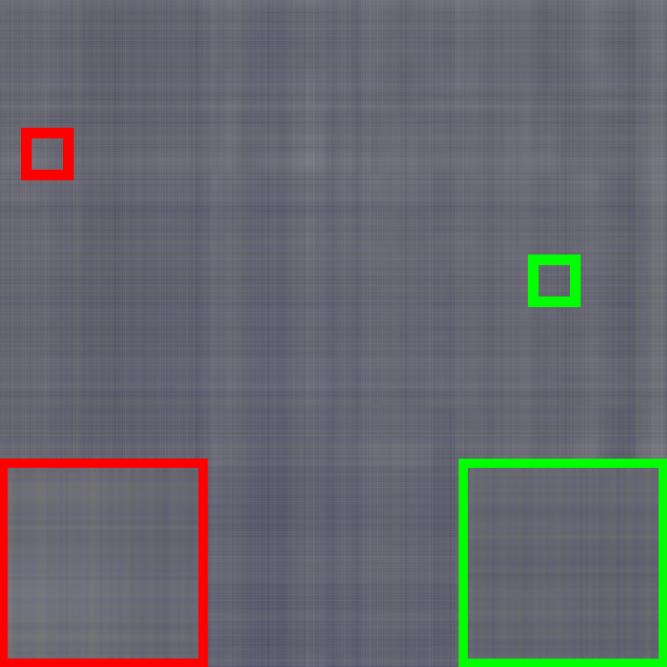}
&
\includegraphics[width=0.586in, height=0.658in]{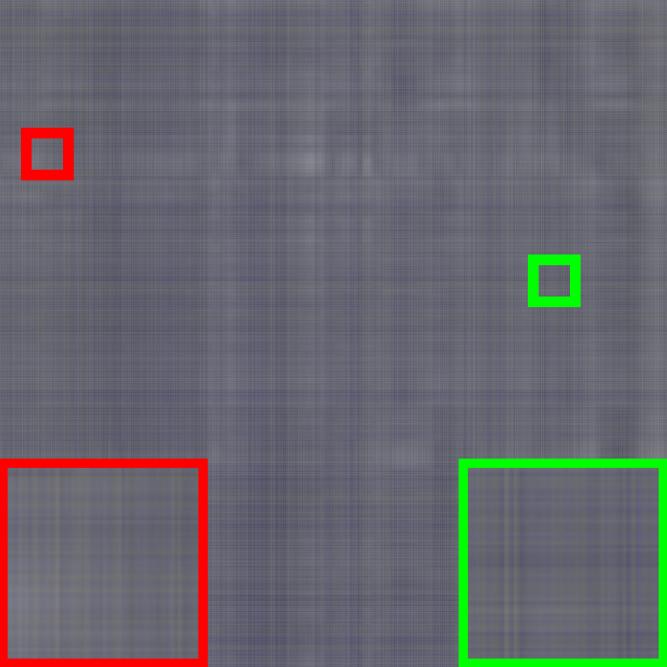}
&
\includegraphics[width=0.586in, height=0.658in]{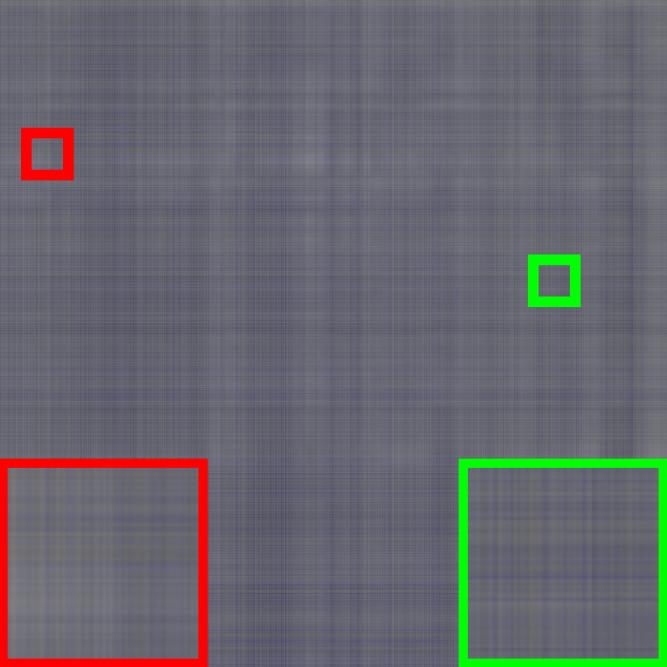}
&
\includegraphics[width=0.586in, height=0.658in]{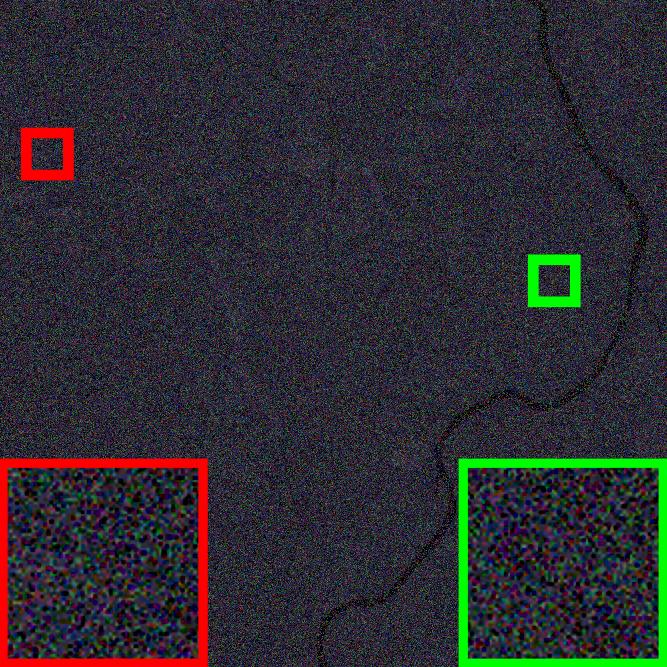}% s4

&
\includegraphics[width=0.586in, height=0.658in]{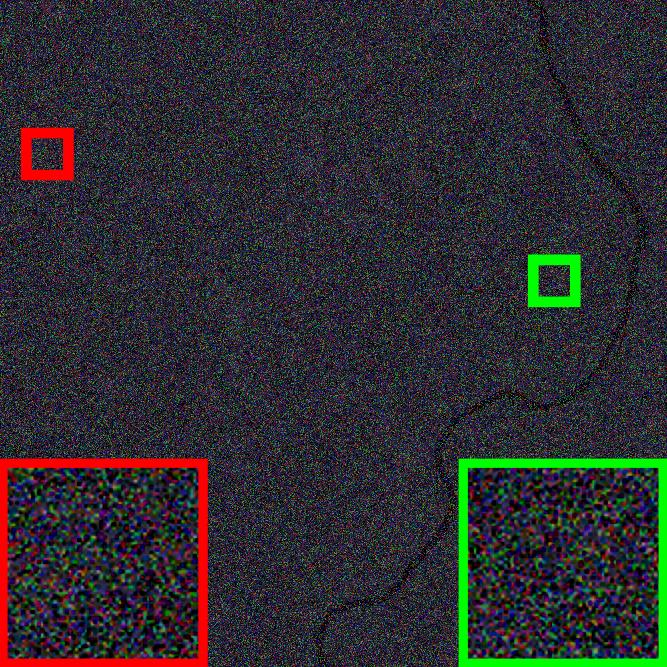} %s5

&
\includegraphics[width=0.586in, height=0.658in]{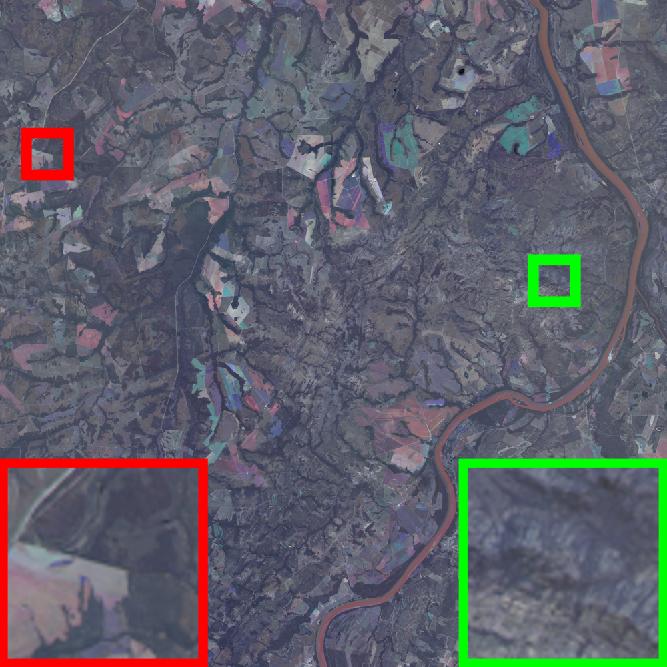}
 \\

 \includegraphics[width=0.586in, height=0.8658in]{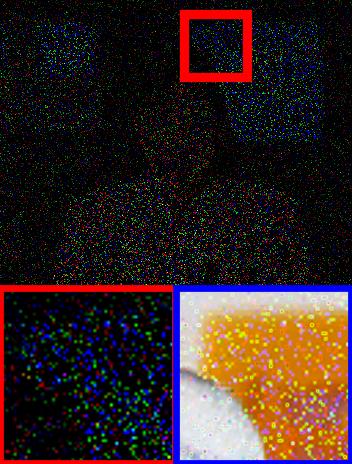}
&
\includegraphics[width=0.586in, height=0.8658in]{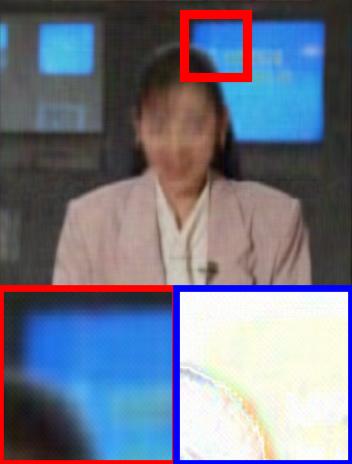}
&
\includegraphics[width=0.586in, height=0.8658in]{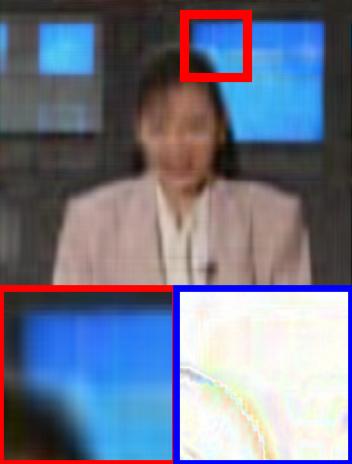}
&
\includegraphics[width=0.586in, height=0.8658in]{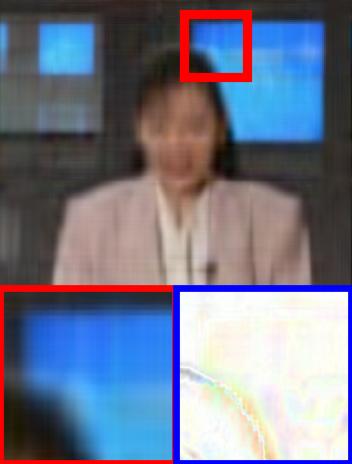}
&
\includegraphics[width=0.586in, height=0.8658in]{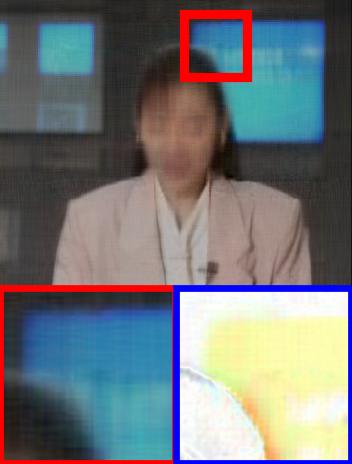}
&
\includegraphics[width=0.586in, height=0.8658in]{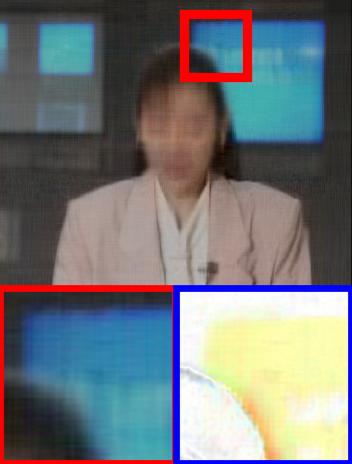} % 4-5 7-8-9
&
%\\
%(g) &
% (h) &
% (i)
% &  (j)
%  &
% (k) &(l)\\
% width=0.586in, height=0.658in8
\includegraphics[width=0.586in, height=0.8658in]{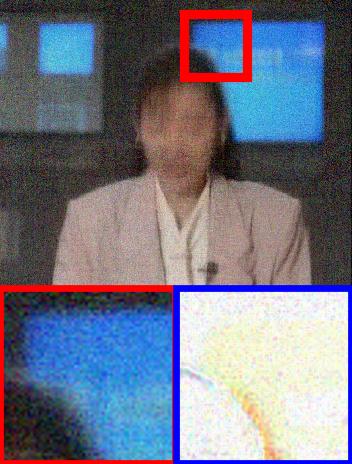}
%  obtc-cvcv-visu003
&
\includegraphics[width=0.586in, height=0.8658in]{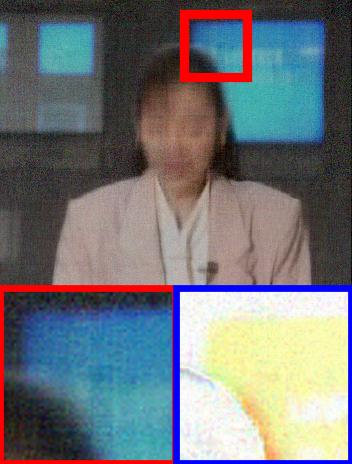}
&
\includegraphics[width=0.586in, height=0.8658in]{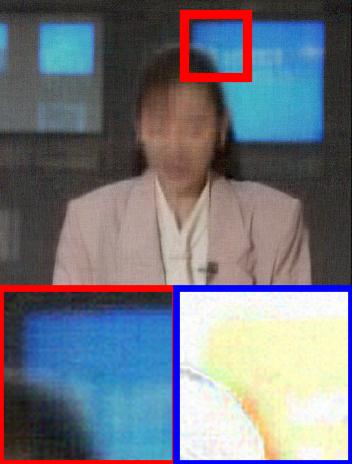}
&
\includegraphics[width=0.586in, height=0.8658in]{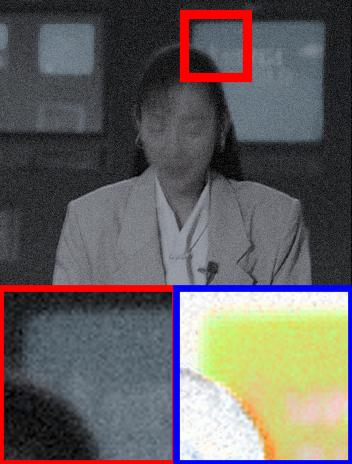}% s4

&
\includegraphics[width=0.586in, height=0.8658in]{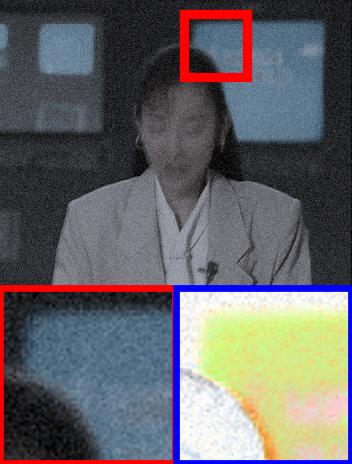} %s5

&
\includegraphics[width=0.586in, height=0.8658in]{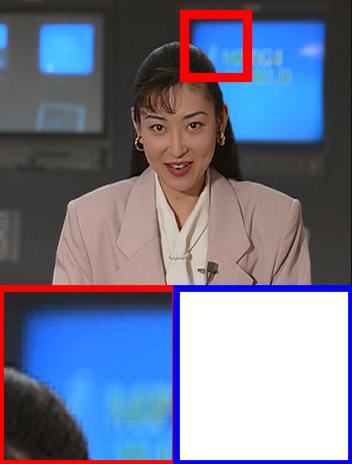}

 \\
 (a)   &
  (b)  & (c) &
(d) & (e)
 &(f)& (g) &
 (h) &
 (i)
 &  (j)
  &
 (k) &(l)
%
%\\
% \toprule
% %\toprule
%\\
\end{tabular}
% \vspace{-0.4cm}
%\vspace{-0.15cm}
\caption{ %\textbf{HSIs-Example2 ---}  %\textbf{HSIs Recovery (Case2):}
Visual comparison of various OBTC methods for  CVs and MRSIs datasets recovery.
 %From top to bottom,
%
The observed CVs  and MRSIs are corrupted by noises generated from Gaussian distribution with mean zero and standard deviation $\sigma=0.2$,
and the sampling rates is set to be $0.1$ (CVs), $0.3$ (MRSIs).
%inpainting.
%From top to bottom, the parameter pair $(SR, NR)$ are set to be (0.1, 0.5) and
%(0.05, 0.5), respectively.
%
%(0.01, 0.5) and (0.01, 0.5), respectively.
%Top row: the (5, 1)-th frame of Landsat-7. Middle row: the (2, 6)-th frame of
%SPOT-5. Bottom row: the (3, 5)-th frame of T22LGN.
% \textbf{M1:}  GNOBHTC, \textbf{M2:} R1-GNOBHTC, \textbf{M3:} R2-GNOBHTC,
% \textbf{M4:} 1BRTC \cite{hou2025robust}, \textbf{M5:} MLE-TNN \cite{hou2021one},
%  \textbf{M6:} MLE-Max \cite{ghadermarzy2019learning}, \textbf{M7:} MLE-Factor \cite{ghadermarzy2019learning},
% \textbf{M8:} MLE-T2M \cite{aidini20181}, \textbf{M9:} L2-Mat \cite{chen2023high}, \textbf{M10:} MLE-Mat \cite{davenport20141}.
From left to right: (a) Observed,
(b) GNOBHTC,  (c) R1-GNOBHTC,  (d) R2-GNOBHTC,  (e) 1BRTC,  (f) MLE-TNN,  (g) MLE-Max, (h) MLE-Factor,   (i) MLE-T2M,   (j)  L2-Mat,
 (k)  MLE-Mat,  (l) Ground-truth.}
\vspace{-0.64cm}
\label{fig_hsi2kkkkkk}
\end{figure*}

%\vspace{-0.15cm}

 \subsubsection {\textbf{Results and Analysis %Results
  on Fourth-Order  MRSIs and CVs, and Fifth-Order LFIs}} \label{cv-mrsi}

  In Tables  \ref{table-mrsi-CV-LFI},  %\ref{table-mrsi}  and \ref{table-cvs},
  we summarize   the mean PSNR, RSE, SSIM values and CPU time of different  RLRTC methods for %four-order
  MRSIs, CVs,  and LFIs,
where %sr = 0.1, 0.2 and τ = 0.3, 0.5.
$SR= 0.1,0.2$, $NR= 1/3,0.5$.
%%%%%%%%%%%%%%%%%%%%%%%%%%%%%%%%%%%%%%%%%%%%%%%%%%%%%%%%%%%%%%%%%%%%%%%%%%%%%%%%%%%%%%%%%%%%%%%%%%%%%%%%%
Figures \ref{fig_MRSI} and    \ref{fig-cvcv} %  fig-lfilfi
% 5 and 6
 present the visual comparisons of the enlarged areas in restored CVs, LFIs,  and MRSIs, % obtained through various RTC methods, respectively.
 which were obtained through various RLRTC methods.
 The conclusions from the robust restoration of MRSIs mirror those of third-order MRIs and HSIs.
 %The conclusions drawn from the robust restoration of MRSIs are highly similar to
 %those obtained from the robust restoration of MRIs and HSIs.
 Namely, our method continues to exhibit a notable superiority over other competing peers. %competing approaches.
 % Namely, our method continues to demonstrate a significant advantage beyond other %compared to
  % competing peers. %approaches.
 % Specifically,
%
In particular, at low sampling rates and high noise levels, the RLRTC models that  purely rely on
$\textbf{L}$ %L-
prior fail to reconstruct missing and  noisy tensor data well, whereas our proposed method performs oppositely.
Although the gap between the proposed method and those purely using $\textbf{L}$ prior in CVs/LFIs restoration %is not as significant as that
  is not as extremely pronounced as that
 observed %in the previous three tensor restoration scenarios,
 in the first three tensor recovery scenarios,
  it still demonstrates good performance in terms of visual quality and quantitative %results.  %
  outcomes.
  Compared with  the highly competitive TCTV-RTC and HWTSN algorithms,
 the improvements of  the proposed GNRHTC method are %achieves
 around  $2$ dB in terms of PSNR values.
 In addition, among all nonconvex RLRTC methods, our randomized approach achieves the shortest CUP %runtime.
 running time.

\vspace{-0.3cm}

\subsection{\textbf{%Comparative
Experiments 3:
One-Bit Tensor Completion in
Quantized Tensor Recovery}}

In this subsection, we compare the proposed GNOBHTC %GNOBHTC
method and its two randomized versions with some state-of-the-art %a number of  state-of-the-art
\textit{one-bit tensor completion} (OBTC) %recovery
  approaches, including  MLE-Mat \cite{davenport20141},
  L2-Mat \cite{chen2023high},   MLE-TNN \cite{hou2021one},
  MLE-T2M \cite{aidini20181},   MLE-Max \cite{ghadermarzy2019learning}, MLE-Factor \cite{ghadermarzy2019learning},
and
$1$BRTC \cite{hou2025robust}.
%In our randomized  versions, one is named  R1-GNOBHTC, integrating the % that integrates %fuses
%fixed-rank LRTC strategy,
%while the other is called  R2-GNOBHTC,  incorporating the %that incorporates
%fixed-accuracy LRTC strategy.
Two randomized versions, which integrate the fixed-rank and fixed-accuracy LRTC strategies,
are named as R1-GNOBHTC and R2-GNOBHTC, respectively.
%Two randomized  versions integrating the fixed-rank and the fixed-accuracy % LRTC
%LRTC strategies are named as R1-GNOBHTC and R2-GNOBHTC, respectively.

Under varying sampling rates and noise levels, Figure \ref{fig-nonconvex-hsi11111111} presents the quantitative results (including CPU time and PSNR) of the proposed and comparative OBTC methods evaluated on LFIs, CVs, and MRSIs datasets.
%
%fig_hsi2kkkkkk
%
The corresponding
visual comparisons %of various OBTC methods for CVs and MRSIs datasets recovery.
are provided in Figure \ref{fig_hsi2kkkkkk}.
Similarly to the aforementioned non-quantized tensor recovery, the proposed deterministic GNOBHTC method achieves higher PSNR values than all competing algorithms.
Additionally, the proposed randomized algorithms demonstrate the lowest computational  cost among all evaluated methods.
This validates the advantages of the proposed nonconvex tensor modeling, $\textbf{L}$+$\textbf{S}$ priors-based representation paradigm, and randomized  strategies in quantized tensor recovery.

\vspace{-0.43cm}
%\subsection{\textcolor[rgb]{0.00,0.00,0.00}{\textbf{Parameters   Analysis}}}
\subsection{\textcolor[rgb]{0.00,0.00,0.00}{\textbf{Related Discussions}}}
In this subsection,
under various types of %various %three types of
real-world  tensors
with different sampling rates and impulse noise levels, we mainly focus on the following key %investigations: %
experiments
 %key experiments
 to illustrate our research motivations:
 %% 非凸正则效果优于凸方法，凸正则差异不大
 %%  横： 非凸正则子    纵：采样率和噪声水平   Z：psnr
 %%  图1： 三阶张量  图二：四阶张量   图三：五阶张量  或者噪声类型： tube slice entry-wise
\begin{itemize}
  \item \textbf{Experiment 1:}
   %1)
  %%%% 讨论非凸函数
       %In this subsection,
%we verify the performance of different nonconvex regularization penalties %different nonconvex functions
%for  $\Phi(\cdot)$ and  $\psi(\cdot)$ in our GNRHTC model,
   %%we mainly
   analyze the performance of
%investigate the influence of
 various nonconvex regularizers $\Phi(\cdot)$+$\psi(\cdot)$
  %combinations,
%i.e.,  $\Phi(\cdot)$+$\psi(\cdot)$ in our GNRHTC model,
%upon recovery performance
% under  different sampling rates and impulse noise levels,
% We analyze the performance of various regulons
on our  proposed % GNRHTC algorithm under deterministic and randomized  patterns;
deterministic and randomized recovery %GNRHTC
algorithms, respectively;

%% 验证光滑先验和低秩共同耦合的优势
\item \textbf{Experiment 2:}
analyze how different prior structures (i.e., $\textbf{L}$+$\textbf{S}$ priors and pure $\textbf{L}$  prior)
%values of p and q $\textbf{L}$+$\textbf{S}$ priors
influence the recovery performance of our proposed recovery %GNRHTC
algorithm in deterministic and randomized  patterns, respectively;

%% 验证提出随机算法的优势
%% 以及比之前TIP29949的优势
%% 一种方式：TIP29949
%%另一种：tucker压缩
\item \textbf{Experiment 3:}
investigate the influence of various  fixed-rank  % adaptive
 randomized  Tucker compression methods
 upon restoration results of our proposed randomized recovery %GNRHTC
  algorithm
 %within
 under   different target rank %block size $b$; %techniques;
 $\bm{r}=(r_1, \cdots, r_d)$;

\item \textbf{Experiment 4:}
 investigate the influence of various  fixed-precision % adaptive
 randomized  Tucker compression approaches
 upon restoration results of our proposed randomized recovery %GNRHTC
 algorithm
 %within
 under   different block size $b$; %techniques;
\end{itemize}
%%%%%%%%%%%%%%%%%%%%
Wherein,  the test tensor data are HSIs, MRI datasets, CVs, MRSIs and LFIs (\textit{please see Table \ref{exp-datasets} for details}).

%
%%%%%%%%%%%%%%%%%%%%%%%%%%%%%%%%%%%%%%%%%%%%%%%%%%%%%%%%%%%%%%%%%%%%%%%%%%%%%%%%%%%%%%%%%%%%%%%%%%%%%%%%%%%%%%%%%%%%%%%%%%%%%%%%%%%%%%%%%%%%%%%%%
\begin{figure*}[!htbp]
\renewcommand{\arraystretch}{0.5}
\setlength\tabcolsep{5pt}
\centering
\begin{tabular}{ccc }
\centering

\includegraphics[width=1.976in, height=1.73in]{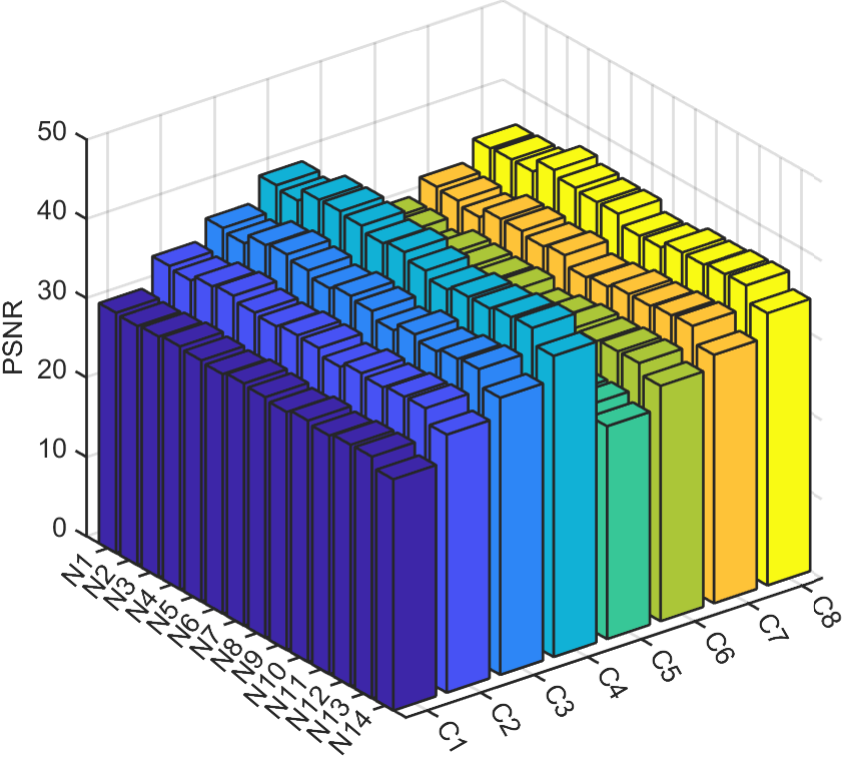}&
\includegraphics[width=1.976in, height=1.73in]{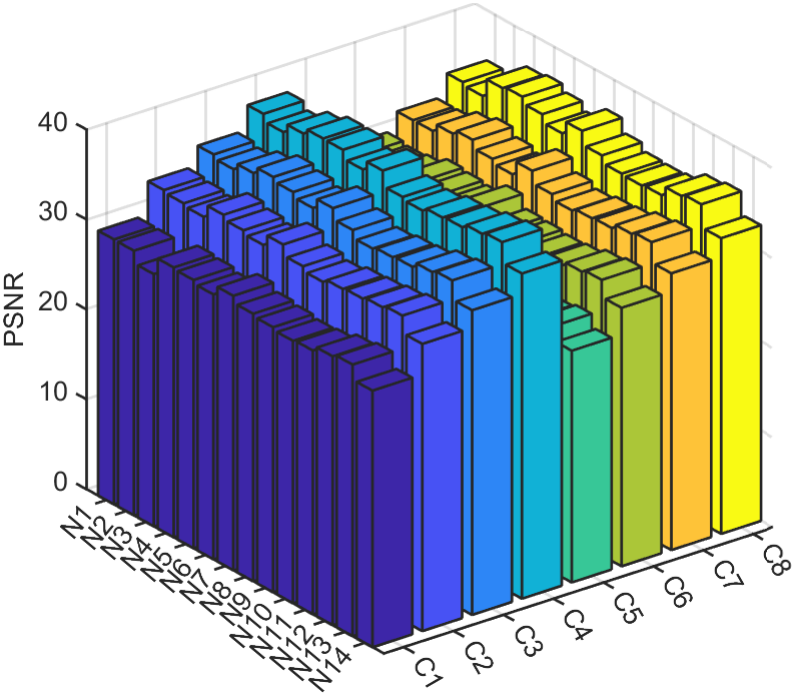}
&
\includegraphics[width=1.976in, height=1.73in]{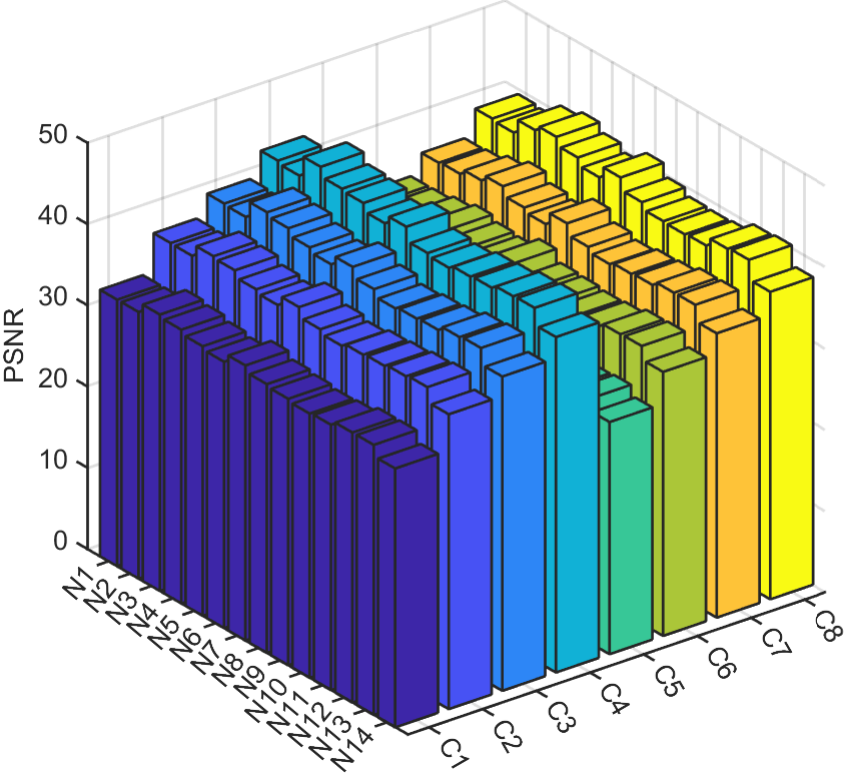}
\\
\footnotesize{{{(a)}} HSIs: $300 \times 300 \times 60$}  &
  \footnotesize{(b) CVs: $288 \times 352 \times 3 \times 60$}  & \footnotesize{(c) LFIs: $200 \times 300 \times 3 \times 15 \times 15$}

\end{tabular}
\vspace{-0.15cm}
\caption{
The influence of %various regularization parameters $\lambda$
different  nonconvex combinations $\Phi(\cdot)$+$\psi(\cdot)$  upon  recovery  performance
of the  proposed deterministic % and randomized
GNRHTC algorithm
under
 various  sampling rates $SR$ and impulse noise levels $NR$. % in which
}
\vspace{-0.2cm}
\label{fig-nonconvex-hsi}
\end{figure*}

%
%%%%%%%%%%%%%%%%%%%%%%%%%%%%%%%%%%%%%%%%%%%%%%%%%%%%%%%%%%%%%%%%%%%%%%%%%%%%%%%%%%%%%%%%%%%%%%%%%%%%%%%%%%%%%%%%%%%%%%%%%%%%%%%%%%%%%%%%%%%%%%%%%
\begin{figure*}[!htbp]
\renewcommand{\arraystretch}{0.5}
\setlength\tabcolsep{10pt}
\centering
\begin{tabular}{ccc }
\centering

\includegraphics[width=1.976in, height=1.573in]{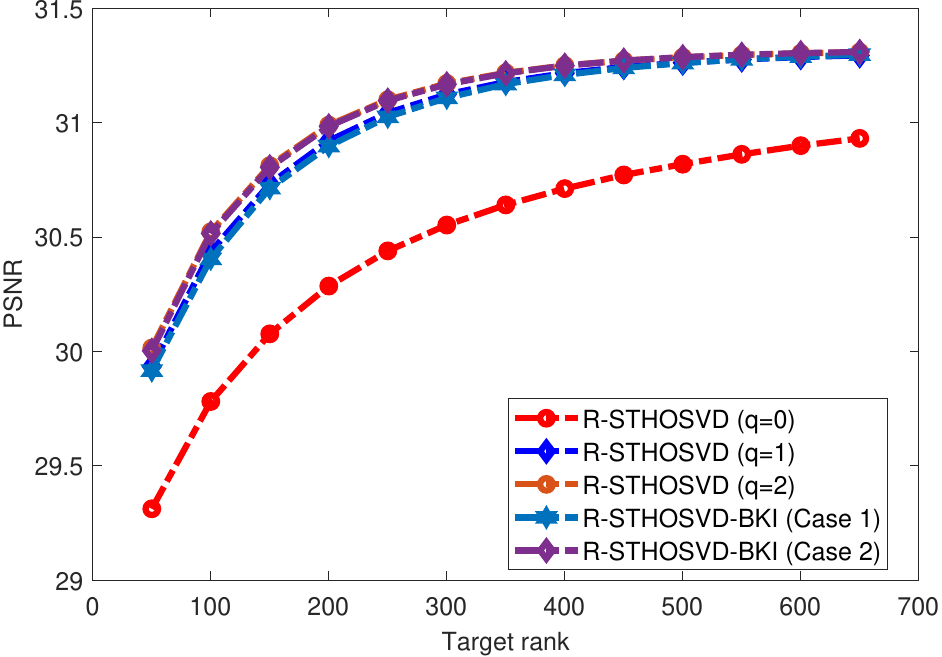}&
\includegraphics[width=1.976in, height=1.573in]{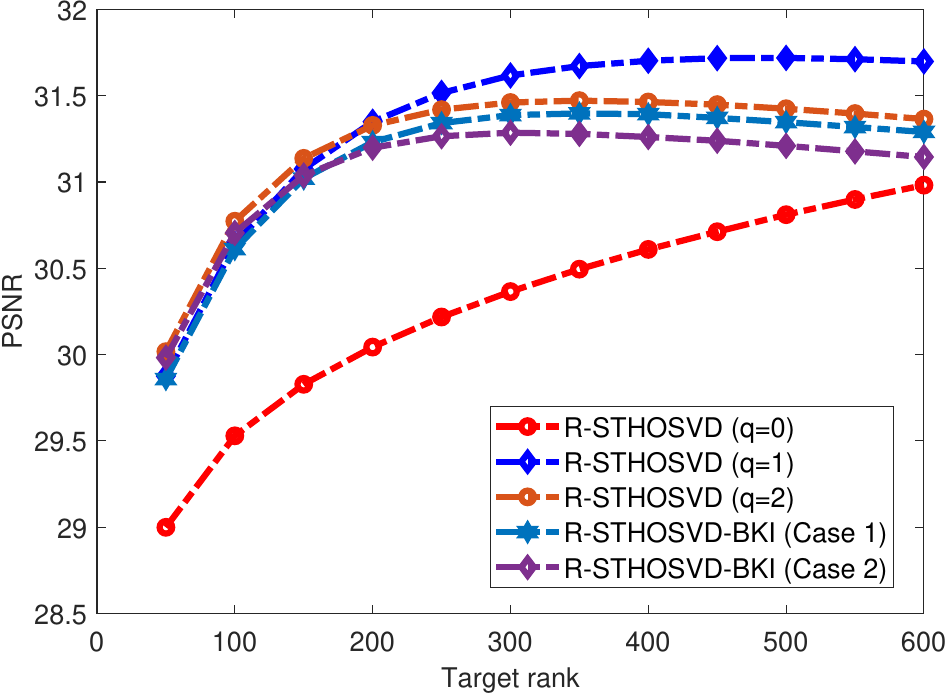}&
\includegraphics[width=1.976in, height=1.573in]{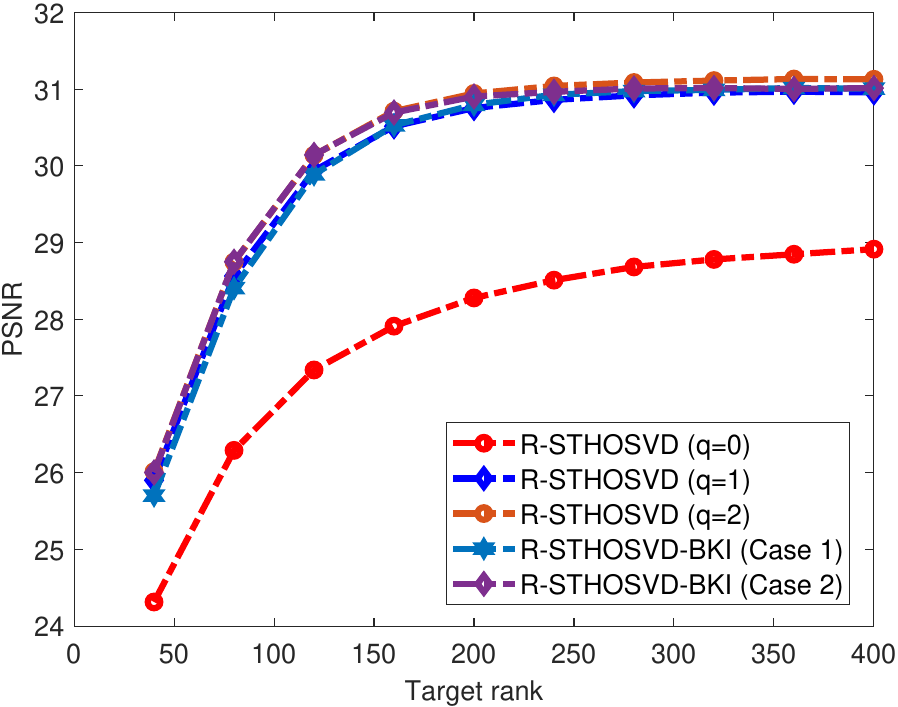} % cv25-psnr1
\\
\includegraphics[width=1.976in, height=1.573in]{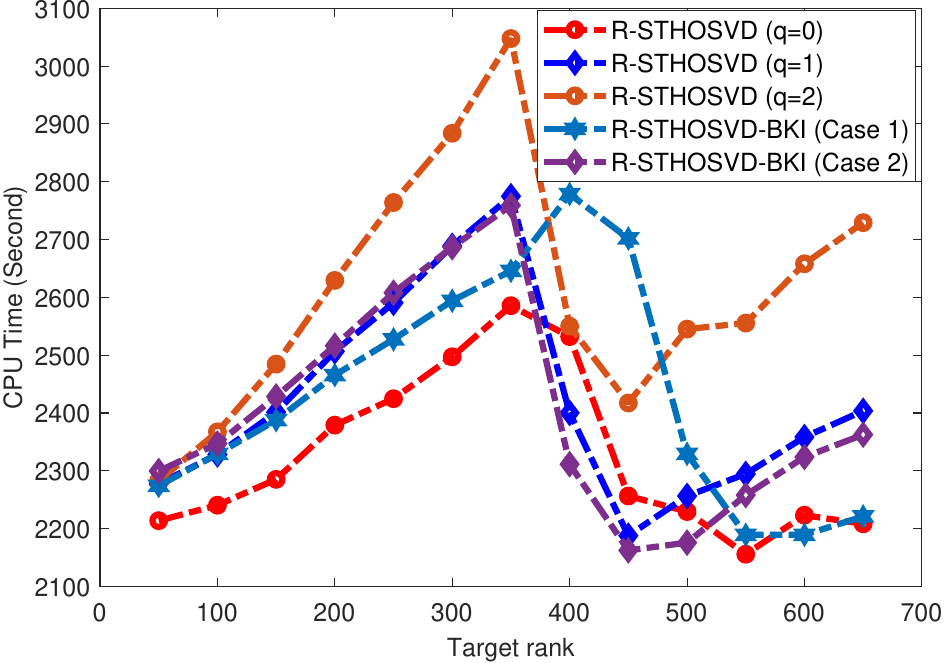}&
\includegraphics[width=1.976in, height=1.573in]{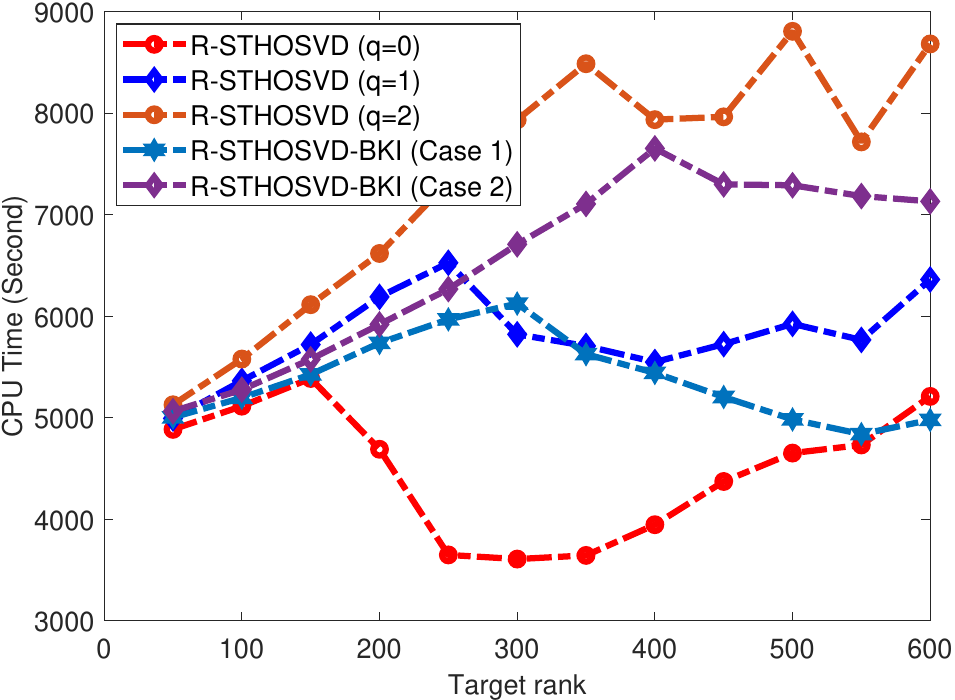}&
\includegraphics[width=1.9876in, height=1.65in]{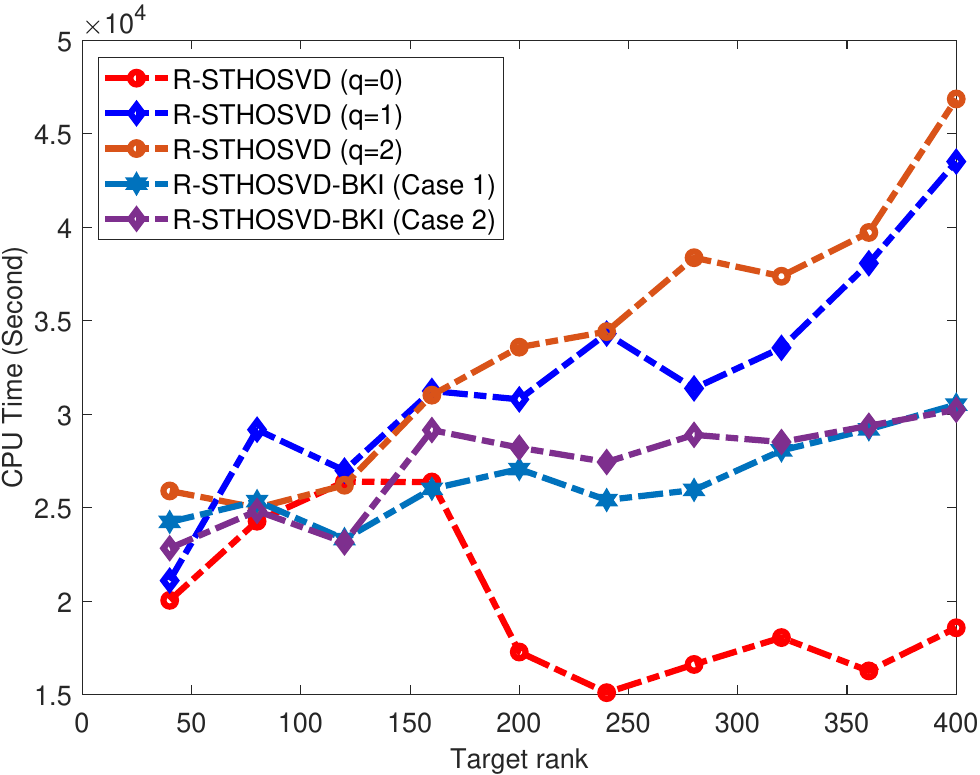} %cv25-time1
\\
 \tabincell{c}{
\footnotesize{{{(a)}} CIs: $5362 \times  6981 \times   3 $,   $\Gamma=\{1,2\}$},\\  \footnotesize{$(\textit{SR},\textit{NR})=(0.1,0.5)$}
}  &
 \tabincell{c}{  \footnotesize{(b) MRSIs: $2500  \times 2500  \times 4  \times 7 $, $\Gamma=$}
\\   \footnotesize{$\{1,2,3,4\}$, $(\textit{SR},\textit{NR})=(0.2,0.5)$} } &
  \tabincell{c}{
  \footnotesize{(c) CVs: $1080 \times 1920  \times 3  \times 100 $, $\Gamma=$} \\ \footnotesize{$\{1,2,4\}$, $(\textit{SR},\textit{NR})=(0.2,0.5)$}
}

\end{tabular}
\vspace{-0.15cm}
\caption{
%The influence of %various regularization parameters $\lambda$
%different  nonconvex combinations $\Phi(\cdot)$+$\psi(\cdot)$  upon  recovery  performance
%under  various  sampling rates $SR$ and impulse noise levels $NR$. % in which
The influence of various  fixed-rank  LRTA % adaptive
 %randomized  Tucker  compression
 methods
 upon restoration performance %results
  of the %our
  proposed randomized GNRHTC algorithm
 %within
 under   different target rank %block size $b$; %techniques;
 $\bm{r}=(r_1, \cdots, r_d)$.
%
%
%The R-STHOSVD algorithm based on  power iteration utilizes   oversampling parameter $p=5$.
%Left --- third-order:  $1000 \times 1000 \times 1000$, Right --- fourth-order: $200 \times 200 \times 200 \times 200$.
%(left --- third-order:  $1000 \times 1000 \times 1000$, right --- fourth-order: $200 \times 200 \times 200 \times 200$).
In the proposed %R-STHOSVD-BKI
 algorithm,
  set $\bm{q}=[4,4,*]$, $\bm{q}=[4,4,*,*]$, or $\bm{q}=[4,4,*,4]$,     $\bm{b}= \lceil\bm{r}./4\rceil$ (\textbf{Case 1}),
 $\bm{b}= \lceil\bm{r}./3\rceil$ (\textbf{Case 2}).
 An asterisk indicates that compression is not applied to this mode.
% an asterisk indicates that this dimension is not compressed
}
\vspace{-0.3cm}
\label{fixed-rank-recovery}
\end{figure*}

%
%%%%%%%%%%%%%%%%%%%%%%%%%%%%%%%%%%%%%%%%%%%%%%%%%%%%%%%%%%%%%%%%%%%%%%%%%%%%%%%%%%%%%%%%%%%%%%%%%%%%%%%%%%%%%%%%%%%%%%%%%%%%%%%%%%%%%%%%%%%%%%%%%
\begin{figure}[!htbp]
\renewcommand{\arraystretch}{0.5}
\setlength\tabcolsep{5pt}
\centering
\begin{tabular}{c }
\centering
\includegraphics[width=3.4in, height=2.5in]{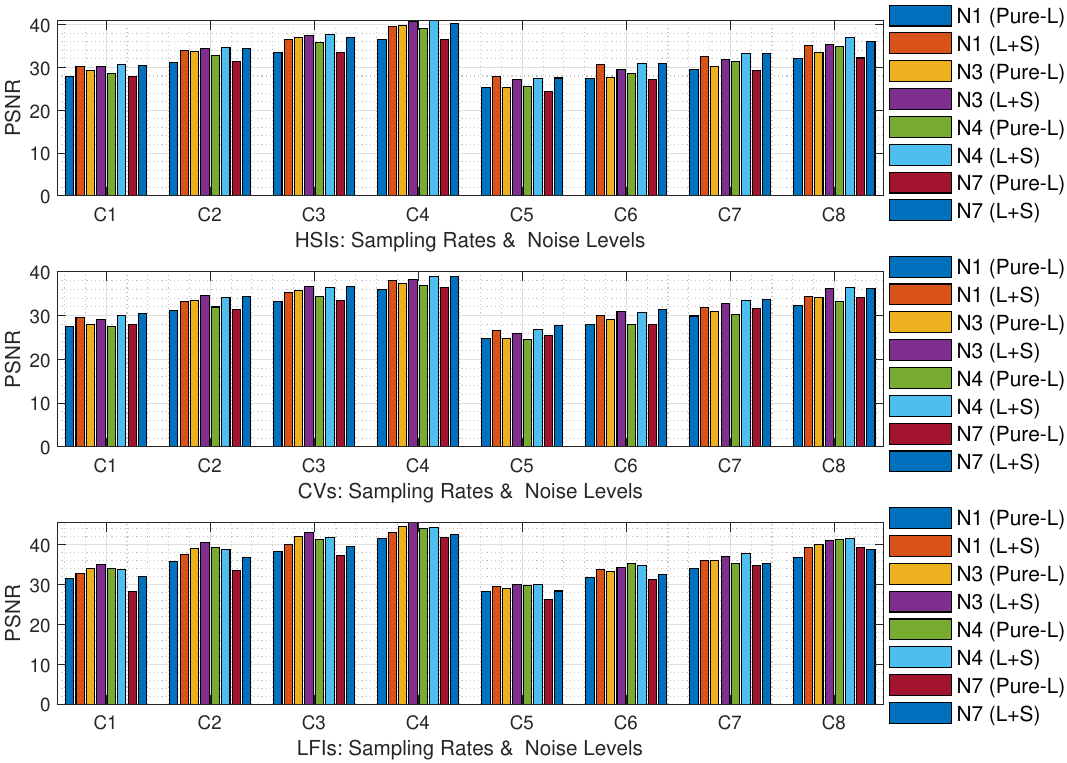} % cvhsi-ls-purelf1
%
%\includegraphics[width=3.4in, height=2.2in]{pureL-LS/hsi-purel-lsv}&
%\includegraphics[width=3.4in, height=2.2in]{pureL-LS/cv-purel-lsv}
%\\
%\footnotesize{{{(a)}} HSIs: $300 \times 300 \times 60$}  &
 % \footnotesize{(b) CVs: $288 \times 352 \times 3 \times 60$}  & \footnotesize{(c) LFIs: $200 \times 300 \times 3 \times 15 \times 15$}

\end{tabular}
\vspace{-0.15cm}
\caption{
The influence of %various regularization parameters $\lambda$
%different  nonconvex combinations $\Phi(\cdot)$+$\psi(\cdot)$
different prior structures (i.e., L+S priors and pure L prior)
upon  recovery  performance
of the  proposed % deterministic  and randomized
GNRHTC algorithm
under  various  sampling rates $SR$ and impulse noise levels $NR$. % in which
%
%
%analyze how different prior structures (i.e., L+S priors and pure L prior) influence the recovery
%performance of our proposed GNRHTC algorithm in deterministic and randomized patterns, respectively;
}
\vspace{-0.23cm}
\label{fig-purel-ls}
\end{figure}

%
%%%%%%%%%%%%%%%%%%%%%%%%%%%%%%%%%%%%%%%%%%%%%%%%%%%%%%%%%%%%%%%%%%%%%%%%%%%%%%%%%%%%%%%%%%%%%%%%%%%%%%%%%%%%%%%%%%%%%%%%%%%%%%%%%%%%%%%%%%%%%%%%%
\begin{figure*}[!htbp]
\renewcommand{\arraystretch}{0.4}
\setlength\tabcolsep{8pt}
\centering
\begin{tabular}{ccc }
\centering

\includegraphics[width=1.976in, height=1.573in]{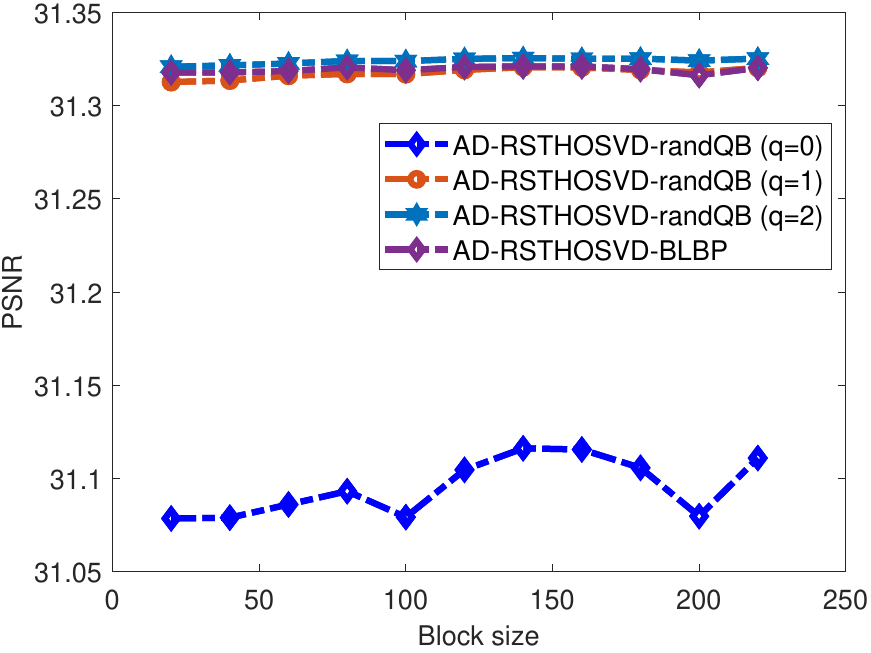}&
\includegraphics[width=1.976in, height=1.573in]{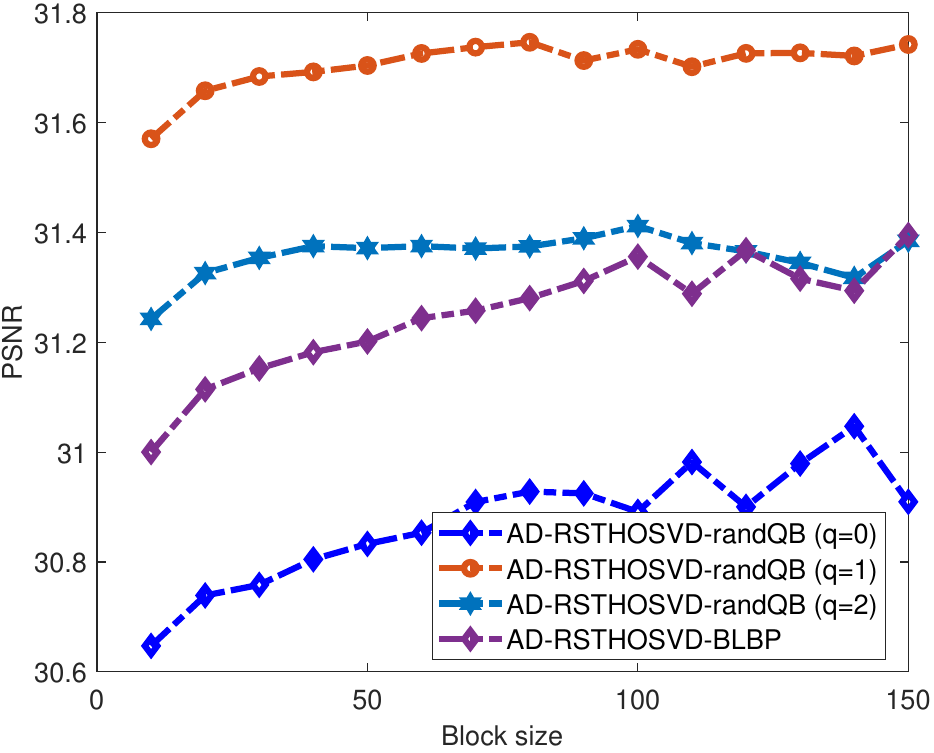}
&
\includegraphics[width=1.976in, height=1.573in]{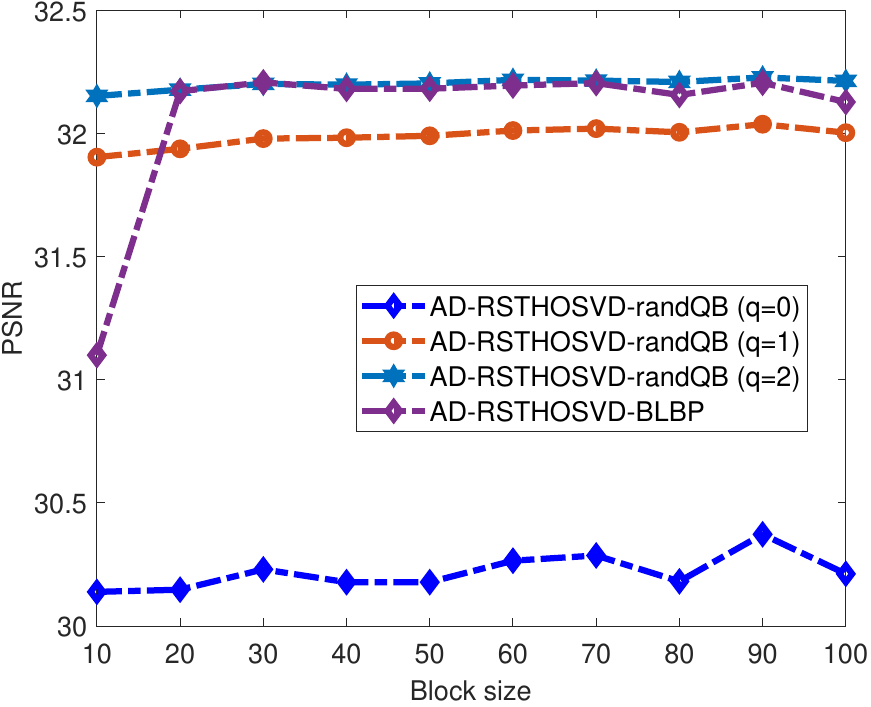}
\\
\includegraphics[width=1.976in, height=1.573in]{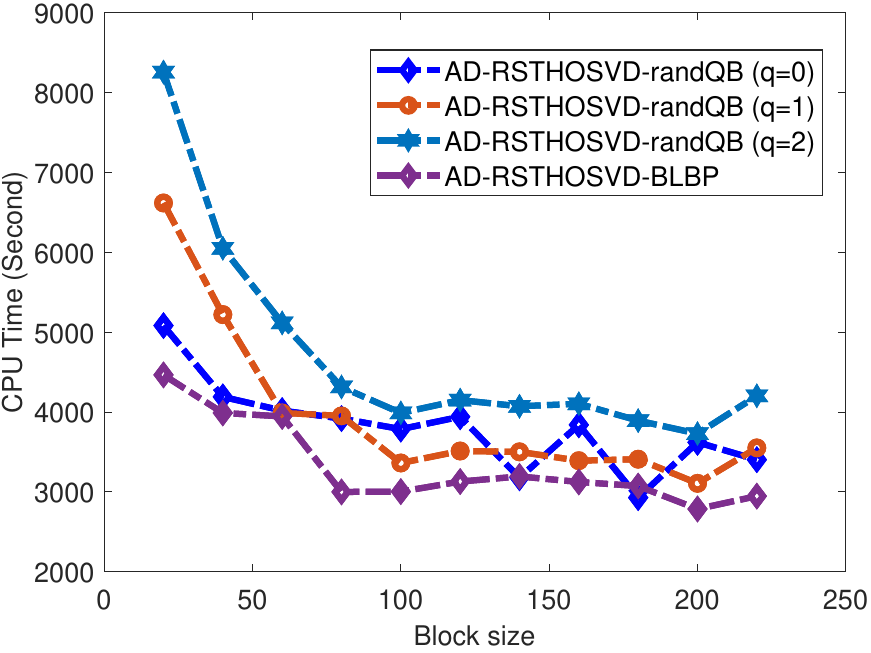}&
\includegraphics[width=1.976in, height=1.573in]{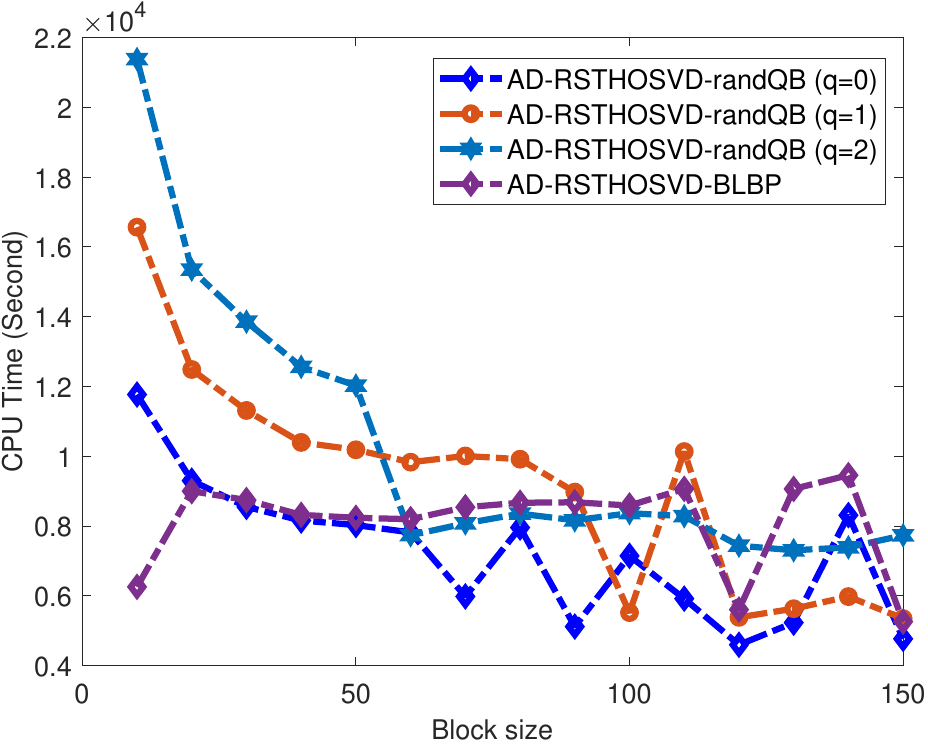}
&
\includegraphics[width=1.976in, height=1.573in]{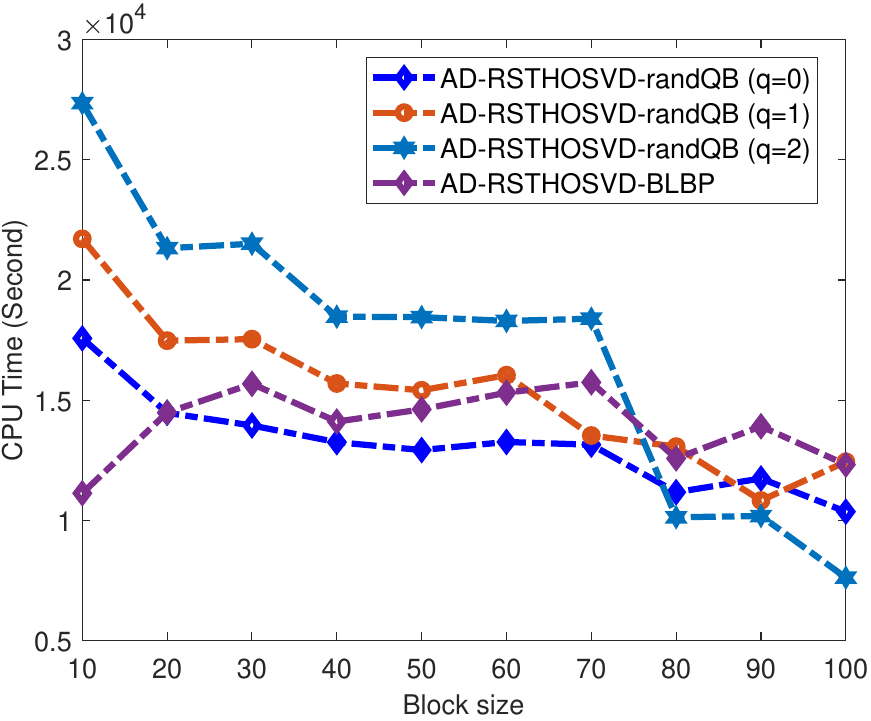}
\\
%\footnotesize{{{(a)}} CIs, $(\textit{SR},\textit{NR})=(0.1,0.5)$}
 \tabincell{c}{
\footnotesize{{{(a)}} CIs: $5362 \times  6981 \times   3 $, $\Gamma=\{1,2\}$,}  \\  \footnotesize{$(\textit{SR},\textit{NR})=(0.1,0.5)$}
}
 &
 % \footnotesize{(b)MRSIs, $(\textit{SR},\textit{NR})=(0.2,0.5)$}
  \tabincell{c}{  \footnotesize{(b) MRSIs: $2500  \times 2500  \times 4  \times 7 $, $\Gamma= $}
\\   \footnotesize{$\{1,2,3,4\}$, $(\textit{SR},\textit{NR})=(0.2,0.5)$} }
   &
%
%  \footnotesize{(c) CVs, $(\textit{SR},\textit{NR})=(0.2,0.5)$}
%%\footnotesize{{{(a)}} Color Image, $(\textit{SR},\textit{NR})=(0.1,0.5)$}  &
%  %\footnotesize{(b) Color Video}  & \footnotesize{(c) MRSI}

 % &
 %&
  \tabincell{c}{
  \footnotesize{(c) CVs: $1080 \times 1920  \times 3  \times 50 $, $\Gamma=$} \\ \footnotesize{$\{1,2,4\}$, $(\textit{SR},\textit{NR})=(0.2,0.5)$}
}

\end{tabular}
\vspace{-0.15cm}
\caption{
%The influence of %various regularization parameters $\lambda$
%different  nonconvex combinations $\Phi(\cdot)$+$\psi(\cdot)$  upon  recovery  performance
%under  various  sampling rates $SR$ and impulse noise levels $NR$. % in which
The influence of various  %fixed-rank
fixed-precision LRTA % adaptive
 %randomized  Tucker  compression
 methods
 upon restoration performance %results
  of the %our
  proposed randomized GNRHTC algorithm
 %within
 %under   different target rank $\bm{r}=(r_1, \cdots, r_d)$. %block size $b$; %techniques;
 %
 %investigate the influence of various  fixed-precision % adaptive
% randomized  Tucker compression approaches
% upon restoration results of our proposed randomized GNRHTC algorithm
 %within
 %under   different block size $b$; %techniques;
 under   different block size $b$.
 %%%%%%%%%%%%%%%%%%%%%%%%%%%%%%%%%%%%%%%%%%%%%%%%%%%%%%%%%%%%%%%%%%%
 Each %fixed-accuracy
 algorithm is run with error tolerance $\epsilon=0.01$,  %target rank $(5,5,5,5,5)$
and the processing order is set to be $\bm{\rho}=\{1,2 \}$. The remaining modes are not compressed.
%%%%%%%%%%%%%%%%%%%%%%%%%%%%%%%%%%%%%%%%%%%%%%%%%%%%%%%%%%%%%%%%%%%%%%%%%%%%%%%%%%%%%%%%%%%%%%%%%%%%%%%%%%%%%%%%
%The last two modes %third mode and
 %are not compressed, indicated here by the asterisk.
}
\vspace{-0.53cm}
\label{fixed-acc-recovery} % \label{fixed-rank-recovery}
\end{figure*}

 \begin{table*}[htbp]
    \caption{
  Quantitative evaluation %PSNR, SSIM, RSE, CPU Time (Second)
  of our   deterministic and randomized methods
  on large-scale CIs, CVs, LFIs and MRSIs.
  In each  method, the values listed from top to bottom  represent the PSNR, SSIM, RSE and CPU Time (Minute), respectively.
  }
  \vspace{-0.15cm}
  \label{large-tensor-experiment}

  \centering
\scriptsize
%\tiny
%\footnotesize
\renewcommand{\arraystretch}{0.8}
\setlength\tabcolsep{3pt}

\begin{tabular}{c c ccccccc  cccc   cccc | c}
 \Xhline{1pt}
\hline
   %  \hline
   %  \hline

     {Data-Type}
     &  \multicolumn{4}{|c|}{CIs
     } &\multicolumn{4}{c|}{HSIs} &\multicolumn{4}{c|}{MRSIs} &\multicolumn{4}{c|}{CVs}  %\\
     & \multirow{3}{*}{
   %Time (s)
   \tabincell{c}{Average\\Value}
   }\\
     \cline{1-1}
     \cline{2-5}
     \cline{6-9}
      \cline{10-13}
      \cline{14-17}
     %\qquad
     $SR$
     &\multicolumn{2}{|c|}{$0.1$}   &\multicolumn{2}{c|}{$0.2$}  &
     \multicolumn{2}{c|}{$0.1$}&\multicolumn{2}{c|}{$0.05$}&
     \multicolumn{2}{c|}{$0.2$}&\multicolumn{2}{c|}{$0.1$}&
     \multicolumn{2}{c|}{$0.1$}&\multicolumn{2}{c|}{$0.2$}\\
     \cline{1-1}
     \cline{2-5}
     \cline{6-9}
      \cline{10-13}
      \cline{14-17}
    %%%
    %\qquad
    $NR$
    &
    \multicolumn{1}{|c|}{$1/3$}&\multicolumn{1}{c|}{$0.5$}  &
    \multicolumn{1}{c|}{$1/3$}&\multicolumn{1}{c|}{$0.5$}  &
    \multicolumn{1}{c|}{$1/3$}&\multicolumn{1}{c|}{$0.5$}  &
    \multicolumn{1}{c|}{$1/3$}&\multicolumn{1}{c|}{$0.5$}  &
    \multicolumn{1}{c|}{$1/3$}&\multicolumn{1}{c|}{$0.5$}  &
    \multicolumn{1}{c|}{$1/3$}&\multicolumn{1}{c|}{$0.5$}  &
    \multicolumn{1}{c|}{$1/3$}&\multicolumn{1}{c|}{$0.5$}  &
    \multicolumn{1}{c|}{$1/3$}&\multicolumn{1}{c|}{$0.5$}
    \\
     \Xhline{1pt}
    \hline
     \hline

     \tabincell{c}{ \textbf{GNRHTC}\\ \textbf{(N1)}}
   &

   \tabincell{c}{  33.099      \\0.8165   \\0.0458   \\   336.15  }   &    \tabincell{c}{  30.929    \\   0.7895    \\     0.0585     \\321.40  }   &
\tabincell{c}{  35.207      \\   0.8600    \\  0.0357    \\321.69   }   &    \tabincell{c}{      34.082  \\ 0.8362  \\  0.0406    \\  332.80}   &

  \tabincell{c}{     31.285    \\   0.9055  \\0.1339 \\  197.09 }   &    \tabincell{c}{    29.935    \\  0.8699     \\  0.1570 \\ 198.36  }   &
  \tabincell{c}{   28.632        \\  0.8310    \\ 0.1817  \\197.63    }   &    \tabincell{c}{ 27.633    \\   0.7862\\0.2038 \\ 190.63 }   &

 \tabincell{c}{   32.998     \\ 0.8274  \\ 0.1571 \\ 604.82    }   &    \tabincell{c}{ 31.636      \\    0.7993     \\   0.1777 \\  616.95  }   &
\tabincell{c}{  31.000        \\ 0.7847   \\ 0.1927 \\ 603.99    }   &    \tabincell{c}{30.212   \\     0.7643 \\   0.2183\\   603.58}   &

%% 165.26
%  31.2949   29.3601   34.5986   28.9671
%  0.0740    0.0924    0.0506    0.0967
%   0.9031    0.8400    0.9220    0.6429
%359.6489  366.8273  369.3418  368.9518
%%
%% 173.10
%30.2118   28.8283   33.4029   32.0305
%0.0838    0.0983    0.0580    0.0680
%0.9027    0.8870    0.9232    0.9090
%478.0646  485.9523  481.8671  481.6970
%%
%418.8568  426.3898  425.6045  425.3244

  \tabincell{c}{31.299    \\    0.9031    \\ 0.0740  \\   478.06}   &    \tabincell{c}{29.361  \\     0.8400 \\  0.0924 \\ 485.95}   &
  \tabincell{c}{  34.598 \\ 0.9220\\ 0.0506  \\  481.86 }   &    \tabincell{c}{ 32.035 \\ 0.9090 \\ 0.0680    \\481.69 }

&  \tabincell{c}{    31.496 \\ 0.8402 \\ 0.1179 \\403.29  }
\\

  \hline

  %  \textbf{R1-GNRHTC}
  \tabincell{c}{ \textbf{R1-GNRHTC}\\ \textbf{(N1)}}
   &

   \tabincell{c}{    33.045       \\  0.8147    \\    0.0460    \\35.024  }   &    \tabincell{c}{    31.155   \\    0.7916   \\  0.0571   \\    37.71   }   &
\tabincell{c}{       34.947   \\     0.8515 \\    0.0368   \\  41.04      }   &    \tabincell{c}{   34.088     \\    0.8331  \\     0.0406  \\ 38.96 }   &

  \tabincell{c}{   28.458   \\ 0.8043    \\ 0.1854  \\   71.01
}   &    \tabincell{c}{ 27.952    \\ 0.7917   \\   0.1965   \\   71.27   }   &
  \tabincell{c}{   27.095   \\ 0.7613  \\  0.2169  \\  68.19      }   &    \tabincell{c}{  26.517    \\   0.7399 \\  0.2318 \\  68.37 }   &

 \tabincell{c}{    33.892    \\ 0.8325   \\0.1231  \\   132.22   }   &    \tabincell{c}{ 32.362       \\  0.8013     \\ 0.1479 \\   132.34   }   &
\tabincell{c}{   31.947     \\  0.7841    \\  0.1545   \\  118.43   }   &    \tabincell{c}{ 29.972   \\   0.7567  \\  0.2207 \\ 91.22 }   &

  \tabincell{c}{ 34.791  \\   0.9199     \\  0.0495  \\ 121.09  }   &    \tabincell{c}{   31.257    \\  0.8701       \\0.0747 \\   111.22  }   &
  \tabincell{c}{ 37.996  \\  0.9370    \\ 0.0342  \\    141.05      }   &    \tabincell{c}{   35.451  \\    0.9162    \\ 0.0458\\  144.67 }

&  \tabincell{c}{
31.932\\
0.8253\\
0.1163\\
88.98
  }
\\

\hline

   % \textbf{R2-GNRHTC}
   \tabincell{c}{ \textbf{R2-GNRHTC}\\ \textbf{(N1)}}
   &

   \tabincell{c}{   33.026    \\  0.8140  \\ 0.0462   \\ 40.28   }   &    \tabincell{c}{   31.187        \\      0.7919      \\    0.0569     \\ 49.65   }   &\tabincell{c}{     34.875   \\  0.8491         \\ 0.0371     \\ 59.54       }   &    \tabincell{c}{ 34.054             \\  0.8318   \\  0.0407  \\59.48}   &

  \tabincell{c}{   28.078 \\ 0.7909   \\ 0.1937    \\   81.18
   }   &    \tabincell{c}{    27.652   \\  0.7808   \\ 0.2239   \\ 81.02    }   &
  \tabincell{c}{ 26.894    \\ 0.7511  \\  0.2034   \\     75.91  }   &    \tabincell{c}{   26.314 \\  0.7325  \\  0.2373 \\  75.35 }   &

 \tabincell{c}{   33.377     \\0.8189    \\0.1280 \\   141.64   }   &    \tabincell{c}{  32.097    \\  0.7947     \\  0.1511 \\ 138.16  }   &
\tabincell{c}{     31.677    \\  0.7789   \\    0.1573 \\   124.11   }   &    \tabincell{c}{  29.837   \\   0.7535   \\   0.2236\\  119.44}   &

 % 167.6905  171.4570  202.4115  201.2602 a=a./120 *90(100)

  \tabincell{c}{ 34.377   \\ 0.9176      \\0.0519  \\ 139.74   }   &    \tabincell{c}{  31.937    \\    0.8915     \\ 0.0687   \\  142.88   }   &
  \tabincell{c}{  37.578   \\  0.9347    \\ 0.0359 \\ 168.67      }   &    \tabincell{c}{  35.887 \\   0.9243      \\ 0.0436  \\ 167.72 }

&  \tabincell{c}{     31.802\\
0.8222\\
0.1187\\
104.04\\
  }
\\
\hline
 \hline
 \Xhline{1pt}
%\hline

     \tabincell{c}{ \textbf{GNRHTC}\\ \textbf{(N4)}}
   &
%%%%%%%

   \tabincell{c}{ 33.385    \\ 0.8150    \\   0.0442 \\  386.81}   &    \tabincell{c}{  28.649   \\0.6988    \\   0.0761        \\  389.52  }   &
\tabincell{c}{    34.723     \\ 0.8378   \\  0.0378     \\ 396.27  }   &    \tabincell{c}{30.799\\  0.7017 \\  0.0597  \\387.06
 }   &

  \tabincell{c}{     33.733    \\  0.9070  \\0.1010  \\233.95
  }   &    \tabincell{c}{  30.768     \\   0.8319     \\ 0.1421 \\    238.29 }   &
  \tabincell{c}{   30.324      \\   0.8664   \\0.1496  \\    241.17 }   &    \tabincell{c}{     28.279   \\  0.7837 \\ 0.1892 \\  240.20}   &

 \tabincell{c}{   34.388    \\ 0.8565   \\ 0.1198   \\  604.69  }   &    \tabincell{c}{   31.335     \\ 0.7628    \\0.1511   \\   609.97 }   &
\tabincell{c}{    31.569     \\  0.7804    \\  0.1651    \\  626.33  }   &    \tabincell{c}{   30.298 \\  0.7389     \\ 0.1899 \\  636.66}   &

 %31.8850   30.0388   35.6824   33.2806
% 32.4696   28.9480   35.4368   22.2520
%
%   0.8921    0.8635    0.9102    0.8625
%   0.8674    0.7245    0.8757    0.2427
%
%0.0691    0.0855    0.0446    0.0589
% 0.0646    0.0969    0.0459    0.2095
%
%   575.5952  584.5368  579.1804  579.0648
%   587.0390  588.9085  589.4786  583.7474

  \tabincell{c}{ 32.469 \\  0.8674    \\  0.0646  \\ 587.03  }   &    \tabincell{c}{  30.038  \\   0.8635      \\ 0.0855 \\ 584.53  }   &
  \tabincell{c}{ 35.682   \\ 0.9102    \\ 0.0446 \\  579.18 }   &    \tabincell{c}{ 33.286 \\   0.8625   \\  0.0589 \\579.06}

&  \tabincell{c}{      31.857\\
0.8177\\
0.1049\\
457.54\\
  }
\\

  \hline

  %  \textbf{R1-GNRHTC}
  \tabincell{c}{ \textbf{R1-GNRHTC}\\ \textbf{(N4)}}
   &

   \tabincell{c}{33.558   \\ 0.8199   \\  0.0433  \\   53.33}   &    \tabincell{c}{ 29.501       \\  0.7363     \\   0.0690   \\     45.51  }   &
\tabincell{c}{    35.336    \\   0.8570  \\   0.0351 \\  62.27     }   &    \tabincell{c}{ 33.468  \\   0.8065 \\       0.0436  \\59.07}   &

  \tabincell{c}{28.989    \\  0.8169   \\  0.1744   \\  68.66 }   &    \tabincell{c}{ 28.394    \\ 0.8077      \\   0.1868    \\   69.58  }   &
  \tabincell{c}{  27.534    \\   0.7756      \\  0.2062   \\   65.38     }   &    \tabincell{c}{ 26.742     \\   0.7545\\  0.2259 \\64.95}   &

 \tabincell{c}{    34.286     \\0.8337   \\ 0.1121    \\ 130.89  }   &    \tabincell{c}{   32.628   \\  0.7896       \\  0.1348  \\  132.66  }   &
\tabincell{c}{    32.163      \\0.7699     \\   0.1408 \\ 116.46    }   &    \tabincell{c}{ 30.127  \\    0.7432 \\  0.1975 \\ 116.83 }   &

 %35.3968         0   38.4886         0
%  0.9132         0    0.9319         0
%   0.0461         0    0.0323         0
%    138.8540  0      174.2273       0
  \tabincell{c}{   35.396  \\ 0.9132    \\ 0.0461  \\ 138.85 }   &    \tabincell{c}{  30.942           \\  0.8511   \\ 0.0770 \\ 125.57  }   &
  \tabincell{c}{  38.488 \\   0.9319    \\0.0323  \\  174.22 }   &    \tabincell{c}{  36.789 \\  0.9228   \\ 0.0393\\ 149.16}

&  \tabincell{c}{      32.146\\
0.8206\\
0.1102\\
98.33\\
 }
\\

\hline

   % \textbf{R2-GNRHTC}
   \tabincell{c}{ \textbf{R2-GNRHTC}\\ \textbf{(N4)}}
   &

   \tabincell{c}{    33.594   \\0.8205    \\ 0.0431     \\ 72.83  }   &    \tabincell{c}{     29.692   \\ 0.7440    \\   0.0675    \\   73.81   }   &
\tabincell{c}{ 35.257       \\0.8545      \\  0.0355     \\76.47 }   &    \tabincell{c}{  33.692  \\0.8129 \\0.0425  \\  69.27  }   &

%

 %0.1839    0.1950    0.2149    0.2321

  \tabincell{c}{    28.525   \\    0.8033    \\ 0.1839       \\  79.05   }   &    \tabincell{c}{   28.192      \\ 0.7955     \\ 0.1950    \\  80.2985    }   &
  \tabincell{c}{   27.754       \\  0.7639    \\0.2149    \\  72.80    }   &    \tabincell{c}{   26.573      \\  0.7461      \\  0.2321    \\71.93 }   &

 \tabincell{c}{  33.973      \\ 0.8267   \\ 0.1139  \\  144.01  }   &    \tabincell{c}{   32.721   \\    0.7955   \\  0.1337 \\ 124.25    }   &
\tabincell{c}{    32.259      \\   0.7766   \\  0.1399 \\  142.01    }   &    \tabincell{c}{ 30.030   \\   0.7420   \\  0.1991 \\  119.84 }   &

  \tabincell{c}{  35.297   \\    0.9171    \\ 0.0467\\ 148.21  }   &    \tabincell{c}{  30.876  \\    0.8571    \\   0.0776 \\    148.83  }   &
  \tabincell{c}{  38.273    \\    0.9334    \\ 0.0331 \\  185.19      }   &    \tabincell{c}{  34.865 \\     0.9032  \\ 0.0490\\   186.93  }

&  \tabincell{c}{     31.973\\
0.8182\\
0.1129\\
112.23
 }
\\
\hline
 \hline
 \Xhline{1pt}
%\hline

\end{tabular}

\vspace{-0.4cm}
\end{table*}

%Randomized-tensor-recovery

%\subsubsection{\textbf{Discussion %Performance
%of different %nonconvex function
%nonconvex function}}
%
%reported result
%%Experimental results of Discussions 1-2
\subsubsection{\textbf{Results and Analysis of Experiments 1-2}}
%

%In Section \ref{gnlstr-oper},
%we list some popular  nonconvex penalty functions  satisfying Assumption \ref{assumpt} and   their corresponding proximity operators.

Note that $\Phi(\cdot)$ and  $\psi(\cdot)$ could be the same.
The nonconvex regularization penalties  %%nonconvex penalty functions
in our experiments include firm,  SCAD, MCP, Log, $\ell_q$, %and
weighted-$\ell_q$ and  capped-$\ell_{q}$,
where $\Phi(\cdot)$ and  $\psi(\cdot)$ are the same nonconvex functions in each case,
namely MCP+MCP (\textbf{N1}), SCAD+SCAD (\textbf{N2}), weighted-$\ell_{q}$+weighted-$\ell_{q}$ (\textbf{N3}),
For different $\Phi(\cdot)$ and  $\psi(\cdot)$, we test ten %eight % five
cases:
%the first three cases are that $\Phi(\cdot)$ is   Log in each case, and $\psi(\cdot)$ are firm,  SCAD and  MCP, respectively
%(called Log+firm, Log+MCP, Log+SCAD, respectively);
%the last three cases are that $\Phi(\cdot)$ is  $\ell_q$ in each case, and $\psi(\cdot)$ are firm,  SCAD and  MCP, respectively
%(called $\ell_q$+firm, $\ell_q$+MCP, $\ell_q$+SCAD, respectively).
capped-$\ell_{q}$+$\ell_q$ (\textbf{N4}),   capped-$\ell_{q}$+MCP (\textbf{N5}),   capped-$\ell_{q}$+SCAD (\textbf{N6}), %capped-$\ell_{q}$+$\ell_q$
Log+$\ell_q$ (\textbf{N7}), Log+MCP (\textbf{N8}), Log+SCAD (\textbf{N9}), $\ell_q$+MCP (\textbf{N10}), $\ell_q$+SCAD (\textbf{N11}),
 MCP+$\ell_q$ (\textbf{N12}), SCAD+$\ell_q$ (\textbf{N13}).
 %%%%%%%%%%%%%%%%%%%%%%%%%%%%%%%%%%%%%%%%%%%%%%%%%%%%%%%%%%%%%%%%%%%%%%%%
 To illustrate the advantages of nonconvex methods, we  also include the  basic convex scenario for comparison,
 i.e., $\ell_{1}$+$\ell_{1}$ (\textbf{N14}).
 We set $(\operatorname{\textit{SR}}, \operatorname{\textit{NR}})$ to be $(0.1,1/3)$ (\textbf{C1}),
 $(0.2,1/3)$ (\textbf{C2}),
 $(0.3,1/3)$ (\textbf{C3}),
 $(0.5,1/3)$ (\textbf{C4}),
 $(0.1,0.5)$ (\textbf{C5}),
 $(0.2,0.5)$ (\textbf{C6}),
 $(0.3,0.5)$ (\textbf{C7}),
 $(0.5,0.5)$ (\textbf{C8}).
 For the  noise/outliers %regularizer %anomaly-sparsity
regularization item,
% we set  $h(\cdot)=\|\cdot\|_{{\mathnormal{F}},1}$.
$h(\cdot)=\|\cdot\|_{\ell_1}$ is defined as the $\ell_{1}$-norm.
%%%%%%%%%%%%%%%%%%%%%%%%%%%%%%%%%%%%%%%%%%%%%%%%%%%%%%%%%%%%%%%%%%%%%%%%%%%%%%%%%%%%
%%%%%%%%%%%%%%%%%%%%%%%%%%%%%%%%%%%%%%%%%%%%%%%%%%%%%%%%%%%%%%%%%%%%%%%%%%%%%%%%%%%%%%%%%%%%%
%
In this experiment, we set $\vartheta=1.1$, $\varpi=10^{-4}$, $\mathfrak{L}=\operatorname{FFT}$, $\mu^{\{0\}}=10^{-3}$, $\mu^{\max}= 10^{10}$;
%$h(\cdot)=\|\cdot\|_{\ell_1}  $;
%$ \Phi(\cdot)= \psi(\cdot)=\operatorname{MCP}$ (\textit{\textbf{MRIs}}),
%$ \Phi(\cdot)$=$\operatorname{capped}$-$\ell_{q}$, $\psi(\cdot)=\ell_q$ (\textit{\textbf{others}});
%
$\Gamma=\{1,2,3\}$ (\textit{\textbf{HSIs}}),  $\Gamma=\{1,2,4\} $ (\textit{\textbf{CVs}}),  $\Gamma=\{1,2,4,5\} $ (\textit{\textbf{LFIs}});
$\lambda=\xi/ (\max{(n_1,n_2)} \cdot  \prod_{i=3}^{d} n_i)^{1/2} $, %n_3
$\xi\in \{2,3,4,5,6,8,10,12,15,18\}$ (\textit{\textbf{N3}}),
$\xi\in \{4, 6,8,10,12,14,16,18,20,25,30,35\}$ (\textit{\textbf{N7-N9}}),
$\xi\in \{0.5,0.8, 1, 1.2, 1.5, 1.8,     2,2.2, 2.5,      3, 3.5,4, 4.5, 5 \}$ (\textit{\textbf{Others}}).
%%%%%%%%%%%%%%%%%%%%%%%%%%%%%%%%%%%%%%%%%%%%%%%%%%%%%%%%%%%%%%%%%%%%%%%%%%%%%%%%%%%%%%%%%%%%%%%%%%%%%%%%%%%%%%%%%%%%%%%

To enhance computational efficiency, we have scaled down the original HSIs and CVs to a certain extent.
The relevant %experimental
results
of \textbf{Experiment 1}
are displayed in Figure \ref{fig-nonconvex-hsi},
from which it can be seen that the N$3$, N$4$ and N$7$ combinations can achieve relatively better recovered results
than other nonconvex combinations in most cases.
%
%from which it can be observed that, in most cases, the N4 and N7 combinations achieve relatively better repair results compared to other non-convex combinations.
Additionally, the PSNR values obtained with the basic %convex
combination N$14$ are lower than those obtained with other combinations,
indicating an advantage of nonconvex regularization over convex regularization in the high-order tensor recovery % RTC
problem.
%%%%%%%%%%%%%%%%%%%%%%%%%%%%%%%%%%%%%%%%%%%%%%%%%%
%Table 1 presents a performance comparison between the proposed deterministic and randomized GNRHTC algorithms, which were experimented on several large-scale tensor datasets.
%
The results of \textbf{Experiment 2} are given  in Figure \ref{fig-purel-ls}, which demonstrate that the GNRHTC %algorithm utilizing
% coupled with the
method modeled by
 joint $\textbf{L}$+$\textbf{S}$ priors  consistently outperforms %outperforms
  the approach  that solely utilizes the $\textbf{L}$ prior
 under different  nonconvex combinations $\Phi(\cdot)$+$\psi(\cdot)$.
 Collectively, these results imply  that the integration of the $\textbf{L}$+$\textbf{S}$ prior %characterization
 strategy with nonconvex regularization for the tensor recovery %RTC
 problem is both effective and advantageous.

Table \ref{large-tensor-experiment} %1
showcases a %comparative analysis of the performance
performance comparison
between the deterministic and randomized versions of the proposed GNRHTC method
across different nonconvex functions. %regularizers.
%which  are
This experiment is  tested on multiple visual large-scale tensor datasets,
i.e., a color image (CI): $5362 \times  6981 \times 3$, a HSI  dataset: $1000  \times  1000 \times 98$,
a MRSI dataset: $2000  \times  2000 \times 4 \times 7$, and a CV dataset: $1080  \times  1920 \times 3 \times 50$.
%%%%%%%%%%%%%%%%%%%%%%%%%%%%%%%%%%%%%%%%%%%%%%%%%%%%%%%%%%%%%%%%%%%%%%%%%%%%%%%%%%%%%%%%%%%%%%%%%%%%%%%%%%%%%%%%%%%%%%%%%%%%%%%
%
%%%%%%%%%%%%%%%%%%%%%%%%%%%%%%%%%%%%%%%%%%%%%%%%%%%%%%%%%%%%%%%%%%%%%%%%%%%%%%%%%%%%%%%%%%%
We can observe that with %a minimal decrease in accuracy
a slight loss % a slight decrease
 in accuracy or comparable accuracy levels maintained, the version incorporating randomized techniques shows a substantial reduction in computational cost compared to its deterministic counterpart.
 %%%%%%%%%%%%%%%%%%%%%%%%%%%%%%%%%%%%%%%%%%%%%%%%%%%%%%%%%%%%%%%%%%%%%%%%%%%%%%
This result indicates that incorporating the randomized LRTA strategy into tensor recovery % RTC
problem can overcome computational bottlenecks, particularly for large-scale high-order tensor data.
%We observe that the version incorporating randomized techniques exhibits a substantial reduction in computational cost compared to its deterministic % counterpart, while maintaining either a slight decrease in accuracy or comparable accuracy levels.

In above-mentioned large-scale experiment, we set
$\mu^{\{0\}}=10^{-3}$,
$\mu^{\max}= 10^{10}$,
 $\mathfrak{L}=\operatorname{FFT}$, $h(\cdot)=\|\cdot\|_{\ell_{1}}$,
 $\vartheta= 1.1$ (\textit{HSIs,  MRSIs}), $1.15$ (\textit{CIs, CVs}), $\varpi=5 \times 10^{-5}$, %1\operatorname{e}-4
$\lambda=\xi/ (\max{(n_1,n_2)} \cdot  \prod_{i=3}^{d} n_i)^{1/2} $,
$\xi \in \{1, 1.2,1.5,1.8, 2\}$ (CIs), $\xi \in \{1, 1.2,  1.5, 2, 2.5,  3\}$ (MRSIs),
 $\xi \in \{1.2, 1.5, 1.8, 2 \}$ (HSIs, deterministic algorithm), $\xi \in \{18, 20,   22, 25, 28  \}$ (HSIs, randomized algorithms),
 $\xi \in \{1.2, 1.5, 1.8, 2, 2.2\}$ (CVs, deterministic algorithm), $\xi \in \{2, 4,6,8,10 \}$ (CVs, randomized algorithms).
For the randomized version using fixed-rank %accuracy
LRTA scheme, we set %$b=40, \epsilon=0.01$,
$\bm{q}=[4,4,*,*]$, $\bm{b}=[60,60,*,*]$ ($\textit{SR}=0.1$),
 $\bm{b}=[80,80,*,*]$ ($\textit{SR}=0.2$) in CVs recovery;
  set %$b=40, \epsilon=0.01$,
  $\bm{q}=[4,4,*,*]$,
$\bm{b}=[80,80,*,*]$ ($\textit{SR}=0.1$),
 $\bm{b}=[100,100,*,*]$ ($\textit{SR}=0.2$) in MRSIs restoration;
 %%%%%%%%%%%%%%%%%%%%%%%%%
 $\bm{q}=[3,3,*]$,
$\bm{b}=[30,30,*]$ ($\textit{SR}=0.05$),
 $\bm{b}=[40,40,*]$ ($\textit{SR}=0.1$) in HSIs recovery;
%%%%%%%%%%%%%%%%%%%%%%%%%%%%%%%%%%
$\bm{q}=[4,4,*]$,
$\bm{b}=[300,300,*]$  in CIs restoration.
%%%%%%%%%%%%%%%%%
For the randomized  version using fixed-accuracy LRTA scheme, we set $\epsilon=0.01$, $b=10$ (HSIs), $b=40$  (CVs, MRSIs), $b=100$  (CIs).
%

%\subsubsection{\textbf{Discussion  of different  regularization %penalties
%items}}
% 讨论低秩+平滑， 单纯低秩
% low-rankness plus smoothness
% low-rank

%\subsubsection{\textbf{Discussion of different   structured noise/outlier}}
%% 几种结构化稀疏异常的讨论：可以以异常检测为例讨论，高光谱，电路板数据

%\subsubsection{\textbf{Discussion of randomized  and deterministic versions}} \label{randomized-diss}

%% 随机算法与确定性方法的对比

%\subsubsection{\textbf{Experimental results of Discussion 3}}
\subsubsection{\textbf{Results and Analysis of Experiment 3-4}}

%\subsubsection{\textbf{Experimental results of Discussion 4}}
%\subsubsection{\textbf{Results and Analysis of Experiment 4}}

%For the  noise/outliers %regularizer %anomaly-sparsity
%regularization item, % we set  $h(\cdot)=\|\cdot\|_{{\mathnormal{F}},1}$.
%$h(\cdot)=\|\cdot\|_{1}$ is defined as the $\ell_{1}$-norm.
%
%For the  noise/outliers %regularizer %anomaly-sparsity
%regularization item, % we set  $h(\cdot)=\|\cdot\|_{{\mathnormal{F}},1}$.
%$h(\cdot)=\|\cdot\|_{1}$ is defined as the $\ell_{1}$-norm.

In this part, we set
$\mathfrak{L}=\operatorname{FFT}$,
 $\vartheta= 1.15$ (\textit{CIs})//$1.2$ (\textit{CVs, MRSIs}), $\varpi=10^{-4}$, %1\operatorname{e}-4
$\lambda=\xi/ (\max{(n_1,n_2)} \cdot  \prod_{i=3}^{d} n_i)^{1/2} $,
$\xi\in \{1, 1.2,1.5,2\}$. Besides, we set
 $\Phi(\cdot)$=$\psi(\cdot)$=MCP,
$h(\cdot)=\|\cdot\|_{\ell_{1}}$,  and a specific set of
$(\operatorname{\textit{SR}}, \operatorname{\textit{NR}})$
 for simplicity.
 %%%%%%%%%%%%%%%%%%%%%  %
 Figure \ref{fixed-rank-recovery}  and Figure \ref{fixed-acc-recovery}  present the impact of different fixed-rank and %different
 fixed-accuracy  LRTA  strategies on the restoration performance of the proposed randomized algorithm, respectively.
 %
 %From Figure 2,
% It is evident
Figure \ref{fixed-rank-recovery} clearly %evidently
shows
 that although R-STHOSVD with $q = 0$ %Method 1
 can achieve the lowest computational cost in most scenarios, it suffers from the worst estimation accuracy.
  In comparison to R-STHOSVD with $q = 1$ and $q = 2$,
  %with q = 0, which uses power iteration,
  our proposed method demonstrates higher
  computational efficiency while maintaining  similar PSNR values.
%
%Overall, %small-block Krylov iteration
Overall, for the randomized algorithm induced by
 R-STHOSVD-BKI strategy, smaller block sizes in Krylov iteration  better enhance recovery efficiency.
%
 %Overall, across different block sizes,
 As can be seen from Figure \ref{fixed-acc-recovery},
 the proposed AD-RSTHOSVD-BLBP scheme demonstrates %lower estimated errors
 %a
 higher PSNR values
 over the AD-RSTHOSVD-randQB method (with power parameter $q = 0$)
 across different block sizes.
 %In terms of speed comparison, AD-RSTHOSVD-BLBP algorithm takes almost exactly the same amount
%of time as AD-RSTHOSVD-randQB method with $q = 0$.
%Besides,
Especially when dealing with  %handling
smaller blocks,
 the
proposed method exhibits %lower %has a
lower computational complexity %while maintaining similar estimated accuracy. %evaluated error and PSNR values.
compared to the subspace-iteration-based counterpart.

\vspace{-0.34cm}

%\newpage
\section{\textbf{Conclusions and  Future Work}}\label{conclusion}

In this article,
%in virtue  of
by using several key  techniques (i.e., random projection, Krylov-subspace iteration, block Lanczos
bidiagonalization process),
we first investigate two novel  %fast and effective %
efficient
%methods for
LRTA %low-rank tensor approximation
algorithms
 under the % by using %both
fixed-precision and fixed-rank paradigms, respectively. Furthermore, we tackle %study
the problem of %fast and efficient
%robust high-order tensor completion
high-order tensor  recovery
 through generalized non-convex modeling and fast %rapid
randomized computation strategies. On the theoretical front, we provide error bound analysis and convergence analysis for the proposed approximation and recovery methods. Experiments on a series of large-scale %real-world
%tensor datasets
multi-dimensional data
have demonstrated the effectiveness and superiority of our proposed  algorithms.
%
 %In the future, we intend to explore deep %generalized
% nonconvex T-CTV regularizer and develop new tensor sketching frameworks
% in a data-driven manner, with the aim of further applying them to high-order tensor approximation and recovery problems.
In the future, we plan to %
devise novel %innovative
nonconvex  sparse regularization %T-CTV regularizers
and %high-efficiency %efficient
quantized   randomized
sketching
frameworks for high-order tensors  through %a data-driven approach,
a model-data dual-driven manner, %approach,
%with the ultimate goal of
with the aim of  leveraging these tools to explore  deep  %efficiently tackle
high-order tensor approximation and recovery approaches %. %problems.
%Leveraging sketching techniques to explore new tensor approximation and recovery methods
from the perspective of continuous  representation \cite{luo2023low, luo2025lowgl}. %modeling. % is also a key focus of our future research.

\ifCLASSOPTIONcaptionsoff
  \newpage
\fi

\ifCLASSOPTIONcaptionsoff
  \newpage
\fi

%\newpage

\bibliographystyle{IEEEtran}
%\bibliography{reference_new1//hyperspectral,reference_new1//various_application,reference_new1//orderp_tensor}
%\bibliography{reference_newnew//hyperspectral,reference_newnew//various_application,reference_newnew//orderp_tensor}
%%%%%%%%%%%
\bibliography{rhtc}
%\bibliography{reference_new//hyperspectral,reference_new//various_application,reference_new//orderp_tensor,reference_newnew//rhtc}

\end{document}